%% file: Bermudez-TIT2026-Gibbs.tex
\documentclass[lettersize,onecolumn]{IEEEtran}
\usepackage[utf8]{inputenc} 
\usepackage[T1]{fontenc}
\usepackage{url}              % provides \url{...}
\usepackage{cite}             % improves presentation of citations
\usepackage[cmex10]{amsmath}	  % Use the [cmex10] option to ensure compliance
\usepackage{amsfonts}
\usepackage{mleftright}       % fix to wrong spacing of \left-,
\usepackage{Bermudez} 
 \newcommand{\Figurespath}{Figures/Journal-TIT2026}
\mleftright

\usepackage{xcolor}

\usepackage{graphicx}
\usepackage{xparse}

\usepackage{orcidlink}
\usepackage{balance}

\begin{document}

%\title{Properties of Gibbs Probability Measures: Applications to Exact Machine Unlearning}
\title{A Unified Framework for Training-Dataset Aggregation and Exact Machine Unlearning via Gibbs Probability Measures}
%\subtitle{}

\author{Yaiza Bermudez {\orcidlink{0009-0005-5231-5906}},
Samir M. Perlaza {\orcidlink{0000-0002-1887-9215}} and
Iñaki Esnaola {\orcidlink{0000-0001-5597-1718}}% 
%Author 4{\orcidlink{}}% <-this % stops a space
\thanks{
Yaiza Bermudez is with the Centre Inria d'Universit\'e C\^ote d'Azur, INRIA, 06902 Sophia Antipolis, France (e-mail: yaiza.bermudez@inria.fr).

Samir M. Perlaza is with the College of Computing, Universit\'e Mohamed VI Polytechnique (UM6P), Ben Guerir, 43150, Morocco; also with the Centre Inria d'Universit\'e C\^ote d'Azur, INRIA, 06902 Sophia Antipolis, France; also with the Department of Electrical and Computer Engineering at  Princeton University, Princeton, NJ 08544 USA; and also with the GAATI Mathematics Laboratory at the University of French Polynesia, 98702 Faaa, French Polynesia (e-mail: samir.perlaza@um6p.ma). 

Iñaki Esnaola is with the School of Electrical and Electronic Engineering, University of Sheffield, Sheffield S1 3JD, U.K.; and also with the Department of Electrical and Computer Engineering, Princeton University, Princeton, NJ 08544 USA (e-mail: esnaola@sheffield.ac.uk).

Parts of this work have been presented at IEEE International Symposium on Information Theory (ISIT), Guangzhou, China, Jun., 2026 \cite{bermudez2026machine, bermudez2026decentralized}; at the IEEE Information Theory Workshop (ITW), Tempe, Arizona, USA, Nov. 2026 \cite{bermudez2026Equivalence}; and appear as Inria Technical Reports in \cite{InriaRR9608, InriaRR9610, InriaRR9613}.
}
}

\maketitle

\begin{abstract}
In this paper, three operations on Gibbs probability measures are studied: renormalization, normalized log-linear combination, and nesting. 
The resulting measures are shown to be Gibbs probability measures with explicit expressions for their objective functions, reference measures, and regularization factors. 
These measures are further characterized as unique solutions to optimization problems involving linear combinations of the original objective functions, subject to either relative entropy regularization or relative entropy constraints. 
When the reference measures are probability measures, geometric interpretations of these three operations are established involving Pythagorean identities and exponential geodesics. 
A central result is the equivalence between nesting and normalized log-linear combinations of Gibbs probability measures. 
These results are then applied to supervised machine learning with Gibbs algorithms, which are represented by Gibbs probability measures.
In this context, either through nesting or log-linear combinations, Gibbs algorithms trained on local datasets are used to form new Gibbs algorithms trained upon the aggregation of all local training datasets. 
Interestingly, the parameters of such operations, determine the weights of each training data point on the aggregated dataset.
%Operations on local Gibbs probability measures are shown to control the contributions of training data. 
This observation is leveraged for achieving: $(a)$ exact machine unlearning in centralized, decentralized, and federated learning systems; and $(b)$ centralized-equivalent performance, in the sense of producing algorithms trained on the aggregation of all local datasets in decentralized and federated learning .
%These results are then applied to supervised machine learning. 
%S uitable choice of the combination and regularization parameters determines the weight of the training data on the aggregated dataset.
%Operations on local Gibbs probability measures are shown to control the contributions of training data. 
%Through these operations, centralized training, in the sense of training on the aggregation of all local datasets; and exact machine unlearning, are shown to be achievable in centralized, decentralized, and federated learning systems.
%
\end{abstract}


%\clearpage
%\tableofcontents
%\clearpage

\section{Introduction}

Gibbs probability measures play an important role in statistical learning theory and information theory, in part due to their versatility for modeling different learning scenarios~\cite{xu2017information,russo2016controlling,medina2022robustness,Bu2024Towards,bu2023generalization,perlaza2023validation,catoni2007pac,ray2023asymptotics,perlaza2024empirical,rodriguezgalvez2024information}.
For instance, under suitable conditions, Gibbs measures describe the long-run behavior of stochastic-gradient-based learning algorithms with time-invariant learning rates~\cite{Azizian2024What}.
Moreover, Gibbs algorithms are represented by data-dependent Gibbs probability measures on the set of models.
Interestingly, these algorithms provide a benchmark for the analysis of the generalization error of learning algorithms~\cite{aminian2021exact,aminian2024information,perlaza2024generalization,perlaza2026corrections,zou2024Generalization}. 
Gibbs probability measures on the set of datasets also characterize worst-case data-generating probability distributions for a fixed model~\cite{zouJSAIT2024}.
More generally, Gibbs probability measures naturally arise as solutions to the minimization or maximization of the expectation of an objective function, such as the empirical risk, subject to a relative entropy regularization. 
In such a case, the relative entropy is defined with respect to a reference measure, and a regularization factor controls the tradeoff between the contribution of the objective function and that of the regularization. 
Similarly, under a specific choice of regularization factor, 
the same Gibbs probability measure also solves the corresponding minimization or maximization problem subject to a relative entropy constraint~\cite{csiszar1975idivergence,perlaza2024empirical,daunas2024equivalence,daunas2025asymmetry,PerlazaEntropy2025}.
These variational characterizations relate an objective function, a reference measure and a regularization factor to the resulting Gibbs probability measures.
This relation allows jointly studying the effects of the regularization factor and the reference measure on the resulting Gibbs algorithm~\cite{perlaza2024empirical,perlaza2023validation,Bu2024Towards}.
It also motivates the study of operations that transform or combine Gibbs probability measures, with the aim of determining how the objective functions, reference measures, and regularization factors affect the resulting measure and its associated optimization problems.
The present work studies three such operations: renormalization, normalized log-linear combinations, and nesting.
The analysis makes explicit how these operations act on the objective functions, reference measures, and regularization factors, and establishes the associated optimization problems and connections between the operations.
Geometric representations are also developed when the reference measures are probability measures.
These results are specialized to supervised machine learning, where connections with weighted empirical risk are established. Particular emphasis is placed on exact machine unlearning and centralized-equivalent performance, in the sense of training on the aggregation of all local datasets in decentralized and federated learning systems.

\subsection{Operations on Gibbs Probability Measures}

The first operation, referred to as \emph{renormalization}, consists of normalizing a power of the Radon--Nikodym derivative of a probability measure.
This operation is frequently encountered in statistical physics and information theory, and the resulting measure is often referred to as an escort distribution. See for instance \cite{bercher2012simple,beck1993thermodynamics,abe2003geometry,ohara2010dually} and references therein.
In this work, the focus is on the renormalization of a given Gibbs probability measure, in the particular case in which the measure used for the normalization is allowed to differ from the reference measure of the given measure.
In this work, Theorem~\ref{TheoremApril16at13h13in2026Antibes} shows that the resulting measure is a Gibbs probability measure with the same objective function, a modified regularization factor, and a possibly different reference measure.
These two changes can be determined independently, and the change of reference measure may restrict the support of the resulting Gibbs measure.
Hence, renormalization describes how a Gibbs probability measure can be transformed through changes in its regularization factor and reference measure.

The second operation, referred to as \emph{normalized log-linear combination}, extends renormalization to several probability measures.
More specifically, this operation consists of normalizing a product of powers of Radon--Nikodym derivatives. Related constructions have been studied beyond Gibbs measures in~\cite{bercher2012simple,lalitha2018social,asadi2025hierarchical}.
In this work, Theorem~\ref{TheoremMay20at13h57in2026Antibes} provides an explicit expression for the normalized log-linear combination of several  Gibbs probability measures.
The resulting measure is a Gibbs probability measure whose objective function is a linear combination of the original objective functions, with coefficients determined by the powers used in the operation and the regularization factors.
This explicit expression is complemented by Theorem~\ref{TheoremApril16at15h23in2026Antibes}, which establishes equivalent formulations for the resulting measure and its role on some specific regularized and constrained optimization problems.

The third operation, referred to as \emph{nesting}, consists of constructing a Gibbs probability measure whose reference measure is itself a Gibbs probability measure.
This construction also appears in sequential updating of Gibbs posteriors in~\cite{perlaza2024empirical} and~\cite{bissiri2016general}.
In this work,  Theorem~\ref{TheoremMay19at113h22in2026Antibes} shows that sequentially nesting several Gibbs probability measures leads to another Gibbs probability measure whose objective function is a linear combination of the objective functions of the original measures.
Theorem~\ref{TheoremMay19at13h57in2026Sophia} establishes the corresponding equivalent variational characterizations of such a resulting measure.
The connection with normalized log-linear combinations is then made explicit by Corollary~\ref{CorollaryApril16at15h20in2026Antibes}, which establishes the equivalence between the two operations under specific conditions.

These operations also admit geometric representations when the reference measures are probability measures.
These representations build on the geometry of escort distributions and relative entropy~\cite{bercher2012simple,csiszar1975idivergence}.
In particular, variations of the parameters of renormalizations and normalized log-linear combinations are shown to induce exponential geodesics.  
For nested Gibbs probability measures, identities relating relative entropies to differences of expectations admit Pythagorean interpretations.
Lemma~\ref{LemmaSept11at15h55in2026Sophia} establishes such an identity for multiple nested Gibbs probability measures.
%
%In the decentralized learning setting, this identity is related to differences in the expected empirical risks associated with the local datasets.

\subsection{Supervised Machine Learning}

In this work, when the operations described above are applied in the context of  supervised machine learning, Gibbs probability measures (algorithms) are defined using empirical risks associated with training datasets as objective functions. 
Normalized log-linear combinations and nesting then yield Gibbs probability measures associated with weighted empirical risks on the aggregation of the corresponding training datasets.
For normalized log-linear combinations, the weights are determined by the coefficients of the combination and the regularization factors of the Gibbs algorithms being combined.
Alternatively, for nested Gibbs algorithms, the weights are determined by the regularization factors, and the dependence on previously used datasets is carried by the data-dependent reference measures.
Thus, the parameters of the corresponding operation determine the weights assigned to the data points within the empirical risk with respect to local datasets.
These parameters can, for instance, be chosen such that the resulting Gibbs probability measure is identical to the measure obtained as the unique solution to an empirical risk minimization with relative entropy regularization (ERM-RER) with respect to the aggregation of the local datasets.
Corollary~\ref{CorollaryDec30at21h00in2025inAntibes} establishes this result in a decentralized learning setting through a sequence of nested Gibbs algorithms.
In such a decentralized setting, each client uses the Gibbs probability measure received from the previous client as its reference measure, and the regularization factors are chosen according to the local dataset sizes.
Alternatively, in a federated learning setting, such a result is  established by Corollary~\ref{CorApril23at18h55in2026Antibes}. In this case, local Gibbs probability measures are combined at a central server through a normalized log-linear combination.
Interestingly, nesting and normalized log-linear combinations yield the same Gibbs probability measure as centralized training, thereby establishing the condition for centralized performance guarantees in decentralized and federated learning systems, respectively.
Nonetheless, the connection with weighted empirical risks is not restricted to reproducing centralized training.
More generally, these operations provide a way to control the weight assigned to each data point through their parameters and its membership to a specific dataset.
In particular, the contribution of selected data points can be canceled through a suitable choice of these parameters, which leads to applications to exact machine unlearning.

%\subsection{Applications to Exact Machine Unlearning}

In this work, the term \emph{machine unlearning} refers to the task of removing the influence of part of the training dataset from a learning algorithm while seeking to avoid retraining from scratch~\cite{Nguyen2025Unlearning,Sai2024Machine,Xu2024Solutions}.
Early formulations of data deletion in learning compare the updated algorithm with the algorithm trained exclusively upon the retained dataset~\cite{ginart2019making,cao2015towards}.
In \emph{exact unlearning}, the probability distributions on the models induced by these two algorithms are required to be identical.
Other formulations allow approximate removal, subject to guarantees on the discrepancy between these distributions~\cite{guo2020certified,bourtoule2021machine}.
This work studies exact unlearning of datasets from a given Gibbs algorithm in centralized learning, as well as in decentralized and federated learning systems. In the decentralized and federated case,  clients collaborate without explicitly exchanging their local datasets~\cite{kairouz2021advances,mcmahan2017communication}.
The connection of nesting and normalized log-linear combinations with weighted empirical risks provides two mechanisms for exact machine unlearning.
In both cases, the parameters of the operation are chosen such that the contribution of the data points to be unlearned is canceled.
Hence, the remaining objective is the empirical risk on the retained dataset, up to a scaling accounted for by the regularization factor.
%
%Exact unlearning follows by identifying the resulting reference measure and regularization factor with those of the ERM-RER problem associated with training exclusively upon the retained dataset.

In this context, the first mechanism uses nesting and consists of maximizing the expected empirical risk on the dataset to be unlearned with relative entropy regularization with respect to the Gibbs algorithm from which data must be unlearnt.
%
%The Gibbs probability measure trained upon the full training dataset is used as the reference measure.
%
Corollary~\ref{CorollaryApril17at18h18in2026Sophia} shows that a particular choice of the regularization factors yields the same Gibbs probability measure as training exclusively upon the retained dataset.
Only the dataset to be unlearned is explicitly used in this optimization problem, while the dependence on the retained dataset is carried by the reference measure.
Together with Corollary~\ref{CorollaryDec30at21h00in2025inAntibes}, this result provides a mechanism for exact unlearning in decentralized learning, where the contribution of a local dataset is canceled from the given Gibbs probability measure.

The second mechanism uses a normalized log-linear combination of the Gibbs probability measure trained upon the full training dataset and a Gibbs probability measure trained upon the dataset to be unlearned.
Corollary~\ref{CorollaryLogLinearExactUnlearning} establishes conditions on the coefficients of the combination and the regularization factors under which the resulting measure achieves exact unlearning.
Alternatively, the unlearning application to federated learning is established by Corollary~\ref{CorFederatedExactUnlearning}.

\subsection{Organization of the Paper}

The remainder of the paper is organized as follows.
Section~\ref{SecPreliminaries} introduces the notation, Gibbs conditional probability measures, and their variational characterizations.
Section~\ref{SecMainResults} studies renormalization, normalized log-linear combinations, and nesting, together with their associated optimization problems and the connections between these operations.
Section~\ref{SecGeometry} develops their geometric representations when the reference measures are probability measures.
Section~\ref{SecApplicationstoSupervisedMachineLearning} specializes these results to supervised machine learning, and Section~\ref{SecUnlearning} presents their applications to exact machine unlearning.
Finally, Section~\ref{SecConclusion} concludes the paper.

\section{Preliminaries}
\label{SecPreliminaries}
This section describes the notation used in this work and formally introduces Gibbs probability measures and Gibbs conditional probability measures. More importantly, the connection between such measures and a particular class of optimization problems involving relative entropies is thoroughly presented.

\subsection{Notation}
Given a set $\set{X} \subseteq \reals^d$, for some $d \in \ints$, the Borel $\sigma$-field defined on $\set{X}$ is denoted by $\mathscr{B}\autoparent{\set{X}}$. 
The set of probability measures that can be defined on the measurable space $\left(\set{X},\mathscr{B}(\set{X})\right)$ is denoted by $\triangle(\set{X})$. 
Given a $\sigma$-finite measure $Q$ on  $\left(\set{X},\mathscr{B}(\set{X})\right)$, the set of probability measures in $\simplex{\set{X}}$ that are absolutely continuous with respect to $Q$ is denoted by $\simplexabs{Q}{\set{X}}$.  
Given a measure $P \in \simplexabs{Q}{\set{X}}$, the notation $P \ll Q$ stands for ``the measure $P$ is absolutely continuous with respect to $Q$''. Thus, 
\begin{IEEEeqnarray}{rCl}
\simplexabs{Q}{\set{X}} & \triangleq & \left\lbrace P \in \simplex{\set{X}} : P \ll Q \right\rbrace.
\end{IEEEeqnarray}
The Radon-Nikodym derivative of $P \in \simplexabs{Q}{\set{X}}$ with respect to $Q$ is a map $\set{X} \to [0,+\infty)$, denoted by $\RND{P}{Q}$. See for instance \cite[Section 2.2]{ash2000probability}.
Using this notation, the relative entropy or Kullback-Leibler (KL) divergence is defined as follows.
\begin{definition}[Relative Entropy]
\label{DefRelativeEntropy}
Given a probability measure $P$ and a $\sigma$-finite measure $Q$, both on the same measurable space, with $P\ll Q$ the relative entropy of $P$ with respect to $Q$ is 
\begin{IEEEeqnarray}{rCl}
\label{EqSept4at10h24in2026Sophia}
\KL{P}{Q} &\triangleq& \int \RND{P}{Q}(x) \log \left(\RND{P}{Q}(x)\right)  \mathrm{d}Q(x).
\end{IEEEeqnarray}
\end{definition}
Note that when $Q$ in Definition~\ref{DefRelativeEntropy} is not a probability measure, the relative entropy might be negative. In particular, when $Q$ is the corresponding Lebesgue measure, the relative entropy $\KL{P}{Q}$ becomes negative Shannon's differential entropy \cite[Section 8.3]{ash2012information}. 

In this work, conditional probability measures are distinguished from probability measures in the following sense.
\begin{definition}[Conditional Probability]
A family $P_{Y|X}\,\triangleq\,(P_{Y|X=x})_{x\in\set{X}}$ of elements of $\triangle\bigl(\set{Y}\bigr)$ indexed by $\set{X}$ is said to be a conditional probability measure if, for all measurable sets $\mathcal{B} \in \mathscr{B}(\set{Y})$, the map
\begin{equation}
\function{\set{X}}{[0,1]}{x}{P_{Y|X=x}(\mathcal{B})}
\end{equation}
is Borel measurable. The set of all such conditional probability measures is denoted by $\triangle(\set{Y} \mid \set{X})$.
\end{definition}
In a nutshell, while $P_{Y|X}$ is a conditional probability measure, the measure $P_{Y|X=x}$, obtained by conditioning on a given $x\in\set{X}$, is a probability measure.

\subsection{Gibbs Conditional Probability Measures}\label{SecGibbsConditionalProbMeasures}

Consider two sets $\set{X} \subset \reals^{d_1}$ and $\set{Y} \subset \reals^{d_2}$, for some given integers $d_1$ and $d_2$. 
Gibbs conditional probability measures in $\simplex{\set{Y} | \set{X}}$ are parametrized by a Borel measurable function
\begin{IEEEeqnarray}{rcl}
\label{EqSeptember7at7h05in2026InTheBusToSophia}
h: \set{X} \times \set{Y} \to \mathbb{R};
\end{IEEEeqnarray}
 a $\sigma$-finite measure $Q$ on the measurable space~$(\set{Y}, \mathscr{B}(\set{Y}))$; and a real $\lambda \in \reals\setminus\lbrace 0 \rbrace$.
Using these parameters, Gibbs conditional probability measures are defined as follows. 
\begin{definition}[Gibbs Conditional Probability Measures]
\label{DefGeneralGibbs}
Given a Borel measurable function~$h : \set{X}\times\set{Y}\to\reals$; a~$\sigma$-finite measure~$Q$ on the measurable space~$(\set{Y}, \mathscr{B}(\set{Y}))$; and
a real~$\lambda \in \reals\setminus\lbrace 0 \rbrace$, the conditional probability measure~$P^{(h, Q, \lambda)}_{Y \mid X}\in \triangle\left(\set{Y} \mid \set{X} \right)$ is said to be an~$(h, Q, \lambda)$-Gibbs conditional probability measure if
\begin{IEEEeqnarray}{rCl}
\label{EqGeneralGibbsIntegrability}
\mbox{$\forall x \in \set{X}$, } 0 < \int \exp\left( -\frac{1}{\lambda} h\autoparent{x,y}\right)\mathrm{d}Q\left( y \right) < + \infty;
\end{IEEEeqnarray}
and for all~$\left(x, y \right) \in \set{X} \times \supp Q$,
\begin{IEEEeqnarray}{rcl}
\label{EqGeneralGibbsRN}
\frac{\mathrm{d}P_{Y\mid X = x}^{(h, Q, \lambda)}}{\mathrm{d}Q } \autoparent{y} & = & \frac{\exp\autoparent{- \frac{1}{\lambda} h\autoparent{x,y}}}{ \displaystyle \int \exp\autoparent{- \frac{1}{\lambda} h\autoparent{x,\tilde{y}}} \mathrm{d}Q\autoparent{\tilde{y}}}.
\end{IEEEeqnarray}
\end{definition}
While $P_{Y|X}^{(h,Q,\lambda)}$ is an $(h, Q, \lambda)$-Gibbs conditional probability measure, the measure $P_{Y|X=x}^{(h,Q,\lambda)}$, obtained by conditioning on a given $x\in\set{X}$, is an $(h, Q, \lambda)$-Gibbs probability measure.

The relevance of $(h, Q, \lambda)$-Gibbs probability measures stems from the fact that under some conditions, they are the unique solutions to particular optimization problems involving relative entropy. The following lemma highlights such a fact.  
\begin{lemma}\label{LemmaJuly28at10h54in2026Sophia}
When~$\lambda > 0$, the $(h, Q, \lambda)$-Gibbs probability measure $P_{Y|X=x}^{(h,Q,\lambda)}$ in~\eqref{EqGeneralGibbsRN} is the unique solution, if it exists, to  
\begin{IEEEeqnarray}{rCl}
\label{EqMarch4at9h04in2026SophiaA}
\min_{P \in \simplexabs{Q}{\set{Y}}} & &  \int h(x,y) \mathrm{d}P(y) + \lambda \KL{P}{Q}.
\end{IEEEeqnarray}
Alternatively, when~$\lambda < 0$, such a measure $P_{Y|X=x}^{(h,Q,\lambda)}$ is the unique solution, if it exists, to 
\begin{IEEEeqnarray}{rCl}
\label{EqMarch4at9h04in2026SophiaB}
\max_{P \in \simplexabs{Q}{\set{Y}}} & &  \int h(x,y) \mathrm{d}P(y) + \lambda \KL{P}{Q}.
\end{IEEEeqnarray}
\end{lemma}
\begin{IEEEproof}
The proof is immediate from \cite[Lemma~1]{PerlazaEntropy2025}.
\end{IEEEproof}

Two optimization problems that are also closely related to the probability measure~$P_{Y\mid X = x}^{(h, Q, \lambda)}$ in~\eqref{EqGeneralGibbsRN} are the following:
\begin{subequations}\label{EqMarch4at9h04in2026SophiaC}
\begin{IEEEeqnarray}{rCl}
\min_{P \in \simplexabs{Q}{\set{Y}}} & &  \int h(x,y) \mathrm{d}P(y) \\
\label{EqMarch4at9h04in2026SophiaC1}
\mbox{s.t.} & \quad & \KL{P}{Q} \leqslant \gamma;
\end{IEEEeqnarray} 
\end{subequations}
and
\begin{subequations}
\label{EqMarch4at9h04in2026SophiaD}
\begin{IEEEeqnarray}{rCl}
\max_{P \in \simplexabs{Q}{\set{Y}}} & &  \int h(x,y) \mathrm{d}P(y) \\
\label{Eq}
\label{EqMarch4at9h04in2026SophiaD1}
\mbox{s.t.}& \quad & \KL{P}{Q} \leq \gamma,
\end{IEEEeqnarray}
\end{subequations}
for some~$\gamma \in \reals$.
The following lemma formalizes these relations.
\begin{lemma}[{\cite[Lemma~4]{PerlazaEntropy2025}}]
\label{LemmaMarch4at9h15in2026inSophia} 
Assume that the optimization problem in~\eqref{EqMarch4at9h04in2026SophiaC} (respectively, in~\eqref{EqMarch4at9h04in2026SophiaD}) admits a solution. Consider the $(h, Q, \lambda)$-Gibbs probability measure~$P_{Y\mid X = x}^{(h, Q, \lambda)}$ in~\eqref{EqGeneralGibbsRN} and assume that $\lambda > 0$ (respectively, $\lambda < 0$) is chosen such that
\begin{IEEEeqnarray}{rCl}
\label{EqJuly14in2026at10h17inAntibes}
 \KL{P_{Y\mid X=x}^{(h, Q, \lambda)}}{Q} & = & \gamma.
\end{IEEEeqnarray}
Then, the $(h, Q, \lambda)$-Gibbs probability measure~$P_{Y\mid X = x}^{(h, Q, \lambda)}$ is the unique solution to~\eqref{EqMarch4at9h04in2026SophiaC} (respectively, to~\eqref{EqMarch4at9h04in2026SophiaD}).
\end{lemma}

A property of the $(h, Q, \lambda)$-Gibbs probability measure~$P_{Y\mid X = x}^{(h, Q, \lambda)}$ in~\eqref{EqGeneralGibbsRN} that plays an important role in this work concerns the variation of the relative entropy $\KL{P^{(h,Q,\lambda)}_{Y\mid X=x}}{Q}$ due to changes in the parameter $\lambda$. The following lemma characterizes such a variation. 
 \begin{lemma}
\label{LemmaMarch10at17h10in2026Sophia}
Consider the $(h,Q,\lambda)$-Gibbs conditional probability measure, $P^{(h,Q,\lambda)}_{Y\mid X} \in \triangle\left(\set{Y} \mid \set{X} \right)$, in Definition~\ref{DefGeneralGibbs}. 
Then, given a fixed $x\in\set{X}$, the map 
\begin{IEEEeqnarray}{rCl}
\lambda \mapsto \KL{P^{(h,Q,\lambda)}_{Y\mid X=x}}{Q}
\end{IEEEeqnarray}
is nonincreasing on all intervals within~$(0,+\infty)$, and nondecreasing on all intervals within~$(-\infty,0)$, for which $\KL{P^{(h,Q,\lambda)}_{Y\mid X=x}}{Q} < +\infty$.
Moreover, if the function $y\mapsto h(x,y)$, with $x$ fixed, is not constant almost surely with respect to $Q$, then the monotonicity is strict.
\end{lemma}
\begin{IEEEproof}
The proof is presented in Appendix~\ref{AppProofOfLemmaMarch10at17h10in2026Sophia}.
\end{IEEEproof}

\section{Main Results}
\label{SecMainResults}
In this section three operations on Gibbs probability measures are thoroughly studied: renormalization, normalized log-linear combinations and nesting. 

\subsection{Renormalized Gibbs Probability Measures}
\label{SecRenormalization}
The normalization of a power of a density of a probability measure, which is often referred to as \emph{renormalization}, is classical in statistical physics and information theory, and is not limited to Gibbs measures. See for instance \cite{bercher2012simple, wilson1983renormalization, asadi2025hierarchical, chhabra1989direct, beck1993thermodynamics, abe2003geometry, ohara2010dually, bercher2009source, tsallis2009introduction, abe2005necessity} and references therein. 
\begin{definition}[Renormalization]
\label{DefRenormalization}
Consider an $\alpha \in \reals\setminus\lbrace 0 \rbrace$, two $\sigma$-finite measures $Q_1$ and $Q_2$, and a probability measure $P$, all on the same measurable space with $P \ll Q_1$ and $Q_2 \ll Q_1$. The $\autoparent{\alpha,Q_1, Q_2}$-renormalization of $P$ is a probability measure,  denoted by $R$, such that for all $y \in \supp Q_2$,
\begin{IEEEeqnarray}{rCl}
\label{EqRenormalization}
\RND{R}{Q_2}\autoparent{y}
&=&
\frac{\autoparent{\RND{P}{Q_1}\autoparent{y}}^{\alpha}}
{\displaystyle\int \autoparent{\RND{P}{Q_1}\autoparent{\tilde{y}}}^{\alpha}\mathrm{d}Q_2\autoparent{\tilde{y}}},
\end{IEEEeqnarray}
provided that 
%\begin{IEEEeqnarray}{rCl}
$0
<
\int \autoparent{\RND{P}{Q_1}\autoparent{y}}^{\alpha}\mathrm{d}Q_2\autoparent{y}
< + \infty.
$
%\end{IEEEeqnarray}
\end{definition}
\begin{remark}
Interestingly, $\autoparent{\alpha,Q, Q}$-renormalizations of a given measure $P$, for a $\sigma$-finite measure $Q$ and $\alpha > 0$, with $\alpha \neq 1$, have a connection with the R\'{e}nyi entropy of order $\alpha$, denoted by $\mathsf{H}_{\alpha}$, of $P$ relative to $Q$, which is defined as follows:
\begin{IEEEeqnarray}{rCl}
\mathsf{H}_{\alpha}(P ; Q) & \triangleq & \frac{1}{1 - \alpha} \log \int \left( \RND{P}{Q}(y) \right)^{\alpha}\mathrm{d} Q (y). 
\end{IEEEeqnarray}
In particular, for all $y \in \supp Q$,
\begin{IEEEeqnarray}{rCl}
\label{EqAugust30in2026at18h42inBenGuerir}
\RND{R}{Q}\autoparent{y}
&=& \autoparent{\RND{P}{Q}\autoparent{y}}^{\alpha} \exp\big( - (1- \alpha) \mathsf{H}_{\alpha}(P ; Q) \big).
\end{IEEEeqnarray}
\end{remark}

The focus of this subsection is on the renormalization of Gibbs probability measures.
More specifically, given a $\sigma$-finite measure $Q_1$ and the $(h,Q,\lambda)$-Gibbs probability measure $P^{(h,Q,\lambda)}_{Y\mid X=x}$ in~\eqref{EqGeneralGibbsRN}, the $(\alpha, Q, Q_1)$-renormalization of $P^{(h,Q,\lambda)}_{Y\mid X=x}$, denoted by $R_1$, satisfies for all $y \in \supp Q_1$,
\begin{IEEEeqnarray}{rCl}
\label{EqApril15at17h43in2026Sophia}
\RND{R_1}{Q_1}\autoparent{y} = \frac{\autoparent{\RND{P^{(h,Q,\lambda)}_{Y\mid X=x}}{Q} \autoparent{y}}^{\alpha}}{\displaystyle\int \autoparent{\RND{P^{(h,Q,\lambda)}_{Y\mid X=x}}{Q}\autoparent{\hat{y}}}^{\alpha} \mathrm{d}Q_1\autoparent{\hat{y}}}.
\end{IEEEeqnarray}
The probability measure $R_1$ in~\eqref{EqApril15at17h43in2026Sophia} is referred to as the ``$\alpha$-escort measure of $P^{(h,Q,\lambda)}_{Y\mid X=x}$ relative to $Q_1$''. The reason why the measure $Q$ is not taken into account within such an appellation, is made clear by the following theorem, which shows that the explicit expression of the Radon-Nikodym derivative of $R_1$ with respect to $Q_1$ does not depend on $Q$. 
 
\begin{theorem}
\label{TheoremApril16at13h13in2026Antibes}
The $(\alpha, Q, Q_1)$-renormalization of the Gibbs probability measure $P^{(h,Q,\lambda)}_{Y\mid X=x}$ in~\eqref{EqGeneralGibbsRN}, namely the measure $R_{1}$ in~\eqref{EqApril15at17h43in2026Sophia}, satisfies for all $y \in \supp Q_1$, 
\begin{IEEEeqnarray}{rCl}
\label{EqApril16at13h03in2026Antibes}
\RND{R_1}{Q_1}\autoparent{y} &=& \frac{\exp\autoparent{-\frac{\alpha}{\lambda} h(x,y)}}{\displaystyle \int \exp\autoparent{-\frac{\alpha}{\lambda} h(x,\hat{y})} \mathrm{d}Q_1(\hat{y})} =  \RND{P^{\autoparent{h,Q_1,\frac{\lambda}{\alpha}}}_{Y\mid X=x}}{Q_1}\autoparent{y}. \quad
\end{IEEEeqnarray}
\end{theorem}
\begin{IEEEproof}
The proof is presented in Appendix~\ref{AppProofOfTheoremApril16at13h13in2026Antibes}.
\end{IEEEproof}

Theorem~\ref{TheoremApril16at13h13in2026Antibes} shows that the $(\alpha, Q, Q_1)$-renormalization of the $(h,Q,\lambda)$-Gibbs probability measure $P^{(h,Q,\lambda)}_{Y\mid X=x}$ in~\eqref{EqGeneralGibbsRN} is identical to the~$\autoparent{h,Q_1,\frac{\lambda}{\alpha}}$-Gibbs probability measure $P^{\autoparent{h,Q_1,\frac{\lambda}{\alpha}}}_{Y\mid X=x}$ in~\eqref{EqApril16at13h03in2026Antibes}. 
More specifically,
%
%\begin{IEEEeqnarray}{rcl}
$\KL{R_1}{P^{\autoparent{h,Q_1,\frac{\lambda}{\alpha}}}_{Y\mid X=x}}  =  0$.
%\end{IEEEeqnarray}
%
In a nutshell, the $\alpha$-escort measure of a Gibbs measure relative to a given reference measure is also a Gibbs probability measure.

Renormalization has a twofold effect on the Gibbs measure $P^{(h,Q,\lambda)}_{Y\mid X=x}$ in~\eqref{EqGeneralGibbsRN}: (i) the regularization factor is changed from $\lambda$ to $\frac{\lambda}{\alpha}$; and (ii) the reference measure is changed from~$Q$ to~$Q_1$.  The second effect might imply a concentration of the measure given that $Q_1 \ll Q$. More specifically, if $\supp Q_1 \subset \supp Q$, then, $\supp P^{\autoparent{h,Q_1,\frac{\lambda}{\alpha}}}_{Y\mid X=x}  = \supp Q_1 \subset \supp  Q =  \supp P^{(h,Q,\lambda)}_{Y\mid X=x}$. 
Interestingly, both effects are not necessarily simultaneously observed, which follows from the fact that $\alpha$ and $Q_1$ can be chosen independently. For instance, if $Q_1$ is chosen to be identical to $Q$, then only the change of the regularization factor is observed. Alternatively, if $\alpha = 1$, only the change of reference measure is observed.
These observations reveal a structural property of Gibbs probability measures, which can be leveraged to characterize the variation of the solution to the optimization problems in \eqref{EqMarch4at9h04in2026SophiaA}, \eqref{EqMarch4at9h04in2026SophiaB}, \eqref{EqMarch4at9h04in2026SophiaC}, and \eqref{EqMarch4at9h04in2026SophiaD} due to changes in the regularization factor $\lambda$ and the reference measure $Q$.
The following corollary, which follows from Lemma~\ref{LemmaJuly28at10h54in2026Sophia} and Lemma~\ref{LemmaMarch4at9h15in2026inSophia}, describes this characterization.
\begin{corollary}
\label{CorApril19at22h17in2026Nice}
When $\frac{\lambda}{\alpha} > 0$, the $(\alpha,Q,Q_1)$-renormalization of the Gibbs probability measure $P_{Y\mid X = x}^{(h,Q,\lambda)}$ in~\eqref{EqGeneralGibbsRN}, namely the measure $R_{1}$ in~\eqref{EqApril15at17h43in2026Sophia}, is the unique solution, if it exists, to the following optimization problems:
\begin{IEEEeqnarray}{rCl}
\label{EqApril19at22h17in2026NiceA}
\min_{P\in\simplexabs{Q_1}{\set{Y}}}
&&
\int h\autoparent{x,y} \mathrm{d}P(y)
\;+\;
\frac{\lambda}{\alpha}\,\KL{P}{Q_1}; \mbox{ and }\\
\nonumber
\min_{P\in\simplexabs{Q_1}{\set{Y}}}
&&
\int h\autoparent{x,y} \mathrm{d}P(y)
\\
\label{EqApril19at22h18in2026NiceA}
\text{s.t.}
& \quad &
\KL{P}{Q_1}
\le
\KL{P_{Y\mid X=x}^{\autoparent{h,Q_1,\frac{\lambda}{\alpha}}}}{Q_1}. 
\end{IEEEeqnarray}
Alternatively, when $\frac{\lambda}{\alpha} < 0$, the measure $R_{1}$ in~\eqref{EqApril15at17h43in2026Sophia}
is the unique solution, if it exists, to the following optimization problems
\begin{IEEEeqnarray}{rCl}
\label{EqApril19at22h17in2026NiceB}
\max_{P\in\simplexabs{Q_1}{\set{Y}}}
&&
\int h\autoparent{x,y} \mathrm{d}P(y)
\;+\;
\frac{\lambda}{\alpha}\,\KL{P}{Q_1}; \mbox{ and }\\
\nonumber
\max_{P\in\simplexabs{Q_1}{\set{Y}}}
&&
\int h\autoparent{x,y} \mathrm{d}P(y)
\\
\label{EqApril19at22h18in2026NiceB}
\text{s.t.}
& \quad &
\KL{P}{Q_1}
\le
\KL{P_{Y\mid X=x}^{\autoparent{h,Q_1,\frac{\lambda}{\alpha}}}}{Q_1}.
\end{IEEEeqnarray}
\end{corollary}

Corollary~\ref{CorApril19at22h17in2026Nice} fully characterizes the effect of changing either the regularization factor $\lambda$ or the reference measure $Q$ in the optimization problems~\eqref{EqMarch4at9h04in2026SophiaA}, \eqref{EqMarch4at9h04in2026SophiaB}, \eqref{EqMarch4at9h04in2026SophiaC}, and \eqref{EqMarch4at9h04in2026SophiaD}, which is a central question  in statistical learning theory. See, for instance, \cite{perlaza2024empirical, bermudez2026machine, perlaza2023validation, bermudez2026decentralized}, and references therein. In a nutshell,  if the solution to the optimization problem in~\eqref{EqMarch4at9h04in2026SophiaA} is known, an $(\alpha, Q, Q_1)$ renormalization of such a solution leads to the solution to the optimization problem in~\eqref{EqApril19at22h17in2026NiceA} in which the regularization factor is $\frac{\lambda}{\alpha}$ and the reference measure is $Q_1$.

\subsection{Normalized Log-Linear Combinations of Gibbs Probability Measures}\label{SecMay27at16h10in2026BusToNice}

The normalization of the product of powers of densities of probability measures, which is often referred to as \emph{normalized log-linear combination of probability measures}, is a generalization of renormalization (Section~\ref{SecRenormalization}). Such a generalization, which is not specific to Gibbs probability measures, has been studied in \cite{asadi2025hierarchical, lalitha2018social, bercher2012simple}, and references therein. The definition of such an operation is as follows:
\begin{definition}[Normalized Log-Linear Combination of Measures]
\label{DefMay19at15h20in2026Sophia}
Consider a vector~$\vect{\alpha} = (\alpha_1, \alpha_2, \ldots, \alpha_K) \in \reals^{K}$, such that  for all $i \inCountK{K}$,  $\alpha_i \in \reals\setminus\lbrace 0 \rbrace$.
Let $P_1, P_2, \ldots, P_K$ be probability measures, let
$Q_1, Q_2, \ldots, Q_K$ be $\sigma$-finite measures, and let $Q$ be a
$\sigma$-finite measure, all on the same measurable space. Assume that for all $i \inCountK{K}$,
$P_i \ll Q_i$ and $Q\ll Q_i$. Then, the
normalized log-linear
combination of $P_1, P_2, \ldots, P_K$ with respect to $Q$ is a probability measure, denoted by
$S_K$, such that for all $y \in \supp Q$,
\begin{IEEEeqnarray}{rCl}
\label{EqMay19at15h20in2026Sophia}
\RND{S_K}{Q}\autoparent{y}
&=&
\frac{\displaystyle \prod_{i=1}^{K} \autoparent{\RND{P_i}{Q_i}\autoparent{y}}^{\alpha_i}}{\displaystyle\int \prod_{i=1}^{K} \autoparent{\RND{P_i}{Q_i}\autoparent{\hat{y}}}^{\alpha_i} \mathrm{d}Q\autoparent{\hat{y}}},
\end{IEEEeqnarray}
provided that 
$0
<
\displaystyle\int \prod_{i=1}^{K} \autoparent{\RND{P_i}{Q_i}\autoparent{\hat{y}}}^{\alpha_i} \mathrm{d}Q\autoparent{\hat{y}} < + \infty$.
\end{definition}

Note that if $K = 1$, the probability measure $S_K$ in \eqref{EqMay19at15h20in2026Sophia} is the $\left( \alpha_1, Q_1, Q \right)$-renormalization of $P_1$, which highlights the generalization with respect to renormalization. 
The following corollary, which follows from Definition~\ref{DefMay19at15h20in2026Sophia}, reinforces this view.
\begin{corollary}
\label{CorollaryAout13at11h08in2026Antibes}
Consider the normalized log-linear combination $S_K$ in
\eqref{EqMay19at15h20in2026Sophia} under the assumptions that for all $i \in \left\lbrace 2, 3, \ldots, K \right\rbrace$ and for all $y \in \supp Q$, 
\begin{IEEEeqnarray}{rCl}
\RND{P_i}{Q_i}\autoparent{y}
&=&
\RND{P_1}{Q_1}\autoparent{y}; \mbox{ and } \\
 \alpha
&\triangleq&
\sum_{i=1}^{K}\alpha_i
\neq 0.
\end{IEEEeqnarray}
Then, the normalized log-linear combination $S_K$ is identical to the
$(\alpha,Q_1,Q)$-renormalization of $P_1$.
\end{corollary}
%Corollary~\ref{CorollaryAout13at11h08in2026Antibes} shows that the $(\alpha,\widetilde{Q},Q)$-renormalization of a probability measure is a particular case of the normalized log-linear combination in~\eqref{EqMay19at15h20in2026Sophia}, obtained when the Radon--Nikodym derivatives of the probability measures involved in the combination coincide with that of a common probability measure $P$ with respect to $\widetilde{Q}$.
%
\begin{remark}
In the case in which $K=2$, with $Q_1=Q_2=Q$ and  $\alpha_1= 1- \alpha_2 = \alpha  \in (0,1)$, for all $y \in \supp Q$ the equality in~\eqref{EqMay19at15h20in2026Sophia} reduces to
\begin{IEEEeqnarray}{rCl}
\label{EqMay20atGeneralTwoMeasures}
\RND{S_2}{Q}\autoparent{y}
&=&
\frac{
\autoparent{\RND{P_1}{Q}\autoparent{y}}^{\alpha}
\autoparent{\RND{P_2}{Q}\autoparent{y}}^{1- \alpha}
}
{
\displaystyle\int
\autoparent{\RND{P_1}{Q}\autoparent{\tilde{y}}}^{\alpha}
\autoparent{\RND{P_2}{Q}\autoparent{\tilde{y}}}^{1-\alpha}
\mathrm{d}Q\autoparent{\tilde{y}}
},
\end{IEEEeqnarray}
where the denominator admits several
connections with classical quantities in information theory.  More specifically, the constant 
\begin{IEEEeqnarray}{rCl}
\label{EqMay20atGeneralHellingerIntegral}
\displaystyle\int
\autoparent{\RND{P_1}{Q}\autoparent{y}}^{\alpha}
\autoparent{\RND{P_2}{Q}\autoparent{y}}^{1-\alpha}
\mathrm{d}Q\autoparent{y}
\end{IEEEeqnarray}
is the Hellinger integral of order $\alpha$ involving $P_1$ and $P_2$, see for instance \cite{Hellinger1909}. Such an integral plays a key role in the definition of the Rényi divergence \cite{Renyi1961,VanErvenHarremoes2014} between the measures $P_1$ and $P_2$, which is defined as follows
\begin{IEEEeqnarray}{rCl}
\label{EqMay20atGeneralRenyiDivergence}
D_{\alpha}\autoparent{P_1\middle\|P_2}
&\triangleq&
\frac{1}{\alpha-1}
\log
\displaystyle\int
\autoparent{\RND{P_1}{Q}\autoparent{y}}^{\alpha}
\autoparent{\RND{P_2}{Q}\autoparent{y}}^{1-\alpha}
\mathrm{d}Q\autoparent{y}.
\end{IEEEeqnarray}
The Hellinger integral involving the measures $P_1$ and $P_2$ also appears in the definition of the Chernoff coefficient,  
\begin{IEEEeqnarray}{rCl}
\label{EqMay20atGeneralChernoff}
\displaystyle\inf_{\alpha\in[0,1]}
\displaystyle\int
\autoparent{\RND{P_1}{Q}\autoparent{y}}^{\alpha}
\autoparent{\RND{P_2}{Q}\autoparent{y}}^{1-\alpha}
\mathrm{d}Q\autoparent{y}, 
\end{IEEEeqnarray} 
and the Chernoff information \cite{Chernoff1952}, 
\begin{IEEEeqnarray}{rCl}
-\log
\displaystyle\inf_{\alpha\in[0,1]}
\displaystyle\int
\autoparent{\RND{P_1}{Q}\autoparent{y}}^{\alpha}
\autoparent{\RND{P_2}{Q}\autoparent{y}}^{1-\alpha}
\mathrm{d}Q\autoparent{y}.
\end{IEEEeqnarray} 
In the case in which $\alpha=\frac{1}{2}$, the quantity in
\eqref{EqMay20atGeneralHellingerIntegral} is the Bhattacharyya coefficient induced by $P_1$ and $P_2$;  and
\begin{IEEEeqnarray}{rCl}
-\log
\displaystyle\int
\autoparent{\RND{P_1}{Q}\autoparent{y}}^{\frac{1}{2}}
\autoparent{\RND{P_2}{Q}\autoparent{y}}^{\frac{1}{2}}
\mathrm{d}Q\autoparent{y},
\end{IEEEeqnarray} 
is the Bhattacharyya distance between $P_1$ and $P_2$, see for instance~\cite{Bhattacharyya1943}.
Finally, it is worth noting that in the case in which $Q \ll P_2$, the integral in
\eqref{EqMay20atGeneralHellingerIntegral} satisfies
\begin{IEEEeqnarray}{rCl}
\label{EqMay20atGeneralMGF}
\displaystyle\int
\autoparent{\RND{P_1}{Q}\autoparent{y}}^{\alpha}
\autoparent{\RND{P_2}{Q}\autoparent{y}}^{1-\alpha}
\mathrm{d}Q\autoparent{y} & = & \displaystyle\int
\exp\autoparent{
\alpha
\log
\RND{P_1}{P_2}\autoparent{y}}
\mathrm{d}P_2\autoparent{y},
\end{IEEEeqnarray}
which is also the moment generating function, evaluated at $\alpha$, of the log-likelihood ratio
$\log
\RND{P_1}{P_2}\autoparent{y}$, when $y$ is sampled from~$P_2$.
These observations are important as they allow, in some specific cases, to write the right-hand side of~\eqref{EqMay20atGeneralTwoMeasures} in terms of information measures, including the R\'{e}nyi entropy, among others.
\end{remark}

In the following, the focus is exclusively on the normalized log-linear combination of $K$ Gibbs
probability measures, which are defined using the following parameters. For all $i\inCountK{K}$, consider a Borel measurable function
\begin{IEEEeqnarray}{rCl}
\label{Eq24aoutat19h32in2026Antibes}
h_i:\set{X}\times\set{Y}\to\reals;
\end{IEEEeqnarray}
a real $\lambda_i\in\reals\setminus\lbrace0\rbrace$; and a $\sigma$-finite
measure $Q_i$ on
$\autoparent{\set{Y},\mathscr{B}\autoparent{\set{Y}}}$. Hence, the
$\autoparent{h_i,Q_i,\lambda_i}$-Gibbs conditional probability measures of interest are
\begin{IEEEeqnarray}{rCl}
\label{EqMay18at19h36in2026AntibesA}
P^{\autoparent{h_i,Q_i,\lambda_i}}_{Y\mid X}
&\in&
\simplex{\set{Y} | \set{X}}.
\end{IEEEeqnarray}

For a fixed vector $\autoparent{x_1, x_2, \ldots, x_K} \in \set{X}^K$, the normalized log-linear combination of the Gibbs probability measures
$P_{Y\mid X=x_1}^{(h_1,Q_1,\lambda_1)},
P_{Y\mid X=x_2}^{(h_2,Q_2,\lambda_2)},\ldots,
P_{Y\mid X=x_K}^{(h_K,Q_K,\lambda_K)}$, with respect to some given $\sigma$-finite measure $Q$, satisfies for all $y \in \supp Q$, 
\begin{IEEEeqnarray}{rCl}
\label{EqMay19at15h40in2026Sophia}
\RND{S_K}{Q}\autoparent{y}
&=&
\frac{
\displaystyle
\prod_{i=1}^{K}
\autoparent{
\RND{
P_{Y\mid X=x_i}^{(h_i,Q_i,\lambda_i)}
}{Q_i}
\autoparent{y}
}^{\alpha_i}
}{
\displaystyle
\int
\prod_{i=1}^{K}
\autoparent{
\RND{
P_{Y\mid X=x_i}^{(h_i,Q_i,\lambda_i)}
}{Q_i}
\autoparent{\tilde{y}}
}^{\alpha_i}
\mathrm{d}Q\autoparent{\tilde{y}}
}.
\end{IEEEeqnarray}
In this section, the vector $\autoparent{x_1, x_2, \ldots, x_K} \in \set{X}^K$ used in \eqref{EqMay19at15h40in2026Sophia} remains fixed. 
Its role becomes apparent in Section~\ref{SecApplicationstoSupervisedMachineLearning}.
%Nonetheless, in the following sections, such a vector plays a central role. 
%
The following theorem provides an explicit expression for the Radon--Nikodym derivative $\RND{S_K}{Q}$ in~\eqref{EqMay19at15h40in2026Sophia}. For this purpose, consider the function
\begin{IEEEeqnarray}{rCl}
\label{EqApril29at13h53in2026Sophia}
\hat{h}_{\vect{\beta}}
&:&
\longfunction{\set{X}^{K}\times\set{Y}}{\reals}
{\autoparent{\vect{x},y}}
{\displaystyle\sum_{i=1}^{K}\beta_i h_i\autoparent{x_i,y},}
\end{IEEEeqnarray}
where
$\vect{\beta}=(\beta_1,\beta_2,\ldots,\beta_K)$ is a vector of
parameters, $h_i$ is the Borel measurable function in
\eqref{Eq24aoutat19h32in2026Antibes}, and
$\vect{x}=(x_1,x_2,\ldots,x_K)\in\set{X}^{K}$.
Using this notation, the following theorem introduces the main result of this subsection. 

\begin{theorem}
\label{TheoremMay20at13h57in2026Antibes}
The normalized log-linear combination 
%of 
%$P^{(h_i,Q_i,\lambda_i)}_{Y\mid X = x_i}$ for all $i \inCountK{K}$ with respect to $Q$, namely the probability measure 
$S_K$ in~\eqref{EqMay19at15h40in2026Sophia}, satisfies for all $y \in \supp Q$,
\begin{IEEEeqnarray}{rCl}
\label{EqMay20at13h58in2026Antibes}
\RND{S_K}{Q}\autoparent{y}
&=&
\frac{
\exp\autoparent{- \displaystyle \sum_{i=1}^{K} \frac{\alpha_i}{\lambda_i}h_i\autoparent{x_i,y}}}
{\displaystyle \int
\exp\autoparent{- \displaystyle \sum_{i=1}^{K} \frac{\alpha_i}{\lambda_i}h_i\autoparent{x_i,\hat{y}}}
\mathrm{d}Q\autoparent{\hat{y}}}\\
\label{EqApril15at22h01in2026AntibesB}
&=& 
\RND{P_{Y\mid \vect{X} = \vect{x}}^{\autoparent{\hat{h}_{\vect{\beta}},Q,1}}}{Q} (y),
\end{IEEEeqnarray}
where $\vect{x} = (x_1,x_2, \ldots, x_K)$;  $\vect{\beta} = \left(\frac{\alpha_1}{\lambda_1}, \frac{\alpha_2}{\lambda_2}, \ldots, \frac{\alpha_K}{\lambda_K} \right)$; the function $\hat{h}_{\vect{\beta}}$ is defined in~\eqref{EqApril29at13h53in2026Sophia}; and the measure $P_{Y\mid \vect{X}= \vect{x}}^{\autoparent{\hat{h}_{\vect{\beta}},Q,1}}$ is an $\autoparent{\hat{h}_{\vect{\beta}},Q,1}$-Gibbs probability measure in $\simplex{\set{Y}}$.
\end{theorem}
\begin{IEEEproof}
The proof is presented in Appendix~\ref{AppProofOfTheoremMay20at13h57in2026Antibes}.
\end{IEEEproof}

Theorem~\ref{TheoremMay20at13h57in2026Antibes} unveils two important observations concerning the effect of log-linearly combining Gibbs probability measures. On one hand, the normalized log-linear combination $S_K$ in~\eqref{EqMay19at15h40in2026Sophia} is identical to the
$\autoparent{\hat{h}_{\vect{\beta}},Q,1}$-Gibbs probability measure
$P_{Y\mid \vect{X} = \vect{x}}^{\autoparent{\hat{h}_{\vect{\beta}},Q,1}}$ in~\eqref{EqApril15at22h01in2026AntibesB}. 
Interestingly, the objective function $\hat{h}_{\vect{\beta}}$  parametrizing the Gibbs conditional probability measure $P_{Y\mid \vect{X}}^{\autoparent{\hat{h}_{\vect{\beta}},Q,1}}$  is the linear combination of the functions $h_1$, $h_2$, $\ldots$, $h_K$ parametrizing the Gibbs conditional probability measures $P_{Y\mid X}^{(h_1,Q_1,\lambda_1)},
P_{Y\mid X}^{(h_2,Q_2,\lambda_2)},\ldots,
P_{Y\mid X}^{(h_K,Q_K,\lambda_K)}$
in~\eqref{EqMay18at19h36in2026AntibesA}, with weights given by $\beta_1 = \frac{\alpha_1}{\lambda_1}$, $\beta_2 = \frac{\alpha_2}{\lambda_2}$, $\ldots$, $\beta_K = \frac{\alpha_K}{\lambda_K}$, respectively.
On the other hand, for all $i\inCountK{K}$, it holds that $Q\ll Q_i$ and thus, 
\begin{IEEEeqnarray}{c}
\supp S_K = \supp Q \subseteq \supp Q_i = \supp P^{(h_i,Q_i,\lambda_i)}_{Y\mid X=x_i}.
\end{IEEEeqnarray}
Essentially, the choice of the reference measure~$Q$ might imply a concentration
of the support of $S_K$ on a support contained within the intersection of the supports of all the Gibbs probability measures taking part in the normalized log-linear combination.

The following theorem further characterizes
the identity in~\eqref{EqApril15at22h01in2026AntibesB} through equivalent
formulations involving the Radon--Nikodym derivative, the relative entropy,
and some optimization problems involving relative entropies.

\begin{theorem}
\label{TheoremApril16at15h23in2026Antibes}
Given the measure $S_K$ in~\eqref{EqMay19at15h40in2026Sophia} and the measure 
$P_{Y\mid \vect{X} = \vect{x}}^{\autoparent{\hat{h}_{\vect{\beta}},Q,1}}$ in~\eqref{EqApril15at22h01in2026AntibesB}, the following statements are equivalent: 
\begin{itemize}
\item[\emph{(i)}]
For all $y\in\supp Q$,
\begin{IEEEeqnarray}{rCl}
\label{EqApril22at22h53in2026Nice}
\RND{S_K}{P_{Y\mid \vect{X} = \vect{x}}^{\autoparent{\hat{h}_{\vect{\beta}},Q,1}}} (y) = 1;
\end{IEEEeqnarray}

\item[\emph{(ii)}]
\begin{IEEEeqnarray}{rCl}
\label{EqApril22at22h54in2026Nice}
\KL{S_K}{P_{Y\mid \vect{X} = \vect{x} }^{\autoparent{\hat{h}_{\vect{\beta}},Q,1}}}
&=&
0;
\end{IEEEeqnarray}

\item[\emph{(iii)}]
The probability measure $S_K$ is the unique solution, if it exists, to the following optimization problem \begin{IEEEeqnarray}{rCl}
\label{EqApril22at22h51in2026Nice}
\min_{P \in \simplexabs{Q}{\set{Y}}} & &
\int \left( \sum_{i = 1}^{K} \frac{\alpha_i}{\lambda_i} h_{i}(x_i, y) \right) \mathrm{d}P(y) + \KL{P}{Q}; \mbox{ and } 
\end{IEEEeqnarray}

\item[\emph{(iv)}]
The probability measure $S_K$ is the unique solution, if it exists, to the following optimization problem \begin{IEEEeqnarray}{rCl}
\nonumber
\min_{P \in \simplexabs{Q}{\set{Y}}} & &
\int  \left( \sum_{i = 1}^{K} \frac{\alpha_i}{\lambda_i} h_{i}(x_i, y) \right)  \mathrm{d}P(y)
\\
\label{EqApril22at22h52in2026Nice}
\text{s.t.} & \quad &
\KL{P}{Q} \le \KL{P_{Y\mid \vect{X} = \vect{x} }^{\autoparent{\hat{h}_{\vect{\beta}},Q,1}}}{Q}.
\end{IEEEeqnarray}
\end{itemize}
\end{theorem}
\begin{IEEEproof}
The proof is presented in Appendix~\ref{AppProofOfTheoremApril16at15h23in2026Antibes}.
\end{IEEEproof}

An interesting observation from Theorem~\ref{TheoremApril16at15h23in2026Antibes} is that the optimization problems in \eqref{EqApril22at22h51in2026Nice} and \eqref{EqApril22at22h52in2026Nice} do not depend on the $\sigma$-finite measures $Q_1$, $Q_2$, $\ldots$, $Q_K$, but exclusively on the $\sigma$-finite measure $Q$.

\subsection{Nested Gibbs Probability Measures}\label{SecNestedGibbsProbMeasures}

Given a Gibbs probability measure, constructing a new Gibbs probability measure by using the given  measure as the reference measure is known as \emph{nesting} Gibbs probability measures. The given Gibbs measure is then said to be nested within the resulting Gibbs measure, yielding a \emph{nested Gibbs probability measure}. Nested Gibbs measures form a special class of Gibbs measures with important applications in statistical learning theory as shown in the remaining of this work.
A formal definition is as follows.
\begin{definition}[Nested Gibbs Probability Measures]
\label{DefNestedGibbsMeasures}
An $(h, Q, \lambda)$-Gibbs probability measure $P^{(h, Q, \lambda)}_{Y \mid X = x}\in \triangle\left(\set{Y} \right)$, of the form in~\eqref{EqGeneralGibbsRN}, is said to be a nested Gibbs probability measure if $Q$ is a Gibbs probability measure.
\end{definition}
For a fixed vector $\autoparent{x_1, x_2} \in \set{X}^2$,  consider the $(h_1,Q_1,\lambda_1)$-Gibbs probability measure $P^{(h_1,Q_1,\lambda_1)}_{Y\mid X  = x_1}$ and the $(h_2,Q_2,\lambda_2)$-Gibbs probability measure $P^{(h_2,Q_2,\lambda_2)}_{Y\mid X  = x_2}$ obtained by conditioning the Gibbs conditional probability measures in~\eqref{EqMay18at19h36in2026AntibesA}. 
%\noindent
%Consider two Borel measurable functions
%\begin{IEEEeqnarray}{rCl}
%\label{Eqh1}
%h_1: & & \set{X}\times\set{Y}\to\reals \mbox{ and }
%\\
%\label{Eqh2}
%h_2: & & \set{X}\times\set{Y}\to\reals.
%\end{IEEEeqnarray}
%Given $x_1\in \set{X}$, consider an $(h_1,Q_1,\lambda_1)$-Gibbs probability measure, denoted by
%\begin{IEEEeqnarray}{rCl}
%\label{EqApril15at18h05in2026Sophia}
%P^{(h_1,Q_1,\lambda_1)}_{Y\mid X  = x_1} \in \simplex{\set{Y}}.
%\end{IEEEeqnarray}
%%
%Given $x_2 \in \set{X}$, consider an $(h_2,Q_2,\lambda_2)$-Gibbs probability measure, denoted by
%\begin{IEEEeqnarray}{rCl}
%\label{EqApril22at21h10in2026Nice}
%P^{(h_2,Q_2,\lambda_2)}_{Y\mid X  = x_2} \in \simplex{\set{Y}}.
%\end{IEEEeqnarray}
%
In the following, the measure $P^{(h_1,Q_1,\lambda_1)}_{Y\mid X  = x_1}$ is said to be nested within $P^{(h_2,Q_2,\lambda_2)}_{Y\mid X  = x_2}$, as $Q_2$ is chosen to be identical to $P^{(h_1,Q_1,\lambda_1)}_{Y\mid X  = x_1}$. This construction yields the $\autoparent{h_2,P^{(h_1,Q_1,\lambda_1)}_{Y\mid X  = x_1},\lambda_2}$-Gibbs probability measure, denoted by
\begin{IEEEeqnarray}{rCl}
\label{EqApril15at18h06in2026Sophia}
P^{\autoparent{h_2,P^{(h_1,Q_1,\lambda_1)}_{Y\mid X =x_1},\lambda_2}}_{Y\mid X =x_2}
\in
\simplex{\set{Y}}.
\end{IEEEeqnarray}
Such a measure is a nested Gibbs probability measure in the sense of
Definition~\ref{DefNestedGibbsMeasures}. Note that $P^{(h_1,Q_1,\lambda_1)}_{Y\mid X  = x_1}$ and $Q_1$ are mutually absolutely continuous \cite[Lemma~3]{perlaza2024empirical}. Then, $\simplexabs{P^{(h_1,Q_1,\lambda_1)}_{Y\mid X  = x_1}}{\set{Y}} = \simplexabs{Q_1}{\set{Y}}$. Using this observation, the following corollary, which follows from Lemma~\ref{LemmaJuly28at10h54in2026Sophia} and Lemma~\ref{LemmaMarch4at9h15in2026inSophia}, unveils an important connection between $P^{\autoparent{h_2,P^{(h_1,Q_1,\lambda_1)}_{Y\mid X =x_1},\lambda_2}}_{Y\mid X =x_2}$ and particular optimization problems involving relative entropies. 
\begin{corollary}
\label{CorMarch17at15h18in2026Sophia}
When $\lambda_2 > 0$, the nested Gibbs probability measure $P^{\autoparent{h_2,P^{(h_1,Q_1,\lambda_1)}_{Y\mid X =x_1},\lambda_2}}_{Y\mid X =x_2}$ in~\eqref{EqApril15at18h06in2026Sophia}
is the unique solution, if it exists, to the following optimization problems:
\begin{IEEEeqnarray}{rCl}
\label{EqMarch5at14h56in2026SophiaA}
\min_{P\in\simplexabs{Q_1}{\set{Y}}} && \int h_2\autoparent{x_2,y} \mathrm{d}P(y) + \lambda_2\,\KL{P}{P^{(h_1,Q_1,\lambda_1)}_{Y\mid X  = x_1}}; \mbox{ and }\\
\nonumber
\min_{P \in \simplexabs{Q_1}{\set{Y}}} & &
\int h_2(x_2,y) \mathrm{d}P(y)
\\
\label{EqMarch18at20h06in2026Nice}
\text{s.t.} & \quad &
\KL{P}{P^{(h_1,Q_1,\lambda_1)}_{Y\mid X  = x_1}} \le \KL{P^{\autoparent{h_2, P^{(h_1,Q_1,\lambda_1)}_{Y\mid X  = x_1}, \lambda_2}}_{Y\mid X =x_2}}{P^{(h_1,Q_1,\lambda_1)}_{Y\mid X  = x_1}}.
\end{IEEEeqnarray}
Alternatively, when $\lambda_2 < 0$, the measure $P^{\autoparent{h_2,P^{(h_1,Q_1,\lambda_1)}_{Y\mid X =x_1},\lambda_2}}_{Y\mid X =x_2}$ in~\eqref{EqApril15at18h06in2026Sophia}
is the unique solution, if it exists, to the following optimization problems
\begin{IEEEeqnarray}{rCl}
\label{EqMarch5at14h56in2026SophiaB}
\max_{P\in\simplexabs{Q_1}{\set{Y}}}
&&
\int h_2\autoparent{x_2,y} \mathrm{d}P(y)
\;+\;\lambda_2\,\KL{P}{P^{(h_1,Q_1,\lambda_1)}_{Y\mid X  = x_1}}; \mbox{ and }\\
\nonumber
\max_{P \in \simplexabs{Q_1}{\set{Y}}} & &
\int h_2(x_2,y) \mathrm{d}P(y)
\\
\label{EqMarch18at20h08in2026Nice}
\text{s.t.} & \quad &
\KL{P}{P^{(h_1,Q_1,\lambda_1)}_{Y\mid X  = x_1}} \le \KL{P^{\autoparent{h_2, P^{(h_1,Q_1,\lambda_1)}_{Y\mid X  = x_1}, \lambda_2}}_{Y\mid X =x_2}}{P^{(h_1,Q_1,\lambda_1)}_{Y\mid X  = x_1}}.
\end{IEEEeqnarray}
\end{corollary}
The following theorem provides an explicit expression for the Radon-Nikodym derivative of the measure $P^{\autoparent{h_2,P^{(h_1,Q_1,\lambda_1)}_{Y\mid X =x_1},\lambda_2}}_{Y\mid X =x_2}$ in~\eqref{EqApril15at18h06in2026Sophia} with respect to the measure~$Q_1$. 
\begin{theorem}\label{TheoremApril16at15h22in2026Antibes}
The nested Gibbs probability measure $P^{\autoparent{h_2,P^{(h_1,Q_1,\lambda_1)}_{Y\mid X =x_1},\lambda_2}}_{Y\mid X =x_2}$ in~\eqref{EqApril15at18h06in2026Sophia} satisfies for all $y\in\supp Q_1$,
\begin{IEEEeqnarray}{rCl}
\label{EqNestedExplicit}
\RND{P^{\autoparent{h_2,P^{(h_1,Q_1,\lambda_1)}_{Y\mid X =x_1},\lambda_2}}_{Y\mid X =x_2}}{Q_1}\autoparent{y}
&=&
\frac{\exp\autoparent{-\frac{1}{\lambda_1} h_1\autoparent{x_1,y} - \frac{1}{\lambda_2} h_2\autoparent{x_2,y}}}
{\displaystyle \int \exp\autoparent{-\frac{1}{\lambda_1} h_1\autoparent{x_1,\tilde{y}} - \frac{1}{\lambda_2} h_2\autoparent{x_2,\tilde{y}}} \mathrm{d}Q_1(\tilde{y})}
\\
\label{EqNestedExplicitB}
&=&
\RND{P_{Y\mid \vect{X} = \vect{x}}^{\autoparent{\hat{h}_{\vect{\beta}},Q_1,1}}}{Q_1} (y),
\end{IEEEeqnarray}
where $\vect{x} = (x_1,x_2)$;  $\vect{\beta} = \left(\frac{1}{\lambda_1}, \frac{1}{\lambda_2}\right)$; the function $\hat{h}_{\vect{\beta}}$ is defined in~\eqref{EqApril29at13h53in2026Sophia}; and the measure $P_{Y\mid \vect{X} = \vect{x}}^{\autoparent{\hat{h}_{\vect{\beta}},Q_1,1}}$ is an $\autoparent{\hat{h}_{\vect{\beta}},Q_1,1}$-Gibbs probability measure in $\simplex{\set{Y}}$.
\end{theorem}
\begin{IEEEproof}
The proof is presented in Appendix~\ref{AppProofOfTheoremMarch9at14h42in2026Sophia}.
\end{IEEEproof}
Theorem~\ref{TheoremApril16at15h22in2026Antibes} shows that the nested Gibbs probability measure $P^{\autoparent{h_2, P^{(h_1,Q_1,\lambda_1)}_{Y\mid X  = x_1}, \lambda_2}}_{Y\mid X =x_2}$ in~\eqref{EqApril15at18h06in2026Sophia} is identical to the $\autoparent{\hat{h}_{\vect{\beta}},Q_1,1}$-Gibbs probability measure $P_{Y\mid \vect{X}=\vect{x}}^{\autoparent{\hat{h}_{\vect{\beta}},Q_1,1}}$ in~\eqref{EqNestedExplicitB}. 
Hence, the nested Gibbs probability measure in~\eqref{EqApril15at18h06in2026Sophia} is itself a  Gibbs probability measure parametrized by the function $\hat{h}_{\vect{\beta}}$ in \eqref{EqApril29at13h53in2026Sophia}, with $\vect{\beta} = \left(\frac{1}{\lambda_1}, \frac{1}{\lambda_2} \right)$, which is a linear combination of the functions $h_1$ and $h_2$, with coefficients $\frac{1}{\lambda_1}$ and $\frac{1}{\lambda_2}$, respectively. 
This observation motivates considering a recursive choice of Gibbs reference
measures to form new nested Gibbs measures. 
For a fixed vector $\autoparent{x_1, x_2, \ldots, x_K} \in \set{X}^K$, consider the $\autoparent{h_k, Q_k, \lambda_k}$-Gibbs probability measure, with $k\in\{1, 2,\ldots,K\}$, 
\begin{subequations}
\label{EqAugust28at11h54in2026Sophia}
\begin{IEEEeqnarray}{rCl}
P^{\autoparent{h_k,Q_k,\lambda_k}}_{Y \mid X = x_k} \in \simplex{\set{Y}}, 
\end{IEEEeqnarray}
where the function $h_k$ is defined in~\eqref{Eq24aoutat19h32in2026Antibes}; $\lambda_k \in \reals\setminus\lbrace 0 \rbrace$; and 
the reference measure $Q_k$ is recursively defined as
\begin{IEEEeqnarray}{rCl}
\label{EqNestedReferenceMeasures}
Q_k
&=&
\left\lbrace
\begin{array}{cc}
Q_1 & \text{if } k = 1\\
P^{\autoparent{h_{k-1},Q_{k-1},\lambda_{k-1}}}_{Y\mid X = x_{k-1}}
& \text{if } k \geqslant 2,
\end{array}
\right.
\end{IEEEeqnarray}
\end{subequations}
for some given $\sigma$-finite measure $Q_1$.

For all $k\in\{2,\ldots,K\}$, the measure $Q_k$ in~\eqref{EqNestedReferenceMeasures} is a Gibbs probability measure. Hence, the $\autoparent{h_k,Q_k,\lambda_k}$-Gibbs probability measure
$P^{\autoparent{h_k,Q_k,\lambda_k}}_{Y\mid X=x_k}$ is a nested Gibbs probability measure in the sense of Definition~\ref{DefNestedGibbsMeasures}.
The following theorem provides an explicit expression for the Radon--Nikodym derivative of the $\autoparent{h_K,Q_K,\lambda_K}$-Gibbs probability measure
$P^{\autoparent{h_K,Q_K,\lambda_K}}_{Y\mid X=x_K}$ in~\eqref{EqAugust28at11h54in2026Sophia} with respect to the measure $Q_1$.
\begin{theorem}
\label{TheoremMay19at113h22in2026Antibes}
The $\autoparent{h_K,Q_K,\lambda_K}$-Gibbs probability measure $P^{\autoparent{h_K,Q_K,\lambda_K}}_{Y\mid X=x_K}$ in~\eqref{EqAugust28at11h54in2026Sophia} satisfies for all $y\in\supp Q_1$,
\begin{IEEEeqnarray}{rCl}
\label{EqMay19at13h30in2026SophiaA}
\RND{P^{\autoparent{h_K,Q_K,\lambda_K}}_{Y \mid X = x_K}}{Q_1}\autoparent{y}
&=&
\frac{\exp\autoparent{- \displaystyle \sum_{i=1}^{K} \frac{1}{\lambda_i} h_i\autoparent{x_i,y}}}
{\displaystyle \int \exp\autoparent{- \displaystyle \sum_{i=1}^{K} \frac{1}{\lambda_i} h_i\autoparent{x_i,\tilde{y}}} \mathrm{d}Q_1(\tilde{y})}
\\
\label{EqMay19at13h30in2026SophiaB}
&=&
\RND{P_{Y\mid \vect{X} = \vect{x}}^{\autoparent{\hat{h}_{\vect{\beta}},Q_1,1}}}{Q_1} (y),
\end{IEEEeqnarray}
where $\vect{x} = (x_1,x_2, \ldots, x_K)$;  
%\begin{IEEEeqnarray}{rCl}
%\label{EqBetaRecursive}
$\vect{\beta}
=
\autoparent{\frac{1}{\lambda_1}, \frac{1}{\lambda_2}, \ldots, \frac{1}{\lambda_K}}$; the function $\hat{h}_{\vect{\beta}}$ is defined in~\eqref{EqApril29at13h53in2026Sophia}; and the measure $P_{Y\mid \vect{X} = \vect{x}}^{\autoparent{\hat{h}_{\vect{\beta}},Q_1,1}}$ is an $\autoparent{\hat{h}_{\vect{\beta}},Q_1,1}$-Gibbs probability measure in $\simplex{\set{Y}}$.
%\end{IEEEeqnarray}
\end{theorem}
\begin{IEEEproof}
The proof is presented in Appendix~\ref{AppProofOfTheoremMay19at113h22in2026Antibes}.
\end{IEEEproof}

Theorem~\ref{TheoremMay19at113h22in2026Antibes} shows that the nested Gibbs probability measure
$P^{\autoparent{h_K,Q_K,\lambda_K}}_{Y\mid X=x_K}$ in~\eqref{EqAugust28at11h54in2026Sophia} is identical to the $\autoparent{\hat{h}_{\vect{\beta}},Q_1,1}$-Gibbs probability measure
$P_{Y\mid \vect{X}=\vect{x}}^{\autoparent{\hat{h}_{\vect{\beta}},Q_1,1}}$
in~\eqref{EqMay19at13h30in2026SophiaB}. Although the reference measures in~\eqref{EqNestedReferenceMeasures} are defined recursively, the Radon--Nikodym derivative in~\eqref{EqMay19at13h30in2026SophiaA} depends on the functions $h_1,h_2,\ldots,h_K$ only through a specific linear combination, i.e., $\hat{h}_{\vect{\beta}}(\vect{x}, y) = \sum_{i=1}^{K}\frac{1}{\lambda_i}h_i\autoparent{x_i,y}$.
The following theorem further characterizes these observations through equivalent formulations involving the Radon--Nikodym derivative, the relative entropy, and some optimization problems involving relative entropies.

\begin{theorem}
\label{TheoremMay19at13h57in2026Sophia}
Given the measure $P^{\autoparent{h_K,Q_K,\lambda_K}}_{Y\mid X=x_K}$ in~\eqref{EqAugust28at11h54in2026Sophia}
%, with $Q_K$ satisfying~\eqref{EqNestedReferenceMeasures} 
and the measure $P_{Y\mid \vect{X }= \vect{x}}^{\autoparent{\hat{h}_{\vect{\beta}},Q_1,1}}$ in~\eqref{EqMay19at13h30in2026SophiaB}, the following statements are equivalent:
\begin{itemize}
\item[\emph{(i)}]
For all $y\in\supp Q_1$,
\begin{IEEEeqnarray}{rCl}
\label{EqMay19at14h14in2026Sophia}
\RND{P^{\autoparent{h_K,Q_K,\lambda_K}}_{Y \mid X = x_K}}{P_{Y\mid \vect{X }= \vect{x} }^{\autoparent{\hat{h}_{\vect{\beta}},Q_1,1}}} (y) &=& 1;
\end{IEEEeqnarray}

\item[\emph{(ii)}] 
\begin{IEEEeqnarray}{rCl}
\label{EqMay19at14h15in2026Sophia}
\KL{P^{\autoparent{h_K,Q_K,\lambda_K}}_{Y \mid X = x_K}}{P_{Y\mid \vect{X }= \vect{x}}^{\autoparent{\hat{h}_{\vect{\beta}},Q_1,1}}} &=& 0;
\end{IEEEeqnarray}

\item[\emph{(iii)}] The probability measure 
$P^{\autoparent{h_K,Q_K,\lambda_K}}_{Y \mid X = x_K}$
is the unique solution, if it exists, to the following optimization problem
\begin{IEEEeqnarray}{rCl}
\label{EqMarch17at6h43in2026BusToSophiaA}
\min_{P \in \simplexabs{Q_1}{\set{Y}}} & &  \int\left( \sum_{i = 1}^{K} \frac{1}{\lambda_i} h_{i}(x_i, y) \right)  \mathrm{d}P(y) + \KL{P}{Q_1}; \mbox{ and }
\end{IEEEeqnarray}
\item[\emph{(iv)}] 
The measure $P^{\autoparent{h_K,Q_K,\lambda_K}}_{Y \mid X = x_K}$
is the unique solution, if it exists, to the following optimization problem
\begin{IEEEeqnarray}{rCl}
\nonumber
\min_{P \in \simplexabs{Q_1}{\set{Y}}} & &
\int \left( \sum_{i = 1}^{K} \frac{1}{\lambda_i} h_{i}(x_i, y) \right) \mathrm{d}P(y)
\\
\label{EqMarch18at9h58in2026SophiaA}
\text{s.t.} & \quad &
\KL{P}{Q_1} \le \KL{P_{Y\mid \vect{X }= \vect{x}}^{\autoparent{\hat{h}_{\vect{\beta}},Q_1,1}}}{Q_1}.
\end{IEEEeqnarray}
\end{itemize}
\end{theorem}
\begin{IEEEproof}
The proof is presented in Appendix~\ref{AppProofOfTheoremMay19at13h57in2026Sophia}.
\end{IEEEproof}

A key observation from Theorem~\ref{TheoremMay19at13h57in2026Sophia} is that the optimization problems in~\eqref{EqMarch17at6h43in2026BusToSophiaA} and
\eqref{EqMarch18at9h58in2026SophiaA} coincide with the optimization problems
in~\eqref{EqApril22at22h51in2026Nice} and
\eqref{EqApril22at22h52in2026Nice}, respectively, when $\alpha_1=\alpha_2 = \ldots = \alpha_K = 1$ and the measures $Q$ and $Q_1$ are identical. Therefore, an equivalence relation between nesting Gibbs probability measures and normalized log-linear combinations can be established as follows. 
\begin{corollary}
\label{CorollaryApril16at15h20in2026Antibes}
Consider $K$ Gibbs probability measures denoted by $P^{\autoparent{h_1, \hat{Q}_1,\lambda_1}}_{Y\mid X = x_1}$, $P^{\autoparent{h_2, \hat{Q}_2,\lambda_2}}_{Y\mid X = x_2}$, $\ldots$, $P^{\autoparent{h_K, \hat{Q}_K,\lambda_K}}_{Y\mid X = x_K}$, where for all $i \inCountK{K}$, the function $h_i$, the nonzero real $\lambda_i$, and the vector $x_i$ are the same as in \eqref{EqAugust28at11h54in2026Sophia}; and the $\sigma$-finite measures $\hat{Q}_{1}$, $\hat{Q}_{2}$, $\ldots$, $\hat{Q}_{K}$ are arbitrary measures, with $\hat{Q}_1$ identical to $Q_1$ in \eqref{EqAugust28at11h54in2026Sophia}. Consider a normalized log-linear combination of such $K$ Gibbs probability measures, denoted by $T \in \simplexabs{Q_1}{\set{Y}}$, such that for all $y \in \supp Q_1$,
\begin{IEEEeqnarray}{rCl}
\label{EqAugust30at22h11in2026BenGuerir}
\RND{T}{Q_1}\autoparent{y}
&=&
\frac{
\displaystyle
\prod_{i=1}^{K}
\RND{
P_{Y\mid X=x_i}^{(h_i,\hat{Q}_i,\lambda_i)}
}{\hat{Q}_i}
\autoparent{y}
}{
\displaystyle
\int
\prod_{i=1}^{K}
\RND{
P_{Y\mid X=x_i}^{(h_i,\hat{Q}_i,\lambda_i)}
}{\hat{Q}_i}
\autoparent{\tilde{y}}
\mathrm{d}Q_1\autoparent{\tilde{y}}
}.
\end{IEEEeqnarray}
Consider also the nested Gibbs probability measure
$P^{\autoparent{h_K,Q_K,\lambda_K}}_{Y\mid X=x_K}$ in \eqref{EqAugust28at11h54in2026Sophia}. Then, 
\begin{IEEEeqnarray}{rcl}
\KL{T}{P^{\autoparent{h_K,Q_K,\lambda_K}}_{Y\mid X=x_K}} & = & 0.
\end{IEEEeqnarray}
\end{corollary}

An interesting observation from Corollary~\ref{CorollaryApril16at15h20in2026Antibes} is that, as long as the measures $\hat{Q}_1$ and $Q_1$ are identical, the measures $\hat{Q}_2$, $\hat{Q}_3$, \ldots, $\hat{Q}_K$ do not influence the normalized log-linear combination $T$ in \eqref{EqAugust30at22h11in2026BenGuerir}, provided that they satisfy the conditions in Definition~\ref{DefMay19at15h20in2026Sophia}.

\section{Geometric Representation}
\label{SecGeometry}

This section provides geometric representations of the renormalized,
log-linearly combined, and nested Gibbs probability measures studied
in Section~\ref{SecMainResults}, under the assumption that the reference
measures are probability measures.
These representations use the connection established by
Lemma~\ref{LemmaMarch4at9h15in2026inSophia} between the Gibbs probability
measure in~\eqref{EqGeneralGibbsRN} and the optimization problems
in~\eqref{EqMarch4at9h04in2026SophiaC}
and~\eqref{EqMarch4at9h04in2026SophiaD}.
In particular,  variations
of some of the parameters of  renormalizations and normalized log-linear
combinations of Gibbs measures are associated with information geometric objects, e.g., exponential geodesics \footnote{In information geometry, an exponential geodesic (or e-geodesic) is a path between probability measures obtained by linearly combining the log densities of the measures involved, subject to a  normalization. See \cite[p.322]{Nielsen2018InformationProjection} and \cite[Sections 2.4 and 2.5]{amari2016information}}, which admit special Pythagorean interpretations.

Consider the $\autoparent{h_k,Q_k,\lambda_k}$-Gibbs
probability measures
$P^{\autoparent{h_k,Q_k,\lambda_k}}_{Y\mid X=x_k}$
in~\eqref{EqGeneralGibbsRN}, with $k\inCountK{K}$.
For all $k\inCountK{K}$, let
\begin{IEEEeqnarray}{rCl}
\label{EqSeptember11at11h28in2026atSophia}
0\leq\gamma_k
&\triangleq&
\KL{P^{\autoparent{h_k,Q_k,\lambda_k}}_{Y\mid X=x_k}}{Q_k}
<+\infty.
\end{IEEEeqnarray}
The reference measure $Q_k$ and $\gamma_k$ in~\eqref{EqSeptember11at11h28in2026atSophia} determine the following
subset of $\simplexabs{Q_k}{\set{Y}}$:
\begin{IEEEeqnarray}{rCl}
\label{EqKLNeighborhood}
\set{G}_{Q_k,\gamma_k}
&\triangleq&
\left\lbrace
P\in\simplexabs{Q_k}{\set{Y}}:
\KL{P}{Q_k}\leq\gamma_k
\right\rbrace.
\end{IEEEeqnarray}
In general, throughout this section, the notation $\set{G}_{Q,\gamma}$ denotes a
set of the form in~\eqref{EqKLNeighborhood}, where the relative entropy is with respect to the reference measure $Q$ and the upper bound on the relative entropy is $\gamma$.
The set $\set{G}_{Q_k,\gamma_k}$ in~\eqref{EqKLNeighborhood} can be
interpreted as a relative-entropy neighborhood around the reference
measure $Q_k$.
Hence, from Lemma~\ref{LemmaMarch4at9h15in2026inSophia}, the measure
$P^{\autoparent{h_k,Q_k,\lambda_k}}_{Y\mid X=x_k}$ is the unique
minimizer when $\lambda_k>0$, or the unique maximizer when $\lambda_k<0$,
of the expectation
$\int h_k\autoparent{x_k,y} \mathrm{d}P\autoparent{y}$,
when $P \in \set{G}_{Q_k,\gamma_k}$.
This observation plays a central role in the remainder of this section.

\subsection{Renormalized Gibbs Probability Measures}

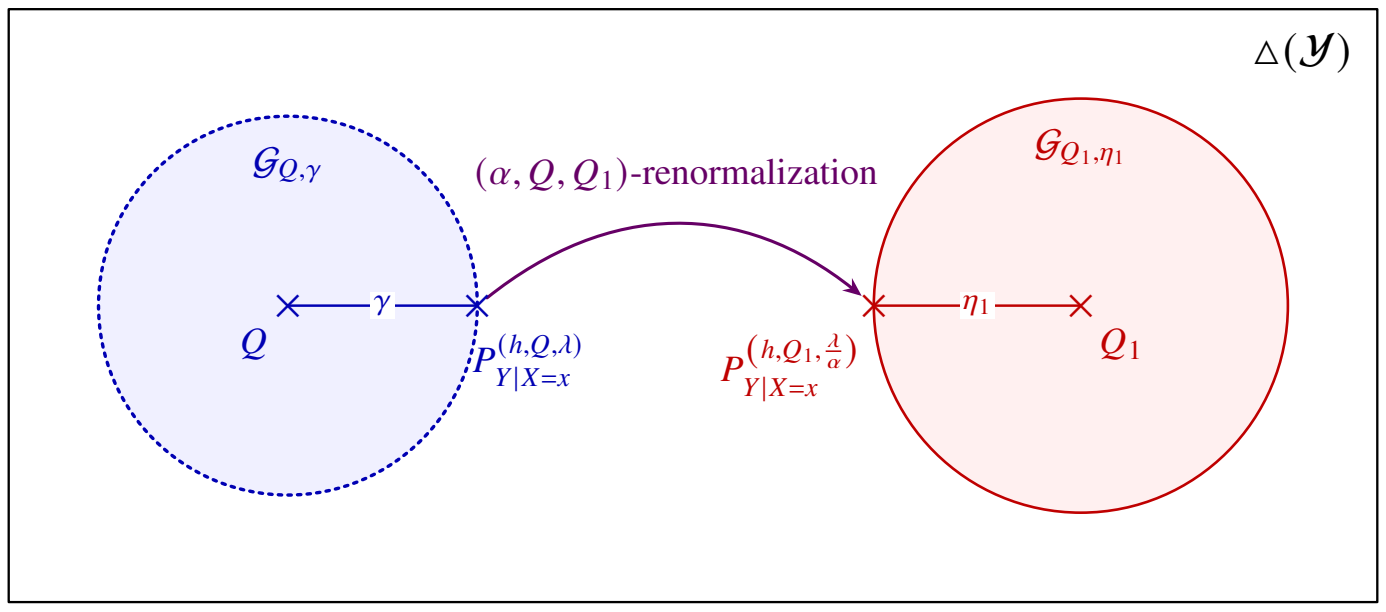
\begin{figure}[h]
\centering
\resizebox{0.5\linewidth}{!}{%
\input{\Figurespath/renormalization_geometry.tikz}}
\caption{Geometric representation of the
$(\alpha,Q,Q_1)$-renormalization of
$P^{\autoparent{h,Q,\lambda}}_{Y\mid X=x}$
in~\eqref{EqGeneralGibbsRN}.
The original Gibbs probability measure
$P^{\autoparent{h,Q,\lambda}}_{Y\mid X=x}$
lies on the boundary of
$\set{G}_{Q,\gamma}$, whereas the renormalized probability measure
$P^{\autoparent{h,Q_1,\frac{\lambda}{\alpha}}}_{Y\mid X=x}$
in~\eqref{EqApril16at13h03in2026Antibes}
lies on the boundary of
$\set{G}_{Q_1,\eta_1}$.
The reals $\gamma$ and $\eta_1$ are defined
in~\eqref{EqRenormalizationRelativeEntropiesA}
and~\eqref{EqRenormalizationRelativeEntropiesB}, respectively.
The purple arrow represents the mapping  from
$P^{\autoparent{h,Q,\lambda}}_{Y\mid X=x}$ to
$P^{\autoparent{h,Q_1,\frac{\lambda}{\alpha}}}_{Y\mid X=x}$, which is induced by the renormalization operation.}
\label{FigRenormalizationGeometry}
\end{figure}

For a fixed $x\in\set{X}$, consider the Gibbs probability measure
$P^{(h,Q,\lambda)}_{Y\mid X=x}$ in~\eqref{EqGeneralGibbsRN} and its
$(\alpha,Q,Q_1)$-renormalization $R_1$
in~\eqref{EqApril15at17h43in2026Sophia}.
Under the assumption that $Q$ and $Q_1$ are probability measures,
consider the reals
\begin{IEEEeqnarray}{rCl}
\label{EqRenormalizationRelativeEntropiesA}
\gamma
&\triangleq&
\KL{P^{(h,Q,\lambda)}_{Y\mid X=x}}{Q}; \mbox{ and }
\\
\label{EqRenormalizationRelativeEntropiesB}
\eta_1
&\triangleq&
\KL{R_1}{Q_1}.
\end{IEEEeqnarray}
The measures $P^{(h,Q,\lambda)}_{Y\mid X=x}$ and $R_1$ are such that  $P^{(h,Q,\lambda)}_{Y\mid X=x} \in \set{G}_{Q,\gamma}$ and $R_1 \in \set{G}_{Q_1,\eta_1}$, with 
$\set{G}_{Q,\gamma}$ and $\set{G}_{Q_1,\eta_1}$ being both sets of the form in~\eqref{EqKLNeighborhood}.
From Theorem~\ref{TheoremApril16at13h13in2026Antibes}, the measure
$R_1$ is identical to the measure
$P^{\autoparent{h,Q_1,\frac{\lambda}{\alpha}}}_{Y\mid X=x}$
in~\eqref{EqApril16at13h03in2026Antibes}.
Hence, Lemma~\ref{LemmaMarch4at9h15in2026inSophia} implies that
$P^{(h,Q,\lambda)}_{Y\mid X=x}$ and $R_1$ optimize the 
expectation
$\int h\autoparent{x,y} \mathrm{d}P\autoparent{y}$,
when $P$ is searched within $\set{G}_{Q,\gamma}$ and
$\set{G}_{Q_1,\eta_1}$, respectively.
The former is a minimizer when $\lambda>0$ and a maximizer
when $\lambda<0$. The latter is a minimizer when
$\frac{\lambda}{\alpha}>0$ and a maximizer when
$\frac{\lambda}{\alpha}<0$.
Thus, renormalization preserves the expectation being optimized, while replacing the reference measure $Q$ by $Q_1$ and the Gibbs parameter $\lambda$ by $\frac{\lambda}{\alpha}$.
Figure~\ref{FigRenormalizationGeometry} provides a geometric
representation of this operation.

A deeper insight is obtained by fixing $h$, $x$, $\lambda$, $Q$, and $Q_1$ and letting $\alpha$ vary.  %and suppose that the mapping $y\mapsto h\autoparent{x,y}$ is finite almost surely with respect to $Q_1$.
From~\eqref{EqApril15at17h43in2026Sophia} and
Theorem~\ref{TheoremApril16at13h13in2026Antibes}, the
Radon--Nikodym derivative of the renormalized probability measure
$R_1$ in~\eqref{EqApril15at17h43in2026Sophia}, or similarly $P^{\autoparent{h,Q_1,\frac{\lambda}{\alpha}}}_{Y\mid X=x}$ in~\eqref{EqApril16at13h03in2026Antibes} ,  with respect to
$Q_1$, satisfies for all $y \in \supp Q_1$,
\begin{IEEEeqnarray}{rCl}
\label{EqSept16at10h03in2026SophiaA}
\log\RND{P^{\autoparent{h,Q_1,\frac{\lambda}{\alpha}}}_{Y\mid X=x}}{Q_1}\autoparent{y}
&=&
\alpha
\log\autoparent{
\RND{
P^{\autoparent{h,Q,\lambda}}_{Y\mid X=x}
}{Q}
\autoparent{y}}
-
\log\autoparent{
\displaystyle
\int
\autoparent{
\RND{
P^{\autoparent{h,Q,\lambda}}_{Y\mid X=x}
}{Q}
\autoparent{\hat{y}}
}^{\alpha}
\mathrm{d}Q_1\autoparent{\hat{y}}}.
%\\
%\label{EqSept16at10h03in2026SophiaB}
%&=&
%-\frac{\alpha}{\lambda}
%h\autoparent{x,y}
%-
%\log\autoparent{
%\displaystyle
%\int
%\exp\autoparent{
%-\frac{\alpha}{\lambda}
%h\autoparent{x,\hat{y}}}
%\mathrm{d}Q_1\autoparent{\hat{y}}}
\end{IEEEeqnarray}
The expression in~\eqref{EqSept16at10h03in2026SophiaA}
implies that, as $\alpha$ varies, the Gibbs probability
measures
$P^{\autoparent{h,Q_1,\frac{\lambda}{\alpha}}}_{Y\mid X=x}$
in~\eqref{EqApril16at13h03in2026Antibes}
describe an exponential geodesic generated by the
Radon--Nikodym derivative
$\RND{P^{\autoparent{h,Q,\lambda}}_{Y\mid X=x}}{Q}$
in~\eqref{EqGeneralGibbsRN}, up to a normalization term
with respect to $Q_1$; see~\cite{nielsen2020elementary}.
In particular, from~\cite[Lemma~5]{perlaza2024empirical},
it follows that this family of probability measures
admits a continuous extension to $\alpha=0$,
which corresponds to the measure $Q_1$.
At $\alpha=1$, the corresponding measure is the
$\autoparent{h,Q_1,\lambda}$-Gibbs probability measure
$P^{\autoparent{h,Q_1,\lambda}}_{Y\mid X=x}$.
For any other value of $\alpha$, the corresponding
measure represents the $\autoparent{\alpha,Q,Q_1}$-renormalization
of the Gibbs probability measure
$P^{\autoparent{h,Q,\lambda}}_{Y\mid X=x}$ i.e., the measure $P^{\autoparent{h,Q_1,\frac{\lambda}{\alpha}}}_{Y\mid X=x}$ in \eqref{EqApril16at13h03in2026Antibes}.
Using these observations, such a  renormalization is shown to exhibit  different interpretations depending on whether  $Q_1=Q$ or $Q_1\neq Q$. Both cases are separately studied. 

First, when $Q_1=Q$, the probability measure $P^{\autoparent{h,Q_1,\frac{\lambda}{\alpha}}}_{Y\mid X=x}$ in~\eqref{EqApril16at13h03in2026Antibes}
at $\alpha=1$ coincides with the Gibbs probability measure
$P^{\autoparent{h,Q,\lambda}}_{Y\mid X=x}$
in~\eqref{EqGeneralGibbsRN}.
Hence, as $\alpha$ varies the renormalized measure $P^{\autoparent{h,Q_1,\frac{\lambda}{\alpha}}}_{Y\mid X=x}$
in~\eqref{EqApril16at13h03in2026Antibes} traces out the exponential geodesic described above passing through $Q_1$ at $\alpha = 0$ and $P^{\autoparent{h,Q_1,\lambda}}_{Y\mid X=x}$ at $\alpha = 1$.
Note also that the inclusion between 
$\set{G}_{Q,\gamma}$ and $\set{G}_{Q,\eta_1}$ is determined by
the ordering of $\gamma$ and $\eta_1$.
For $\alpha>0$, the parameters $\lambda$ and
$\frac{\lambda}{\alpha}$ have the same sign.
From Lemma~\ref{LemmaMarch10at17h10in2026Sophia}, the relative
entropy is nonincreasing with respect to the Gibbs parameter
on positive intervals and nondecreasing on negative intervals.
Consequently, for either sign of $\lambda$,
\begin{IEEEeqnarray}{rCl}
\set{G}_{Q,\gamma}
&\subseteq&
\set{G}_{Q,\eta_1},
\quad \text{if } \alpha>1;
\\
\set{G}_{Q,\eta_1}
&\subseteq&
\set{G}_{Q,\gamma},
\quad \text{if } 0<\alpha<1.
\end{IEEEeqnarray}
At $\alpha=1$, the two measures coincide, and hence
$\set{G}_{Q,\gamma}=\set{G}_{Q,\eta_1}$.
For $\alpha<0$, the parameters $\lambda$ and
$\frac{\lambda}{\alpha}$ have opposite signs.
The monotonicity properties in
Lemma~\ref{LemmaMarch10at17h10in2026Sophia} apply separately
on positive and negative intervals and therefore do not
provide an ordering of $\gamma$ and $\eta_1$ in this case.
The neighborhoods remain ordered by inclusion, but the
direction of inclusion depends on which of these relative
entropies, $\gamma$ and $\eta_1$, is larger.

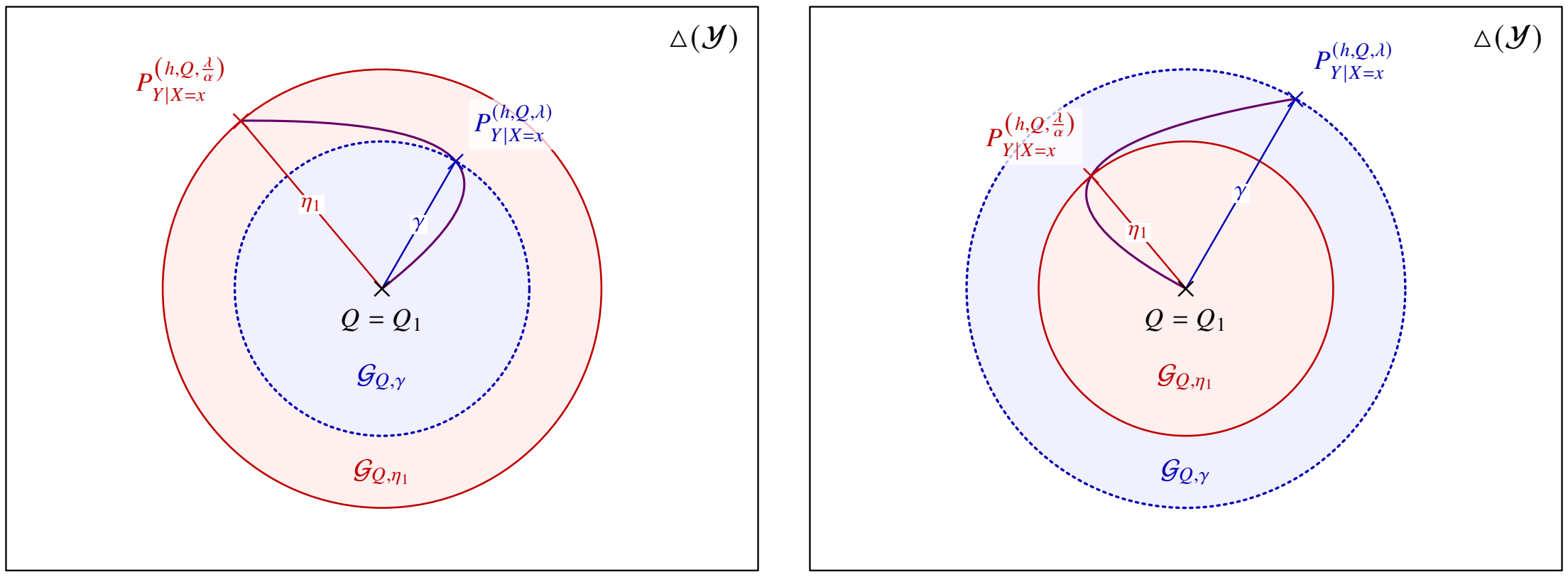
\begin{figure}[h]
\centering
\input{\Figurespath/renormalization_geodesic_commonQ_two_cases.tikz}
\resizebox{0.8\linewidth}{!}{%
  \RenormCommonPanel{3.50}{2.35}{(a) $\gamma < \eta_1$}%
  \hspace{8mm}%
  \RenormCommonPanel{2.35}{3.50}{(b) $\eta_1 < \gamma$}%
}
\caption{Geometric representation of the
$(\alpha,Q,Q)$-renormalization of
$P^{\autoparent{h,Q,\lambda}}_{Y\mid X=x}$
in~\eqref{EqGeneralGibbsRN}.
The solid purple curve represents the exponential geodesic generated by the measure $P^{\autoparent{h,Q,\frac{\lambda}{\alpha}}}_{Y\mid X=x}$ in~\eqref{EqApril16at13h03in2026Antibes} as  $\alpha$ varies. 
The panels show
(a) $\gamma<\eta_1$, with
$\set{G}_{Q,\gamma}\subseteq\set{G}_{Q,\eta_1}$, and
(b) $\eta_1<\gamma$, with
$\set{G}_{Q,\eta_1}\subseteq\set{G}_{Q,\gamma}$.
The sets $\set{G}_{Q,\gamma}$ and
$\set{G}_{Q,\eta_1}$ are of the form
in~\eqref{EqKLNeighborhood}, with $\gamma$ and $\eta_1$
in~\eqref{EqRenormalizationRelativeEntropiesA}
and~\eqref{EqRenormalizationRelativeEntropiesB}, respectively.}
\label{FigRenormalizationGeometryCommonQ}
\end{figure}

Figure~\ref{FigRenormalizationGeometryCommonQ} provides a
geometric representation of the two possible orderings of
$\gamma$ and $\eta_1$.

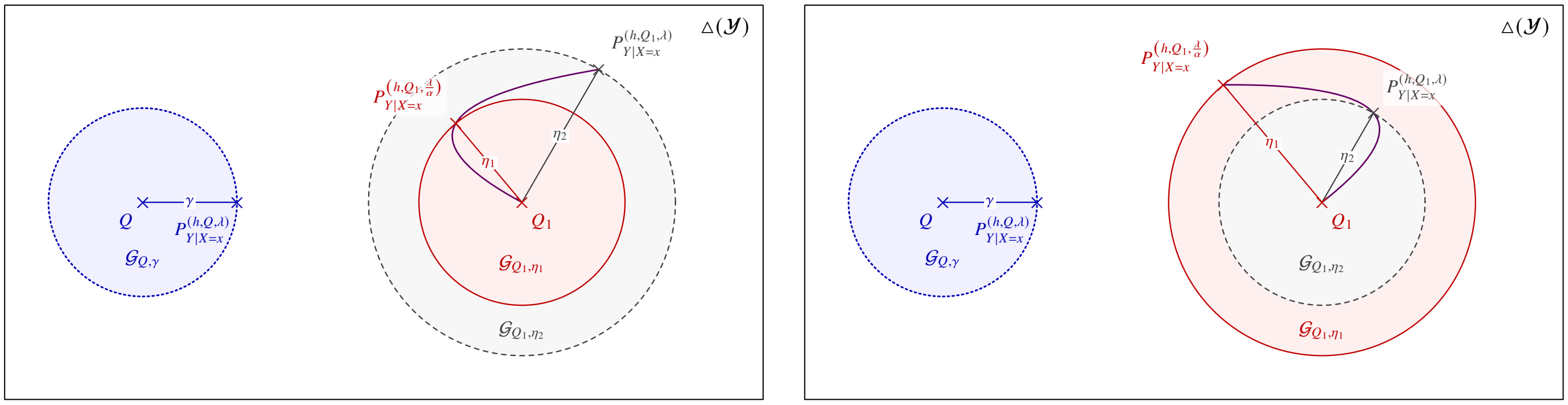
\begin{figure}[h]
\centering
\input{\Figurespath/renormalization_geodesic_two_cases.tikz}
\resizebox{\linewidth}{!}{%
  \RenormPanel{2.35}{3.50}{(a) $\eta_1<\eta_2$}%
  \hspace{8mm}%
  \RenormPanel{3.50}{2.35}{(b) $\eta_1>\eta_2$}%
}
\caption{Geometric representation of the
$(\alpha,Q,Q_1)$-renormalization of
$P^{\autoparent{h,Q,\lambda}}_{Y\mid X=x}$
in~\eqref{EqGeneralGibbsRN}, with $Q_1\neq Q$.
The solid purple curve represents the exponential geodesic generated by the measure $P^{\autoparent{h,Q_1,\frac{\lambda}{\alpha}}}_{Y\mid X=x}$ in~\eqref{EqApril16at13h03in2026Antibes}  when  $\alpha$ varies.  
The panels show
(a) $\eta_1<\eta_2$, with
$\set{G}_{Q_1,\eta_1}\subseteq\set{G}_{Q_1,\eta_2}$, and
(b) $\eta_1>\eta_2$, with
$\set{G}_{Q_1,\eta_2}\subseteq\set{G}_{Q_1,\eta_1}$.
The sets $\set{G}_{Q,\gamma}$,
$\set{G}_{Q_1,\eta_1}$, and
$\set{G}_{Q_1,\eta_2}$ are of the form
in~\eqref{EqKLNeighborhood}, with $\gamma$, $\eta_1$, and $\eta_2$ defined 
in~\eqref{EqRenormalizationRelativeEntropiesA},
\eqref{EqRenormalizationRelativeEntropiesB}, and
\eqref{EqRenormalizationRelativeEntropiesC}, respectively.}

\label{FigRenormalizationGeometryDifferentQ}
\end{figure}

On the other hand, when $Q_1\neq Q$, the intersection
between the sets $\set{G}_{Q,\gamma}$ and
$\set{G}_{Q_1,\eta_1}$ might be empty, as shown in
Figure~\ref{FigRenormalizationGeometry}.
Nonetheless, the same interpretation of the exponential
geodesic described above holds.
As $\alpha$ varies, the renormalized probability measure
$P^{\autoparent{h,Q_1,\frac{\lambda}{\alpha}}}_{Y\mid X=x}$
in~\eqref{EqApril16at13h03in2026Antibes}
traces out this geodesic, passing through $Q_1$
at $\alpha=0$, by continuous extension, and
$P^{\autoparent{h,Q_1,\lambda}}_{Y\mid X=x}$
at $\alpha=1$, as shown in
Figure~\ref{FigRenormalizationGeometryDifferentQ},
where $\eta_2$ is defined as follows:
\begin{IEEEeqnarray}{rCl}
\label{EqRenormalizationRelativeEntropiesC}
\eta_2
&\triangleq&
\KL{
P^{\autoparent{h,Q_1,\lambda}}_{Y\mid X=x}
}{
Q_1
}.
\end{IEEEeqnarray}
Interestingly, the Gibbs probability measure
$P^{\autoparent{h,Q,\lambda}}_{Y\mid X=x}$ in~\eqref{EqGeneralGibbsRN},
as well as the reference measure $Q$, do not necessarily
belong to such an exponential geodesic.
Finally, the inclusion relations between
$\set{G}_{Q_1,\eta_1}$ and $\set{G}_{Q_1,\eta_2}$
in~\eqref{EqKLNeighborhood} follow from the same argument
as in the case $Q_1=Q$, with $\gamma$ replaced by
$\eta_2$ in~\eqref{EqRenormalizationRelativeEntropiesC}.
Figure~\ref{FigRenormalizationGeometryDifferentQ}
provides a geometric representation of the two possible
orderings of $\eta_1$ and $\eta_2$.

\subsection{Normalized Log-Linear Combinations of Gibbs Probability Measures}

For $K=2$ and a fixed
$\vect{x}=\autoparent{x_1,x_2}\in\set{X}^2$, consider the Gibbs
probability measures
$P^{\autoparent{h_1,Q_1,\lambda_1}}_{Y\mid X=x_1}$ and
$P^{\autoparent{h_2,Q_2,\lambda_2}}_{Y\mid X=x_2}$
in~\eqref{EqGeneralGibbsRN}.
These measures satisfy, with equality, the relative entropy
constraints defining the sets  $\set{G}_{Q_1,\gamma_1}$ and
$\set{G}_{Q_2,\gamma_2}$ in~\eqref{EqKLNeighborhood},
respectively, where $\gamma_1$ and $\gamma_2$ are defined
in~\eqref{EqSeptember11at11h28in2026atSophia}.
The normalized log-linear combination of the densities of these measures with
respect to $Q$ and coefficients $(\alpha_1 , \alpha_2) \in \reals^2$, coincides, from
Theorem~\ref{TheoremMay20at13h57in2026Antibes}, with the Gibbs
probability measure
$P_{Y\mid\vect{X}=\vect{x}}^{
\autoparent{\hat{h}_{\vect{\beta}},Q,1}}$
in~\eqref{EqApril15at22h01in2026AntibesB}, where $\vect{\beta}
=
\autoparent{
\frac{\alpha_1}{\lambda_1},
\frac{\alpha_2}{\lambda_2}}$
and $\hat{h}_{\vect{\beta}}$ is the linear combination
in~\eqref{EqApril29at13h53in2026Sophia}.

For a fixed pair $\autoparent{\alpha_1,\alpha_2} \in\autoparent{\reals\setminus\lbrace0\rbrace}^2$, and under the assumption that $Q$ is a probability measure,
consider the real  
\begin{IEEEeqnarray}{rCl}
\label{EqLogLinearRelativeEntropy}
\gamma_0
&\triangleq&
\KL{
P_{Y\mid\vect{X}=\vect{x}}^{
\autoparent{\hat{h}_{\vect{\beta}},Q,1}}
}{
Q
}.
\end{IEEEeqnarray}
From Theorem~\ref{TheoremApril16at15h23in2026Antibes}, it follows that 
the probability measure
$P_{Y\mid\vect{X}=\vect{x}}^{
\autoparent{\hat{h}_{\vect{\beta}},Q,1}}$
in~\eqref{EqApril15at22h01in2026AntibesB} is the unique
minimizer of the expectation
$\int
\hat{h}_{\vect{\beta}}\autoparent{\vect{x},y}
 \mathrm{d}P\autoparent{y}$
when $P$ is searched within
$\set{G}_{Q,\gamma_0}$ in~\eqref{EqKLNeighborhood}.
Beyond this observation, the normalized log-linear combination $P_{Y\mid\vect{X}=\vect{x}}^{
\autoparent{\hat{h}_{\vect{\beta}},Q,1}}$
also admits a geometric interpretation obtained by fixing
$h_1$, $h_2$, $x_1$, $x_2$, $\lambda_1$, $\lambda_2$,
$Q_1$, $Q_2$, and $Q$, and varying the coefficients
$\alpha_1$ and $\alpha_2$. 
From~\eqref{EqApril15at22h01in2026AntibesB}, the
Radon--Nikodym derivative of
$P^{\autoparent{\hat{h}_{\vect{\beta}},Q,1}}_
{Y\mid\vect{X}=\vect{x}}$
with respect to $Q$ satisfies for all $y \in \supp Q$,
\begin{IEEEeqnarray}{rCl}
\label{EqSept16at11h43in2026SophiaA}
\log
\RND{P_{Y\mid\vect{X}=\vect{x}}^{\autoparent{\hat{h}_{\vect{\beta}},Q,1}}}{Q}
\autoparent{y}
&=& \alpha_1 \log \autoparent{\RND{P_{Y\mid X=x_1}^{(h_1,Q_1,\lambda_1)}}{Q_1}
\autoparent{y} } + \alpha_2 \log \autoparent{\RND{P_{Y\mid X=x_2}^{(h_2,Q_2,\lambda_2)}}{Q_2}
\autoparent{y} } - \log \autoparent{\displaystyle \int \autoparent{\RND{P_{Y\mid X=x_1}^{(h_1,Q_1,\lambda_1)}}{Q_1}
\autoparent{\tilde{y}} } ^{\alpha_1} \autoparent{\RND{P_{Y\mid X=x_2}^{(h_2,Q_2,\lambda_2)}}{Q_2}
\autoparent{\tilde{y}} }^{\alpha_2} \mathrm{d}Q\autoparent{\tilde{y}}}\middlesqueezeequ\spnum\\
\label{EqSept16at11h43in2026SophiaB}
&=&
-\frac{\alpha_1}{\lambda_1}h_1\autoparent{x_1,y}
-\frac{\alpha_2}{\lambda_2}h_2\autoparent{x_2,y}
-
\log\int
\exp\autoparent{
-\frac{\alpha_1}{\lambda_1}h_1\autoparent{x_1,\hat{y}}
-\frac{\alpha_2}{\lambda_2}h_2\autoparent{x_2,\hat{y}}
}
 \mathrm{d}Q\autoparent{\hat{y}}.
\end{IEEEeqnarray}
Hence, as $\alpha_1$ and $\alpha_2$ vary over values for which
the normalization term 
in~\eqref{EqSept16at11h43in2026SophiaA} is finite, the Gibbs
probability measure
$P^{\autoparent{\hat{h}_{\vect{\beta}},Q,1}}_
{Y\mid\vect{X}=\vect{x}}$
in~\eqref{EqApril15at22h01in2026AntibesB}
forms an exponential family with natural parameter
$\autoparent{-\frac{\alpha_1}{\lambda_1}, -\frac{\alpha_2}{\lambda_2}} \in \reals^2$.
For the ease of presentation, consider $\alpha_1=\alpha$ and
$\alpha_2=1-\alpha$, for some $\alpha\in\autoparent{0,1}$.
The expression in~\eqref{EqSept16at11h43in2026SophiaA}
implies that, as $\alpha$ varies, the Gibbs probability
measures
$P_{Y\mid\vect{X}=\vect{x}}^{
\autoparent{\hat{h}_{\vect{\beta}},Q,1}}$
in~\eqref{EqApril15at22h01in2026AntibesB}
describe a segment of an exponential geodesic, up to a normalization with respect to $Q$.
Note that at $\alpha=0$, the probability measure $P_{Y\mid\vect{X}=\vect{x}}^{
\autoparent{\hat{h}_{\vect{\beta}},Q,1}}$
in~\eqref{EqApril15at22h01in2026AntibesB} is identical to  the
$\autoparent{h_2,Q,\lambda_2}$-Gibbs probability measure
$P^{\autoparent{h_2,Q,\lambda_2}}_{Y\mid X=x_2}$. Alternatively,  at $\alpha=1$, such a measure is identical to  the
$\autoparent{h_1,Q,\lambda_1}$-Gibbs probability measure
$P^{\autoparent{h_1,Q,\lambda_1}}_{Y\mid X=x_1}$.
These endpoints are understood as continuous extensions of
\eqref{EqSept16at11h43in2026SophiaB}, taking into account that 
Definition~\ref{DefMay19at15h20in2026Sophia} assumes that
the coefficients, $\alpha_1$ and $\alpha_2$, are nonzero.

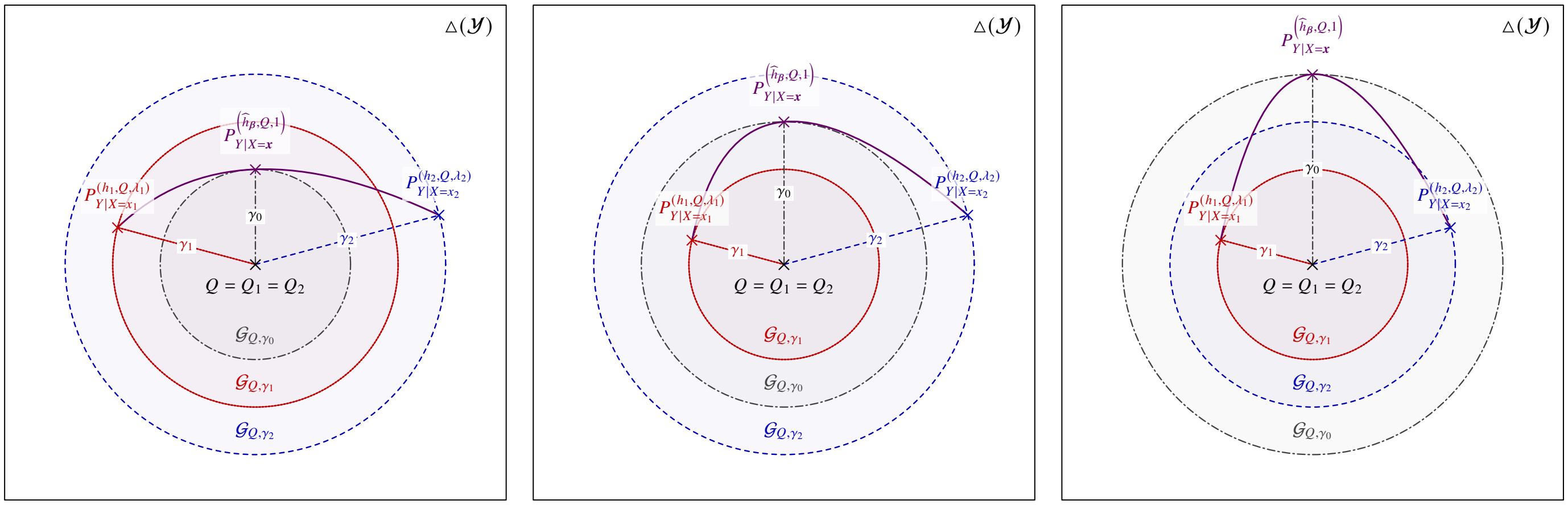
\begin{figure*}[t]
\centering
\input{\Figurespath/log-linear-geodesic_commonQ.tikz}
\resizebox{\linewidth}{!}{%
  \LogLinearCommonPanel{3.30}{2.20}{4.40}
    {(a) $\gamma_0<\gamma_1<\gamma_2$}%
  \hspace{6mm}%
  \LogLinearCommonPanel{2.20}{3.30}{4.40}
    {(b) $\gamma_1<\gamma_0<\gamma_2$}%
  \hspace{6mm}%
  \LogLinearCommonPanel{2.20}{4.40}{3.30}
    {(c) $\gamma_1<\gamma_2<\gamma_0$}%
}
\caption{Geometric representation of the normalized log-linear
combination
$P_{Y\mid\vect{X}=\vect{x}}^{
\autoparent{\hat{h}_{\vect{\beta}},Q,1}}$
in~\eqref{EqApril15at22h01in2026AntibesB}, with
$Q_1=Q_2=Q$ and
$\vect{\beta} = \autoparent{\frac{\alpha}{\lambda_1},\frac{1-\alpha}{\lambda_2}}$,
where $0<\alpha<1$.
The solid purple curve represents the exponential segment obtained
by varying $\alpha$ over $[0,1]$.
Its endpoints are
$P^{\autoparent{h_2,Q,\lambda_2}}_{Y\mid X=x_2}$
at $\alpha=0$ and
$P^{\autoparent{h_1,Q,\lambda_1}}_{Y\mid X=x_1}$
at $\alpha=1$, and it passes through the normalized log-linear
combination at the chosen value of $\alpha$.
The neighborhoods
$\set{G}_{Q,\gamma_1}$,
$\set{G}_{Q,\gamma_2}$, and
$\set{G}_{Q,\gamma_0}$
are defined by~\eqref{EqKLNeighborhood}, with
$\gamma_1$ and $\gamma_2$ given
in~\eqref{EqSeptember11at11h28in2026atSophia}
and $\gamma_0$ given in~\eqref{EqLogLinearRelativeEntropy}.
Under the illustrated ordering $\gamma_1<\gamma_2$, the panels show
(a) $\gamma_0<\gamma_1<\gamma_2$,
(b) $\gamma_1<\gamma_0<\gamma_2$, and
(c) $\gamma_1<\gamma_2<\gamma_0$.}
\label{FigGeodesicLogLinearSameQ}
\end{figure*}

When $Q_1=Q_2=Q$, these endpoints   coincide with
the original Gibbs probability measures
$P^{\autoparent{h_2,Q_2,\lambda_2}}_{Y\mid X=x_2}$ and
$P^{\autoparent{h_1,Q_1,\lambda_1}}_{Y\mid X=x_1}$,
respectively.
Hence, the exponential geodesic obtained by varying
$\alpha$ over $[0,1]$ connects the two original Gibbs
probability measures.
%
%The normalized log-linear combination $P^{\autoparent{\hat{h}_{\vect{\beta}},Q,1}}_{Y\mid\vect{X}=\vect{x}}$ in~\eqref{EqApril15at22h01in2026AntibesB} belongs to this exponential segment at the chosen value of $\alpha$.
%
%If $y\mapsto \frac{1}{\lambda_1}h_1\autoparent{x_1,y}- \frac{1}{\lambda_2}h_2\autoparent{x_2,y}$ is constant almost surely with respect to $Q$, the two endpoint measures coincide and the normalized log-linear combination is independent of $\alpha$.
%
Figure~\ref{FigGeodesicLogLinearSameQ} provides a geometric representation
of this case.

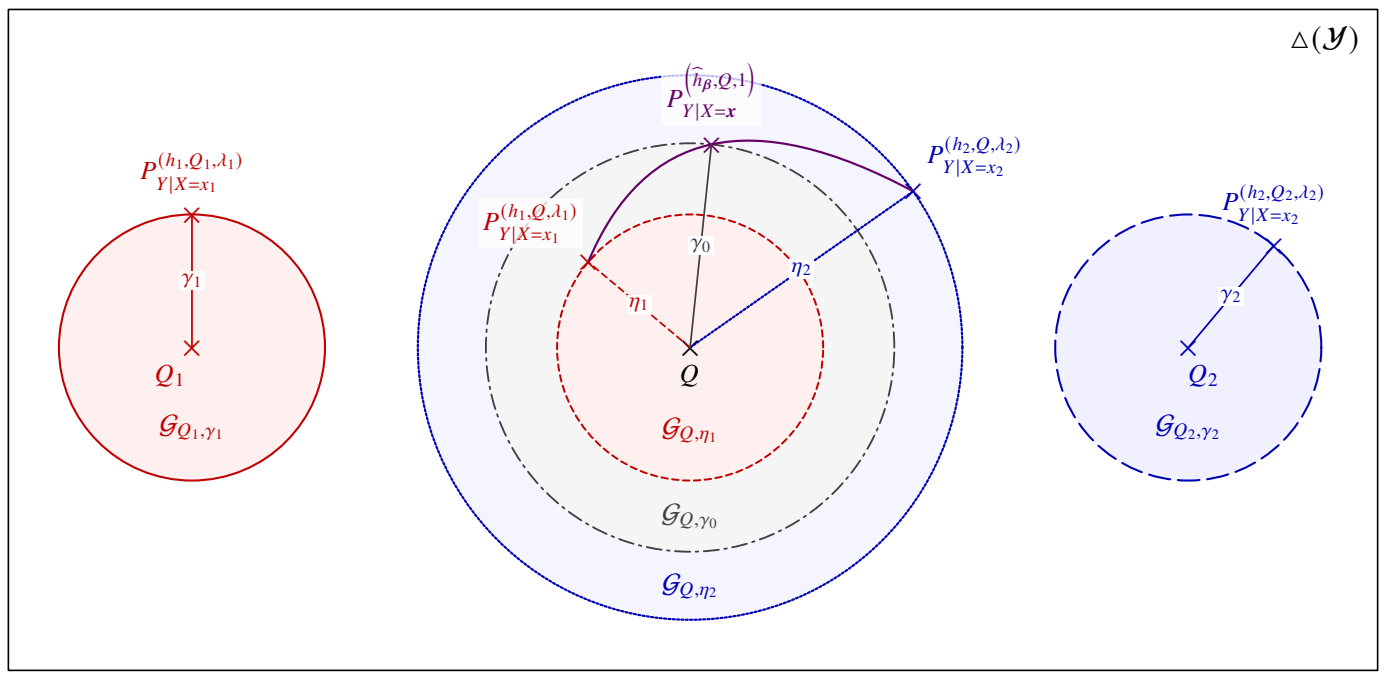
\begin{figure}[h]
\centering
\resizebox{0.5\linewidth}{!}{%
\input{\Figurespath/log-linear-geodesic_diffQ.tikz}}
\caption{Geometric representation of the normalized log-linear
combination
$P_{Y\mid\vect{X}=\vect{x}}^{
\autoparent{\hat{h}_{\vect{\beta}},Q,1}}$
in~\eqref{EqApril15at22h01in2026AntibesB}, with
$\vect{\beta} =
\autoparent{\alpha/\lambda_1,(1-\alpha)/\lambda_2}$
and $0<\alpha<1$.
The solid purple curve represents the exponential geodesic obtained
by varying $\alpha$ over $[0,1]$.
Its endpoints are
$P^{\autoparent{h_2,Q,\lambda_2}}_{Y\mid X=x_2}$
at $\alpha=0$ and
$P^{\autoparent{h_1,Q,\lambda_1}}_{Y\mid X=x_1}$
at $\alpha=1$.
The original Gibbs probability measures
$P^{\autoparent{h_1,Q_1,\lambda_1}}_{Y\mid X=x_1}$
and
$P^{\autoparent{h_2,Q_2,\lambda_2}}_{Y\mid X=x_2}$
are represented separately on the boundaries of
$\set{G}_{Q_1,\gamma_1}$ and
$\set{G}_{Q_2,\gamma_2}$, respectively, and do not, in general,
belong to the exponential geodesic.
The sets
$\set{G}_{Q,\eta_1}$,
$\set{G}_{Q,\eta_2}$, and
$\set{G}_{Q,\gamma_0}$
contain the two endpoints and the normalized log-linear combination,
respectively, with $\eta_1$, $\eta_2$, and $\gamma_0$ defined
in~\eqref{EqLogLinearRelativeEntropiesA},
\eqref{EqLogLinearRelativeEntropiesB}, and
\eqref{EqLogLinearRelativeEntropy}.}

\label{FigGeodesicLogLinearDiffQ}
\end{figure}

When $Q_1$, $Q_2$, and $Q$ are different, the Gibbs
probability measures
$P^{\autoparent{h_1,Q,\lambda_1}}_{Y\mid X=x_1}$ and
$P^{\autoparent{h_2,Q,\lambda_2}}_{Y\mid X=x_2}$
at the endpoints of the exponential geodesic do not, in general,
coincide with the original Gibbs probability measures
$P^{\autoparent{h_1,Q_1,\lambda_1}}_{Y\mid X=x_1}$ and
$P^{\autoparent{h_2,Q_2,\lambda_2}}_{Y\mid X=x_2}$,
respectively.
Consider the values 
\begin{IEEEeqnarray}{rCl}
\label{EqLogLinearRelativeEntropiesA}
\eta_1
&\triangleq&
\KL{
P^{\autoparent{h_1,Q,\lambda_1}}_{Y\mid X=x_1}
}{
Q
}; \mbox{ and }
\\
\label{EqLogLinearRelativeEntropiesB}
\eta_2
&\triangleq&
\KL{
P^{\autoparent{h_2,Q,\lambda_2}}_{Y\mid X=x_2}
}{
Q
}.
\end{IEEEeqnarray}
The endpoint measures therefore satisfy, with equality, the
relative entropy constraints defining
$\set{G}_{Q,\eta_1}$ and $\set{G}_{Q,\eta_2}$
in~\eqref{EqKLNeighborhood}, respectively.
Hence, varying $\alpha$ over $[0,1]$ describes the
exponential geodesic connecting
$P^{\autoparent{h_2,Q,\lambda_2}}_{Y\mid X=x_2}$ ($\alpha=0$) and
$P^{\autoparent{h_1,Q,\lambda_1}}_{Y\mid X=x_1}$ ($\alpha=1$).
Figure~\ref{FigGeodesicLogLinearDiffQ} provides a geometric
representation of this case.

\subsection{Nested Gibbs Probability Measures}

Consider the recursively nested Gibbs probability measures $P^{\autoparent{h_k,Q_k,\lambda_k}}_{Y \mid X = x_k}$
in~\eqref{EqAugust28at11h54in2026Sophia} with $k\inCountK{K}$ and the recursive choice of the reference measures
in~\eqref{EqNestedReferenceMeasures}, namely,
$Q_{k+1}=P^{\autoparent{h_k,Q_k,\lambda_k}}_{Y\mid X=x_k}$
for $k\in\{1,2,\ldots,K-1\}$.
In this case, it suffices to assume that $Q_1$ is a probability measure.
In addition to the values $\gamma_k$
in~\eqref{EqSeptember11at11h28in2026atSophia}, consider the real  
\begin{IEEEeqnarray}{rCl}
\label{EqSept11at15h42in2026Sophia}
\eta
&\triangleq&
\KL{P^{\autoparent{h_K,Q_K,\lambda_K}}_{Y\mid X=x_K}}{Q_1},
\end{IEEEeqnarray}
and, for all $i\in\{1,2,\ldots,K-1\}$, consider also the reals
\begin{IEEEeqnarray}{rCl}
\label{EqSept11at15h41in2026Sophia}
\delta_i
&\triangleq&
\frac{1}{\lambda_i}
\autoparent{
\int h_i\autoparent{x_i,y}
 \mathrm{d}
P^{\autoparent{h_K,Q_K,\lambda_K}}_{Y\mid X=x_K}
\autoparent{y}
-
\int h_i\autoparent{x_i,y}
 \mathrm{d}Q_{i+1}\autoparent{y}
},
\end{IEEEeqnarray}
where both expectations are assumed to be finite.

\begin{figure*}[t]
\centering
\input{\Figurespath/nesting_geometric_inclusions.tikz}
\resizebox{0.8\linewidth}{!}{%
  \NestingPanel{2.15}{3.05}{(a) $\gamma_1<\eta$}%
  \hspace{8mm}%
  \NestingPanel{3.05}{2.15}{(b) $\eta<\gamma_1$}%
}
\caption{Geometric representation of nested Gibbs probability
measures illustrating
(a) $\mathcal{G}_{Q_1,\gamma_1}\subseteq\mathcal{G}_{Q_1,\eta}$
and
(b) $\mathcal{G}_{Q_1,\eta}\subseteq\mathcal{G}_{Q_1,\gamma_1}$.
The dotted neighborhood $\mathcal{G}_{Q_2,\gamma_2}$
has reference measure $Q_2$ in both cases.}
\label{FigNestingGeometricInclusions}
\end{figure*}
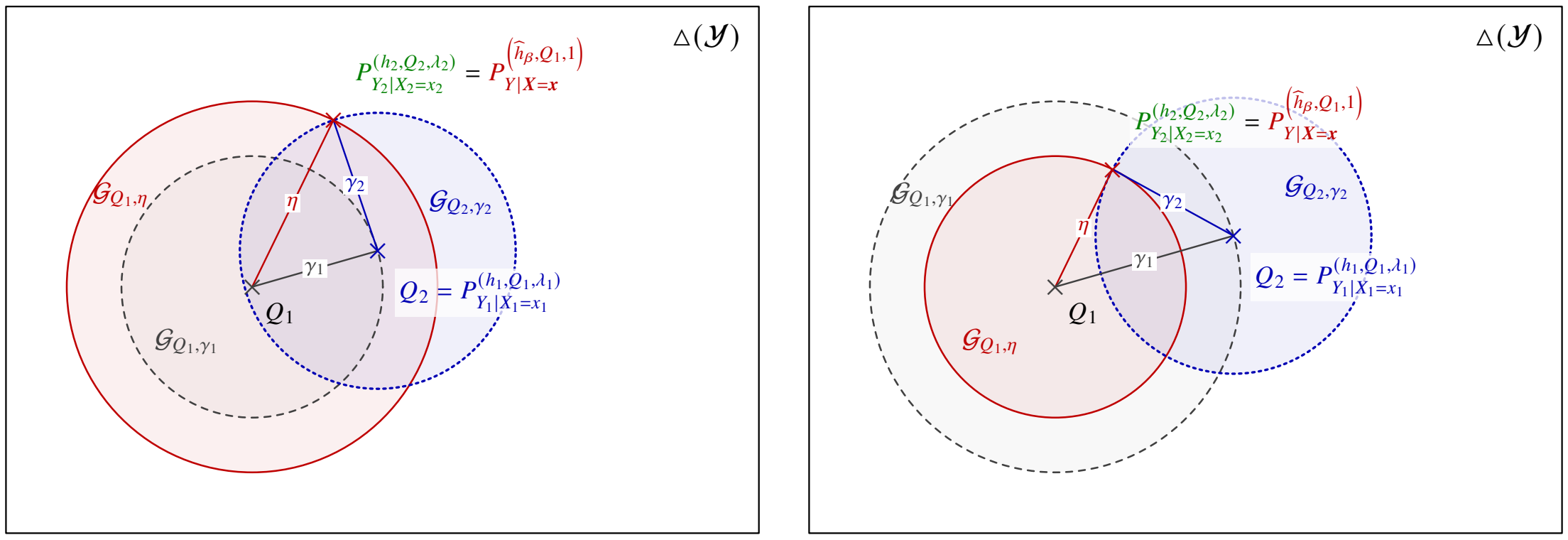

For $K=2$, from Lemma~\ref{LemmaMarch4at9h15in2026inSophia}, it follows
that the Gibbs probability measure
$P^{\autoparent{h_1,Q_1,\lambda_1}}_{Y\mid X=x_1}$ and the nested
Gibbs probability measure
$P^{\autoparent{h_2,Q_2,\lambda_2}}_{Y\mid X=x_2}$ are optimizers
(minimizer or maximizer depending on the sign of $\lambda_1$ and
$\lambda_2$, respectively) of the expectations
$\int h_1\autoparent{x_1,y} \mathrm{d}P_1\autoparent{y}$ and
$\int h_2\autoparent{x_2,y} \mathrm{d}P_2\autoparent{y}$,
when $P_1$ and $P_2$ are respectively searched within the
neighborhoods $\set{G}_{Q_1, \gamma_1}$ and $\set{G}_{Q_2, \gamma_2}$ both of the form in \eqref{EqKLNeighborhood}.
Similarly, using Lemma~\ref{LemmaMarch4at9h15in2026inSophia}, the
measure
$P_{Y\mid\vect{X}=\vect{x}}^{\autoparent{\hat{h}_{\vect{\beta}},Q_1,1}}$
in~\eqref{EqNestedExplicitB}, which coincides with the nested Gibbs
probability measure
$P^{\autoparent{h_2,Q_2,\lambda_2}}_{Y\mid X=x_2}$, can be identified
as a minimizer of the expectation
$\int \hat{h}_{\vect{\beta}}\autoparent{(x_1,x_2),y}
 \mathrm{d}P_3\autoparent{y}$,
when $P_3$ is searched within the neighborhood
$\set{G}_{Q_1,\eta}$.
Consequently,
\begin{IEEEeqnarray}{rCl}
\label{EqApril1at10h48in2026Sophia}
\left\lbrace
P_{Y\mid\vect{X}=\vect{x}}^{\autoparent{\hat{h}_{\vect{\beta}},Q_1,1}}
\right\rbrace
&=&
\displaystyle\arg\min_{P\in\set{G}_{Q_1,\eta}}
\int\hat{h}_{\vect{\beta}}\autoparent{(x_1,x_2),y}
 \mathrm{d}P\autoparent{y}
\\
&=&
\left\lbrace
\begin{array}{ll}
\displaystyle\arg\min_{P\in\set{G}_{Q_2,\gamma_2}}
\int h_2\autoparent{x_2,y} \mathrm{d}P\autoparent{y},
&\lambda_2>0;\\
\displaystyle\arg\max_{P\in\set{G}_{Q_2,\gamma_2}}
\int h_2\autoparent{x_2,y} \mathrm{d}P\autoparent{y},
&\lambda_2<0,
\end{array}
\right.
\\
&=&
\left\lbrace
P^{\autoparent{h_2,Q_2,\lambda_2}}_{Y\mid X=x_2}
\right\rbrace.
\end{IEEEeqnarray}
The measures
$P^{\autoparent{h_1,Q_1,\lambda_1}}_{Y\mid X=x_1}$,
$P^{\autoparent{h_2,Q_2,\lambda_2}}_{Y\mid X=x_2}$, and
$P_{Y\mid\vect{X}=\vect{x}}^{\autoparent{\hat{h}_{\vect{\beta}},Q_1,1}}$
can be placed within the sets
$\set{G}_{Q_1,\gamma_1}$, $\set{G}_{Q_2,\gamma_2}$, and
$\set{G}_{Q_1,\eta}$, respectively, as in
Figure~\ref{FigNestingGeometricInclusions}.
%
%Note that in such a representation, it has been arbitrarily assumed
%that $\gamma_1<\gamma_2<\eta$, which justifies the inclusion
%$\set{G}_{Q_1,\gamma_1}\subseteq\set{G}_{Q_1,\eta}$.

%\begin{figure}[t]
%\centering
%\resizebox{0.6\textwidth}{!}{%
%\input{\Figurespath/general_geometric_interpretation.tikz}}
%\caption{Representation of $\set{G}_{Q_1,\gamma_1}$ as a solid black
%circle, $\set{G}_{Q_2,\gamma_2}$ as a dotted blue circle, and
%$\set{G}_{Q_1,\eta}$ as a dashed red circle, under the assumption that
%$\gamma_1<\gamma_2<\eta$.}
%\label{FigGeomNestedGibbsGeneral}
%\end{figure}

Beyond this geometrical representation, it is important to highlight
that $\gamma_1$ and $\gamma_2$
in~\eqref{EqSeptember11at11h28in2026atSophia}, and $\eta$
in~\eqref{EqSept11at15h42in2026Sophia} exhibit an important relation
as shown by the following lemma.
\begin{lemma}
\label{LemmaSeptember9at9h48in2026inSophiaAntipolis}
The measures
$P^{(h_1,Q_1,\lambda_1)}_{Y\mid X=x_1}$ in \eqref{EqMay18at19h36in2026AntibesA},
$P^{\autoparent{h_2,Q_2,\lambda_2}}_{Y\mid X=x_2}$
in~\eqref{EqApril15at18h06in2026Sophia}, and $P_{Y\mid\vect{X}=\vect{x}}^{\autoparent{\hat{h}_{\vect{\beta}},Q_1,1}}$
in~\eqref{EqNestedExplicitB} satisfy
\begin{IEEEeqnarray}{rCl}
\label{EqMethodOfGapsGeometric}
\eta
+
\frac{1}{\lambda_1}
\left(
\int h_1\autoparent{x_1,y}
 \mathrm{d}
P^{\autoparent{h_2,Q_2,\lambda_2}}_{Y\mid X=x_2}
\autoparent{y}
-
\int h_1\autoparent{x_1,y}
 \mathrm{d}
P^{\autoparent{h_1,Q_1,\lambda_1}}_{Y\mid X=x_1}
\autoparent{y}
\right)
&=&
\gamma_1+\gamma_2,
\end{IEEEeqnarray}
where $\gamma_1$ and $\gamma_2$ are defined
in~\eqref{EqSeptember11at11h28in2026atSophia}, and $\eta$
in~\eqref{EqSept11at15h42in2026Sophia}.
\end{lemma}
\begin{IEEEproof}
The proof follows from \cite[Lemma~3]{PerlazaEntropy2025}.
\end{IEEEproof}

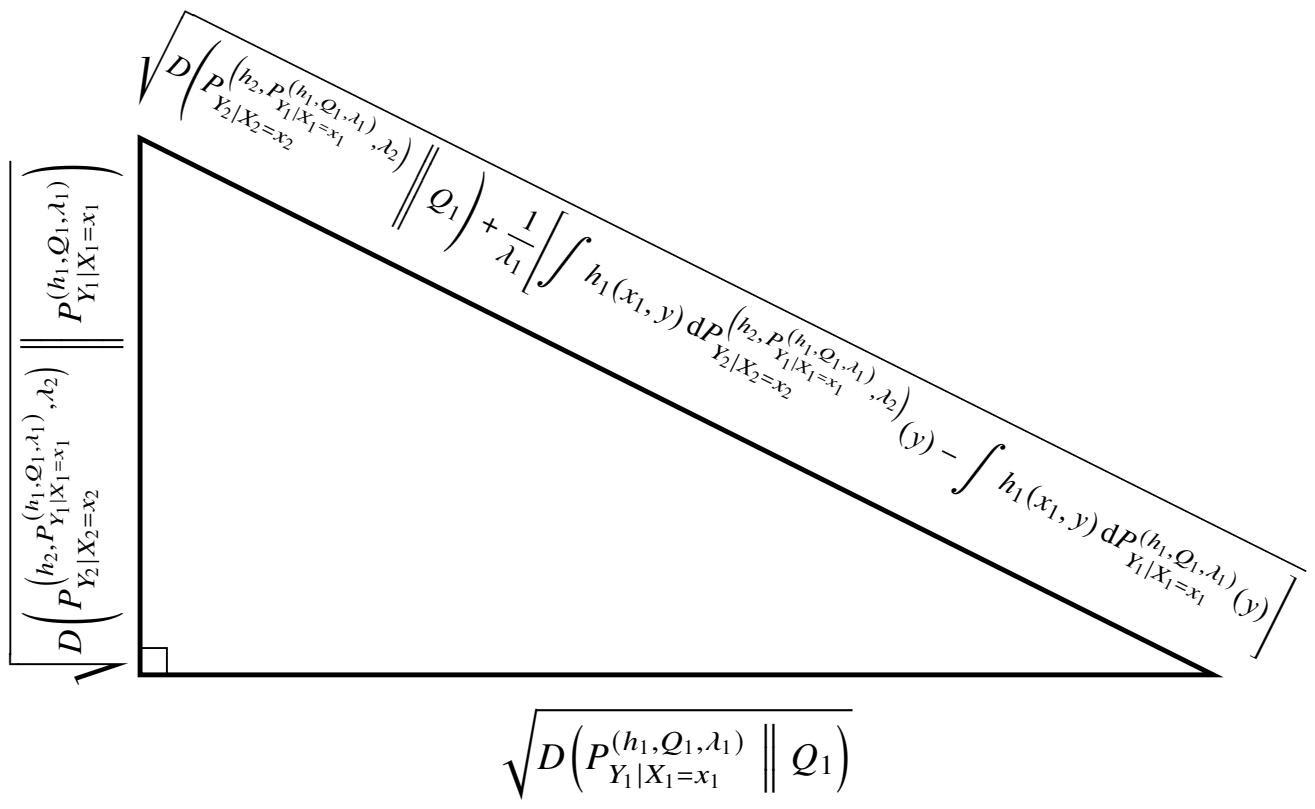
\begin{figure}[h]
\centering
\resizebox{0.5\linewidth}{!}{%
\input{\Figurespath/Triangle_Geo_rep.tikz}}
\caption{Pythagorean interpretation of
Lemma~\ref{LemmaSeptember9at9h48in2026inSophiaAntipolis}.}
\label{FigTriangle}
\end{figure}

Note that under the assumption that $Q_1$ is a probability measure,
both $\gamma_1$ and $\gamma_2$ are nonnegative, and thus, the left-hand
side of~\eqref{EqMethodOfGapsGeometric} is nonnegative as well.
These observations lead to interpreting~\eqref{EqMethodOfGapsGeometric}
from the perspective of the converse Pythagorean
theorem~\cite[Proposition~48]{heath1956thirteen}.
From such a perspective, a right-angle triangle can be formed as shown
in Figure~\ref{FigTriangle}.
This provides yet another intuition on the structure governing the
change in the sizes of the neighborhoods
$\set{G}_{Q_1,\gamma_1}$, $\set{G}_{Q_2,\gamma_2}$, and
$\set{G}_{Q_1,\eta}$; and the variation of the expectation of
$h_1(x_1,y)$ when $y$ is sampled from a measure that varies from
$P^{\autoparent{h_1,Q_1,\lambda_1}}_{Y\mid X=x_1}$ to
$P^{\autoparent{h_2,Q_2,\lambda_2}}_{Y\mid X=x_2}$.

The arguments above can be extended to the case in which $K$ Gibbs
probability measures are recursively nested. The corresponding identity is given by the following
lemma.
\begin{lemma}
\label{LemmaSept11at15h55in2026Sophia}
The nested Gibbs probability measures
$P^{\autoparent{h_1,Q_1,\lambda_1}}_{Y\mid X=x_1}$,
$P^{\autoparent{h_2,Q_2,\lambda_2}}_{Y\mid X=x_2}$,
$\ldots$,
$P^{\autoparent{h_K,Q_K,\lambda_K}}_{Y\mid X=x_K}$
in~\eqref{EqAugust28at11h54in2026Sophia} satisfy
\begin{IEEEeqnarray}{rCl}
\label{EqSumGammaEtaDelta}
\sum_{k=1}^{K}
\KL{
P^{\autoparent{h_k,Q_k,\lambda_k}}_{Y\mid X=x_k}
}{Q_k}
&=&
\sum_{i=1}^{K-1}\delta_i
+
\KL{
P^{\autoparent{h_K,Q_K,\lambda_K}}_{Y\mid X=x_K}
}{Q_1},
\end{IEEEeqnarray}
where $\delta_i$ is defined
in~\eqref{EqSept11at15h41in2026Sophia}.
\end{lemma}
\begin{IEEEproof}
The proof is presented in Appendix~\ref{AppProofOfLemmaSept11at15h55in2026Sophia}.
\end{IEEEproof}

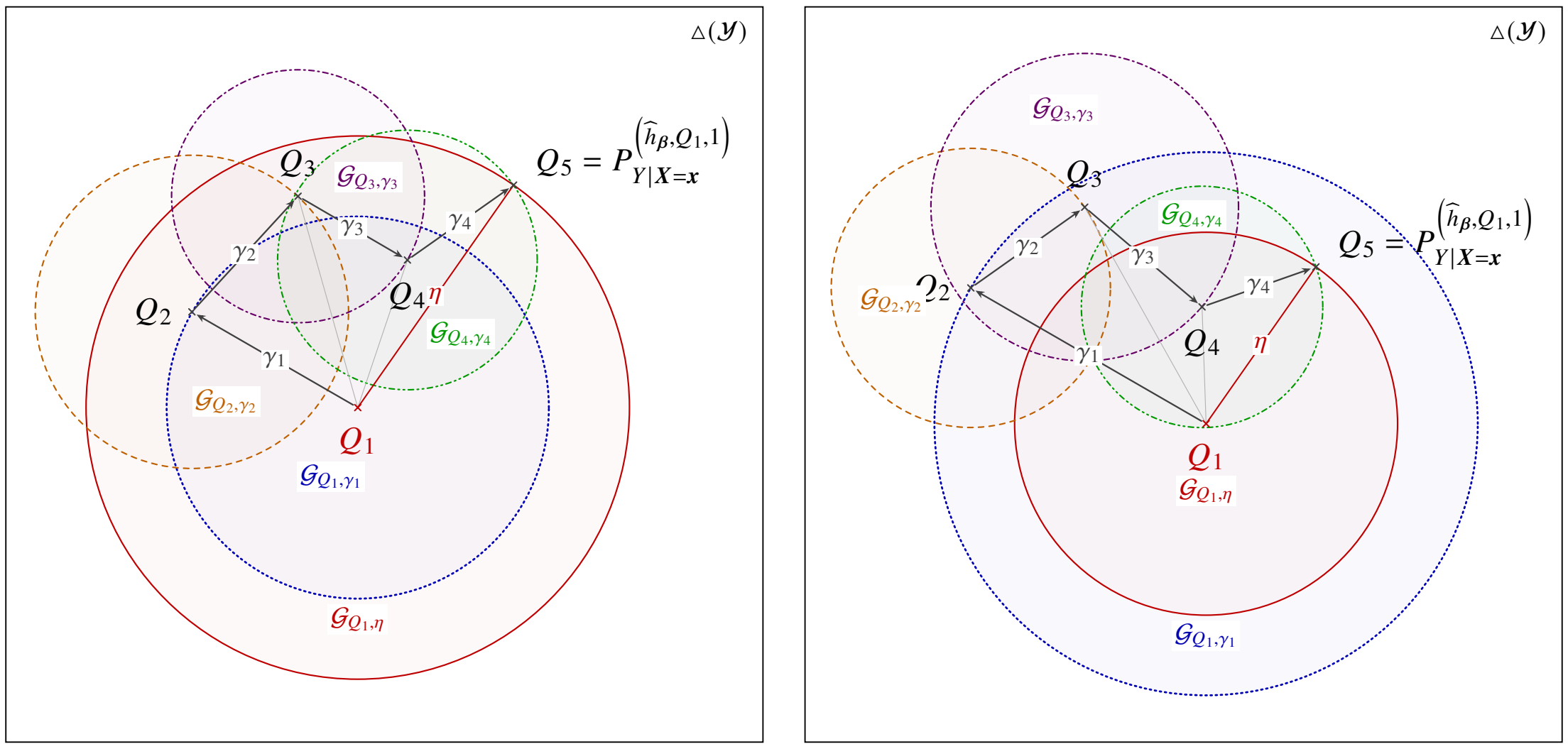
\begin{figure*}[t]
\centering
\input{\Figurespath/Nested_Decentralized/recursive_nesting_inclusions.tikz}
\resizebox{0.8\linewidth}{!}{%
  \RecursiveNestingPanel{2.15}{3.05}{(a) $\gamma_1<\eta$}%
  \hspace{8mm}%
  \RecursiveNestingPanel{3.05}{2.15}{(b) $\gamma_1>\eta$}%
}
\caption{Geometric representation of recursively nested Gibbs
probability measures illustrating
(a) $\mathcal{G}_{Q_1,\gamma_1}\subseteq\mathcal{G}_{Q_1,\eta}$
and
(b) $\mathcal{G}_{Q_1,\eta}\subseteq\mathcal{G}_{Q_1,\gamma_1}$.}
\label{FigRecursiveNestingInclusions}
\end{figure*}

In Lemma~\ref{LemmaSept11at15h55in2026Sophia},
the sum $\sum_{k=1}^{K} \KL{
P^{\autoparent{h_k,Q_k,\lambda_k}}_{Y\mid X=x_k}
}{Q_k}$ can be viewed as the length of a path from the
reference measure $Q_1$ to the final nested Gibbs probability measure
$P^{\autoparent{h_K,Q_K,\lambda_K}}_{Y\mid X=x_K}$, passing successively
through the reference measures $Q_2,Q_3,\ldots,Q_K$.
Figure~\ref{FigRecursiveNestingInclusions} provides a
geometric representation of this path for $K=4$.

%\begin{figure*}[t]
%\centering
%\resizebox{0.4\textwidth}{!}{%
%  \input{\Figurespath/Nested_Decentralized/geometric_interpretation.tikz}}
%\caption{Geometric representation of
%Lemma~\ref{LemmaSept11at15h55in2026Sophia} for $K=4$.}
%\label{FigRecursiveNestingGeometricRepresentation}
%\end{figure*}

\begin{figure}[t]
\centering
\resizebox{0.5\linewidth}{!}{%
  \input{\Figurespath/Triangle_Geo_repGen.tikz}}
\caption{Pythagorean interpretation of
Lemma~\ref{LemmaSept11at15h55in2026Sophia}.}
\label{FigTriangleGen}
\end{figure}
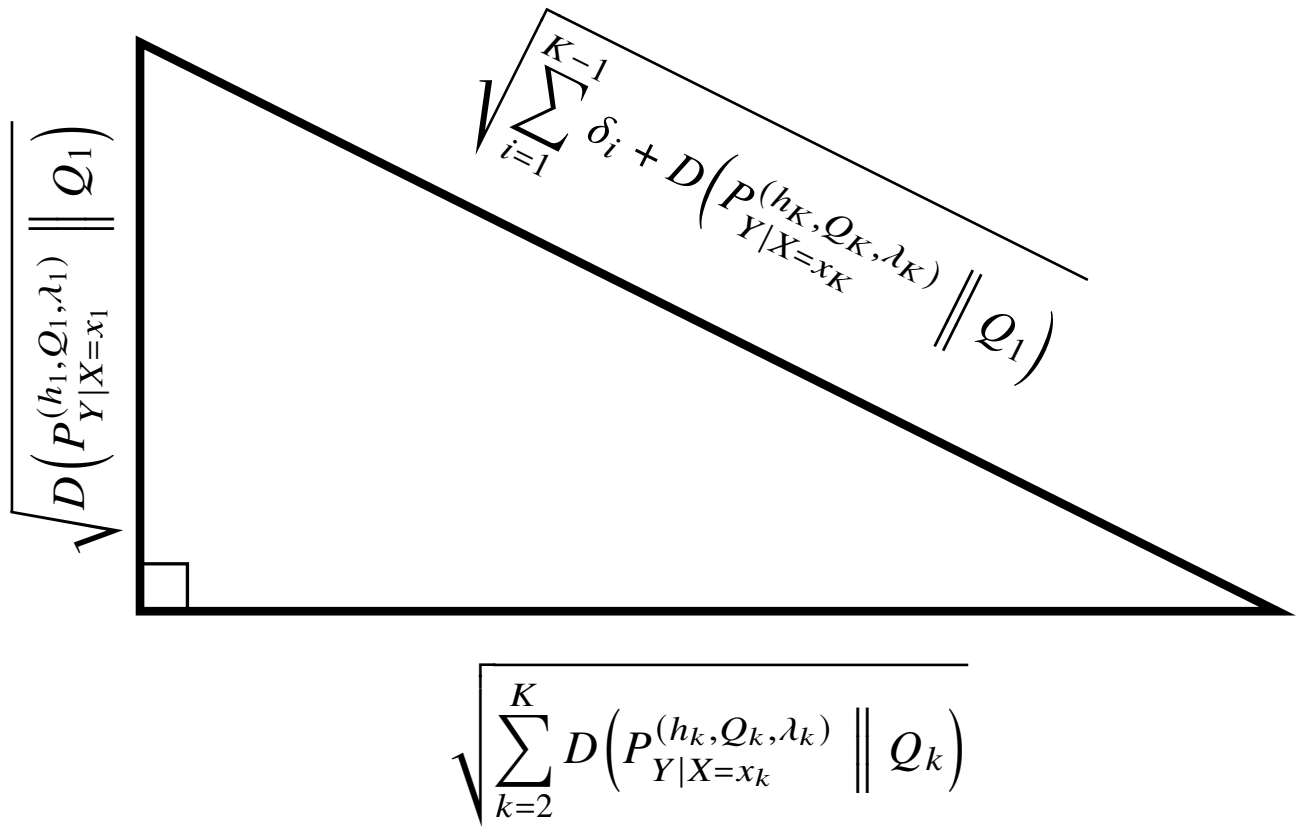

An equivalent form of the identity
in~\eqref{EqSumGammaEtaDelta} is
\begin{IEEEeqnarray}{rCl}
\label{EqPythagoreanRecursiveNesting}
\gamma_1+\displaystyle\sum_{i=2}^{K}\gamma_i
&=&
\displaystyle\sum_{i=1}^{K-1}\delta_i+\eta.
\end{IEEEeqnarray}
Under the assumption that $Q_1$ is a probability measure, the reals $\gamma_1,\gamma_2,\ldots,\gamma_K$ are nonnegative.
Consequently, from~\eqref{EqPythagoreanRecursiveNesting}, it follows that
\begin{IEEEeqnarray}{rCl}
\displaystyle\sum_{i=1}^{K-1}\delta_i+\eta
&\geqslant&
0,
\end{IEEEeqnarray}
which allows interpreting~
\eqref{EqPythagoreanRecursiveNesting} from the
perspective of the converse Pythagorean theorem~\cite[Proposition~48]{heath1956thirteen}.
Figure~\ref{FigTriangleGen} shows the corresponding right-angle triangle.

\section{Gibbs Measures in Supervised Machine Learning}
\label{SecApplicationstoSupervisedMachineLearning}

This section studies normalized log-linear combinations of Gibbs probability measures (Definition~\ref{DefMay19at15h20in2026Sophia}) and nested Gibbs probability measures (Definition~\ref{DefNestedGibbsMeasures}) in the context of supervised machine learning. The first subsection introduces a formal formulation of such a problem and establishes the notation used throughout the section. 
The following subsections leverage the results in Section~\ref{SecMainResults} to characterize the solutions to optimizations of weighted empirical risks, subject to relative entropy regularizations, as results of log-linear combinations of Gibbs measures and/or nesting multiple Gibbs probability measures. 

\subsection{The Problem of Supervised Machine Learning}
Let $\set{X}$ and $\set{Y}$ be sets of \emph{patterns} and \emph{labels}, respectively. Let $\set{M}\subseteq\reals^{d}$, with $d\in\ints$, denote a set of \emph{models}. 
A data point~$(x,y)$ is an element of the set $\set{Z} \triangleq \set{X} \times \set{Y}$. 
A training dataset $\vect{z}$ consists of a tuple of $n \in \ints$ data points,  explicitly written as
\begin{IEEEeqnarray}{rCl}
\label{EqDataset}
\vect{z} \;\triangleq\; \autoparent{(x_{1},y_{1}),(x_{2},y_{2}),\ldots,(x_{n},y_{n})} \in\set{Z}^{n}.
\end{IEEEeqnarray}
Given a model $\vect{\theta}\in\set{M}$, the loss induced by such a model on a data point~$(x,y)\in\set{Z}$ is denoted by $\ell(x,y,\vect{\theta})$, where
\begin{IEEEeqnarray}{rCl}
\label{EqLossFunction}
\ell:\set{Z}\times\set{M}\to[0,+\infty)
\end{IEEEeqnarray}
is referred to as the \emph{loss function}, which is assumed to be Borel measurable.
The \emph{empirical risk} induced by a model with respect to an $n$-length dataset is determined by the function
\begin{IEEEeqnarray}{rCl}
\label{EqEmpiricalRiskLn}
\mathsf{L}_n:\function{\set{Z}^{n}\times\set{M}}{[0,+\infty)}{\autoparent{\vect{z},\vect{\theta}}}
{\frac{1}{n}\sum_{i=1}^{n}\ell\autoparent{x_i,y_i,\vect{\theta}}.}
\end{IEEEeqnarray}

The \emph{expected empirical risk} with respect to a fixed dataset $\vect{z}$ of the form in~\eqref{EqDataset}, when models are sampled from a given probability measure, is determined by the functional 
\begin{IEEEeqnarray}{rCl}
\label{EqExpEmpRiskModels}
\mathsf{R}_{\vect{z}}:\function{\simplex{\set{M}}}{[0,+\infty)}{P}
{\displaystyle \int \EmpRisk{n}{\vect{z}}{\vect{\theta}} \mathrm{d}P\autoparent{\vect{\theta}}.}
\end{IEEEeqnarray}

In this work, a supervised learning algorithm, trained upon $n$-length datasets, is represented by a conditional probability measure in $\simplex{\set{M} \mid \set{Z}^n}$ as defined hereunder.

\begin{definition}[Algorithm]
\label{DefAlgorithmSL}
A conditional probability measure
$P_{\vect{\Theta}\mid\vect{Z}}\in \simplex{\set{M}\mid\set{Z}^{n}}$
is said to represent a supervised machine learning algorithm with $n$-length training datasets.
\end{definition}

The
instance of such an algorithm trained upon the dataset $\vect{z}$ in~\eqref{EqDataset} is denoted by $P_{\vect{\Theta} \mid \vect{Z} = \vect{z}}$, which is simply a probability measure in $\simplex{\set{M}}$.
%Given $P_{\vect{\Theta}\mid\vect{Z}}\in \simplex{\set{M}\mid\set{Z}^{n}}$ and a dataset $\vect{z}\in\set{Z}^{n}$, the corresponding
%instance is the probability measure $P_{\vect{\Theta}\mid\vect{Z}=\vect{z}}\in\simplex{\set{M}}$.
Given a $\sigma$-finite measure $Q$ defined on the set $\set{M}$ and a real $\lambda \in \reals\setminus\lbrace 0 \rbrace$, a central class of algorithms known as Gibbs algorithms, are represented by $(\mathsf{L}_n, Q,\lambda)$-Gibbs conditional probability measures (Definition~\ref{DefGeneralGibbs}). An example of such a class of algorithms can be denoted by
\begin{IEEEeqnarray}{rCl}
\label{EqGibbsRNDFixedDataset_SL}
P^{(\mathsf{L}_n, Q,\lambda)}_{\vect{\Theta}\mid\vect{Z}} \in \simplex{\set{M}\mid\set{Z}^{n}}.
\end{IEEEeqnarray}
Given a fixed dataset $\vect{z} \in \set{Z}^{n}$, the instance of the Gibbs algorithm  in~\eqref{EqGibbsRNDFixedDataset_SL} obtained by training it upon the dataset $\vect{z}$ is represented by the measure
\begin{subequations}
\label{EqMarch31at15h49in2026Nice}
\begin{IEEEeqnarray}{rcl}
P^{(\mathsf{L}_n, Q,\lambda)}_{\vect{\Theta}\mid\vect{Z} = \vect{z}} \in \simplexabs{Q}{\set{M}}, 
\end{IEEEeqnarray}
whose Radon-Nikodym derivative with respect to $Q$ satisfies
\begin{IEEEeqnarray}{rCl}
\RND{P^{(\mathsf{L}_n, Q,\lambda)}_{\vect{\Theta}\mid\vect{Z} = \vect{z}}}{Q} (\vect{\theta}) &=& \frac{\exp\autoparent{- \frac{1}{\lambda} \EmpRisk{n}{\vect{z}}{\vect{\theta}}}}{\displaystyle \int \exp\autoparent{- \frac{1}{\lambda} \EmpRisk{n}{\vect{z}}{\vect{\nu}} } \mathrm{d}Q\autoparent{\vect{\nu}}}.
\end{IEEEeqnarray}
\end{subequations}
%With this choice of the function $h$ in Definition~\ref{DefGeneralGibbs}, under the assumptions that $Q$ is a probability measure and $\lambda > 0$, the condition in~\eqref{EqGeneralGibbsIntegrability} is always met. Thus, under such assumptions, a conditional measure that satisfies~\eqref{EqGeneralGibbsRN} is always a Gibbs conditional probability measure.
%%
%Alternatively, when $\lambda < 0$, the condition in~\eqref{EqGeneralGibbsIntegrability} is more restrictive and might impose particular constraints on the function $h$. 

%The following lemmas show that this measure  can be obtained as the unique solution to two different optimization problems. 
The Gibbs probability measure
$P^{(\mathsf{L}_n,Q,\lambda)}_{\vect{\Theta}\mid\vect{Z}=\vect{z}}$
in~\eqref{EqMarch31at15h49in2026Nice} admits two complementary characterizations in terms of empirical-risk optimization problems.
The first characterization consists of an empirical-risk optimization (minimization or maximization) with relative-entropy regularization:
\begin{IEEEeqnarray}{rCl}
\label{EqMarch4at14h00in2026SophiaA}
\min_{P\in\triangle_Q\autoparent{\set{M}}}
&&
\mathsf{R}_{\vect{z}}\autoparent{P} + \lambda\,\KL{P}{Q},
\qquad \text{with } \lambda>0; \ \text{and} \\
\label{EqMarch4at14h00in2026SophiaB}
\max_{P\in\triangle_Q\autoparent{\set{M}}}
&&
\mathsf{R}_{\vect{z}}\autoparent{P} + \lambda\,\KL{P}{Q},
\qquad \text{with } \lambda<0,
\end{IEEEeqnarray}
where $\mathsf{R}_{\vect{z}}$ is defined in~\eqref{EqExpEmpRiskModels}. 
These problems have been thoroughly studied in~\cite{perlaza2024empirical, perlaza2024generalization, daunas2025asymmetry}.
The following lemma establishes their connection with Gibbs algorithms.

\begin{lemma}\label{LemmaSolution_SL}
Assume that the optimization problem in~\eqref{EqMarch4at14h00in2026SophiaA}
(respectively, in~\eqref{EqMarch4at14h00in2026SophiaB}) admits a solution.
Then, if $\lambda>0$ (respectively, if $\lambda<0$), the Gibbs probability measure
$P^{(\mathsf{L}_n, Q,\lambda)}_{\vect{\Theta}\mid\vect{Z}=\vect{z}}$ in~\eqref{EqMarch31at15h49in2026Nice}
is the unique solution to~\eqref{EqMarch4at14h00in2026SophiaA}
(respectively, to~\eqref{EqMarch4at14h00in2026SophiaB}).
\end{lemma}
\begin{IEEEproof}
The proof is immediate from Lemma~\ref{LemmaJuly28at10h54in2026Sophia}.
\end{IEEEproof}

The optimization problems in~\eqref{EqMarch4at14h00in2026SophiaA} and~\eqref{EqMarch4at14h00in2026SophiaB} provide a classical method for constructing Gibbs algorithms~\cite{perlaza2024empirical}. The minimization is motivated by building an algorithm that minimizes a linear combination of the expectation of the empirical risk and a relative entropy term.
The first term represents the objective of searching for probability measures concentrated on the minimizers of the empirical risk with respect to the training dataset $\vect{z}$, i.e., the set $\arg \min_{\vect{\theta} \in \set{M}} \mathsf{L}_n(\vect{z}, \vect{\theta})$. The second term represents the aim of constraining such a search to measures that have small relative entropy with respect to the reference measure $Q$. That is, measures that ``look like'' or are ``close to'' the reference measure, in the sense of relative entropy.
This suggests that $\lambda$ deals with the trade-off between the solution being driven by the training data, for small $\abs{\lambda}$, or by the reference measure, for large $\abs{\lambda}$. Interestingly, this justifies why $Q$ is often seen as a prior in the existing literature. See~\cite{aminian2024information,perlaza2024empirical,xu2017information} and references therein.
While the minimization in~\eqref{EqMarch4at14h00in2026SophiaA} concentrates most of the attention in the existing literature, the maximization in~\eqref{EqMarch4at14h00in2026SophiaB} is shown later in this work to be important for machine unlearning.

The second characterization consists of an empirical-risk optimization (minimization or maximization) subject to a relative entropy constraint:
\begin{subequations}\label{EqMarch4at14h35in2026SophiaA}
\begin{IEEEeqnarray}{rCl}
\min_{P\in\triangle_Q\autoparent{\set{M}}} && \mathsf{R}_{\vect{z}}\autoparent{P} \\
\text{s.t.} & \quad & \KL{P}{Q}\le \gamma, 
\end{IEEEeqnarray}
and
\end{subequations}
\begin{subequations}\label{EqMarch4at14h35in2026SophiaB}
\begin{IEEEeqnarray}{rCl}
\max_{P\in\triangle_Q\autoparent{\set{M}}} && \mathsf{R}_{\vect{z}}\autoparent{P} \\
\text{s.t.} & \quad & \KL{P}{Q}\le \gamma, 
\end{IEEEeqnarray}
\end{subequations}
where $\gamma \in \reals$. These problems have been thoroughly studied in \cite{PerlazaEntropy2025}.
The following lemma establishes their connection with Gibbs algorithms.
\begin{lemma}\label{LemmaMarch4at14h48in2026Sophia}
Assume that~$\lambda > 0$ (respectively,~$\lambda < 0$) is such that the $(\mathsf{L}_n, Q, \lambda)$ Gibbs probability measure $P^{(\mathsf{L}_n, Q,\lambda)}_{\vect{\Theta}\mid\vect{Z} = \vect{z}}$ in~\eqref{EqMarch31at15h49in2026Nice} satisfies
\begin{IEEEeqnarray}{rCl}
\label{EqMarch19at16h20in2025Nice}
\KL{P^{(\mathsf{L}_n, Q,\lambda)}_{\vect{\Theta}\mid\vect{Z}=\vect{z}}}{Q} &=& \gamma,
\end{IEEEeqnarray}
for some given $\gamma \in \reals$. 
Then, such a measure $P^{(\mathsf{L}_n, Q,\lambda)}_{\vect{\Theta}\mid\vect{Z}=\vect{z}}$ 
is the unique solution to~\eqref{EqMarch4at14h35in2026SophiaA}
(respectively, to~\eqref{EqMarch4at14h35in2026SophiaB}).
\end{lemma}
\begin{IEEEproof}
The proof is immediate from  Lemma~\ref{LemmaMarch4at9h15in2026inSophia}.
\end{IEEEproof}

When $Q$ is a probability measure, the optimization problems in~\eqref{EqMarch4at14h35in2026SophiaA} and~\eqref{EqMarch4at14h35in2026SophiaB} respectively consist in the minimization and maximization of the empirical risk within a relative entropy neighborhood $\set{G}_{Q, \gamma}$ in \eqref{EqKLNeighborhood}, with $\gamma$ in \eqref{EqMarch19at16h20in2025Nice}, around the reference measure $Q$. 
Hence, the $(\mathsf{L}_n, Q, \lambda)$ Gibbs probability measure
$P^{(\mathsf{L}_n, Q,\lambda)}_{\vect{\Theta}\mid\vect{Z}=\vect{z}}$ in~\eqref{EqMarch31at15h49in2026Nice} is the minimizer ($\lambda > 0$) or the maximizer ($\lambda < 0$), of the expected empirical risk $\mathsf{R}_{\vect{z}}$ within $\set{G}_{Q, \gamma}$. 
  
\subsection{Normalized Log-Linear Combination of Gibbs Algorithms}
\label{SubSecNormalizedLogComGibbsAlg}
This subsection leverages the results on normalized log-linear combinations of Gibbs probability measures presented in Section~\ref{SecMay27at16h10in2026BusToNice}. In the supervised learning setting, these results reveal a direct connection between normalized log-linear combinations of Gibbs algorithms and weighted empirical risks.
Consider two datasets
\begin{subequations}\label{EqApril1at7h46in2026Sophia}
\begin{IEEEeqnarray}{rCl}
\label{Datasetz1}
 \vect{z}_1 & \triangleq & \autoparent{(x_{1,1}, y_{1,1}), (x_{1,2}, y_{1,2}), \ldots, (x_{1,n_1}, y_{1,n_1})} \in \set{Z}^{n_1}\mbox{ and }\\
\label{Datasetz2}
 \vect{z}_2 & \triangleq & \autoparent{(x_{2,1}, y_{2,1}), (x_{2,2}, y_{2,2}), \ldots, (x_{2,n_2}, y_{2,n_2})}\in \set{Z}^{n_2},
\end{IEEEeqnarray}
\end{subequations}
with $n_1$ and $n_2$ data points, respectively. Let $\lambda_1 \in \reals\setminus\lbrace 0 \rbrace$ and $\lambda_2 \in \reals\setminus\lbrace 0 \rbrace$ be two given reals; and let $Q_1$ and $Q_2$ be two given $\sigma$-finite measures. 
Consider also an $\autoparent{\mathsf{L}_{n_1},Q_1,\lambda_1}$-Gibbs probability measure denoted by \begin{IEEEeqnarray}{rCl}
\label{EqMay26at14h04in2026inAntibes}
P_{\vect{\Theta} \mid\vect{Z} = \vect{z}_1}^{\autoparent{\mathsf{L}_{n_1},Q_1,\lambda_1}} \in \simplexabs{Q_1}{\set{M}}
\end{IEEEeqnarray}
 and an $(\mathsf{L}_{n_2},Q_2,\lambda_2)$-Gibbs probability measure denoted by \begin{IEEEeqnarray}{rCl}
\label{EqMay26at14h05in2026inAntibes}
 P_{\vect{\Theta} \mid\vect{Z} = \vect{z}_2}^{\autoparent{\mathsf{L}_{n_2},Q_2,\lambda_2}} \in \simplexabs{Q_2}{\set{M}},
 \end{IEEEeqnarray}
 where the functions $\mathsf{L}_{n_1}$ and $\mathsf{L}_{n_2}$ are defined as in~\eqref{EqEmpiricalRiskLn}.
Given $\alpha_1 \in \reals\setminus\lbrace 0 \rbrace$, $\alpha_2 \in \reals\setminus\lbrace 0 \rbrace$ and a $\sigma$-finite measure $Q$ on $\set{M}$, such that $Q\ll Q_1$ and $Q\ll Q_2$, let the normalized log-linear combination of the Gibbs probability measures in~\eqref{EqMay26at14h04in2026inAntibes} and~\eqref{EqMay26at14h05in2026inAntibes} with respect to $Q$ be denoted by $S_2 \in \simplex{\set{M}}$, such that  for all $\vect{\theta} \in \supp Q$,
\begin{IEEEeqnarray}{rCl}
\label{EqMay26at14h34in2026Antibes}
\RND{S_2}{Q}(\vect{\theta}) = \frac{\autoparent{\RND{P_{\vect{\Theta} \mid\vect{Z} = \vect{z}_1}^{\autoparent{\mathsf{L}_{n_1},Q_1,\lambda_1}}}{Q_1}\autoparent{\vect{\theta}}}^{\alpha_1} \autoparent{\RND{P_{\vect{\Theta} \mid\vect{Z} = \vect{z}_2}^{\autoparent{\mathsf{L}_{n_2},Q_2,\lambda_2}}}{Q_2}(\vect{\theta})}^{\alpha_2}}{\displaystyle \int \autoparent{\RND{P_{\vect{\Theta} \mid\vect{Z} = \vect{z}_1}^{\autoparent{\mathsf{L}_{n_1},Q_1,\lambda_1}}}{Q_1}\autoparent{\vect{\nu}}}^{\alpha_1} \autoparent{\RND{P_{\vect{\Theta} \mid\vect{Z} = \vect{z}_2}^{\autoparent{\mathsf{L}_{n_2},Q_2,\lambda_2}}}{Q_2}(\vect{\nu})}^{\alpha_2} \mathrm{d}Q(\vect{\nu})}.
\end{IEEEeqnarray}

%%%%%%%%%%%%%%%%%%%%%%%%%%%%%%%%%%%%
%
For ease of presentation, consider a positive integer $m\in\ints$, and the function
\begin{IEEEeqnarray}{rCl}
\label{EqMarch22at12h49in2026Nice}
\widebar{\mathsf{L}}_m:
\function{\reals^m \times \set{Z}^{m}\times\set{M}}{\reals}{\autoparent{\vect{\omega}, \vect{z},\vect{\theta}}}
{\sum_{t=1}^{m}\omega_t \ell\autoparent{x_t,y_t,\vect{\theta}},}
\end{IEEEeqnarray}
where the vector $\vect{\omega} \in \reals^{m}$ is of the form $\vect{\omega} = \autoparent{\omega_1, \omega_2, \ldots, \omega_m}$; the $m$-length dataset $\vect{z}$ is of the form in~\eqref{EqDataset}; and the function $\ell$ is defined in~\eqref{EqLossFunction}.
Note that $\widebar{\mathsf{L}}_m\autoparent{\vect{\omega}, \vect{z},\vect{\theta}}$ is a weighted empirical risk. More specifically, each data point $(x_t, y_t)$ in the dataset $\vect{z}$ is associated with a weight $\omega_t$, with $t\inCountK{m}$.
Note that, if for all $t\inCountK{m}$, $\omega_t=\frac{1}{m}$, the function $\widebar{\mathsf{L}}_m$ in~\eqref{EqMarch22at12h49in2026Nice} satisfies that $\widebar{\mathsf{L}}_m\autoparent{\vect{\omega}, \vect{z},\vect{\theta}} = \mathsf{L}_{m}\autoparent{\vect{z},\vect{\theta}}$, where the function $\mathsf{L}_{m}$ is the empirical risk function in~\eqref{EqEmpiricalRiskLn}.

Let the aggregation of the datasets $\vect{z}_1$ and $\vect{z}_2$ in~\eqref{EqApril1at7h46in2026Sophia} be denoted by
\begin{subequations}
\label{EqMarch10at20h02in2026Sophia}
\begin{IEEEeqnarray}{rCl}
\vect{z}_0
&\triangleq&
\autoparent{\vect{z}_1,\vect{z}_2}\\
& = & \autoparent{({x}_{0,1},{y}_{0,1}),({x}_{0,2},{y}_{0,2}),\ldots,({x}_{0,n_0},{y}_{0,n_0})}  \in \set{Z}^{n_0},
\end{IEEEeqnarray}
where
\begin{IEEEeqnarray}{rCl}
n_0
&\triangleq&
n_1 + n_2.
\end{IEEEeqnarray}
\end{subequations}
Using this notation, Theorem~\ref{TheoremMay20at13h57in2026Antibes} leads to an explicit expression for the measure $S_2$ in~\eqref{EqMay26at14h34in2026Antibes}, as shown by the following lemma.

\begin{lemma}\label{LemmaMay26at15h45in2026Antibes}
The normalized log-linear combination $S_2$ in~\eqref{EqMay26at14h34in2026Antibes} satisfies for all~$\vect{\theta}\in\supp Q$,
\begin{IEEEeqnarray}{rCl}
\label{EqMarch22at13h24in2026Nice}
\RND{S_2}{Q} (\vect{\theta})
& = &
\frac{\exp\autoparent{-\widebar{\mathsf{L}}_{n_0}\autoparent{\vect{\omega}, \vect{z}_0,\vect{\theta}}}}
{\displaystyle \int \exp\autoparent{-\widebar{\mathsf{L}}_{n_0}\autoparent{\vect{\omega}, \vect{z}_0, \vect{\nu}}} \mathrm{d}Q(\vect{\nu})},
\end{IEEEeqnarray}
where the function $\widebar{\mathsf{L}}_{n_0}$ is defined in~\eqref{EqMarch22at12h49in2026Nice}; the dataset $\vect{z}_0$ is defined in~\eqref{EqMarch10at20h02in2026Sophia}; and the vector $\vect{\omega}=\autoparent{\omega_1,\omega_2,\ldots,\omega_{n_0}}$ satisfies, for all $t\inCountK{n_0}$,
\begin{IEEEeqnarray}{rCl}
\label{EqMay27at19h29in2026Antibes}
\omega_t
& = &
\left\{
\begin{array}{cl}
\frac{\alpha_1}{n_1\lambda_1} & \text{if } t \leqslant n_1,\\[0.2cm]
\frac{\alpha_2}{n_2\lambda_2} & \text{if } n_1 < t \leqslant n_0.
\end{array}
\right.
\end{IEEEeqnarray} 
\end{lemma}
\begin{IEEEproof}
The proof follows from Theorem~\ref{TheoremMay20at13h57in2026Antibes} by observing that
\begin{IEEEeqnarray}{rCl}
\label{EqMay26at15h48in2026Antibes}
\widebar{\mathsf{L}}_{n_0}\autoparent{\vect{\omega}, \vect{z}_0,\vect{\theta}}
& = &
\frac{\alpha_1}{\lambda_1}\mathsf{L}_{n_1}\autoparent{\vect{z}_1, \vect{\theta}}
+
\frac{\alpha_2}{\lambda_2}\mathsf{L}_{n_2}\autoparent{\vect{z}_2, \vect{\theta}}, 
\end{IEEEeqnarray}
which completes the proof. 
\end{IEEEproof}
Lemma~\ref{LemmaMay26at15h45in2026Antibes} shows that the normalized log-linear combination of the two Gibbs probability measures in~\eqref{EqMay26at14h04in2026inAntibes} and~\eqref{EqMay26at14h05in2026inAntibes} is itself a Gibbs probability measure whose objective function is a weighted empirical risk over the aggregated dataset $\vect{z}_0$. More precisely, for all $i\inCountTwo$, the contribution associated with the dataset $\vect{z}_i$ is jointly determined by the coefficient $\alpha_i$ of the normalized log-linear combination and the regularization factor $\lambda_i$ of the corresponding Gibbs probability measure.
This Gibbs representation also provides an optimization interpretation of the normalized log-linear combination.
From Lemma~\ref{LemmaJuly28at10h54in2026Sophia}, the probability measure $S_2$
in~\eqref{EqMay26at14h34in2026Antibes} is the unique solution to the
weighted empirical-risk minimization with relative-entropy regularization:
\begin{IEEEeqnarray}{rcl}
\label{EqMay17at19h32in2026Antibes}
\min_{P \in \simplexabs{Q}{\set{M}}}
\int \widebar{\mathsf{L}}_{n_0}\autoparent{\vect{\omega}, \vect{z}_0, \vect{\theta}}
\mathrm{d} P(\vect{\theta})
+ \KL{P}{Q},
\end{IEEEeqnarray}
where the vector of weights $\vect{\omega}$ satisfies
\eqref{EqMay27at19h29in2026Antibes}.

For some fixed $m \in \ints$ and $\vect{\omega} \in \reals^{m}$, let the notation $\widebar{\mathsf{L}}_{m}(\vect{\omega}, \cdot, \cdot)$, with the function $\widebar{\mathsf{L}}_{m}$ defined as in~\eqref{EqMarch22at12h49in2026Nice}, identify the mapping
\begin{IEEEeqnarray}{rCl}
\label{EqSeptember3at6h58in2026Benguerir}
\widebar{\mathsf{L}}_{m}(\vect{\omega}, \cdot, \cdot):
\function{\set{Z}^{m}\times\set{M}}{\reals}{\autoparent{\vect{z},\vect{\theta}}}
{\sum_{t=1}^{m}\omega_t \ell\autoparent{x_t,y_t,\vect{\theta}}.}
\end{IEEEeqnarray}

%\textcolor{red}{Samir: is this needed ? also there is problem in how it is defined}
%\begin{IEEEeqnarray}{rCl}
%\label{EqSeptember3at6h58in2026Benguerir}
%\widebar{\mathsf{L}}_{\tilde{n}}(\vect{\omega}, \cdot, \cdot):
%\function{\set{Z}^{\tilde{n}}\times\set{M}}{\reals}{\autoparent{\vect{\omega}, \vect{z},\vect{\theta}}}
%{\sum_{t=1}^{\tilde{n}}\omega_t \ell\autoparent{x_t,y_t,\vect{\theta}}.}
%\end{IEEEeqnarray}
Using this notation, Lemma~\ref{LemmaMay26at15h45in2026Antibes} and Definition~\ref{DefGeneralGibbs} imply that the probability measure $S_2$ in~\eqref{EqMay26at14h34in2026Antibes} is identical to  an $(\widebar{\mathsf{L}}_{n_0}(\vect{\omega}, \cdot, \cdot), Q, 1)$-Gibbs probability measure denoted by 
\begin{IEEEeqnarray}{rCl}
\label{EqMay27at19h33in2026Antibes}
P_{\vect{\Theta}\mid\vect{Z} = \vect{z}_0}^{(\widebar{\mathsf{L}}_{n_0}(\vect{\omega}, \cdot, \cdot), Q, 1)} \in \simplexabs{Q}{\set{M}},
\end{IEEEeqnarray}
which, from Lemma~\ref{LemmaJuly28at10h54in2026Sophia} is the unique solution to~\eqref{EqMay17at19h32in2026Antibes}.
The following corollary formalizes this
observation.
\begin{corollary}
The probability measure $S_2$ in~\eqref{EqMay26at14h34in2026Antibes} and the
$(\widebar{\mathsf{L}}_{n_0}(\vect{\omega}, \cdot, \cdot), Q, 1)$-Gibbs
probability measure $P_{\vect{\Theta}\mid\vect{Z} = \vect{z}_0}^{(\widebar{\mathsf{L}}_{n_0}(\vect{\omega}, \cdot, \cdot), Q, 1)} $ in~\eqref{EqMay27at19h33in2026Antibes} satisfy
\begin{IEEEeqnarray}{rCl}
\KL{S_2}
{P_{\vect{\Theta}\mid\vect{Z} = \vect{z}_0}^{(\widebar{\mathsf{L}}_{n_0}(\vect{\omega}, \cdot, \cdot), Q, 1)}}
& = & 0.
\end{IEEEeqnarray}
\end{corollary}

Lemma~\ref{LemmaMay26at15h45in2026Antibes} reveals that the weight associated to each data point on the measure $S_2$ in \eqref{EqMarch22at13h24in2026Nice} depends on whether it is part of $\vect{z}_1$ or $\vect{z}_2$.   
This naturally raises the question of whether the parameters of the normalized log-linear combination can be chosen such that all data points exhibit the same weight in the weighted empirical risk in~\eqref{EqMay17at19h32in2026Antibes}. 

\begin{lemma}\label{LemmaMay29at20h00in2026Antibes}
Consider a real $\lambda_0\in\reals\setminus\lbrace 0 \rbrace$. Assume also that the
regularization factors $\lambda_1$ and $\lambda_2$ of the Gibbs probability measures in
\eqref{EqMay26at14h34in2026Antibes} satisfy
\begin{IEEEeqnarray}{rCl}
\label{EqMay29at11h01in2026Antibes}
\lambda_1
& = &
\frac{n_0\lambda_0\alpha_1}{n_1};
\qquad
\lambda_2
=
\frac{n_0\lambda_0\alpha_2}{n_2}.
\end{IEEEeqnarray}
Then, the probability measure $S_2$ in
\eqref{EqMay26at14h34in2026Antibes} satisfies, for all
$\vect{\theta}\in\supp Q$,
\begin{IEEEeqnarray}{rCl}
\label{EqMay29at11h02in2026Antibes}
\RND{S_2}{Q}(\vect{\theta})
& = &
\frac{
\exp\autoparent{
-\frac{1}{\lambda_0}
\mathsf{L}_{n_0}\autoparent{\vect{z}_0,\vect{\theta}}}}
{
\int
\exp\autoparent{
-\frac{1}{\lambda_0}
\mathsf{L}_{n_0}\autoparent{\vect{z}_0,\vect{\nu}}}
\mathrm{d}Q(\vect{\nu})}
\\
\label{EqMay29at11h03in2026Antibes}
& = &
\RND{
P_{\vect{\Theta}\mid\vect{Z}=\vect{z}_0}^{(\mathsf{L}_{n_0},Q,\lambda_0)}
}{Q}
(\vect{\theta}),
\end{IEEEeqnarray}
where the probability measure
$P_{\vect{\Theta}\mid\vect{Z}=\vect{z}_0}^{(\mathsf{L}_{n_0},Q,\lambda_0)}$
is an $(\mathsf{L}_{n_0},Q,\lambda_0)$-Gibbs probability measure, with the function
$\mathsf{L}_{n_0}$ defined in~\eqref{EqEmpiricalRiskLn}.
\end{lemma}

\begin{IEEEproof}
The proof follows by observing that under the choice of $\lambda_1$ and $\lambda_2$ in
\eqref{EqMay29at11h01in2026Antibes}, the entries of the vector
$\vect{\omega}=\autoparent{\omega_1,\omega_2,\ldots,\omega_{n_0}}$ in
\eqref{EqMay27at19h29in2026Antibes} satisfy, for all
$t\inCountK{n_0}$,
\begin{IEEEeqnarray}{rCl}
\omega_t
& = &
\frac{1}{n_0\lambda_0}.
\end{IEEEeqnarray}
Therefore, for all $\vect{\theta}\in\set{M}$,
\begin{IEEEeqnarray}{rCl}
\widebar{\mathsf{L}}_{n_0}
\autoparent{\vect{\omega},\vect{z}_0,\vect{\theta}}
& = &
\displaystyle\sum_{t=1}^{n_0}
\frac{1}{n_0\lambda_0}
\ell\autoparent{x_{0,t},y_{0,t},\vect{\theta}}
\\
& = &
\frac{1}{\lambda_0}
\mathsf{L}_{n_0}\autoparent{\vect{z}_0,\vect{\theta}},
\end{IEEEeqnarray}
where the function $\widebar{\mathsf{L}}_{n_0}$ is defined in
\eqref{EqMarch22at12h49in2026Nice}. The equality in
\eqref{EqMay29at11h02in2026Antibes} follows from
Lemma~\ref{LemmaMay26at15h45in2026Antibes}, and the equality in
\eqref{EqMay29at11h03in2026Antibes} follows from
Definition~\ref{DefGeneralGibbs}, which completes the proof.
\end{IEEEproof}

Lemma~\ref{LemmaMay29at20h00in2026Antibes} shows a particular choice of the regularization factors $\lambda_1$ and $\lambda_2$ for which the normalized log-linear combination $S_2$ in~\eqref{EqMay26at14h34in2026Antibes} is identical to the $(\mathsf{L}_{n_0},Q,\lambda_0)$-Gibbs probability measure trained upon the aggregated dataset $\vect{z}_0$. Under this choice, the weighted empirical risk associated with $S_2$ satisfies
\begin{IEEEeqnarray}{rCl}
\widebar{\mathsf{L}}_{n_0}\autoparent{\vect{\omega},\vect{z}_0,\vect{\theta}}
&=&
\frac{1}{\lambda_0}
\mathsf{L}_{n_0}\autoparent{\vect{z}_0,\vect{\theta}}.
\end{IEEEeqnarray}
Hence, the optimization problem in~\eqref{EqMay17at19h32in2026Antibes} is equivalent to
\begin{IEEEeqnarray}{rCl}
\left\{
\begin{array}{cl}
\displaystyle
\min_{P\in\triangle_Q\autoparent{\set{M}}}
&
\displaystyle
\int
\mathsf{L}_{n_0}\autoparent{\vect{z}_0,\vect{\theta}}
 \mathrm{d}P\autoparent{\vect{\theta}}
+
\lambda_0\KL{P}{Q},
\quad \mbox{if } \lambda_0>0,
\\[1.5ex]
\displaystyle
\max_{P\in\triangle_Q\autoparent{\set{M}}}
&
\displaystyle
\int
\mathsf{L}_{n_0}\autoparent{\vect{z}_0,\vect{\theta}}
 \mathrm{d}P\autoparent{\vect{\theta}}
+
\lambda_0\KL{P}{Q},
\quad \mbox{if } \lambda_0<0,
\end{array}
\right.
\end{IEEEeqnarray}
where equivalence follows by multiplying the objective function in~\eqref{EqMay17at19h32in2026Antibes} by $\lambda_0$. Thus, the choice of $\lambda_1$ and $\lambda_2$ in Lemma~\ref{LemmaMay29at20h00in2026Antibes} makes all the data points in the aggregated dataset $\vect{z}_0$ contribute uniformly to the empirical risk and yields the same Gibbs probability measure as that obtained by training directly upon $\vect{z}_0$ with reference measure $Q$ and regularization factor $\lambda_0$.

When the datasets $\vect{z}_1$ and $\vect{z}_2$ share common data points, the contributions induced by the two Gibbs probability measures are combined. The following lemma makes this effect explicit.

\begin{lemma}\label{LemmaMarch23at16h09in2026BusToNice}
Assume that the datasets $\vect{z}_1$ and $\vect{z}_2$ in the normalized
log-linear combination $S_2$ in~\eqref{EqMay26at14h34in2026Antibes} satisfy
\begin{subequations}
\label{EqMay29at10h00in2026Antibes}
\begin{IEEEeqnarray}{rCl}
\vect{z}_1
& = &
\autoparent{\vect{z}_{1,1}, \vect{z}_{1,2}}
\in \set{Z}^{n_{1,1}} \times \set{Z}^{n_{1,2}}; \mbox{ and }\\
\vect{z}_2
& = &
\autoparent{\vect{z}_{2,1}, \vect{z}_{1,2}}
\in \set{Z}^{n_{2,1}} \times \set{Z}^{n_{1,2}},
\end{IEEEeqnarray}
and
\begin{IEEEeqnarray}{rCl}
n_1 & = & n_{1,1} + n_{1,2}; \mbox{ and }\\
n_2 & = & n_{2,1} + n_{1,2},
\end{IEEEeqnarray}
\end{subequations}
where all the data points common to $\vect{z}_1$ and $\vect{z}_2$ are in
$\vect{z}_{1,2}$. 
Then, the
$(\widebar{\mathsf{L}}_{n_0}(\vect{\omega}, \cdot, \cdot), Q, 1)$-Gibbs
probability measure in~\eqref{EqMay27at19h33in2026Antibes} is such that the vector
$\vect{\omega}=\autoparent{\omega_1,\omega_2,\ldots,\omega_{n_0}}$ satisfies, for
all $t\inCountK{n_0}$,
\begin{IEEEeqnarray}{CCl}
\label{EqMay29at10h02in2026Antibes}
\omega_t
& = &
\left\{
\begin{array}{cl}
\frac{\alpha_1}{n_1\lambda_1}
& \text{if } t \leqslant n_{1,1},\\[0.2cm]
\frac{\alpha_1}{n_1\lambda_1}
+
\frac{\alpha_2}{n_2\lambda_2}
& \text{if } n_{1,1} < t \leqslant n_1,\\[0.2cm]
\frac{\alpha_2}{n_2\lambda_2}
& \text{if } n_1 < t \leqslant n_1 + n_{2,1},\\[0.2cm]
0 & \text{if } n_1 + n_{2,1} < t \leqslant n_0.
\end{array}
\right.
\end{IEEEeqnarray}
\end{lemma}

\begin{IEEEproof}
Given a fixed model $\vect{\theta}\in\set{M}$, the empirical risk functions
$\mathsf{L}_{n_1}$ and $\mathsf{L}_{n_2}$ in~\eqref{EqEmpiricalRiskLn} satisfy
\begin{IEEEeqnarray}{rCl}
\mathsf{L}_{n_1}\autoparent{\vect{z}_1, \vect{\theta}}
& = &
\frac{n_{1,1}}{n_1}
\mathsf{L}_{n_{1,1}}\autoparent{\vect{z}_{1,1}, \vect{\theta}}
+
\frac{n_{1,2}}{n_1}
\mathsf{L}_{n_{1,2}}\autoparent{\vect{z}_{1,2}, \vect{\theta}};
\mbox{ and }
\\
\mathsf{L}_{n_2}\autoparent{\vect{z}_2, \vect{\theta}}
& = &
\frac{n_{2,1}}{n_2}
\mathsf{L}_{n_{2,1}}\autoparent{\vect{z}_{2,1}, \vect{\theta}}
+
\frac{n_{1,2}}{n_2}
\mathsf{L}_{n_{1,2}}\autoparent{\vect{z}_{1,2}, \vect{\theta}},
\end{IEEEeqnarray}
which follows from~\cite[Lemma~4]{perlaza2023validation}.
Moreover, the function $\widebar{\mathsf{L}}_{n_0}$ in
\eqref{EqMarch22at12h49in2026Nice} satisfies
\begin{IEEEeqnarray}{rCl}
\widebar{\mathsf{L}}_{n_0}\autoparent{\vect{\omega}, \vect{z}_0, \vect{\theta}}
& = &
\frac{\alpha_1}{\lambda_1}
\mathsf{L}_{n_1}\autoparent{\vect{z}_1, \vect{\theta}}
+
\frac{\alpha_2}{\lambda_2}
\mathsf{L}_{n_2}\autoparent{\vect{z}_2, \vect{\theta}}
\\
& = &
\frac{\alpha_1 n_{1,1}}{n_1\lambda_1}
\mathsf{L}_{n_{1,1}}\autoparent{\vect{z}_{1,1}, \vect{\theta}}
+
\frac{n_{1,2}\alpha_1}{n_1\lambda_1}
\mathsf{L}_{n_{1,2}}\autoparent{\vect{z}_{1,2}, \vect{\theta}}
+
\frac{\alpha_2 n_{2,1}}{n_2\lambda_2}
\mathsf{L}_{n_{2,1}}\autoparent{\vect{z}_{2,1}, \vect{\theta}}
+ 
\frac{n_{1,2}\alpha_2}{n_2\lambda_2}
\mathsf{L}_{n_{1,2}}\autoparent{\vect{z}_{1,2}, \vect{\theta}}\\
%& = &
%\frac{\alpha_1 n_{1,1}}{n_1\lambda_1}
%\mathsf{L}_{n_{1,1}}\autoparent{\vect{z}_{1,1}, \vect{\theta}}
%+
%n_{1,2}
%\autoparent{
%\frac{\alpha_1}{n_1\lambda_1}
%+
%\frac{\alpha_2}{n_2\lambda_2}}
%\mathsf{L}_{n_{1,2}}\autoparent{\vect{z}_{1,2}, \vect{\theta}}
%+
%\frac{\alpha_2 n_{2,1}}{n_2\lambda_2}
%\mathsf{L}_{n_{2,1}}\autoparent{\vect{z}_{2,1}, \vect{\theta}}
%\\
& = &
\displaystyle\sum_{t=1}^{n_{1,1}}
\frac{\alpha_1}{n_1\lambda_1}
\ell\autoparent{x_{0,t}, y_{0,t}, \vect{\theta}}
+
\hspace{-1ex}\displaystyle\sum_{t=n_{1,1}+1}^{n_{1}}
\autoparent{
\frac{\alpha_1}{n_1\lambda_1}
+
\frac{\alpha_2}{n_2\lambda_2}}
\ell\autoparent{x_{0,t}, y_{0,t}, \vect{\theta}}
+
\hspace{-1ex}\displaystyle\sum_{t=n_1+1}^{n_1+n_{2,1}}
\frac{\alpha_2}{n_2\lambda_2}
\ell\autoparent{x_{0,t}, y_{0,t}, \vect{\theta}}
+
\hspace{-1ex}
\displaystyle\sum_{t=n_1+n_{2,1} +1}^{n_0} \hspace{-2ex}
0
\ell\autoparent{x_{0,t}, y_{0,t}, \vect{\theta}} \squeezeequ \spnum
\\
\label{EqMay29at10h03in2026Antibes}
& = &
\displaystyle\sum_{t=1}^{n_0}
\omega_t
\ell\autoparent{x_{0,t}, y_{0,t}, \vect{\theta}},
\end{IEEEeqnarray}
where the dataset $\vect{z}_0$ is defined in
\eqref{EqMarch10at20h02in2026Sophia}, and the vector
$\vect{\omega}=\autoparent{\omega_1,\omega_2,\ldots,\omega_{n_0}}$
satisfies~\eqref{EqMay29at10h02in2026Antibes}, which completes the proof.
\end{IEEEproof}

A fundamental observation from Lemma~\ref{LemmaMarch23at16h09in2026BusToNice} is that, under the decomposition of the datasets in~\eqref{EqMay29at10h00in2026Antibes}, the weight assigned to a data point depends on whether it appears exclusively in one of the two datasets; or whether it is common to both datasets. More specifically, for all $i \inCountTwo$, the data points in $\vect{z}_{i,1}$, which are exclusively in the dataset~$\vect{z}_i$, exhibit weight $\frac{\alpha_i}{n_i\lambda_i}$ in~\eqref{EqMay29at10h02in2026Antibes}, and thus their contribution is exclusively controlled by $\alpha_i$ and the parameters of the Gibbs probability measure
$P_{\vect{\Theta} \mid\vect{Z} = \vect{z}_i}^{\autoparent{\mathsf{L}_{n_i},Q_i,\lambda_i}}$
in~\eqref{EqMay26at14h04in2026inAntibes} or~\eqref{EqMay26at14h05in2026inAntibes}. 
 In contrast, the data points in $\vect{z}_{1,2}$, which are common to the datasets $\vect{z}_1$ and $\vect{z}_2$, exhibit weight
$\frac{\alpha_1}{n_1\lambda_1}+\frac{\alpha_2}{n_2\lambda_2}$ in~\eqref{EqMay29at10h02in2026Antibes}. Hence, their contribution to the weighted empirical risk is jointly determined by the coefficients $\alpha_1$ and $\alpha_2$ of the normalized log-linear combination $S_2$ in~\eqref{EqMay26at14h34in2026Antibes} and by the regularization factors $\lambda_1$ and $\lambda_2$ of the Gibbs probability measures in~\eqref{EqMay26at14h04in2026inAntibes} and~\eqref{EqMay26at14h05in2026inAntibes}, respectively. 
This observation plays a crucial role in the remainder of this work. Essentially, the weights of data points that appear in both training datasets, namely $\vect{z}_1$ and $\vect{z}_2$ can be made equal to zero by choosing the parameters such that $\frac{\alpha_1}{n_1\lambda_1} + \frac{\alpha_2}{n_2\lambda_2} =0 $. Note that the weights $\frac{\alpha_i}{n_i\lambda_i}$ of data points that appear exclusively on one training dataset, with $i \inCountTwo$, cannot be made equal to zero. 
Interestingly, a zero-weight associated to a data point eliminates its effect on the training process, which resonates with the objective of exact machine unlearning, as discussed later in this work.

\subsection{Nested Gibbs Algorithms with Data-dependent Reference Measures}
\label{SubSecNestedGibbsAlg}

This subsection leverages the results on nested Gibbs probability measures presented in Section~\ref{SecNestedGibbsProbMeasures} in the context of supervised machine learning. 
In particular, in Section~\ref{SecNestedGibbsProbMeasures}, an equivalence between nesting Gibbs probability measures and log-linear combinations of Gibbs measures was established under some specific choice of parameters (Corollary~\ref{CorollaryApril16at15h20in2026Antibes}). Consequently, under such conditions, all comments in Section~\ref{SubSecNormalizedLogComGibbsAlg} also apply to nested Gibbs probability measures in supervised machine learning.
Nonetheless, for completeness, some of those comments are revisited when new relevant observations arise.
%As established in Section~\ref{SecMainResults}, nesting can also be viewed as a particular normalized log-linear combination. Consequently, the weighted empirical-risk interpretation developed in Section~\ref{SubSecNormalizedLogComGibbsAlg} applies directly to nested Gibbs probability measures.
%

In the following, the measure
$P_{\vect{\Theta} \mid\vect{Z} = \vect{z}_1}^{\autoparent{\mathsf{L}_{n_1},Q_1,\lambda_1}}$
in~\eqref{EqMay26at14h04in2026inAntibes} is nested within
$P_{\vect{\Theta} \mid\vect{Z} = \vect{z}_2}^{\autoparent{\mathsf{L}_{n_2},Q_2,\lambda_2}}$
in~\eqref{EqMay26at14h05in2026inAntibes} by choosing
$Q_2 =
P_{\vect{\Theta} \mid\vect{Z} = \vect{z}_1}^{\autoparent{\mathsf{L}_{n_1},Q_1,\lambda_1}}$.
This yields the nested Gibbs probability measure denoted by
\begin{IEEEeqnarray}{rCl}
\label{EqMay30at13h09in2026Antibes}
P^{\autoparent{\mathsf{L}_{n_2},P_{\vect{\Theta} \mid\vect{Z} = \vect{z}_1}^{\autoparent{\mathsf{L}_{n_1},Q_1,\lambda_1}},\lambda_2}}_{\vect{\Theta} \mid\vect{Z} = \vect{z}_2}
\in
\simplexabs{Q_1}{\set{M}},
\end{IEEEeqnarray}
which represents a nested Gibbs algorithm. 
A central observation is that the nested Gibbs probability measure $P^{\autoparent{\mathsf{L}_{n_2},P_{\vect{\Theta} \mid\vect{Z} = \vect{z}_1}^{\autoparent{\mathsf{L}_{n_1},Q_1,\lambda_1}},\lambda_2}}_{\vect{\Theta} \mid\vect{Z} = \vect{z}_2}$ in
\eqref{EqMay30at13h09in2026Antibes} depends explicitly on the training dataset $\vect{z}_2$, whereas its dependence on the training dataset $\vect{z}_1$ is carried by the data-dependent reference measure $P_{\vect{\Theta}\mid\vect{Z}=\vect{z}_1}^{(\mathsf{L}_{n_1},Q_1,\lambda_1)}$ in
\eqref{EqMay26at14h04in2026inAntibes}.
The following corollary, which follows from Lemma~\ref{LemmaMay26at15h45in2026Antibes} and Corollary~\ref{CorollaryApril16at15h20in2026Antibes}, reinforces this observation.
\begin{corollary}
\label{CorollaryMay30at13h11in2026Antibes}
The nested Gibbs probability measure $P^{\autoparent{\mathsf{L}_{n_2},P_{\vect{\Theta} \mid\vect{Z} = \vect{z}_1}^{\autoparent{\mathsf{L}_{n_1},Q_1,\lambda_1}},\lambda_2}}_{\vect{\Theta} \mid\vect{Z} = \vect{z}_2}$ in
\eqref{EqMay30at13h09in2026Antibes} satisfies, for all
$\vect{\theta}\in\supp Q_1$,
\begin{IEEEeqnarray}{rCl}
\label{EqMay30at13h12in2026Antibes}
\RND{P^{\autoparent{\mathsf{L}_{n_2},P_{\vect{\Theta} \mid\vect{Z} = \vect{z}_1}^{\autoparent{\mathsf{L}_{n_1},Q_1,\lambda_1}},\lambda_2}}_{\vect{\Theta} \mid\vect{Z} = \vect{z}_2}}{Q_1}(\vect{\theta})
& = &
\frac{
\exp\autoparent{-\widebar{\mathsf{L}}_{n_0}
\autoparent{\vect{\omega},\vect{z}_0,\vect{\theta}}}}
{\displaystyle \int
\exp\autoparent{-\widebar{\mathsf{L}}_{n_0}
\autoparent{\vect{\omega},\vect{z}_0,\vect{\nu}}}
\mathrm{d}Q_1(\vect{\nu})} = \RND{P_{\vect{\Theta}\mid\vect{Z} = \vect{z}_0}^{\autoparent{
\widebar{\mathsf{L}}_{n_0}\autoparent{\vect{\omega},\cdot,\cdot},Q_1,1}}}{Q_1}\autoparent{\vect{\theta}},
\end{IEEEeqnarray}
where the function $\widebar{\mathsf{L}}_{n_0}$ is defined in
\eqref{EqMarch22at12h49in2026Nice}; the dataset $\vect{z}_0$ is defined in
\eqref{EqMarch10at20h02in2026Sophia}; the vector
$\vect{\omega}=\autoparent{\omega_1,\omega_2,\ldots,\omega_{n_0}}$ satisfies,
for all $t\inCountK{n_0}$,
\begin{IEEEeqnarray}{rCl}
\label{EqMay30at13h13in2026Antibes}
\omega_t
& = &
\left\{
\begin{array}{cl}
\frac{1}{n_1\lambda_1}
& \text{if } t \leqslant n_1,\\[0.2cm]
\frac{1}{n_2\lambda_2}
& \text{if } n_1 < t \leqslant n_0;
\end{array}
\right.
\end{IEEEeqnarray}
and $P_{\vect{\Theta}\mid\vect{Z} = \vect{z}_0}^{\autoparent{
\widebar{\mathsf{L}}_{n_0}\autoparent{\vect{\omega},\cdot,\cdot},Q_1,1}}$ is an $\autoparent{
\widebar{\mathsf{L}}_{n_0}\autoparent{\vect{\omega},\cdot,\cdot},Q_1,1}$ Gibbs probability measure. 
\end{corollary}

Corollary~\ref{CorollaryMay30at13h11in2026Antibes} shows that the nested Gibbs probability measure $P^{\autoparent{\mathsf{L}_{n_2},P_{\vect{\Theta} \mid\vect{Z} = \vect{z}_1}^{\autoparent{\mathsf{L}_{n_1},Q_1,\lambda_1}},\lambda_2}}_{\vect{\Theta} \mid\vect{Z} = \vect{z}_2}$ in
\eqref{EqMay30at13h09in2026Antibes}  is itself a Gibbs probability measure whose objective function is a weighted empirical risk, with weights $\omega_1$, $\omega_2$, $\ldots$, $\omega_{n_0}$ that satisfy~\eqref{EqMay30at13h13in2026Antibes}, with respect to the aggregated dataset $\vect{z}_0$, i.e., the function  $\widebar{\mathsf{L}}_{n_0}(\vect{\omega}, \cdot,\cdot)$. In particular, the data points within $\vect{z}_1$ are assigned weight $\frac{1}{n_1\lambda_1}$, whereas those within $\vect{z}_2$ are assigned weight $\frac{1}{n_2\lambda_2}$. 
Hence, although the dependence on the dataset $\vect{z}_1$ is carried by the reference measure in the nested Gibbs algorithm, the contribution of $\vect{z}_1$ appears explicitly in the weighted empirical risk.
Beyond this important observation, Corollary~\ref{CorollaryMay30at13h11in2026Antibes} also implies the following identity.
\begin{corollary}
\label{CorollaryMay30at13h19in2026Antibes}
The probability measures $P^{\autoparent{\mathsf{L}_{n_2},P_{\vect{\Theta} \mid\vect{Z} = \vect{z}_1}^{\autoparent{\mathsf{L}_{n_1},Q_1,\lambda_1}},\lambda_2}}_{\vect{\Theta} \mid\vect{Z} = \vect{z}_2}$ in~\eqref{EqMay30at13h09in2026Antibes} and $P_{\vect{\Theta}\mid\vect{Z} = \vect{z}_0}^{\autoparent{
\widebar{\mathsf{L}}_{n_0}\autoparent{\vect{\omega},\cdot,\cdot},Q_1,1}}$ in~\eqref{EqMay30at13h12in2026Antibes} satisfy
\begin{IEEEeqnarray}{rCl}
\label{EqMay30at13h20in2026Antibes}
\KL{P^{\autoparent{\mathsf{L}_{n_2},P_{\vect{\Theta} \mid\vect{Z} = \vect{z}_1}^{\autoparent{\mathsf{L}_{n_1},Q_1,\lambda_1}},\lambda_2}}_{\vect{\Theta} \mid\vect{Z} = \vect{z}_2}}
{P_{\vect{\Theta}\mid\vect{Z} = \vect{z}_0}^{\autoparent{
\widebar{\mathsf{L}}_{n_0}\autoparent{\vect{\omega},\cdot,\cdot},Q_1,1}}}
& = & 0.
\end{IEEEeqnarray}
\end{corollary}

Corollary~\ref{CorollaryMay30at13h19in2026Antibes} and Lemma~\ref{LemmaSolution_SL} allow establishing connections between the nested Gibbs probability measure $P^{\autoparent{\mathsf{L}_{n_2},P_{\vect{\Theta} \mid\vect{Z} = \vect{z}_1}^{\autoparent{\mathsf{L}_{n_1},Q_1,\lambda_1}},\lambda_2}}_{\vect{\Theta} \mid\vect{Z} = \vect{z}_2}$ in
\eqref{EqMay30at13h09in2026Antibes} and two different optimization problems involving a relative entropy. 
On one hand, from Lemma~\ref{LemmaSolution_SL}, it follows that the nested Gibbs probability measure
$P^{\autoparent{\mathsf{L}_{n_2},P_{\vect{\Theta} \mid\vect{Z} = \vect{z}_1}^{\autoparent{\mathsf{L}_{n_1},Q_1,\lambda_1}},\lambda_2}}_{\vect{\Theta} \mid\vect{Z} = \vect{z}_2}$
in~\eqref{EqMay30at13h09in2026Antibes} is the unique solution, if it exists, to 
\begin{IEEEeqnarray}{rCl}
\label{EqMay30at13h17in2026Antibes}
\left\lbrace
\begin{array}{rcl}
\displaystyle\min_{P \in \triangle_{Q_1}\autoparent{\set{M}}}
\int
\EmpRisk{n_2}{\vect{z}_2}{\vect{\theta}}
\mathrm{d}P(\vect{\theta})
+
\lambda_2
\KL{P}{P_{\vect{\Theta} \mid\vect{Z} = \vect{z}_1}^{\autoparent{\mathsf{L}_{n_1},Q_1,\lambda_1}}}
& \mbox{if} & \lambda_2 > 0,
\\[0.2cm]
\displaystyle\max_{P \in \triangle_{Q_1}\autoparent{\set{M}}}
\int
\EmpRisk{n_2}{\vect{z}_2}{\vect{\theta}}
\mathrm{d}P(\vect{\theta})
+
\lambda_2
\KL{P}{P_{\vect{\Theta} \mid\vect{Z} = \vect{z}_1}^{\autoparent{\mathsf{L}_{n_1},Q_1,\lambda_1}}}
& \mbox{if} & \lambda_2 < 0.
\end{array}
\right.
\end{IEEEeqnarray}
On the other hand, Lemma~\ref{LemmaSolution_SL} shows that the measure $P_{\vect{\Theta}\mid\vect{Z} = \vect{z}_0}^{\autoparent{
\widebar{\mathsf{L}}_{n_0}\autoparent{\vect{\omega},\cdot,\cdot},Q_1,1}}$ in \eqref{EqMay30at13h12in2026Antibes} is the unique solution, if it exists, to  
\begin{IEEEeqnarray}{c}
\label{EqMay30at13h15in2026Antibes}
\min_{P \in \simplexabs{Q_1}{\set{M}}}
\int
\widebar{\mathsf{L}}_{n_0}
\autoparent{\vect{\omega},\vect{z}_0,\vect{\theta}}
\mathrm{d}P(\vect{\theta})
+
\KL{P}{Q_1},
\end{IEEEeqnarray}
where the vector of weights $\vect{\omega}$ satisfies
\eqref{EqMay30at13h13in2026Antibes}.
Hence, from the fact that the nested Gibbs probability measure $P^{\autoparent{\mathsf{L}_{n_2},P_{\vect{\Theta} \mid\vect{Z} = \vect{z}_1}^{\autoparent{\mathsf{L}_{n_1},Q_1,\lambda_1}},\lambda_2}}_{\vect{\Theta} \mid\vect{Z} = \vect{z}_2}$ in~\eqref{EqMay30at13h09in2026Antibes} is identical to the
$\autoparent{\widebar{\mathsf{L}}_{n_0}
\autoparent{\vect{\omega},\cdot,\cdot},Q_1,1}$-Gibbs probability measure $P_{\vect{\Theta}\mid\vect{Z} = \vect{z}_0}^{\autoparent{
\widebar{\mathsf{L}}_{n_0}\autoparent{\vect{\omega},\cdot,\cdot},Q_1,1}}$ in~\eqref{EqMay30at13h12in2026Antibes}, as established by Corollary~\ref{CorollaryMay30at13h19in2026Antibes}, it follows that
\begin{IEEEeqnarray}{rCl}
\label{EqMay30at13h18in2026Antibes}
\autobrace{P^{\autoparent{\mathsf{L}_{n_2},P_{\vect{\Theta} \mid\vect{Z} = \vect{z}_1}^{\autoparent{\mathsf{L}_{n_1},Q_1,\lambda_1}},\lambda_2}}_{\vect{\Theta} \mid\vect{Z} = \vect{z}_2}} & = &
\left\lbrace
\begin{array}{rcl}
\displaystyle\argmin_{P \in \triangle_{Q_1}\autoparent{\set{M}}}
\int
\EmpRisk{n_2}{\vect{z}_2}{\vect{\theta}}
\mathrm{d}P(\vect{\theta})
+
\lambda_2
\KL{P}{P_{\vect{\Theta} \mid\vect{Z} = \vect{z}_1}^{\autoparent{\mathsf{L}_{n_1},Q_1,\lambda_1}}}
& \mbox{if} & \lambda_2 > 0,
\\[0.2cm]
\displaystyle\argmax_{P \in \triangle_{Q_1}\autoparent{\set{M}}}
\int
\EmpRisk{n_2}{\vect{z}_2}{\vect{\theta}}
\mathrm{d}P(\vect{\theta})
+
\lambda_2
\KL{P}{P_{\vect{\Theta} \mid\vect{Z} = \vect{z}_1}^{\autoparent{\mathsf{L}_{n_1},Q_1,\lambda_1}}}
& \mbox{if} & \lambda_2 < 0.
\end{array}
\right. \\
&=&
\displaystyle\argmin_{P \in \triangle_{Q_1}\autoparent{\set{M}}}
\int
\widebar{\mathsf{L}}_{n_0}
\autoparent{\vect{\omega},\vect{z}_0,\vect{\theta}}
\mathrm{d}P(\vect{\theta})
+
\KL{P}{Q_1}
\\
&=&
\autobrace{
P_{\vect{\Theta}\mid\vect{Z} = \vect{z}_0}^{\autoparent{
\widebar{\mathsf{L}}_{n_0}\autoparent{\vect{\omega},\cdot,\cdot},Q_1,1}}
}.
\end{IEEEeqnarray}
which provides two equivalent interpretations of the nested Gibbs probability measure
$P^{\autoparent{\mathsf{L}_{n_2},P_{\vect{\Theta} \mid\vect{Z} = \vect{z}_1}^{\autoparent{\mathsf{L}_{n_1},Q_1,\lambda_1}},\lambda_2}}_{\vect{\Theta} \mid\vect{Z} = \vect{z}_2}$
in~\eqref{EqMay30at13h09in2026Antibes}. On one hand, it is the solution to an empirical-risk optimization with respect to the dataset $\vect{z}_2$, with a regularization by a relative entropy with respect to 
$P_{\vect{\Theta} \mid\vect{Z} = \vect{z}_1}^{\autoparent{\mathsf{L}_{n_1},Q_1,\lambda_1}}$. On the other hand, it is the solution to the minimization of a weighted empirical-risk with respect to the aggregated dataset $\vect{z}_0$, with a regularization by relative entropy with respect to the reference measure $Q_1$.

The following corollary, which follows from Corollary~\ref{CorollaryMay30at13h11in2026Antibes}, considers a particular choice of the regularization factors $\lambda_1$ and $\lambda_2$ that leads to a case in which all the data points in the aggregated dataset $\vect{z}_0$ exhibit the same weight in the weighted empirical risk in~\eqref{EqMay30at13h15in2026Antibes}.

\begin{corollary}
\label{CorollaryMay30at13h23in2026Antibes}
Consider a real $\lambda_0\in\reals\setminus\lbrace 0 \rbrace$. Assume that the regularization factors $\lambda_1$ and $\lambda_2$ satisfy
\begin{IEEEeqnarray}{rCl}
\label{EqMay30at13h24in2026Antibes}
\lambda_1
& = &
\frac{n_0\lambda_0}{n_1} 
\qquad \mbox{ and }  \qquad 
\lambda_2 = 
\frac{n_0\lambda_0}{n_2}.
\end{IEEEeqnarray}
Then, the nested Gibbs probability measure $P^{\autoparent{\mathsf{L}_{n_2},P_{\vect{\Theta} \mid\vect{Z} = \vect{z}_1}^{\autoparent{\mathsf{L}_{n_1},Q_1,\lambda_1}},\lambda_2}}_{\vect{\Theta} \mid\vect{Z} = \vect{z}_2}$ in
\eqref{EqMay30at13h09in2026Antibes} satisfies, for all
$\vect{\theta}\in\supp Q_1$,
\begin{IEEEeqnarray}{rCl}
\label{EqMay30at13h25in2026Antibes}
\RND{P^{\autoparent{\mathsf{L}_{n_2},P_{\vect{\Theta} \mid\vect{Z} = \vect{z}_1}^{\autoparent{\mathsf{L}_{n_1},Q_1,\lambda_1}},\lambda_2}}_{\vect{\Theta} \mid\vect{Z} = \vect{z}_2}}{Q_1}(\vect{\theta})
& = &
\frac{
\exp\autoparent{
-\frac{1}{\lambda_0}
\mathsf{L}_{n_0}\autoparent{\vect{z}_0,\vect{\theta}}}}
{
\int
\exp\autoparent{
-\frac{1}{\lambda_0}
\mathsf{L}_{n_0}\autoparent{\vect{z}_0,\vect{\nu}}}
\mathrm{d}Q_1(\vect{\nu})}
\\
\label{EqMay30at13h26in2026Antibes}
& = &
\RND{
P_{\vect{\Theta}\mid\vect{Z}=\vect{z}_0}^{\autoparent{\mathsf{L}_{n_0},Q_1,\lambda_0}}
}{Q_1}
(\vect{\theta}),
\end{IEEEeqnarray}
where the probability measure
$P_{\vect{\Theta}\mid\vect{Z}=\vect{z}_0}^{\autoparent{\mathsf{L}_{n_0},Q_1,\lambda_0}}$
is an $\autoparent{\mathsf{L}_{n_0},Q_1,\lambda_0}$-Gibbs probability measure,
with the function $\mathsf{L}_{n_0}$ defined in~\eqref{EqEmpiricalRiskLn}.
\end{corollary}

Corollary~\ref{CorollaryMay30at13h23in2026Antibes} shows a particular choice of the regularization factors $\lambda_1$ and $\lambda_2$ for which the nested Gibbs probability measure in~\eqref{EqMay30at13h09in2026Antibes} is identical to the $(\mathsf{L}_{n_0},Q_1,\lambda_0)$-Gibbs probability measure trained upon the aggregated dataset $\vect{z}_0$. Under this choice, the weighted empirical risk associated with the nested Gibbs probability measure satisfies for all $\vect{\theta} \in \set{M}$, 
\begin{IEEEeqnarray}{rCl}
\widebar{\mathsf{L}}_{n_0}\autoparent{\vect{\omega},\vect{z}_0,\vect{\theta}}
&=&
\frac{1}{\lambda_0}
\mathsf{L}_{n_0}\autoparent{\vect{z}_0,\vect{\theta}}.
\end{IEEEeqnarray}
Hence, the choice of $\lambda_1$ and $\lambda_2$ in~\eqref{EqMay30at13h24in2026Antibes} makes all the data points in the aggregated dataset $\vect{z}_0$ contribute uniformly to the empirical risk and yields the same Gibbs probability measure as that obtained by training directly upon $\vect{z}_0$ with reference measure $Q_1$ and regularization factor $\lambda_0$.
%
%Thus, this choice extends the training dataset from $\vect{z}_2$ to $\vect{z}_0$ while assigning uniform weight to all data points.

The following corollary, which follows from Corollary~\ref{CorollaryMay30at13h11in2026Antibes},  highlights the fact that when datasets share common data points, those common data points exhibit weights that are dependent on a larger number of parameters, when compared to data points that appear exclusively in one of the two datasets.

\begin{corollary}
\label{CorollaryMay30at13h21in2026Antibes}
Assume that the datasets $\vect{z}_1$ and $\vect{z}_2$ satisfy
\eqref{EqMay29at10h00in2026Antibes}, where all the data points common to
$\vect{z}_1$ and $\vect{z}_2$ are in $\vect{z}_{1,2}$.
Then, the nested Gibbs probability measure $P^{\autoparent{\mathsf{L}_{n_2},P_{\vect{\Theta} \mid\vect{Z} = \vect{z}_1}^{\autoparent{\mathsf{L}_{n_1},Q_1,\lambda_1}},\lambda_2}}_{\vect{\Theta} \mid\vect{Z} = \vect{z}_2}$ in
\eqref{EqMay30at13h09in2026Antibes} is such that the vector
$\vect{\omega}=\autoparent{\omega_1,\omega_2,\ldots,\omega_{n_0}}$ satisfies,
for all $t\inCountK{n_0}$,
\begin{IEEEeqnarray}{CCl}
\label{EqMay30at13h22in2026Antibes}
\omega_t
& = &
\left\{
\begin{array}{cl}
\frac{1}{n_1\lambda_1}
& \text{if } t \leqslant n_{1,1},\\[0.2cm]
\frac{1}{n_1\lambda_1}
+
\frac{1}{n_2\lambda_2}
& \text{if } n_{1,1} < t \leqslant n_1,\\[0.2cm]
\frac{1}{n_2\lambda_2}
& \text{if } n_1 < t \leqslant n_1+n_{2,1},\\[0.2cm]
0
& \text{if } n_1+n_{2,1} < t \leqslant n_0.
\end{array}
\right.
\end{IEEEeqnarray}
\end{corollary}

Corollary~\ref{CorollaryMay30at13h21in2026Antibes} highlights a fundamental observation. In particular, the weight associated to common data points in the  weighted empirical-risk optimization with relative-entropy regularization problem in~\eqref{EqMay30at13h15in2026Antibes} is determined by the regularization factors $\lambda_1$ and $\lambda_2$.  That is, $\frac{1}{n_1\lambda_1}+\frac{1}{n_2\lambda_2}$. Alternatively, the weights of data points that appear exclusively on dataset $i$, with $i \inCountTwo$, are determined by $\lambda_i$, that is, $\frac{1}{n_i\lambda_i}$.
%
%More specifically, the data points in $\vect{z}_{1,1}$, which appear exclusively in $\vect{z}_1$, exhibit weight $\frac{1}{n_1\lambda_1}$, whereas the data points in $\vect{z}_{2,1}$, which appear exclusively in $\vect{z}_2$, exhibit weight $\frac{1}{n_2\lambda_2}$. In contrast, the data points in $\vect{z}_{1,2}$, which are common to both datasets, exhibit weight $\frac{1}{n_1\lambda_1}+\frac{1}{n_2\lambda_2}$.
%
Hence, nesting causes the contributions associated with a data point appearing in both training datasets to accumulate in the resulting weighted empirical risk. 

%A key feature of nesting is that it allows: $(a)$ extending the training dataset through the incorporation of additional data points by means of the data-dependent reference measure; and $(b)$ modulating, through the choice of the regularization factors $\lambda_1$ and $\lambda_2$ in~\eqref{EqMay30at13h09in2026Antibes}, the contribution of specific data points to the weighted empirical-risk optimization with relative-entropy regularization problem in~\eqref{EqMay30at13h15in2026Antibes}.
%In particular, the contribution of data points appearing in both training datasets vanishes when
%\begin{IEEEeqnarray}{rCl}
%\frac{1}{n_1\lambda_1}
%+
%\frac{1}{n_2\lambda_2}
%&=&
%0.
%\end{IEEEeqnarray}
More importantly, the weight of data points appearing in both datasets can be set to satisfy
\begin{IEEEeqnarray}{rCl}
\frac{1}{n_1\lambda_1}
+
\frac{1}{n_2\lambda_2}
&=&
0,
\end{IEEEeqnarray}
whereas the weights of data points appearing exclusively on one dataset $\vect{z}_i$ satisfy $\frac{1}{n_i\lambda_i} \neq 0$.
Thus, the regularization factors can be chosen not only to modulate the contribution of specific data points, but also to cancel the contribution of data points common to both datasets. This property is leveraged in the following section in the context of exact machine unlearning.

\section{Exact Machine Unlearning}
\label{SecUnlearning}
\label{SecMachineUnlearning}

This section leverages the properties of nested Gibbs probability measures and normalized log-linear combinations of Gibbs probability measures established in Section~\ref{SecMainResults}, together with their weighted empirical-risk interpretations in supervised machine learning developed in Section~\ref{SecApplicationstoSupervisedMachineLearning}, to study exact machine unlearning.
In particular, Section~\ref{SubSecNormalizedLogComGibbsAlg} shows that, for normalized log-linear combinations, the weights of the resulting weighted empirical risk are jointly determined by the coefficients of the combination and the regularization factors. Alternatively, Section~\ref{SubSecNestedGibbsAlg} shows that nesting Gibbs algorithms induces a weighted empirical risk whose weights are exclusively determined by the corresponding regularization factors.
%
%Order changed
%In particular, Section~\ref{SubSecNestedGibbsAlg} shows that nesting Gibbs algorithms induces a weighted empirical risk whose weights are determined by the corresponding regularization factors, whereas Section~\ref{SubSecNormalizedLogComGibbsAlg} shows that, for normalized log-linear combinations, these weights are jointly determined by the coefficients of the combination and the regularization factors.
%
In both cases, exact unlearning is achieved by choosing the corresponding parameters such that the contribution of the dataset to be unlearned is associated with a zero weight. %, which implies that it is eliminated from the resulting weighted empirical risk.

Consider the datasets~$\vect{z}_1\in\set{Z}^{n_1}$ and
$\vect{z}_2\in\set{Z}^{n_2}$ in~\eqref{EqApril1at7h46in2026Sophia}, and let
their aggregation be the dataset~$\vect{z}_0\in\set{Z}^{n_0}$
in~\eqref{EqMarch10at20h02in2026Sophia}, with
$n_0=n_1+n_2$.
Assume that the dataset~$\vect{z}_1$ is the dataset to be unlearned and that
the dataset~$\vect{z}_2$ is the dataset to be retained.
Consider also the $(\mathsf{L}_{n_0},Q,\lambda_0)$-Gibbs probability measure
\begin{IEEEeqnarray}{rCl}
\label{EqApril17at18h10in2026Sophia}
P^{(\mathsf{L}_{n_0},Q,\lambda_0)}_{\vect{\Theta}\mid\vect{Z}=\vect{z}_0}
&\in&
\simplexabs{Q}{\set{M}},
\end{IEEEeqnarray}
with $\lambda_0>0$, which represents a Gibbs algorithm trained upon the
aggregated dataset~$\vect{z}_0$.

The objective is to transform the
$(\mathsf{L}_{n_0},Q,\lambda_0)$-Gibbs probability measure
$P^{(\mathsf{L}_{n_0},Q,\lambda_0)}_{\vect{\Theta}\mid\vect{Z}=\vect{z}_0}$
into an $(\mathsf{L}_{n_2},Q,\lambda_2)$-Gibbs probability measure
\begin{IEEEeqnarray}{rCl}
\label{EqApril17at18h11in2026Sophia}
P^{(\mathsf{L}_{n_2},Q,\lambda_2)}_{\vect{\Theta}\mid\vect{Z}=\vect{z}_2}
&\in&
\simplexabs{Q}{\set{M}},
\end{IEEEeqnarray}
for some $\lambda_2>0$, which represents a Gibbs algorithm trained exclusively
upon the retained dataset~$\vect{z}_2$.
This is referred to as \emph{exact unlearning} and is defined for this
particular setting as follows.

\begin{definition}[Exact unlearning]
\label{DefApril17at18h12in2026Sophia}
Consider the algorithm represented by the
$(\mathsf{L}_{n_0},Q,\lambda_0)$-Gibbs probability measure
$P^{(\mathsf{L}_{n_0},Q,\lambda_0)}_{\vect{\Theta}\mid\vect{Z}=\vect{z}_0}$
in~\eqref{EqApril17at18h10in2026Sophia}, and the datasets~$\vect{z}_1$ and
$\vect{z}_2$ to be unlearned and retained, respectively.
An algorithm $P\in\simplexabs{Q}{\set{M}}$ is said to achieve exact unlearning
of~$\vect{z}_1$ from
$P^{(\mathsf{L}_{n_0},Q,\lambda_0)}_{\vect{\Theta}\mid\vect{Z}=\vect{z}_0}$
if, for all~$\vect{\theta}\in\supp Q$,
\begin{IEEEeqnarray}{rCl}
\label{EqApril17at18h13in2026Sophia}
\RND{P}
{P^{(\mathsf{L}_{n_2},Q,\lambda_2)}_{\vect{\Theta}\mid\vect{Z}=\vect{z}_2}}
\autoparent{\vect{\theta}}
&=&
1,
\end{IEEEeqnarray}
where the probability measure
$P^{(\mathsf{L}_{n_2},Q,\lambda_2)}_{\vect{\Theta}\mid\vect{Z}=\vect{z}_2}$
is defined in~\eqref{EqApril17at18h11in2026Sophia}, for some $\lambda_2>0$.
\end{definition}

The equality in~\eqref{EqApril17at18h13in2026Sophia} implies that the measures
$P$ and
$P^{(\mathsf{L}_{n_2},Q,\lambda_2)}_{\vect{\Theta}\mid\vect{Z}=\vect{z}_2}$
are identical. See~\cite[Theorem~3]{InriaRR9591}.
In the following, two mechanisms for achieving exact unlearning are studied.
The first relies on nested Gibbs algorithms, whereas the second relies on
normalized log-linear combinations of Gibbs algorithms.

\subsection{Exact Unlearning with Nested Gibbs Algorithms}
\label{SecExactUnlearningNested}

This subsection leverages the results on nested Gibbs algorithms developed in
Sections~\ref{SecNestedGibbsProbMeasures} and~\ref{SubSecNestedGibbsAlg} to study unlearning in both centralized and decentralized learning systems.
%
%The key observation is that using a Gibbs probability measure as reference
%measure induces a weighted empirical risk in which the contribution of the
%dataset used to construct that reference measure can be modulated through the
%regularization factors.
%%
%In particular, exact unlearning is obtained when the contribution associated
%with the dataset to be unlearned vanishes.

\vspace{2ex}
\subsubsection{Centralized Learning Systems}
\label{SubSecUnlearningCentralized}

This section shows that exact unlearning of $\vect{z}_1$ from $P^{(\mathsf{L}_{n_0},Q,\lambda_0)}_{\vect{\Theta}\mid\vect{Z}=\vect{z}_0}$
in~\eqref{EqApril17at18h10in2026Sophia} is achieved by solving the optimization problem 
\begin{IEEEeqnarray}{rCl}
\label{EqApril17at18h14in2026Sophia}
\max_{P\in\simplexabs{Q}{\set{M}}}
&&
\mathsf{R}_{\vect{z}_1}\autoparent{P}
+
\lambda_1
\KL{P}
{P^{(\mathsf{L}_{n_0},Q,\lambda_0)}_{\vect{\Theta}\mid\vect{Z}=\vect{z}_0}},
\end{IEEEeqnarray}
for some $\lambda_1<0$, where the functional
$\mathsf{R}_{\vect{z}_1}$ is defined in~\eqref{EqExpEmpRiskModels}.
This setting is associated to a central learning system, capable of using all the dataset $\vect{z}_0$ for training, that receives the request to unlearn a particular dataset $\vect{z}_1$. 
From Corollary~\ref{CorMarch17at15h18in2026Sophia}, the unique solution
to~\eqref{EqApril17at18h14in2026Sophia}, if it exists, is the nested Gibbs
probability measure
\begin{IEEEeqnarray}{rCl}
\label{EqApril17at18h15in2026Sophia}
P^{\autoparent{
\mathsf{L}_{n_1},
P^{(\mathsf{L}_{n_0},Q,\lambda_0)}_{\vect{\Theta}\mid\vect{Z}=\vect{z}_0},
\lambda_1
}}_{\vect{\Theta}\mid\vect{Z}=\vect{z}_1}
&\in&
\simplexabs{Q}{\set{M}}.
\end{IEEEeqnarray}

The following corollary of
Lemma~\ref{LemmaMarch23at16h09in2026BusToNice} makes explicit the reweighting
induced by the use of the Gibbs probability measure
$P^{(\mathsf{L}_{n_0},Q,\lambda_0)}_{\vect{\Theta}\mid\vect{Z}=\vect{z}_0}$
in~\eqref{EqApril17at18h10in2026Sophia} as reference measure.

\begin{corollary}
\label{LemmaApril17at18h16in2026Sophia}
The
$\autoparent{
\mathsf{L}_{n_1},
P^{(\mathsf{L}_{n_0},Q,\lambda_0)}_{\vect{\Theta}\mid\vect{Z}=\vect{z}_0},
\lambda_1
}$-Gibbs probability measure
$P^{\autoparent{
\mathsf{L}_{n_1},
P^{(\mathsf{L}_{n_0},Q,\lambda_0)}_{\vect{\Theta}\mid\vect{Z}=\vect{z}_0},
\lambda_1
}}_{\vect{\Theta}\mid\vect{Z}=\vect{z}_1}$
in~\eqref{EqApril17at18h15in2026Sophia}, with $\lambda_0>0$ and
$\lambda_1\in\reals\setminus\lbrace0\rbrace$, satisfies, for all
$\vect{\theta}\in\supp Q$,
\begin{IEEEeqnarray}{rCl}
\label{EqApril17at18h17in2026Sophia}
\RND{
P^{\autoparent{
\mathsf{L}_{n_1},
P^{(\mathsf{L}_{n_0},Q,\lambda_0)}_{\vect{\Theta}\mid\vect{Z}=\vect{z}_0},
\lambda_1
}}_{\vect{\Theta}\mid\vect{Z}=\vect{z}_1}
}{Q}
\autoparent{\vect{\theta}}
&=&
\frac{
\exp\autoparent{
-\autoparent{
\frac{1}{\lambda_1} +\frac{n_1}{n_0\lambda_0}}
\mathsf{L}_{n_1}\autoparent{\vect{z}_1,\vect{\theta}}-\frac{n_2}{n_0\lambda_0}\mathsf{L}_{n_2}\autoparent{\vect{z}_2,\vect{\theta}}}}{
\displaystyle
\int\exp\autoparent{-\autoparent{\frac{1}{\lambda_1}+\frac{n_1}{n_0\lambda_0}}\mathsf{L}_{n_1}\autoparent{\vect{z}_1,\vect{\nu}}
-
\frac{n_2}{n_0\lambda_0}
\mathsf{L}_{n_2}\autoparent{\vect{z}_2,\vect{\nu}}}\mathrm{d}Q(\vect{\nu})}.
\end{IEEEeqnarray}
\end{corollary}

Corollary~\ref{LemmaApril17at18h16in2026Sophia} shows that the use of the Gibbs
probability measure
$P^{(\mathsf{L}_{n_0},Q,\lambda_0)}_{\vect{\Theta}\mid\vect{Z}=\vect{z}_0}$
as reference measure induces a weighted combination of the empirical risks
associated with the datasets~$\vect{z}_1$ and~$\vect{z}_2$.
More precisely, the coefficient of the empirical risk induced by the dataset
to be unlearned is
$\frac{1}{\lambda_1}+\frac{n_1}{n_0\lambda_0}$, whereas the coefficient of the
empirical risk induced by the dataset to be retained is
$\frac{n_2}{n_0\lambda_0}$.
Hence, exact unlearning is obtained by choosing~$\lambda_1$ such that the first
coefficient is zero in \eqref{EqApril17at18h17in2026Sophia}.
The following corollary shows that a specific choice of $\lambda_1$ and $\lambda_2$ makes the~$\autoparent{\mathsf{L}_{n_1}, P_{\vect{\Theta}\mid\vect{Z}=\vect{z}_0}^{(\mathsf{L}_{n_0},Q,\lambda_0)},\lambda_1}$-Gibbs probability  measure in~\eqref{EqApril17at18h15in2026Sophia}  identical to the probability measure~$P^{(\mathsf{L}_{n_2}, Q,\lambda_2)}_{\vect{\Theta}\mid \vect{Z}=\vect{z}_2}$ in~\eqref{EqApril17at18h11in2026Sophia},  in the sense of \cite[Theorem~3]{InriaRR9591}.

\begin{corollary}
\label{CorollaryApril17at18h18in2026Sophia}
Assume that
\begin{IEEEeqnarray}{rCl}
\label{EqMarch10at20h16in2026Sophia}
\lambda_1
&=&
-\frac{n_0}{n_1}\lambda_0;
\qquad
\lambda_2
=
\frac{n_0}{n_2}\lambda_0,
\end{IEEEeqnarray}
for some~$\lambda_0>0$.
Then, the
$\autoparent{
\mathsf{L}_{n_1},
P_{\vect{\Theta}\mid\vect{Z}=\vect{z}_0}^{(\mathsf{L}_{n_0},Q,\lambda_0)},
\lambda_1
}$-Gibbs probability measure in~\eqref{EqApril17at18h15in2026Sophia}
and the $(\mathsf{L}_{n_2},Q,\lambda_2)$-Gibbs probability measure
in~\eqref{EqApril17at18h11in2026Sophia} satisfy, for all
$\vect{\theta}\in\supp Q$,
\begin{IEEEeqnarray}{rCl}
\label{EqMarch10at20h17in2026Sophia}
\RND{
P^{\autoparent{
\mathsf{L}_{n_1},
P_{\vect{\Theta}\mid\vect{Z}=\vect{z}_0}^{(\mathsf{L}_{n_0},Q,\lambda_0)},
\lambda_1
}}_{\vect{\Theta}\mid\vect{Z}=\vect{z}_1}
}{
P^{(\mathsf{L}_{n_2},Q,\lambda_2)}_{\vect{\Theta}\mid\vect{Z}=\vect{z}_2}
}
\autoparent{\vect{\theta}}
&=&
1.
\end{IEEEeqnarray}
\end{corollary}

%\begin{IEEEproof}
%The proof is presented in
%Appendix~\ref{AppProofOfTheoremApril17at18h18in2026Sophia}.
%\end{IEEEproof}

Corollary~\ref{CorollaryApril17at18h18in2026Sophia} shows that the probability
measure
$P_{\vect{\Theta}\mid\vect{Z}=\vect{z}_0}^{(\mathsf{L}_{n_0},Q,\lambda_0)}$
in~\eqref{EqApril17at18h10in2026Sophia} can be transformed into
$P^{(\mathsf{L}_{n_2},Q,\lambda_2)}_{\vect{\Theta}\mid\vect{Z}=\vect{z}_2}$
in~\eqref{EqApril17at18h11in2026Sophia} via the optimization problem
in~\eqref{EqApril17at18h14in2026Sophia}, achieving exact unlearning in the
sense of Definition~\ref{DefApril17at18h12in2026Sophia}.
An important observation is that only the dataset to be unlearned,
i.e.,~$\vect{z}_1$, is explicitly used in the optimization problem
in~\eqref{EqApril17at18h14in2026Sophia}.
The dependence on the dataset to be retained, i.e.,~$\vect{z}_2$, is carried by
the reference measure
$P_{\vect{\Theta}\mid\vect{Z}=\vect{z}_0}^{(\mathsf{L}_{n_0},Q,\lambda_0)}$,
which represents the Gibbs algorithm from which~$\vect{z}_1$ is to be
unlearned.
Hence, unlearning~$\vect{z}_1$ is achieved by training a new Gibbs algorithm
upon the dataset to be unlearned while using the original Gibbs algorithm $P_{\vect{\Theta}\mid\vect{Z}=\vect{z}_0}^{(\mathsf{L}_{n_0},Q,\lambda_0)}$ as
reference measure.
Under the scaling in~\eqref{EqMarch10at20h16in2026Sophia}, the resulting
nested Gibbs probability measure coincides with the Gibbs probability measure
that would have been obtained by training exclusively upon the retained
dataset~$\vect{z}_2$.

Another consequence of
Corollary~\ref{CorollaryApril17at18h18in2026Sophia}, together with the uniqueness
of the solutions implied by Lemma~\ref{LemmaSolution_SL}, is
\begin{IEEEeqnarray}{rCl}
\nonumber
&&
\argmax_{P \in \simplexabs{Q}{\set{M}}}
\mathsf{R}_{\vect{z}_1}\autoparent{P}
-
\frac{n_0}{n_1}\lambda_0
\KL{
P
}{
P_{\vect{\Theta}\mid\vect{Z}=\vect{z}_0}^{(\mathsf{L}_{n_0},Q,\lambda_0)}
}
\\
\label{EqApril17at18h30in2026Sophia}
&=&
\argmin_{P \in \simplexabs{Q}{\set{M}}}
\mathsf{R}_{\vect{z}_2}\autoparent{P}
+
\frac{n_0}{n_2}\lambda_0
\KL{P}{Q}.
\end{IEEEeqnarray}

From the perspective of
Lemma~\ref{LemmaMarch4at14h48in2026Sophia},
when $Q$ is a probability measure, the equality
in~\eqref{EqApril17at18h30in2026Sophia}
admits a geometric representation in terms of the
relative-entropy neighborhoods in~\eqref{EqKLNeighborhood}.
Under the assumption in~\eqref{EqMarch10at20h16in2026Sophia},
consider the values 
\begin{IEEEeqnarray}{rCl}
\label{EqUnlearningGeometryGammaOne}
\gamma_1
&\triangleq&
\KL{
P_{\vect{\Theta}\mid\vect{Z}=\vect{z}_0}^{
(\mathsf{L}_{n_0},Q,\lambda_0)}
}{Q},
\\
\label{EqUnlearningGeometryGammaTwo}
\gamma_2
&\triangleq&
\KL{
P^{\autoparent{
\mathsf{L}_{n_1},
P_{\vect{\Theta}\mid\vect{Z}=\vect{z}_0}^{
(\mathsf{L}_{n_0},Q,\lambda_0)},
\lambda_1
}}_{\vect{\Theta}\mid\vect{Z}=\vect{z}_1}
}{
P_{\vect{\Theta}\mid\vect{Z}=\vect{z}_0}^{
(\mathsf{L}_{n_0},Q,\lambda_0)}
},
\\
\label{EqUnlearningGeometryEta}
\eta
&\triangleq&
\KL{
P_{\vect{\Theta}\mid\vect{Z}=\vect{z}_2}^{
(\mathsf{L}_{n_2},Q,\lambda_2)}
}{Q}.
\end{IEEEeqnarray}
Consider the sets~$\set{G}_{Q,\gamma_1}$,  $\set{G}_{Q,\eta}$ and~$\set{G}_{
P_{\vect{\Theta}\mid\vect{Z}=\vect{z}_0}^{
(\mathsf{L}_{n_0},Q,\lambda_0)},\gamma_2}$ defined as in~\eqref{EqKLNeighborhood}. 
Using this notation, it follows from Corollary~\ref{CorollaryApril17at18h18in2026Sophia} that 
the nested Gibbs probability measure
in~\eqref{EqApril17at18h15in2026Sophia}
is identical to the Gibbs probability measure
in~\eqref{EqApril17at18h11in2026Sophia}.
Hence, the equality in~\eqref{EqApril17at18h30in2026Sophia}
implies that the probability measure that minimizes
the expected empirical risk with respect to the dataset
to be retained within $\set{G}_{Q,\eta}$
is identical to the probability measure that maximizes
the expected empirical risk with respect to the dataset
to be unlearned within
$\set{G}_{
P_{\vect{\Theta}\mid\vect{Z}=\vect{z}_0}^{
(\mathsf{L}_{n_0},Q,\lambda_0)},\gamma_2}$.
More precisely,
\begin{IEEEeqnarray}{rCl}
\label{EqUnlearningGeometryOptimizers}
\left\lbrace
P_{\vect{\Theta}\mid\vect{Z}=\vect{z}_2}^{
(\mathsf{L}_{n_2},Q,\lambda_2)}
\right\rbrace
&=&
\displaystyle\arg\min_{P\in\set{G}_{Q,\eta}}
\mathsf{R}_{\vect{z}_2}\autoparent{P}
\\
&=&
\displaystyle\arg\max_{
P\in\set{G}_{
P_{\vect{\Theta}\mid\vect{Z}=\vect{z}_0}^{
(\mathsf{L}_{n_0},Q,\lambda_0)},\gamma_2}}
\mathsf{R}_{\vect{z}_1}\autoparent{P}
\\
&=&
\left\lbrace
P^{\autoparent{
\mathsf{L}_{n_1},
P_{\vect{\Theta}\mid\vect{Z}=\vect{z}_0}^{
(\mathsf{L}_{n_0},Q,\lambda_0)},
\lambda_1
}}_{\vect{\Theta}\mid\vect{Z}=\vect{z}_1}
\right\rbrace.
\end{IEEEeqnarray}

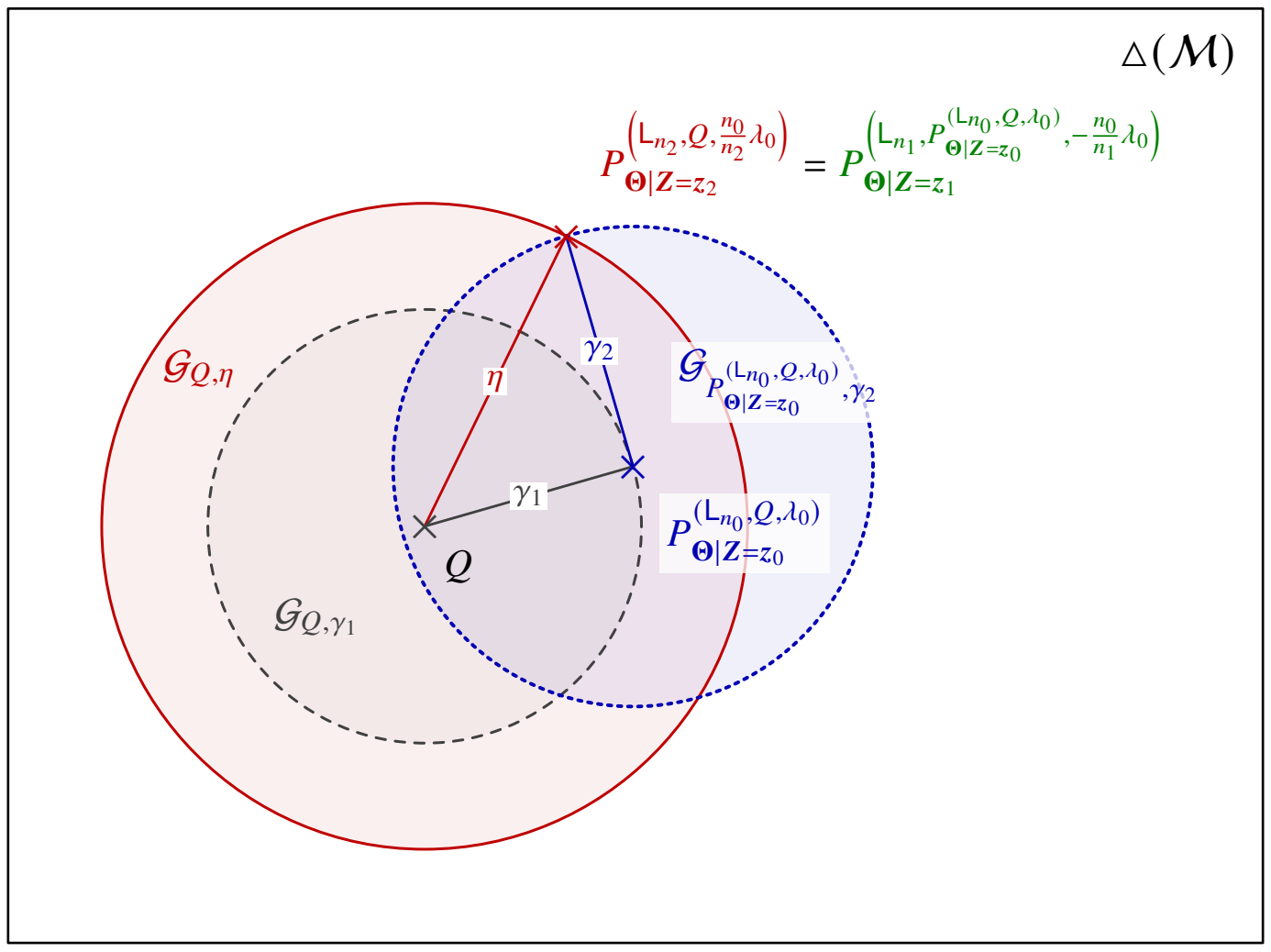
\begin{figure}[t]
\centering
\resizebox{0.5\textwidth}{!}{%
\input{\Figurespath/geometric_interpretation.tikz}}
\caption{Geometric representation of exact unlearning
under the assumption in~\eqref{EqMarch10at20h16in2026Sophia},
with $\gamma_1<\eta$.
The neighborhoods
$\set{G}_{Q,\gamma_1}$,
$\set{G}_{Q,\eta}$, and
$\set{G}_{
P_{\vect{\Theta}\mid\vect{Z}=\vect{z}_0}^{
(\mathsf{L}_{n_0},Q,\lambda_0)},\gamma_2}$
are defined as in~\eqref{EqKLNeighborhood},
with $\gamma_1$, $\gamma_2$, and $\eta$
in~\eqref{EqUnlearningGeometryGammaOne},
\eqref{EqUnlearningGeometryGammaTwo},
and~\eqref{EqUnlearningGeometryEta}, respectively.
The Gibbs probability measure
in~\eqref{EqApril17at18h11in2026Sophia}
and the nested Gibbs probability measure
in~\eqref{EqApril17at18h15in2026Sophia}
coincide and satisfy
\eqref{EqUnlearningGeometryOptimizers}.}
\label{FigGeomNestedGibbs}
\end{figure}

Figure~\ref{FigGeomNestedGibbs} provides a geometric
representation of~\eqref{EqUnlearningGeometryOptimizers}.
The figure shows the sets 
$\set{G}_{Q,\gamma_1}$, $\set{G}_{Q,\eta}$, and
$\set{G}_{
P_{\vect{\Theta}\mid\vect{Z}=\vect{z}_0}^{
(\mathsf{L}_{n_0},Q,\lambda_0)},\gamma_2}$.
The original Gibbs probability measure
$P_{\vect{\Theta}\mid\vect{Z}=\vect{z}_0}^{
(\mathsf{L}_{n_0},Q,\lambda_0)}$
in~\eqref{EqApril17at18h10in2026Sophia}
is the minimizer of the expected empirical risk
with respect to $\vect{z}_0$ within
$\set{G}_{Q,\gamma_1}$ and is the reference measure used to form the set $\set{G}_{
P_{\vect{\Theta}\mid\vect{Z}=\vect{z}_0}^{
(\mathsf{L}_{n_0},Q,\lambda_0)},\gamma_2}$.
The nested Gibbs probability measure
in~\eqref{EqApril17at18h15in2026Sophia}
and the Gibbs probability measure
in~\eqref{EqApril17at18h11in2026Sophia}
are represented by the same point, taking into account that they are identical
under the assumption in~\eqref{EqMarch10at20h16in2026Sophia}.
Note that in this representation, it has been 
assumed that $\gamma_1<\eta$, which justifies the inclusion
$\set{G}_{Q,\gamma_1}\subseteq\set{G}_{Q,\eta}$.

Finally, the parameter~$\lambda_0>0$ remains free and can be further optimized,
for instance with the objective of improving the generalization properties of
the resulting Gibbs algorithm.
See, for instance,
\cite{perlaza2024generalization,aminian2024information,zouJSAIT2024,aminian2021exact,xu2017information}
and the references therein.

\vspace{2ex}
\subsubsection{Decentralized Learning Systems}
\label{SecDecentralizedExactUnlearning}

The centralized exact-unlearning mechanism in the preceding subsection
requires the availability of a Gibbs probability measure trained upon the
entire dataset from which a subset is to be unlearned.
In a decentralized learning system, the local datasets are distributed among
the clients and such a probability measure is not obtained by directly
training upon their aggregation.
The purpose of this subsection is first to show that the nested Gibbs
construction developed in Section~\ref{SubSecNestedGibbsAlg} can produce the
same Gibbs probability measure as in a centralized system  through a peer-to-peer communication protocol
without explicitly aggregating the local datasets.
Once this centralized-equivalent Gibbs probability measure is available, the
exact-unlearning mechanism established in
Corollary~\ref{CorollaryApril17at18h18in2026Sophia} can be applied to the dataset
of any client to be unlearned.

Consider a decentralized learning system in which $K\in\ints$ clients collaboratively tune their learning algorithms by communicating with each other.
Assume that all clients use the same model space~$\set{M}\subseteq\reals^d$, for some $d\in\ints$, and the same data space~$\set{Z}\triangleq\set{X}\times\set{Y}$.
For all $k\inCountK{K}$, let the training dataset available at client~$k$ be denoted by
\begin{IEEEeqnarray}{rCl}
\label{EqDatasetK}
\vect{z}_k
&\triangleq&
\autoparent{(x_{k,1},y_{k,1}),(x_{k,2},y_{k,2}),\ldots,(x_{k,n_k},y_{k,n_k})}
\in \set{Z}^{n_k},
\end{IEEEeqnarray}
for some $n_k\in\ints$.
The aggregation of all local datasets is denoted by
\begin{IEEEeqnarray}{rCl}
\label{EqDatasetZero}
\vect{z}_0
&\triangleq&
\autoparent{\vect{z}_1,\vect{z}_2,\ldots,\vect{z}_K}
\\
\label{EqDatasetZeroExplicit}
&=&
\autoparent{(x_{0,1},y_{0,1}),(x_{0,2},y_{0,2}),\ldots,(x_{0,n_0},y_{0,n_0})}
\in \set{Z}^{n_0},
\end{IEEEeqnarray}
where
\begin{IEEEeqnarray}{rCl}
\label{EqnZero}
n_0
&\triangleq&
\sum_{k=1}^{K} n_k.
\end{IEEEeqnarray}

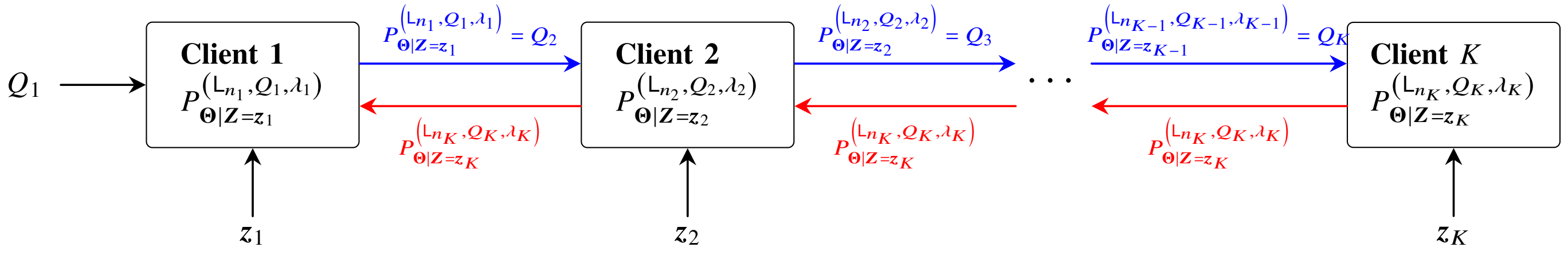
\begin{figure*}[t]
\centering
\resizebox{\textwidth}{!}{%
  \input{\Figurespath/Nested_Decentralized/Nested_structure.tikz}}
\caption{Nested structure induced by the forward--backward peer-to-peer communication protocol: for all $k \in \{1,2,\ldots,K-1\}$, the Gibbs probability measure produced by client~$k$ becomes the reference measure used by client~$k+1$.}
\label{FigNestedStructure}
\end{figure*}

Consider, for all $k \inCountK{K}$, the $(\mathsf{L}_{n_k},Q_{k},\lambda_k)$-Gibbs probability measure
\begin{subequations}\label{EqJan20at16h44in2026Sophia}
\begin{IEEEeqnarray}{rcl}
P^{\autoparent{\mathsf{L}_{n_k},Q_k,\lambda_k}}_{\vect{\Theta} \mid \vect{Z} = \vect{z}_k}
&\in&
\simplex{\set{M}},
\end{IEEEeqnarray}
with $\lambda_k>0$, which represents the Gibbs algorithm of client~$k$ trained upon its local dataset~$\vect{z}_k$ in~\eqref{EqDatasetK}. The corresponding reference measure $Q_k$ is specified recursively as
\begin{IEEEeqnarray}{rcl}
\label{EqDec27at10h57in2025inMadrid}
Q_k &=& \left\lbrace
\begin{array}{cc}
Q_1, & \text{if } k = 1,\\
P^{\autoparent{\mathsf{L}_{n_{k-1}}, Q_{k-1},\lambda_{k-1}}}_{\vect{\Theta} \mid \vect{Z} = \vect{z}_{k-1}}, & \text{if } k \geqslant 2,
\end{array}
\right.
\end{IEEEeqnarray}
which implies that,  for all $k \in \{1,2,\ldots,K-1\}$, the Gibbs probability measure produced by client~$k$ becomes the reference measure used by client~$k+1$.
More specifically, from Definition~\ref{DefGeneralGibbs}, for all $k \inCountK{K}$ and for all $\vect{\theta}\in\supp Q_1$, it follows that
\begin{IEEEeqnarray}{rCl}
\label{EqJan20at16h45in2026Sophia}
\RND{P^{\autoparent{\mathsf{L}_{n_k},Q_k,\lambda_k}}_{\vect{\Theta}\mid\vect{Z}=\vect{z}_k}}{Q_k}\autoparent{\vect{\theta}}
&=&
\frac{\exp\autoparent{-\frac{1}{\lambda_k}\mathsf{L}_{n_k}\autoparent{\vect{z}_k,\vect{\theta}}}}
{\displaystyle \int \exp\autoparent{-\frac{1}{\lambda_k}\mathsf{L}_{n_k}\autoparent{\vect{z}_k,\vect{\nu}}} \mathrm{d}Q_k\autoparent{\vect{\nu}}},
\end{IEEEeqnarray}
\end{subequations}
which shows that the $\autoparent{\mathsf{L}_{n_k},Q_k,\lambda_k}$-Gibbs probability measure in~\eqref{EqJan20at16h44in2026Sophia} is a nested Gibbs probability measure in the sense of Definition~\ref{DefNestedGibbsMeasures}.
From Lemma~\ref{LemmaSolution_SL}, for all $k\inCountK{K}$, such a Gibbs probability measure is the unique solution to the optimization problem
\begin{IEEEeqnarray}{rCl}
\label{EqJul25at9h55in2025NiceA}
\min_{P\in\simplexabs{Q_k}{\set{M}}}
&&
\mathsf{R}_{\vect{z}_k}\autoparent{P}
+
\lambda_k\KL{P}{Q_k},
\end{IEEEeqnarray}
where the functional $\mathsf{R}_{\vect{z}_k}$ is defined in~\eqref{EqExpEmpRiskModels}.
Equivalently, from Lemma~\ref{LemmaMarch4at14h48in2026Sophia}, it is also the unique solution to
\begin{IEEEeqnarray}{rCl}
\nonumber
\min_{P\in\simplexabs{Q_k}{\set{M}}}
&&
\mathsf{R}_{\vect{z}_k}\autoparent{P}
\\
\label{EqJan20at16h46in2026Sophia}
\text{s.t.}
&& \quad 
\KL{P}{Q_k}\le \KL{P^{\autoparent{\mathsf{L}_{n_k},Q_k,\lambda_k}}_{\vect{\Theta}\mid\vect{Z}=\vect{z}_k}}{Q_k}.
\end{IEEEeqnarray}
Such a nested Gibbs probability measure is obtained through the forward--backward peer-to-peer communication protocol depicted in Figure~\ref{FigNestedStructure} and described hereunder.

\paragraph{Communication protocol}
\label{SubsecComm}

Figure~\ref{FigNestedStructure} depicts the forward--backward peer-to-peer communication protocol used in this section.
In the forward direction (blue arrows), for all $k \in \{1,2,\ldots,K-1\}$, client~$k$ transmits to client~$k+1$ the $\autoparent{\mathsf{L}_{n_k},Q_k,\lambda_k}$-Gibbs probability measure
$P^{\autoparent{\mathsf{L}_{n_k},Q_k,\lambda_k}}_{\vect{\Theta}\mid \vect{Z} = \vect{z}_{k}}$
in~\eqref{EqJan20at16h44in2026Sophia}, obtained as the solution to the optimization problem in~\eqref{EqJul25at9h55in2025NiceA}.
Client~$k+1$ adopts this transmitted measure as its reference measure~$Q_{k+1}$ in~\eqref{EqJul25at9h55in2025NiceA}.
This choice induces a nested structure of the reference measure, as
$Q_{k+1} = P^{\autoparent{\mathsf{L}_{n_k},Q_k,\lambda_k}}_{\vect{\Theta}\mid \vect{Z} = \vect{z}_{k}}$,
for all $k \in \{1,2,\ldots,K-1\}$.
This is formalized later in Corollary~\ref{CorollaryDec21at18h00in2025inMadrid}.
The backward direction (red arrows) disseminates the final Gibbs probability measure.
More specifically, once client~$K$ has computed its $\autoparent{\mathsf{L}_{n_K},Q_K,\lambda_K}$-Gibbs probability measure
$P^{\autoparent{\mathsf{L}_{n_K},Q_K,\lambda_K}}_{\vect{\Theta}\mid \vect{Z} = \vect{z}_{K}}$
in~\eqref{EqJan20at16h44in2026Sophia}, this measure is transmitted back along the chain, from client~$K$ to client~$K-1$, then from client~$K-1$ to client~$K-2$, and so on until client~$1$.
This backward transmission provides all clients with access to the same final Gibbs algorithm.

\paragraph{Nested algorithms}

The following corollary is a direct consequence of Theorem~\ref{TheoremMay19at113h22in2026Antibes}, applied to the nested structure induced by~\eqref{EqDec27at10h57in2025inMadrid}.

\begin{corollary}
\label{CorollaryDec21at18h00in2025inMadrid}
For all $k\inCountK{K}$, let the probability measure
$P^{\autoparent{\mathsf{L}_{n_k},Q_k,\lambda_k}}_{\vect{\Theta}\mid\vect{Z}=\vect{z}_k}$
be defined as in~\eqref{EqJan20at16h44in2026Sophia}, with the reference measures satisfying~\eqref{EqDec27at10h57in2025inMadrid}. Then, for all $\vect{\theta}\in\supp Q_1$,
\begin{IEEEeqnarray}{rCl}
\label{EqDec30at11h25in2025Antibes}
\RND{P^{\autoparent{\mathsf{L}_{n_k},Q_k,\lambda_k}}_{\vect{\Theta}\mid\vect{Z}=\vect{z}_k}}{Q_1}\autoparent{\vect{\theta}}
&=&
\frac{\exp\autoparent{-\displaystyle\sum_{j=1}^{k}\frac{1}{\lambda_j}\mathsf{L}_{n_j}\autoparent{\vect{z}_j,\vect{\theta}}}}
{\displaystyle \int
\exp\autoparent{-\displaystyle\sum_{j=1}^{k}\frac{1}{\lambda_j}\mathsf{L}_{n_j}\autoparent{\vect{z}_j,\vect{\nu}}}
 \mathrm{d}Q_1(\vect{\nu})}.
\end{IEEEeqnarray}
\end{corollary}

The choice of reference measure $Q_k$ in~\eqref{EqDec27at10h57in2025inMadrid} induces a nested structure. Under this structure, the training performed by client~$k$ uses only its local dataset, while the influence of the previous clients' datasets is carried through the reference measure~$Q_k$.
Corollary~\ref{CorollaryDec21at18h00in2025inMadrid} makes this dependence explicit: although the reference measures are recursively defined, the Radon--Nikodym derivative in~\eqref{EqDec30at11h25in2025Antibes} depends on the datasets $\vect{z}_1,\vect{z}_2,\ldots,\vect{z}_k$ through the weighted sum of their empirical risks, with coefficients $\frac{1}{\lambda_1},\frac{1}{\lambda_2},\ldots,\frac{1}{\lambda_k}$, respectively.
Moreover, from Theorem~\ref{TheoremMay19at13h57in2026Sophia}, the probability measure
$P^{\autoparent{\mathsf{L}_{n_k},Q_k,\lambda_k}}_{\vect{\Theta}\mid\vect{Z}=\vect{z}_k}$ in~\eqref{EqJan20at16h44in2026Sophia}
is the unique solution, if it exists, to the following optimization problem:
\begin{IEEEeqnarray}{rCl}
\label{EqDec28at23h24in2025FlyingToNice}
\min_{P \in \simplexabs{Q_1}{\set{M}}}
&&
\int
\autoparent{\sum_{j=1}^{k}\frac{1}{\lambda_j}
\EmpRisk{n_j}{\vect{z}_j}{\vect{\theta}}}
\mathrm{d}P\autoparent{\vect{\theta}}
+
\KL{P}{Q_1}.
\end{IEEEeqnarray}

Hence, the probability measure
$P^{\autoparent{\mathsf{L}_{n_k},Q_k,\lambda_k}}_{\vect{\Theta}\mid\vect{Z}=\vect{z}_k}$ in~\eqref{EqJan20at16h44in2026Sophia}
admits two equivalent representations. On one hand, it is obtained by training client~$k$ exclusively upon its local dataset~$\vect{z}_k$, using the Gibbs probability measure produced by the preceding client as reference measure. On the other hand, it is the solution to the weighted empirical-risk optimization in~\eqref{EqDec28at23h24in2025FlyingToNice}, which depends explicitly on the datasets of all clients from $1$ to $k$ and uses the initial reference measure~$Q_1$.

The following corollary formalizes this observation.

\begin{corollary}
\label{CorJanuary1at20h49in2026HomeNice}
Under the assumption that, for all $k \inCountK{K}$, the measure $Q_k$ satisfies~\eqref{EqDec27at10h57in2025inMadrid}, the solutions to the optimization problems in~\eqref{EqJul25at9h55in2025NiceA} and~\eqref{EqDec28at23h24in2025FlyingToNice} are unique and coincide with the $\autoparent{\mathsf{L}_{n_k},Q_k,\lambda_k}$-Gibbs probability measure
$P^{\autoparent{\mathsf{L}_{n_k},Q_k,\lambda_k}}_{\vect{\Theta}\mid\vect{Z}=\vect{z}_k}$
in~\eqref{EqDec30at11h25in2025Antibes}.
\end{corollary}

Given $k > 1$, the optimization problem in~\eqref{EqJul25at9h55in2025NiceA} depends exclusively on the local training dataset~$\vect{z}_k$. While the reference measure~$Q_k$ in~\eqref{EqDec27at10h57in2025inMadrid} depends on the training datasets of the $k-1$ previous clients, the dependence is not direct. Solving~\eqref{EqJul25at9h55in2025NiceA} requires access to the probability measure $Q_k \in \simplex{\set{M}}$, but not to the training datasets of all previous clients. This is because, for fixed training datasets, $Q_k$ is simply a probability measure on the model space.
In contrast, the optimization problem in~\eqref{EqDec28at23h24in2025FlyingToNice} depends on the training datasets of client~$j$, for all $j \inCountK{k}$, while the reference measure $Q_1$ does not depend on any training dataset.
The fact that problems~\eqref{EqJul25at9h55in2025NiceA} and~\eqref{EqDec28at23h24in2025FlyingToNice} share the same solution unveils an important observation: providing client~$k$ with a reference measure $Q_k$ of the form in~\eqref{EqDec27at10h57in2025inMadrid} reproduces the effect of having access to the training datasets of the $k-1$ previous clients, without transmitting those datasets.

\paragraph{Centralized-performance guarantee}

An important observation is that a strategic choice of $\lambda_1$, $\lambda_2$, $\ldots$, $\lambda_K$ in~\eqref{EqDec30at11h25in2025Antibes} can lead to the same Gibbs probability measure as in a setting in which the training datasets of all clients are available at all clients.
This describes a decentralized system whose distributed nature does not prevent achieving the same Gibbs algorithm that one would have obtained if all the training data were available at all clients.
The following corollary formalizes this observation.

\begin{corollary}
\label{CorollaryDec30at21h00in2025inAntibes}
Assume that, for all $k \inCountK{K}$,
\begin{IEEEeqnarray}{rCl}
\label{EqJanuary1at15h16in2026atHomeNice}
\lambda_k
&=&
\frac{n_0\lambda_0}{n_k},
\end{IEEEeqnarray}
for some $\lambda_0 > 0$.
Consider the measure $Q_K$ satisfying~\eqref{EqDec27at10h57in2025inMadrid}.
Then, for all $\vect{\theta}\in\supp Q_1$, it follows that
\begin{IEEEeqnarray}{rCl}
\label{EqJanuary1at16h30in2026atHomeNice}
\RND{P^{\autoparent{\mathsf{L}_{n_K},Q_K,\lambda_K}}_{\vect{\Theta}\mid \vect{Z} = \vect{z}_{K}}}{P^{(\mathsf{L}_{n_0},Q_1,\lambda_0)}_{\vect{\Theta}\mid \vect{Z} = \vect{z}_{0}}}\autoparent{\vect{\theta}}
&=&
1,
\end{IEEEeqnarray}
where the probability measure $P^{\autoparent{\mathsf{L}_{n_K},Q_K,\lambda_K}}_{\vect{\Theta}\mid \vect{Z} = \vect{z}_{K}}$ is defined in~\eqref{EqDec30at11h25in2025Antibes}, and the measure $P^{(\mathsf{L}_{n_0},Q_1,\lambda_0)}_{\vect{\Theta}\mid \vect{Z} = \vect{z}_{0}}$ satisfies, for all $\vect{\theta}\in\supp Q_1$,
\begin{IEEEeqnarray}{rCl}
\label{EqDec30at15h43in2025Antibes}
\RND{P^{(\mathsf{L}_{n_0},Q_1,\lambda_0)}_{\vect{\Theta}\mid \vect{Z} = \vect{z}_{0}}}{Q_1}\autoparent{\vect{\theta}}
&=&
\frac{\exp\autoparent{-\frac{1}{\lambda_0}\mathsf{L}_{n_0}\autoparent{\vect{z}_0,\vect{\theta}}}}
{\displaystyle \int \exp\autoparent{-\frac{1}{\lambda_0}\mathsf{L}_{n_0}\autoparent{\vect{z}_0,\vect{\nu}}}\mathrm{d} Q_1(\vect{\nu})}.
\end{IEEEeqnarray}
\end{corollary}
%\begin{IEEEproof}
%The proof is presented in Appendix~\ref{AppProofOfEqDec30at21h00in2025inAntibes}.
%\end{IEEEproof}

Corollary~\ref{CorollaryDec30at21h00in2025inAntibes} establishes the
centralized-performance guarantee required for exact unlearning: under the
choice in~\eqref{EqJanuary1at15h16in2026atHomeNice}, the last nested Gibbs
probability measure is identical to the Gibbs probability measure that would
have been obtained by training directly upon~$\vect{z}_0$.
The relevance of this identity is further highlighted by the following
observations.
Under the assumptions of Corollary~\ref{CorollaryDec30at21h00in2025inAntibes}, the equality in~\eqref{EqJanuary1at15h16in2026atHomeNice} together with~\cite[Lemma~4]{perlaza2023validation} allow rewriting the optimization problem in~\eqref{EqDec28at23h24in2025FlyingToNice} as
\begin{IEEEeqnarray}{rCl}
\label{EqJanuary1at15h20in2026atHomeNice}
\min_{P \in \simplexabs{Q_1}{\set{M}}}
&&
\int \mathsf{L}_{n_0}\autoparent{\vect{z}_0,\vect{\theta}}
 \mathrm{d}P\autoparent{\vect{\theta}} + \lambda_0\KL{P}{Q_1},
\end{IEEEeqnarray}
which requires access to the training datasets of all clients.
Interestingly, from Lemma~\ref{LemmaSolution_SL}, it follows that the probability measure
$P^{(\mathsf{L}_{n_0},Q_1,\lambda_0)}_{\vect{\Theta}\mid \vect{Z} = \vect{z}_{0}}$
in~\eqref{EqDec30at15h43in2025Antibes} is the unique solution to~\eqref{EqJanuary1at15h20in2026atHomeNice}.

In the following, assume that $Q_1$ is a probability measure and that
the regularization factors satisfy
\eqref{EqJanuary1at15h16in2026atHomeNice}.
For all $k\inCountK{K}$, consider the values 
\begin{IEEEeqnarray}{rCl}
\label{EqSept22at19h45in2026}
\label{EqDecentralizedGeometryGamma}
\gamma_k
&\triangleq&
\KL{
P^{\autoparent{\mathsf{L}_{n_k},Q_k,\lambda_k}}_{
\vect{\Theta}\mid\vect{Z}=\vect{z}_k}
}{Q_k}
<+\infty,
\end{IEEEeqnarray}
and the corresponding neighborhoods
\begin{IEEEeqnarray}{rCl}
\label{EqJan20at16h47in2026Sophia}
\set{G}_{Q_k,\gamma_k}
&\triangleq&
\left\lbrace
P\in\simplexabs{Q_k}{\set{M}}:
\KL{P}{Q_k}\leqslant\gamma_k
\right\rbrace,
\end{IEEEeqnarray}
which are reminiscent of the set in~\eqref{EqKLNeighborhood}.
Taking into account that $\lambda_k>0$,
Lemma~\ref{LemmaMarch4at14h48in2026Sophia}
implies that the Gibbs probability measure
$P^{\autoparent{\mathsf{L}_{n_k},Q_k,\lambda_k}}_{
\vect{\Theta}\mid\vect{Z}=\vect{z}_k}$
in~\eqref{EqJan20at16h44in2026Sophia}
is the unique minimizer of the expected empirical risk
$\mathsf{R}_{\vect{z}_k}$ in~\eqref{EqExpEmpRiskModels}
within $\set{G}_{Q_k,\gamma_k}$.

Similarly, consider the finite relative entropy
\begin{IEEEeqnarray}{rCl}
\label{EqDecentralizedGeometryEta}
\eta
&\triangleq&
\KL{
P^{(\mathsf{L}_{n_0},Q_1,\lambda_0)}_{
\vect{\Theta}\mid\vect{Z}=\vect{z}_0}
}{Q_1}
<+\infty.
\end{IEEEeqnarray}
Taking into account that $\lambda_0>0$,
Lemma~\ref{LemmaMarch4at14h48in2026Sophia}
identifies the probability measure
$P^{(\mathsf{L}_{n_0},Q_1,\lambda_0)}_{
\vect{\Theta}\mid\vect{Z}=\vect{z}_0}$
in~\eqref{EqDec30at15h43in2025Antibes}
as the unique solution to
\begin{subequations}
\label{EqJanuary1at15h37in2026atHomeNice}
\begin{IEEEeqnarray}{rCl}
\min_{P\in\simplexabs{Q_1}{\set{M}}}
&&
\mathsf{R}_{\vect{z}_0}\autoparent{P}
\\
\text{s.t.}
&\quad&
\KL{P}{Q_1}\leqslant\eta,
\end{IEEEeqnarray}
\end{subequations}
that is, the minimizer of the expected empirical risk
with respect to the aggregated dataset $\vect{z}_0$
within $\set{G}_{Q_1,\eta}$.

From~\eqref{EqJanuary1at16h30in2026atHomeNice},
the probability measure $P^{(\mathsf{L}_{n_0},Q_1,\lambda_0)}_{
\vect{\Theta}\mid\vect{Z}=\vect{z}_0}$ is identical to the Gibbs probability measure
$P^{\autoparent{\mathsf{L}_{n_K},Q_K,\lambda_K}}_{
\vect{\Theta}\mid\vect{Z}=\vect{z}_K}$ of the form in \eqref{EqJan20at16h44in2026Sophia}.
Hence, after $K-1$ forward messages
(blue arrows in Figure~\ref{FigNestedStructure}),
client~$K$ obtains the same Gibbs probability measure
as a centralized system in which all training datasets
are available at client~$K$.
The backward dissemination
(red arrows in Figure~\ref{FigNestedStructure})
then provides each client with access to such a Gibbs algorithm.

\begin{figure*}[t]
\centering
\resizebox{0.5\textwidth}{!}{%
\input{\Figurespath/Nested_Decentralized/geometric_interpretation.tikz}}
\caption{Geometric representation of recursively nested
Gibbs probability measures for $K=4$, under the assumption
in~\eqref{EqJanuary1at15h16in2026atHomeNice}.
The neighborhoods $\set{G}_{Q_k,\gamma_k}$,
with $k\inCountK{4}$, and $\set{G}_{Q_1,\eta}$
are defined as in~\eqref{EqKLNeighborhood},
with $\gamma_k$ and $\eta$
in~\eqref{EqDecentralizedGeometryGamma}
and~\eqref{EqDecentralizedGeometryEta}, respectively.
The points $Q_{k+1}$ represent the Gibbs probability
measures in~\eqref{EqDecentralizedGeometryPoints}.
The final point $Q_5$
in~\eqref{EqDecentralizedGeometryEndpoint}
coincides with the Gibbs probability measure
in~\eqref{EqDec30at15h43in2025Antibes}
obtained by training upon the aggregated dataset
$\vect{z}_0$.}
\label{FigGeometricInter}
\end{figure*}
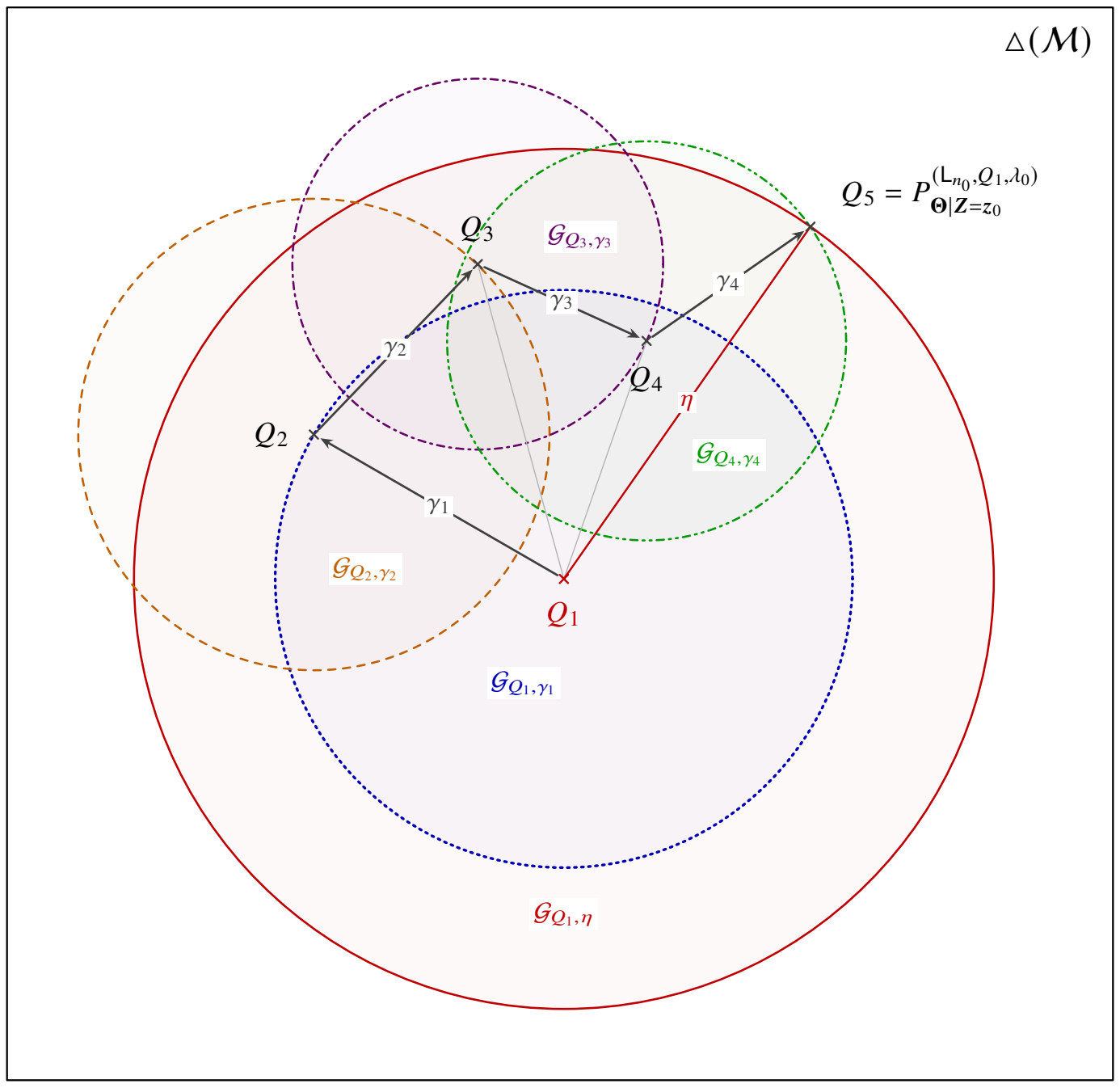

Figure~\ref{FigGeometricInter} provides a geometric
representation of this recursive nesting for $K=4$.
The points $Q_{k+1}$ represent
\begin{IEEEeqnarray}{rCl}
\label{EqDecentralizedGeometryPoints}
Q_{k+1}
&=&
P^{\autoparent{\mathsf{L}_{n_k},Q_k,\lambda_k}}_{
\vect{\Theta}\mid\vect{Z}=\vect{z}_k},
\qquad k\inCountK{K},
\end{IEEEeqnarray}
where the notation $Q_{K+1}$ is also used for the final
nested Gibbs probability measure.
For each $k\in\{1,2,\ldots,K-1\}$, the measure obtained
by client~$k$ becomes the reference measure for client~$k+1$,
yielding the sequence $Q_1,Q_2,\ldots,Q_{K+1}$.
The figure shows the corresponding neighborhoods
$\set{G}_{Q_k,\gamma_k}$
in~\eqref{EqJan20at16h47in2026Sophia},
together with $\set{G}_{Q_1,\eta}$,
where $\eta$ is defined
in~\eqref{EqDecentralizedGeometryEta}.

For $K=4$, the final Gibbs probability measures satisfies
\begin{IEEEeqnarray}{rCl}
\label{EqDecentralizedGeometryEndpoint}
Q_5
&=&
P^{\autoparent{\mathsf{L}_{n_4},Q_4,\lambda_4}}_{
\vect{\Theta}\mid\vect{Z}=\vect{z}_4}
\\
&=&
P^{(\mathsf{L}_{n_0},Q_1,\lambda_0)}_{
\vect{\Theta}\mid\vect{Z}=\vect{z}_0},
\end{IEEEeqnarray}
which follows from the equality in~\eqref{EqJanuary1at16h30in2026atHomeNice}.
Thus, the measure $Q_5$
in~\eqref{EqDecentralizedGeometryEndpoint}
represents both the minimizer of
$\mathsf{R}_{\vect{z}_4}$ within
$\set{G}_{Q_4,\gamma_4}$ and the minimizer of
$\mathsf{R}_{\vect{z}_0}$ within $\set{G}_{Q_1,\eta}$.

Beyond this geometric representation,
Lemma~\ref{LemmaSept11at15h55in2026Sophia}
relates the reals
$\gamma_k$ in~\eqref{EqDecentralizedGeometryGamma}
and $\eta$ in~\eqref{EqDecentralizedGeometryEta}
to the variation of the expected empirical risks
associated with the local datasets, as shown
by the following corollary.

\begin{corollary}
\label{CorollaryDecentralizedTriangleEquality}
Under the assumptions of
Corollary~\ref{CorollaryDec30at21h00in2025inAntibes}, the reals $\gamma_1, \gamma_2, \cdots, \gamma_{K}$ 
in~\eqref{EqDecentralizedGeometryGamma}
and $\eta$ in~\eqref{EqDecentralizedGeometryEta} satisfy 
\begin{IEEEeqnarray}{rCl}
\label{EqDecentralizedTriangleEquality}
\sum_{k=1}^{K}\gamma_k-\eta
&=&
\frac{1}{n_0\lambda_0}
\sum_{i=1}^{K-1}n_i
\Bigg(
\mathsf{R}_{\vect{z}_i}\autoparent{
P^{(\mathsf{L}_{n_0},Q_1,\lambda_0)}_{
\vect{\Theta}\mid\vect{Z}=\vect{z}_0}}-
\mathsf{R}_{\vect{z}_i}\autoparent{
P^{\autoparent{\mathsf{L}_{n_i},Q_i,\lambda_i}}_{
\vect{\Theta}\mid\vect{Z}=\vect{z}_i}}
\Bigg),
\end{IEEEeqnarray}
where $\mathsf{R}_{\vect{z}_i}$ is defined
in~\eqref{EqExpEmpRiskModels}.
\end{corollary}

For all $i\in\{1,2,\ldots,K-1\}$, the difference
on the right-hand side
of~\eqref{EqDecentralizedTriangleEquality}
compares the expected empirical risk on $\vect{z}_i$
with respect to the Gibbs probability measure
$P^{(\mathsf{L}_{n_0},Q_1,\lambda_0)}_{
\vect{\Theta}\mid\vect{Z}=\vect{z}_0}$
in~\eqref{EqDec30at15h43in2025Antibes},
trained upon the aggregated dataset $\vect{z}_0$,
with that under the Gibbs probability measure
$P^{\autoparent{\mathsf{L}_{n_i},Q_i,\lambda_i}}_{
\vect{\Theta}\mid\vect{Z}=\vect{z}_i}$
in~\eqref{EqJan20at16h44in2026Sophia},
obtained by client~$i$.

The following lemma expresses
\eqref{EqDecentralizedTriangleEquality}
in terms of the expected empirical risk on the aggregated
dataset and the expected empirical risks on the local datasets.

\begin{lemma}
\label{LemmaSept24at20h24in2026Antibes}
Under the assumptions of
Corollary~\ref{CorollaryDecentralizedTriangleEquality},
assume also that the expected empirical risks below are finite.
Then,
\begin{IEEEeqnarray}{rCl}
\label{EqDecentralizedPythagoreanEquality}
\sum_{k=1}^{K}\gamma_k-\eta
&=&
\frac{1}{\lambda_0}
\Bigg(
\mathsf{R}_{\vect{z}_0}\autoparent{
P^{(\mathsf{L}_{n_0},Q_1,\lambda_0)}_{
\vect{\Theta}\mid\vect{Z}=\vect{z}_0}}
-
\sum_{i=1}^{K}\frac{n_i}{n_0}
\mathsf{R}_{\vect{z}_i}\autoparent{
P^{\autoparent{\mathsf{L}_{n_i},Q_i,\lambda_i}}_{
\vect{\Theta}\mid\vect{Z}=\vect{z}_i}}
\Bigg).
\end{IEEEeqnarray}
\end{lemma}

\begin{IEEEproof}
The proof is presented in Appendix~\ref{AppProofOfLemmaSept24at20h24in2026Antibes}
\end{IEEEproof}

Lemma~\ref{LemmaSept24at20h24in2026Antibes}
shows that the difference $\sum_{k=1}^{K}\gamma_k-\eta$
is $\frac{1}{\lambda_0}$ times the difference between
the expected empirical risk on the aggregated dataset
$\vect{z}_0$ with respect to
$P^{(\mathsf{L}_{n_0},Q_1,\lambda_0)}_{
\vect{\Theta}\mid\vect{Z}=\vect{z}_0}$
in~\eqref{EqDec30at15h43in2025Antibes}
and the weighted sum of the expected empirical risks
on the local datasets $\vect{z}_i$ with respect to
$P^{\autoparent{\mathsf{L}_{n_i},Q_i,\lambda_i}}_{
\vect{\Theta}\mid\vect{Z}=\vect{z}_i}$
in~\eqref{EqJan20at16h44in2026Sophia},
with weights $\frac{n_i}{n_0}$.
In particular, when these quantities are equal,
$\sum_{k=1}^{K}\gamma_k=\eta$, the training error $\mathsf{R}_{\vect{z}_0}\autoparent{
P^{(\mathsf{L}_{n_0},Q_1,\lambda_0)}_{
\vect{\Theta}\mid\vect{Z}=\vect{z}_0}}$, which is  induced by the Gibbs algorithm trained upon the aggregation of all data, i.e., $
P^{(\mathsf{L}_{n_0},Q_1,\lambda_0)}_{
\vect{\Theta}\mid\vect{Z}=\vect{z}_0}$, is identical to the convex combination of the training error induced by Gibbs algorithms $P^{\autoparent{\mathsf{L}_{n_i},Q_i,\lambda_i}}_{
\vect{\Theta}\mid\vect{Z}=\vect{z}_i}$, trained exclusively on local training datasets $\vect{z}_{i}$, with $i \inCountK{K}$.

\paragraph{Exact unlearning}

The centralized exact-unlearning result in Corollary~\ref{CorollaryApril17at18h18in2026Sophia} requires the availability of the Gibbs probability measure trained upon the aggregated dataset~$\vect{z}_0$, namely
$P^{(\mathsf{L}_{n_0},Q_1,\lambda_0)}_{\vect{\Theta}\mid \vect{Z}=\vect{z}_0}$
in~\eqref{EqDec30at15h43in2025Antibes}.
In the decentralized setting, such a Gibbs probability measure can be obtained without explicit data aggregation by using the nested structure induced by the recursive choice of the reference measures in~\eqref{EqDec27at10h57in2025inMadrid}.
More specifically, under the assumptions in~\eqref{EqJanuary1at15h16in2026atHomeNice}, Corollary~\ref{CorollaryDec30at21h00in2025inAntibes} shows that the final Gibbs probability measure obtained through the forward pass of the communication protocol coincides with the Gibbs probability measure that would have been obtained by training upon the aggregated dataset~$\vect{z}_0$.

Once the Gibbs probability measure in~\eqref{EqDec30at15h43in2025Antibes} is available, Corollary~\ref{CorollaryApril17at18h18in2026Sophia} can be applied in a decentralized manner.
Fix $j\inCountK{K}$ and consider the deletion request of client~$j$, that is, the dataset~$\vect{z}_j\in\set{Z}^{n_j}$ is to be unlearned from the Gibbs probability measure
$P^{(\mathsf{L}_{n_0},Q_1,\lambda_0)}_{\vect{\Theta}\mid \vect{Z}=\vect{z}_0}$.
To do so, client~$j$ solves the following optimization problem:
\begin{IEEEeqnarray}{rCl}
\label{EqDecentralizedUnlearningClientj}
\max_{P \in \simplexabs{Q_1}{\set{M}}}
&&
\mathsf{R}_{\vect{z}_j}\autoparent{P}
-\frac{n_0}{n_j}\lambda_0
\KL{P}{P^{(\mathsf{L}_{n_0},Q_1,\lambda_0)}_{\vect{\Theta}\mid \vect{Z}=\vect{z}_0}},
\end{IEEEeqnarray}
where the functional~$\mathsf{R}_{\vect{z}_j}$ is defined in~\eqref{EqExpEmpRiskModels}.
This optimization problem requires access only to the dataset~$\vect{z}_j$ and to the Gibbs probability measure in~\eqref{EqDec30at15h43in2025Antibes}, and does not require explicit access to the datasets of the retained clients.

From Corollary~\ref{CorollaryApril17at18h18in2026Sophia}, solving~\eqref{EqDecentralizedUnlearningClientj} admits an exact-unlearning interpretation: the contribution of the dataset~$\vect{z}_j$ is canceled, and the updated probability measure coincides with the Gibbs probability measure that would have been obtained by training only upon the datasets of the retained clients.
Moreover, such a retained-data Gibbs probability measure can itself be obtained in a decentralized manner by applying again the nested decentralized construction above, but only over the retained clients.

Hence, provided that the Gibbs probability measure associated with the aggregated dataset~$\vect{z}_0$ can be obtained in a decentralized manner through the nested structure induced by~\eqref{EqDec27at10h57in2025inMadrid}, the exact-unlearning mechanism established above extends naturally to decentralized learning systems.
More specifically, this extension preserves the same operational advantages: the datasets remain local, the communication protocol only requires transmitting Gibbs probability measures, and exact unlearning is achieved without retraining from scratch upon the retained clients.

The same principle extends to multiple deletion requests of whole local datasets.
If $\mathcal{J}\subseteq\{1,\dots,K\}$ denotes the set of clients whose training dataset is to be forgotten, the corresponding updates can be performed sequentially, each time using as reference measure the Gibbs probability measure obtained after the previous update.
Hence, the exact-unlearning mechanism in Section~\ref{SubSecUnlearningCentralized} extends to multiple deletion requests in the decentralized setting of whole or part of local training datasets, provided that the Gibbs probability measure used at each update is obtained through the corresponding nested decentralized construction.

\subsection{Exact Unlearning with Normalized Log-Linear Combinations}
\label{SecExactUnlearningLogLinear}

This subsection leverages the results on normalized log-linear combinations of Gibbs algorithms developed in Section~\ref{SubSecNormalizedLogComGibbsAlg}.
As shown therein, the contribution of a data point to the weighted empirical risk associated with a normalized log-linear combination is jointly determined by the coefficients of the combination and the regularization factors of the corresponding Gibbs probability measures.

\vspace{2ex}
\subsubsection{Centralized Learning Systems}

Consider the $(\mathsf{L}_{n_1},Q,\lambda_1)$-Gibbs probability measure
\begin{IEEEeqnarray}{rCl}
\label{EqLogLinearUnlearningGibbsZ1}
P^{(\mathsf{L}_{n_1},Q,\lambda_1)}_{\vect{\Theta}\mid\vect{Z}=\vect{z}_1}
&\in&
\simplexabs{Q}{\set{M}},
\end{IEEEeqnarray}
which represents a Gibbs algorithm exclusively trained upon the dataset to be unlearned~$\vect{z}_1$.

Consider the normalized log-linear combination of
$P^{(\mathsf{L}_{n_0},Q,\lambda_0)}_{\vect{\Theta}\mid\vect{Z}=\vect{z}_0}$
in~\eqref{EqApril17at18h10in2026Sophia} and
$P^{(\mathsf{L}_{n_1},Q,\lambda_1)}_{\vect{\Theta}\mid\vect{Z}=\vect{z}_1}$
in~\eqref{EqLogLinearUnlearningGibbsZ1}, with coefficients $\alpha_0 \in\mathbb{R}\setminus\{0\}$ and $\alpha_1 \in\mathbb{R}\setminus\{0\}$, respectively, denoted by $S_2$. That is,
 for all $\vect{\theta} \in \supp Q$,
\begin{IEEEeqnarray}{rCl}
\label{EqOctober2at17h29in2026Benguerir}
\RND{S_2}{Q}(\vect{\theta}) = \frac{\autoparent{\RND{P_{\vect{\Theta} \mid\vect{Z} = \vect{z}_0}^{\autoparent{\mathsf{L}_{n_0},Q,\lambda_0}}}{Q}\autoparent{\vect{\theta}}}^{\alpha_0} \autoparent{\RND{P_{\vect{\Theta} \mid\vect{Z} = \vect{z}_1}^{\autoparent{\mathsf{L}_{n_1},Q,\lambda_1}}}{Q}(\vect{\theta})}^{\alpha_1}}{\displaystyle \int\autoparent{\RND{P_{\vect{\Theta} \mid\vect{Z} = \vect{z}_0}^{\autoparent{\mathsf{L}_{n_0},Q,\lambda_0}}}{Q}\autoparent{\vect{\nu}}}^{\alpha_0} \autoparent{\RND{P_{\vect{\Theta} \mid\vect{Z} = \vect{z}_1}^{\autoparent{\mathsf{L}_{n_1},Q,\lambda_1}}}{Q}(\vect{\nu})}^{\alpha_1} \mathrm{d}Q(\vect{\nu})}.
\end{IEEEeqnarray}
The following corollary of Lemma~\ref{LemmaMarch23at16h09in2026BusToNice} makes explicit the reweighting induced by this normalized log-linear combination.

\begin{corollary}
\label{CorollaryLogLinearUnlearningReweighting}
The normalized log-linear combination $S_2$ in~\eqref{EqOctober2at17h29in2026Benguerir} satisfies, for all $\vect{\theta}\in\supp Q$,
\begin{IEEEeqnarray}{rCl}
\label{EqLogLinearUnlearningDensity}
\RND{S_2}{Q}\autoparent{\vect{\theta}}
&=&
\frac{\exp\autoparent{-\autoparent{\frac{\alpha_0n_1}{n_0\lambda_0}+\frac{\alpha_1}{\lambda_1}}\mathsf{L}_{n_1}\autoparent{\vect{z}_1,\vect{\theta}}
-\frac{\alpha_0n_2}{n_0\lambda_0}\mathsf{L}_{n_2}\autoparent{\vect{z}_2,\vect{\theta}}}}
{\displaystyle\int
\exp\autoparent{-\autoparent{\frac{\alpha_0n_1}{n_0\lambda_0}+\frac{\alpha_1}{\lambda_1}}\mathsf{L}_{n_1}\autoparent{\vect{z}_1,\vect{\nu}}
-\frac{\alpha_0n_2}{n_0\lambda_0}\mathsf{L}_{n_2}\autoparent{\vect{z}_2,\vect{\nu}}}
 \mathrm{d}Q\autoparent{\vect{\nu}}}.
\end{IEEEeqnarray}
\end{corollary}

Corollary~\ref{CorollaryLogLinearUnlearningReweighting} shows that the normalized log-linear combination $S_2$ induces a linear combination of the empirical risks associated with the datasets $\vect{z}_1$ and $\vect{z}_2$.
More precisely, the coefficient of the empirical risk induced by the dataset to be unlearned is
\begin{IEEEeqnarray}{rCl}
\label{EqLogLinearUnlearningCoefficient}
\frac{\alpha_0n_1}{n_0\lambda_0}
+\frac{\alpha_1}{\lambda_1},
\end{IEEEeqnarray}
whereas the coefficient of the empirical risk induced by the dataset to be retained is
$\frac{\alpha_0n_2}{n_0\lambda_0}$.
Hence, exact unlearning is obtained by choosing the parameters such that the first coefficient is zero.
The following theorem shows that a joint choice of $\lambda_1$ and $\lambda_2$, $\alpha_0$ and $\alpha_1$, makes the normalized log-linear combination $S_2$ in~\eqref{EqOctober2at17h29in2026Benguerir}  be identical to the probability measure
$P^{(\mathsf{L}_{n_2},Q,\lambda_2)}_{\vect{\Theta}\mid\vect{Z}=\vect{z}_2}$
in~\eqref{EqApril17at18h11in2026Sophia}, in the sense of Definition~\ref{DefApril17at18h12in2026Sophia}.

\begin{corollary}
\label{CorollaryLogLinearExactUnlearning}
Assume that
\begin{IEEEeqnarray}{rCl}
\label{EqLogLinearUnlearningParametersA}
\frac{\lambda_1}{\alpha_1}
&=&
-\frac{n_0}{\alpha_0n_1}\lambda_0, \mbox{ and }
\\
\label{EqLogLinearUnlearningParametersB}
\lambda_2
&=&
\frac{n_0}{\alpha_0n_2}\lambda_0,
\end{IEEEeqnarray}
with $\lambda_0>0$ and $\lambda_2>0$.
Then, the normalized log-linear combination $S_2$ in~\eqref{EqOctober2at17h29in2026Benguerir}  and the $(\mathsf{L}_{n_2},Q,\lambda_2)$-Gibbs probability measure
$P^{(\mathsf{L}_{n_2},Q,\lambda_2)}_{\vect{\Theta}\mid\vect{Z}=\vect{z}_2}$
in~\eqref{EqApril17at18h11in2026Sophia} satisfy, for all $\vect{\theta}\in\supp Q$,
\begin{IEEEeqnarray}{rCl}
\label{EqLogLinearExactUnlearning}
\RND{S_2}{P^{(\mathsf{L}_{n_2},Q,\lambda_2)}_{\vect{\Theta}\mid\vect{Z}=\vect{z}_2}}
\autoparent{\vect{\theta}}
&=&
1.
\end{IEEEeqnarray}
\end{corollary}

Corollary~\ref{CorollaryLogLinearExactUnlearning} shows that the probability measure
$P^{(\mathsf{L}_{n_0},Q,\lambda_0)}_{\vect{\Theta}\mid\vect{Z}=\vect{z}_0}$
in~\eqref{EqApril17at18h10in2026Sophia} can be transformed into
$P^{(\mathsf{L}_{n_2},Q,\lambda_2)}_{\vect{\Theta}\mid\vect{Z}=\vect{z}_2}$
in~\eqref{EqApril17at18h11in2026Sophia} through a normalized log-linear combination with the Gibbs probability measure
$P^{(\mathsf{L}_{n_1},Q,\lambda_1)}_{\vect{\Theta}\mid\vect{Z}=\vect{z}_1}$
in~\eqref{EqLogLinearUnlearningGibbsZ1}, achieving exact unlearning in the sense of Definition~\ref{DefApril17at18h12in2026Sophia}.
An important observation is that, besides the original Gibbs probability measure, only the Gibbs probability measure trained upon the dataset to be unlearned, i.e., dataset~$\vect{z}_1$, is involved in the normalized log-linear combination.
The dependence on the dataset to be retained, i.e., dataset~$\vect{z}_2$, is carried by the original Gibbs probability measure
$P^{(\mathsf{L}_{n_0},Q,\lambda_0)}_{\vect{\Theta}\mid\vect{Z}=\vect{z}_0}$.
Under the choice in~\eqref{EqLogLinearUnlearningParametersA}--\eqref{EqLogLinearUnlearningParametersB}, the contribution associated with $\vect{z}_1$ is eliminated due to a zero weight. Alternatively, the coefficient associated with $\vect{z}_2$ becomes $\frac{1}{\lambda_2}$.
Consequently, the resulting probability measure coincides with the Gibbs probability measure that would have been obtained by training exclusively upon the retained dataset~$\vect{z}_2$.

Unlike the nested construction, in which the cancellation of the contribution associated with $\vect{z}_1$ is controlled by the regularization factor of the nested Gibbs probability measure, the normalized log-linear combination provides an additional degree of freedom.
More specifically, from~\eqref{EqLogLinearUnlearningParametersA}, the cancellation is jointly determined by the coefficient $\alpha_1$ and the parameter~$\lambda_1$ of the Gibbs probability measure trained upon $\vect{z}_1$.
Interestingly, from~\eqref{EqLogLinearUnlearningParametersB}, it follows that  the parameter $\alpha_0$ controls both  the coefficient of the retained dataset, which allows achieving unlearning; and also determines the parameter~$\lambda_2$ of the resulting Gibbs algorithm obtained after exact unlearning.

\vspace{2ex}
\subsubsection{Federated Learning}
\label{SecFederatedExactUnlearning}

Consider the same collection of $K$ local datasets
$\vect{z}_1,\vect{z}_2,\ldots,\vect{z}_K$ in~\eqref{EqDatasetK}, with
aggregation~$\vect{z}_0$ in~\eqref{EqDatasetZero}.
Consider a federated learning system in which the $K$ clients collaboratively
tune their local learning algorithms by communicating with a common server.
For all $k\inCountK{K}$, consider the
$(\mathsf{L}_{n_k},Q_k,\lambda_k)$-Gibbs probability measure
\begin{IEEEeqnarray}{rCl}
\label{EqGibbsClientKFixedDataset}
P^{(\mathsf{L}_{n_k},Q_k,\lambda_k)}
_{\vect{\Theta}\mid\vect{Z}=\vect{z}_k}
&\in&
\simplexabs{Q_k}{\set{M}},
\end{IEEEeqnarray}
with $\lambda_k>0$, which represents the Gibbs algorithm of client~$k$
trained upon its local dataset~$\vect{z}_k$.

\paragraph{Communication protocol}

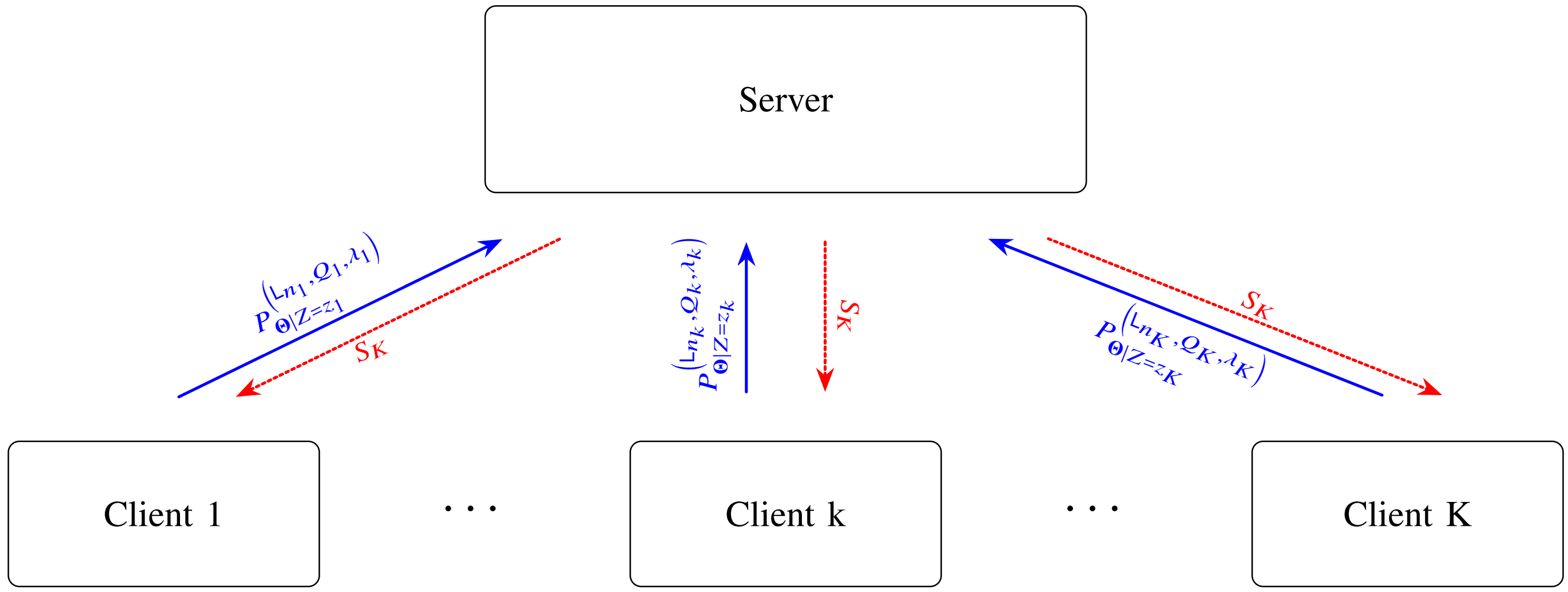
\begin{figure}[t]
\centering
\resizebox{\textwidth}{!}{
\input{\Figurespath/FL.tikz}}
\caption{One-shot client--server communication system. For all
$k\inCountK{K}$, the Gibbs algorithm of client~$k$,
$P^{(\mathsf{L}_{n_k},Q_k,\lambda_k)}
_{\vect{\Theta}\mid\vect{Z}=\vect{z}_k}$,
is defined in~\eqref{EqGibbsClientKFixedDataset}, and the probability measure
representing the aggregation at the server, $S_K$, is defined
in~\eqref{EqApril23at18h56in2026Antibes}.}
\label{FigFLGibbs}
\end{figure}

Consider the one-shot federated learning system depicted in
Figure~\ref{FigFLGibbs}.
In the first communication phase, the clients transmit their local Gibbs
algorithms to the server.
During the second phase, the server aggregates these local algorithms by means
of a normalized log-linear combination and broadcasts the resulting
probability measure to all clients.

\paragraph{Normalized log-linear combination}

The server constructs a probability measure 
$S_K\in\simplexabs{Q}{\set{M}}$ such that, for all
$\vect{\theta}\in\supp Q$,
\begin{IEEEeqnarray}{rCl}
\label{EqApril23at18h56in2026Antibes}
\RND{S_K}{Q}\autoparent{\vect{\theta}}
&=&
\frac{
\displaystyle
\prod_{k=1}^{K}
\autoparent{
\RND{
P^{(\mathsf{L}_{n_k},Q_k,\lambda_k)}
_{\vect{\Theta}\mid\vect{Z}=\vect{z}_k}
}{Q_k}
\autoparent{\vect{\theta}}
}^{\alpha_k}
}{
\displaystyle
\int
\prod_{k=1}^{K}
\autoparent{
\RND{
P^{(\mathsf{L}_{n_k},Q_k,\lambda_k)}
_{\vect{\Theta}\mid\vect{Z}=\vect{z}_k}
}{Q_k}
\autoparent{\vect{\nu}}
}^{\alpha_k}
 \mathrm{d}Q(\vect{\nu})
}
\\
\label{EqApril23at18h31in2026Antibes}
&=&
\frac{
\exp\autoparent{
-\displaystyle\sum_{k=1}^{K}
\frac{\alpha_k}{\lambda_k}
\mathsf{L}_{n_k}
\autoparent{\vect{z}_k,\vect{\theta}}
}
}{
\displaystyle
\int
\exp\autoparent{
-\displaystyle\sum_{k=1}^{K}
\frac{\alpha_k}{\lambda_k}
\mathsf{L}_{n_k}
\autoparent{\vect{z}_k,\vect{\nu}}
}
 \mathrm{d}Q(\vect{\nu})
},
\end{IEEEeqnarray}
where, for all $k\inCountK{K}$, it is assumed that $Q\ll Q_k$.
The equality in~\eqref{EqApril23at18h31in2026Antibes} follows from
Theorem~\ref{TheoremMay20at13h57in2026Antibes}.
Moreover, from Lemma~\ref{LemmaJuly28at10h54in2026Sophia}, the probability
measure $S_K$ is the unique solution to
\begin{IEEEeqnarray}{rCl}
\label{EqApril26at20h06in2026Antibes}
\min_{P\in\simplexabs{Q}{\set{M}}}
&&
\int
\autoparent{
\sum_{k=1}^{K}
\frac{\alpha_k}{\lambda_k}
\mathsf{L}_{n_k}
\autoparent{\vect{z}_k,\vect{\theta}}
}
 \mathrm{d}P(\vect{\theta})
+
\KL{P}{Q}.
\end{IEEEeqnarray}
Hence, although each client trains exclusively upon its local dataset, the
normalized log-linear combination performed at the server induces a weighted
empirical risk that depends on the datasets of all clients.

\paragraph{Centralized-performance guarantee}

A particular choice of the regularization factors and the coefficients of the
normalized log-linear combination makes the weighted empirical risk
in~\eqref{EqApril26at20h06in2026Antibes} be identical to the empirical risk
induced by the aggregation of all local datasets.
The following corollary formalizes this observation.

\begin{corollary}
\label{CorApril23at18h55in2026Antibes}
Assume that, for all $k\inCountK{K}$,
\begin{IEEEeqnarray}{rCl}
\lambda_k
&=&
\frac{1}{n_k}
\mbox{ and } 
\alpha_k
=
\frac{1}{n_0\lambda_0},
\end{IEEEeqnarray}
for some $\lambda_0>0$.
Then, for all $\vect{\theta}\in\supp Q$,
\begin{IEEEeqnarray}{rCl}
\label{EqApril23at18h47in2026Antibes}
\RND{S_K}{Q}\autoparent{\vect{\theta}}
&=&
\frac{
\exp\autoparent{
-\frac{1}{\lambda_0}
\mathsf{L}_{n_0}
\autoparent{\vect{z}_0,\vect{\theta}}
}
}{
\displaystyle
\int
\exp\autoparent{
-\frac{1}{\lambda_0}
\mathsf{L}_{n_0}
\autoparent{\vect{z}_0,\vect{\nu}}
}
 \mathrm{d}Q(\vect{\nu})
}
\\
&=&
\RND{
P^{(\mathsf{L}_{n_0},Q,\lambda_0)}
_{\vect{\Theta}\mid\vect{Z}=\vect{z}_0}
}{Q}
\autoparent{\vect{\theta}}.
\end{IEEEeqnarray}
\end{corollary}

Corollary~\ref{CorApril23at18h55in2026Antibes} shows that the normalized
log-linear combination of the $K$ Gibbs algorithms independently obtained by
the clients is identical to the Gibbs probability measure that would have
been obtained by training directly upon the aggregated dataset~$\vect{z}_0$.
More generally, the relevant condition  to observe~\eqref{EqApril23at18h47in2026Antibes} is for all $k\inCountK{K}$,
\begin{IEEEeqnarray}{rCl}
\label{EqFederatedCentralizedCondition}
\frac{\alpha_k}{\lambda_k}
&=&
\frac{n_k}{n_0\lambda_0}.
\end{IEEEeqnarray}
The particular choice in Corollary~\ref{CorApril23at18h55in2026Antibes}
allows each client to choose $\lambda_k$ based only on the size of its local
dataset, whereas the server uses the same coefficient $\alpha_k$ for all
clients.
Hence, the Gibbs probability measure required as the starting point of the
centralized exact-unlearning mechanism is available at the server without
explicitly aggregating the local datasets.

\paragraph{Exact unlearning}

Once the Gibbs probability measure
$P^{(\mathsf{L}_{n_0},Q,\lambda_0)}
_{\vect{\Theta}\mid\vect{Z}=\vect{z}_0}$
is available through the normalized log-linear combination in
Corollary~\ref{CorApril23at18h55in2026Antibes}, the exact-unlearning mechanism
developed in Section~\ref{SecExactUnlearningLogLinear} can be applied in the
federated setting.

Fix $j\inCountK{K}$ and consider a deletion request from client~$j$, that is,
the dataset~$\vect{z}_j\in\set{Z}^{n_j}$ is to be unlearned from the
algorithm~$P^{(\mathsf{L}_{n_0},Q,\lambda_0)}
_{\vect{\Theta}\mid\vect{Z}=\vect{z}_0}$ in~\eqref{EqApril23at18h47in2026Antibes}.
Denote by
\begin{IEEEeqnarray}{rCl}
\label{EqFederatedRetainedDataset}
\vect{z}_{-j}
&\triangleq&
\autoparent{
\vect{z}_1,\ldots,\vect{z}_{j-1},
\vect{z}_{j+1},\ldots,\vect{z}_K
}
\in\set{Z}^{n_{-j}},
\end{IEEEeqnarray}
the aggregation of the datasets to be retained, where
\begin{IEEEeqnarray}{rCl}
\label{EqFederatedRetainedDatasetSize}
n_{-j}
&\triangleq&
n_0-n_j.
\end{IEEEeqnarray}
The subscript $-j$ in~\eqref{EqFederatedRetainedDataset} indicates that the dataset of client~$j$ is excluded
from the aggregation of all datasets, i.e., $\vect{z}_0$ in \eqref{EqDatasetZero}.

Consider the normalized log-linear combination of~$P^{(\mathsf{L}_{n_0},Q,\lambda_0)}
_{\vect{\Theta}\mid\vect{Z}=\vect{z}_0}$, or simply $S_K$ for short,
in~\eqref{EqApril23at18h47in2026Antibes} and
$P^{(\mathsf{L}_{n_j},Q_j,\lambda_j)}
_{\vect{\Theta}\mid\vect{Z}=\vect{z}_j}$ of the form in \eqref{EqMarch31at15h49in2026Nice},
with coefficients $1$ and $-\alpha_j$, respectively, denoted by
$\widetilde{S}_{K,j}\in\simplexabs{Q}{\set{M}}$.
That is, for all $\vect{\theta}\in\supp Q$,
\begin{IEEEeqnarray}{rCl}
\label{EqFederatedUnlearningCombination}
\RND{\widetilde{S}_{K,j}}{Q}\autoparent{\vect{\theta}}
&=&
\frac{
\RND{P^{(\mathsf{L}_{n_0},Q,\lambda_0)}
_{\vect{\Theta}\mid\vect{Z}=\vect{z}_0}}{Q}\autoparent{\vect{\theta}}
\autoparent{
\RND{
P^{(\mathsf{L}_{n_j},Q_j,\lambda_j)}
_{\vect{\Theta}\mid\vect{Z}=\vect{z}_j}
}{Q_j}
\autoparent{\vect{\theta}}
}^{-\alpha_j}
}{
\displaystyle
\int
\RND{P^{(\mathsf{L}_{n_0},Q,\lambda_0)}
_{\vect{\Theta}\mid\vect{Z}=\vect{z}_0}}{Q}\autoparent{\vect{\nu}}
\autoparent{
\RND{
P^{(\mathsf{L}_{n_j},Q_j,\lambda_j)}
_{\vect{\Theta}\mid\vect{Z}=\vect{z}_j}
}{Q_j}
\autoparent{\vect{\nu}}
}^{-\alpha_j}
\mathrm{d}Q(\vect{\nu})
}.
\end{IEEEeqnarray}
The following corollary of
Theorem~\ref{TheoremMay20at13h57in2026Antibes}
makes explicit the reweighting induced by this normalized
log-linear combination.

\begin{corollary}
\label{CorFederatedUnlearningReweighting}
The normalized log-linear combination $\widetilde{S}_{K,j}$
in~\eqref{EqFederatedUnlearningCombination} satisfies,
for all $\vect{\theta}\in\supp Q$,
\begin{IEEEeqnarray}{rCl}
\label{EqFederatedUnlearningReweighting}
\RND{\widetilde{S}_{K,j}}{Q}\autoparent{\vect{\theta}}
&=&
\frac{
\exp\autoparent{
-\autoparent{
\frac{n_j}{n_0\lambda_0}
-\frac{\alpha_j}{\lambda_j}
}
\mathsf{L}_{n_j}\autoparent{\vect{z}_j,\vect{\theta}}
-\frac{n_{-j}}{n_0\lambda_0}
\mathsf{L}_{n_{-j}}\autoparent{\vect{z}_{-j},\vect{\theta}}
}
}{
\displaystyle
\int
\exp\autoparent{
-\autoparent{
\frac{n_j}{n_0\lambda_0}
-\frac{\alpha_j}{\lambda_j}
}
\mathsf{L}_{n_j}\autoparent{\vect{z}_j,\vect{\nu}}
-\frac{n_{-j}}{n_0\lambda_0}
\mathsf{L}_{n_{-j}}\autoparent{\vect{z}_{-j},\vect{\nu}}
}
\mathrm{d}Q(\vect{\nu})
},
\end{IEEEeqnarray}
where the dataset $\vect{z}_{-j}$ is defined in~\eqref{EqFederatedRetainedDataset} and $n_{-j}$ is defined in~\eqref{EqFederatedRetainedDatasetSize}.
\end{corollary}

Corollary~\ref{CorFederatedUnlearningReweighting} shows that the
normalized log-linear combination $\widetilde{S}_{K,j}$ in~\eqref{EqFederatedUnlearningCombination} induces a
linear combination of the empirical risks associated with the datasets
$\vect{z}_j$  in~\eqref{EqDatasetK} and $\vect{z}_{-j}$ in~\eqref{EqFederatedRetainedDataset}.
More precisely, the coefficient of the empirical risk induced by the
dataset to be unlearned is
\begin{IEEEeqnarray}{rCl}
\label{EqFederatedUnlearningCoefficient}
\frac{n_j}{n_0\lambda_0}
-\frac{\alpha_j}{\lambda_j},
\end{IEEEeqnarray}
whereas the coefficient of the empirical risk induced by the dataset
to be retained is $\frac{n_{-j}}{n_0\lambda_0}$.
Under the condition in~\eqref{EqFederatedCentralizedCondition},
\begin{IEEEeqnarray}{rCl}
\label{EqFederatedUnlearningZeroCoefficient}
\frac{n_j}{n_0\lambda_0}
-\frac{\alpha_j}{\lambda_j}
&=&
0.
\end{IEEEeqnarray}
Hence, the contribution associated with the dataset~$\vect{z}_j$
is eliminated due to a zero weight.
The following corollary shows that the normalized log-linear combination
$\widetilde{S}_{K,j}$ is identical to the Gibbs probability measure
trained exclusively upon the retained dataset~$\vect{z}_{-j}$ in~\eqref{EqFederatedRetainedDatasetSize},
in the sense of Definition~\ref{DefApril17at18h12in2026Sophia}.

\begin{corollary}
\label{CorFederatedExactUnlearning}
Assume that the condition in~\eqref{EqFederatedCentralizedCondition} holds,
and define
\begin{IEEEeqnarray}{rCl}
\label{EqFederatedUnlearningLambda}
\lambda_{-j}
&\triangleq&
\frac{n_0}{n_{-j}}\lambda_0,
\end{IEEEeqnarray}
where $n_{-j}$ is defined
in~\eqref{EqFederatedRetainedDatasetSize}.
Then, the normalized log-linear combination $\widetilde{S}_{K,j}$
in~\eqref{EqFederatedUnlearningCombination} and the
$(\mathsf{L}_{n_{-j}},Q,\lambda_{-j})$-Gibbs probability measure
$P^{(\mathsf{L}_{n_{-j}},Q,\lambda_{-j})}
_{\vect{\Theta}\mid\vect{Z}=\vect{z}_{-j}}$,
with $\vect{z}_{-j}$ in~\eqref{EqFederatedRetainedDataset},
satisfy, for all $\vect{\theta}\in\supp Q$,
\begin{IEEEeqnarray}{rCl}
\label{EqFederatedExactUnlearning}
\RND{
\widetilde{S}_{K,j}
}{
P^{(\mathsf{L}_{n_{-j}},Q,\lambda_{-j})}
_{\vect{\Theta}\mid\vect{Z}=\vect{z}_{-j}}
}
\autoparent{\vect{\theta}}
&=&
1.
\end{IEEEeqnarray}
\end{corollary}

Corollary~\ref{CorFederatedExactUnlearning} shows that the normalized
log-linear combination $\widetilde{S}_{K,j}$ in~\eqref{EqFederatedUnlearningCombination}
achieves exact unlearning of the dataset~$\vect{z}_j$.
Under~\eqref{EqFederatedCentralizedCondition}, the coefficient associated
with the empirical risk induced by $\vect{z}_j$ is zero,
as shown in~\eqref{EqFederatedUnlearningZeroCoefficient},
whereas the coefficient associated with the empirical risk induced by
$\vect{z}_{-j}$ satisfies
\begin{IEEEeqnarray}{rCl}
\label{EqFederatedRetainedCoefficient}
\frac{n_{-j}}{n_0\lambda_0}
&=&
\frac{1}{\lambda_{-j}}.
\end{IEEEeqnarray}
Consequently, the resulting probability measure $\widetilde{S}_{K,j}$ is identical to the
Gibbs probability measure $P^{(\mathsf{L}_{n_{-j}},Q,\lambda_{-j})}
_{\vect{\Theta}\mid\vect{Z}=\vect{z}_{-j}}$ that would have been obtained by training
exclusively upon the retained dataset~$\vect{z}_{-j}$, with
regularization factor~$\lambda_{-j}$
in~\eqref{EqFederatedUnlearningLambda}.
An important observation is that, once $P^{(\mathsf{L}_{n_0},Q,\lambda_0)}
_{\vect{\Theta}\mid\vect{Z}=\vect{z}_0}$ and the Gibbs probability
measure of client~$j$, $P^{(\mathsf{L}_{n_j},Q_j,\lambda_j)}
_{\vect{\Theta}\mid\vect{Z}=\vect{z}_j}$,  are available at the server, the normalized
log-linear combination in~\eqref{EqFederatedUnlearningCombination}
does not require explicit access to the datasets of the retained clients.
The dependence on the dataset to be retained, i.e.,
dataset~$\vect{z}_{-j}$, is carried by the original Gibbs probability
measure~$P^{(\mathsf{L}_{n_0},Q,\lambda_0)}
_{\vect{\Theta}\mid\vect{Z}=\vect{z}_0}$.
Hence, the same centralized exact-unlearning mechanism developed above
is implemented in the federated system using the probability measures
sent by the clients and the probability measure obtained at the server.

\subsection{Discussion}

The results in this section establish exact machine unlearning through two
closely related mechanisms.
With nested Gibbs algorithms, the Gibbs probability measure trained upon the
original dataset is used as a data-dependent reference measure, and the
regularization factor associated with the dataset to be unlearned is chosen
such that its contribution to the resulting weighted empirical risk vanishes.
This mechanism extends to decentralized learning because the Gibbs probability
measure associated with the aggregation of the local datasets can itself be
obtained through a sequence of nested Gibbs algorithms.

With normalized log-linear combinations, the contributions of the Gibbs
probability measures are combined directly through their Radon--Nikodym
derivatives.
Exact unlearning is obtained by assigning a compensating contribution to the
Gibbs probability measure associated with the dataset to be unlearned.
In the federated setting, this allows the contribution of a given client to be
canceled directly from the normalized log-linear combination produced at the
server.

Hence, although nesting and normalized log-linear combinations operate
differently on Gibbs probability measures, their exact-unlearning
interpretations rely on the same property established in
Section~\ref{SecApplicationstoSupervisedMachineLearning}: the parameters of
the corresponding construction determine the weights of the training
data points, and exact unlearning is achieved when the weights associated with
the data points to be unlearned are set to zero.

\section{Conclusion}
\label{SecConclusion}

Renormalization, normalized log-linear combinations, and nesting of Gibbs probability measures have been studied.
Under suitable conditions, the resulting measures have been shown to be Gibbs probability measures with new objective functions, reference measures, and regularization factors. 
More specifically, renormalization preserves the objective function while allowing changes in the regularization factor and reference measure. In contrast, normalized log-linear combinations and nesting lead to objective functions consisting of linear combinations of the original objective functions.
Using these observations, an equivalence between nesting and normalized log-linear combinations is established.
Geometric representations complement these results when the reference measures are probability measures, establishing connections with exponential geodesics in information geometry and interpreting the constrained optimization problems through relative entropy neighborhoods.

In supervised machine learning, normalized log-linear combinations and nesting of Gibbs probability measures yield Gibbs probability measures representing Gibbs algorithms associated with weighted empirical risks with respect to aggregated training datasets.
This connection provides a way to control the contribution of each data point to the resulting Gibbs algorithm through the parameters of the corresponding operations.
For instance, these parameters can be chosen such that the resulting algorithm is identical to the Gibbs algorithm obtained by training upon the aggregation of all local datasets, which has been referred to as \emph{centralized learning} or \emph{centralized-equivalent performance}. That is, the case in which local datasets are aggregated to serve as the training dataset.
Consequently, nesting and normalized log-linear combinations have been shown to achieve ccentralized-equivalent performance in decentralized and federated learning systems, respectively.

The same control of the weights provides two mechanisms for exact machine unlearning of specific datasets from Gibbs algorithms. 
For nesting, this is achieved by maximizing the expected empirical risk with respect to the dataset to be unlearned subject to a regularization by  relative entropy with respect to the Gibbs algorithm from which data must be unlearnt.
Alternatively, for normalized log-linear combinations, the algorithm from which data must be unlearnt is combined with a Gibbs algorithm trained exclusively on the dataset to be unlearned.
In both cases, a suitable choice of the parameters yields the same Gibbs algorithm as if it would be training exclusively upon the retained dataset.
Together with the constructions that yield the centralized-equivalent performance, these results on (centralized) unlearning establish exact unlearning of a client's dataset in decentralized and federated learning systems.
Thus, centralized-equivalent performance and exact machine unlearning are obtained as particular applications of the control of training-data contributions through operations on Gibbs algorithms.
%
%\triggeratref{24}
\bibliographystyle{IEEEtranlink}%tran
\bibliography{Merged_BERMUDEZ_ref.bib}

\appendices

\section{Proof of Lemma~\ref{LemmaMarch10at17h10in2026Sophia}}
\label{AppProofOfLemmaMarch10at17h10in2026Sophia}
For notational convenience, consider the $(h,Q,\lambda)$-Gibbs probability measure, $P^{(h,Q,\lambda)}_{Y\mid X = x} \in \triangle\left(\set{Y}  \right)$ and define the function
\begin{IEEEeqnarray}{cCl}
\label{EqGeneralCumulant}
\mathsf{K}_{h,Q,x} :\longfunction{ \reals}{\reals}{t}{\log\left( \int \exp\left( t \, h\autoparent{x,y}\right)\mathrm{d}Q\left( y \right) \right),}
\end{IEEEeqnarray}
which is a differentiable infinitely many times~\cite[Lemma~15]{PerlazaEntropy2025} in the interior of $\left\lbrace t \in \reals: \int \exp\left( t \, h\autoparent{x,y}\right)\mathrm{d}Q\left( y \right)  < +\infty \right\rbrace$.
From \cite[Lemma~2]{PerlazaEntropy2025}, it follows that 
\begin{IEEEeqnarray}{rCl}
\label{EqApril30at15h51in2026Sophia}
\KL{P^{(h,Q,\lambda)}_{Y\mid X=x}}{Q}   &=& -\frac{1}{\lambda} \int h(x,y) \mathrm{d}P^{(h,Q,\lambda)}_{Y\mid X=x}(y)  - \mathsf{K}_{h,Q,x}\autoparent{-\frac{1}{\lambda}}. 
\end{IEEEeqnarray}
The $n$-th derivative of the function $\mathsf{K}_{h,Q,x}$ in \eqref{EqGeneralCumulant} is  denoted by $\mathsf{K}_{h,Q,x}^{(n)}$. 
Hence, from \cite[Lemma~17]{perlaza2024empirical}, the first derivative of the function $\mathsf{K}_{h,Q,x}$, evaluated at $-\frac{1}{\lambda}$, satisfies 
\begin{IEEEeqnarray}{rCl}
\label{EqApril30at15h52in2026Sophia}
\mathsf{K}^{(1)}_{h,Q,x}\autoparent{-\frac{1}{\lambda}} &=& \int h(x,y) \mathrm{d}P^{(h,Q,\lambda)}_{Y\mid X=x}(y).
\end{IEEEeqnarray}
Hence, combining~\eqref{EqApril30at15h51in2026Sophia} and~\eqref{EqApril30at15h52in2026Sophia} yields
\begin{IEEEeqnarray}{rCl}
\label{EqApril30at17h06in2026Antibes}
\KL{P^{(h,Q,\lambda)}_{Y\mid X=x}}{Q}
&=&
-\,\mathsf{K}_{h,Q,x}\autoparent{-\frac{1}{\lambda}}
-\frac{1}{\lambda}\mathsf{K}^{(1)}_{h,Q,x}\autoparent{-\frac{1}{\lambda}}.
\end{IEEEeqnarray}
The derivative of the right-hand side of~\eqref{EqApril30at17h06in2026Antibes}
with respect to~$\lambda$ yields
\begin{IEEEeqnarray}{rCl}
\label{EqApril30at17h07in2026Antibes}
\frac{\mathrm{d}}{\mathrm{d}\lambda}
\KL{P^{(h,Q,\lambda)}_{Y\mid X=x}}{Q}
&=&
-\frac{1}{\lambda^3}\mathsf{K}^{(2)}_{h,Q,x}\autoparent{-\frac{1}{\lambda}}.
\end{IEEEeqnarray}
Moreover, from \cite[Lemma~17]{perlaza2024empirical}, the second derivative of the function $\mathsf{K}_{h,Q,x}$, evaluated at $-\frac{1}{\lambda}$, satisfies 
\begin{IEEEeqnarray}{rCl}
\label{EqApril30at17h08in2026Antibes}
\mathsf{K}^{(2)}_{h,Q,x}\autoparent{-\frac{1}{\lambda}}
&=&
\int \autoparent{h(x,y)-\mathsf{K}^{(1)}_{h,Q,x}\autoparent{-\frac{1}{\lambda}}}^{2}
 \mathrm{d}P^{(h,Q,\lambda)}_{Y\mid X=x}(y)
\;\geq\; 0.
\end{IEEEeqnarray}
Therefore,~\eqref{EqApril30at17h07in2026Antibes} and~\eqref{EqApril30at17h08in2026Antibes} imply that
\begin{IEEEeqnarray}{rCl}
\label{EqSeptember7at6h30in2026InTheBusToSophiaA}
\lambda>0
&\Rightarrow&
\frac{\mathrm{d}}{\mathrm{d}\lambda}
\KL{P^{(h,Q,\lambda)}_{Y\mid X=x}}{Q}
\leq 0,
\\
\label{EqSeptember7at6h30in2026InTheBusToSophiaB}
\lambda<0
&\Rightarrow&
\frac{\mathrm{d}}{\mathrm{d}\lambda}
\KL{P^{(h,Q,\lambda)}_{Y\mid X=x}}{Q}
\geq 0.
\end{IEEEeqnarray}
This proves that the mapping
\begin{IEEEeqnarray}{rCl}
\lambda \mapsto \KL{P^{(h,Q,\lambda)}_{Y\mid X=x}}{Q}
\end{IEEEeqnarray}
is nonincreasing on all intervals within~$(0,+\infty)$, and nondecreasing on all intervals within~$(-\infty,0)$, for which
\begin{IEEEeqnarray}{rCl}
\KL{P^{(h,Q,\lambda)}_{Y\mid X=x}}{Q}<+\infty .
\end{IEEEeqnarray}

The proof continues by identifying the conditions under which $\mathsf{K}^{(2)}_{h,Q,x}\autoparent{-\frac{1}{\lambda}}$ in \eqref{EqApril30at17h07in2026Antibes} is nonzero. That is,  when the inequalities in~\eqref{EqSeptember7at6h30in2026InTheBusToSophiaA} and \eqref{EqSeptember7at6h30in2026InTheBusToSophiaB} are strict. 
For doing so,  the notion of separable functions \cite[Definition~5]{perlaza2024empirical} is introduced. 
\begin{definition}[Separable Functions]
Given a fixed $x$, the function $h$ in \eqref{EqSeptember7at7h05in2026InTheBusToSophia} is said to be separable with respect to a probability measure $P \in \simplex{\set{Y}}$, if there exists two subsets 
\begin{IEEEeqnarray}{rcl}
\set{A}_{-}& =& \left\lbrace y \in \set{Y}:  h(x,y) > c \right\rbrace; and \\
\set{A}_{+} & = & \left\lbrace y \in \set{Y}:  h(x,y) < c \right\rbrace,
\end{IEEEeqnarray}
that are nonnegligible with respect to $P$, 
for some $c \in \reals$.
\end{definition}
Note that if the function $h$ in \eqref{EqSeptember7at7h05in2026InTheBusToSophia} is nonseparable with respect to a measure $P$, it is a constant almost surely with respect to $P$.  That is,
\begin{IEEEeqnarray}{rcl}
P\autoparent{\left\lbrace  y \in \set{Y}: h(x,y) = c \right\rbrace} & = & 1,
\end{IEEEeqnarray}
for some $c \in \reals$.
Using this observation, the following lemma characterizes the conditions under which $\mathsf{K}^{(2)}_{h,Q,x}\autoparent{-\frac{1}{\lambda}}$ in \eqref{EqApril30at17h07in2026Antibes} is nonzero.

\begin{lemma}\label{LemmaSeptember7at14h56in2026Sophia}
The value $\mathsf{K}^{(2)}_{h,Q,x}\autoparent{-\frac{1}{\lambda}}$ in \eqref{EqApril30at17h08in2026Antibes} satisfies
\begin{IEEEeqnarray}{rcl}
\mathsf{K}^{(2)}_{h,Q,x}\autoparent{-\frac{1}{\lambda}}& = & 0,
\end{IEEEeqnarray}
if and only if the function $h$ is nonseparable with respect to $Q$.
\end{lemma}
\begin{IEEEproof}
The proof uses the following sets:
\begin{subequations}\label{EqSeptember7at11h23in2026SophiaAntipolis}
\begin{IEEEeqnarray}{rcl}
\set{A}_{-}& =& \left\lbrace y \in \set{Y}:  h(x,y) > c \right\rbrace;\\
\set{A}_{0}& =& \left\lbrace y \in \set{Y}:  h(x,y) = c \right\rbrace; and \\
\set{A}_{+} & = & \left\lbrace y \in \set{Y}:  h(x,y) < c \right\rbrace,
\end{IEEEeqnarray}
with 
\begin{IEEEeqnarray}{rcl}
c & \triangleq &\int h(x,y)\mathrm{d}P^{(h,Q,\lambda)}_{Y\mid X=x}(y)\\
& = &  \mathsf{K}^{(1)}_{h,Q,x}\autoparent{-\frac{1}{\lambda}}.
\end{IEEEeqnarray}
\end{subequations}
The main argument of the proof consists of the following equalities:
\begin{IEEEeqnarray}{rcl}
\label{EqSeptember7at11h25in2026SophiaAntipolisA}
\mathsf{K}^{(2)}_{h,Q,x}\autoparent{-\frac{1}{\lambda}}
& = & \int \autoparent{h(x,y)-\mathsf{K}^{(1)}_{h,Q,x}\autoparent{-\frac{1}{\lambda}}}^{2}
 \mathrm{d}P^{(h,Q,\lambda)}_{Y\mid X=x}(y) \\
\nonumber
& = & \int_{\set{A}_{-}} \autoparent{h(x,y)-\mathsf{K}^{(1)}_{h,Q,x}\autoparent{-\frac{1}{\lambda}}}^{2}
 \mathrm{d}P^{(h,Q,\lambda)}_{Y\mid X=x}(y) +
\int_{\set{A}_{0}} \autoparent{h(x,y)-\mathsf{K}^{(1)}_{h,Q,x}\autoparent{-\frac{1}{\lambda}}}^{2}
 \mathrm{d}P^{(h,Q,\lambda)}_{Y\mid X=x}(y)\\
\label{EqSeptember7at11h25in2026SophiaAntipolisB}
& & + \int_{\set{A}_{+}} \autoparent{h(x,y)-\mathsf{K}^{(1)}_{h,Q,x}\autoparent{-\frac{1}{\lambda}}}^{2}
 \mathrm{d}P^{(h,Q,\lambda)}_{Y\mid X=x}(y)\\
\label{EqSeptember7at11h25in2026SophiaAntipolisC}
& = & \int_{\set{A}_{-}} \autoparent{h(x,y)-\mathsf{K}^{(1)}_{h,Q,x}\autoparent{-\frac{1}{\lambda}}}^{2}
 \mathrm{d}P^{(h,Q,\lambda)}_{Y\mid X=x}(y) + \int_{\set{A}_{+}} \autoparent{h(x,y)-\mathsf{K}^{(1)}_{h,Q,x}\autoparent{-\frac{1}{\lambda}}}^{2}
 \mathrm{d}P^{(h,Q,\lambda)}_{Y\mid X=x}(y),\end{IEEEeqnarray}
where the equality in~\eqref{EqSeptember7at11h25in2026SophiaAntipolisA} follows from~\cite[Lemma~17]{perlaza2024empirical}; 
the equality in~\eqref{EqSeptember7at11h25in2026SophiaAntipolisB} follows from the definition of the sets in~\eqref{EqSeptember7at11h23in2026SophiaAntipolis}; and  
the equality in~\eqref{EqSeptember7at11h25in2026SophiaAntipolisC} follows from observing that
\begin{IEEEeqnarray}{rcl}
\int_{\set{A}_{0}} \autoparent{h(x,y)-\mathsf{K}^{(1)}_{h,Q,x}\autoparent{-\frac{1}{\lambda}}}^{2}
 \mathrm{d}P^{(h,Q,\lambda)}_{Y\mid X=x}(y) & = & 0.
\end{IEEEeqnarray}

The proof is divided into two parts. In the first part it is shown that if $\mathsf{K}^{(2)}_{h,Q,x}\autoparent{-\frac{1}{\lambda}} =  0$, then the function $h$ is nonseparable with respect to $Q$, given $x$. 
The second part shows that if the function $h$ is nonseparable with respect to $Q$, given $x$, then $\mathsf{K}^{(2)}_{h,Q,x}\autoparent{-\frac{1}{\lambda}} =  0$.

The first part is as follows. Assume that 
\begin{IEEEeqnarray}{rcl}
\mathsf{K}^{(2)}_{h,Q,x}\autoparent{-\frac{1}{\lambda}}& = & 0, 
\end{IEEEeqnarray}
which, together with \eqref{EqSeptember7at11h25in2026SophiaAntipolisC},  imply that
\begin{IEEEeqnarray}{rcl}
\label{EqSeptember7at14h50in2026SophiaAntipolisC}
0 & = &  \int_{\set{A}_{-}} \autoparent{h(x,y)-\mathsf{K}^{(1)}_{h,Q,x}\autoparent{-\frac{1}{\lambda}}}^{2}
 \mathrm{d}P^{(h,Q,\lambda)}_{Y\mid X=x}(y) + \int_{\set{A}_{+}} \autoparent{h(x,y)-\mathsf{K}^{(1)}_{h,Q,x}\autoparent{-\frac{1}{\lambda}}}^{2}
 \mathrm{d}P^{(h,Q,\lambda)}_{Y\mid X=x}(y).\end{IEEEeqnarray}
Note that the integrands in~\eqref{EqSeptember7at14h50in2026SophiaAntipolisC} are nonnegative. Thus, from \cite[Theorem~1.6.6b]{ash2000probability}, it follows that
\begin{IEEEeqnarray}{rcl}
P^{(h,Q,\lambda)}_{Y\mid X=x} \autoparent{\left\lbrace y \in \set{Y}:  h(x,y) - \mathsf{K}^{(1)}_{h,Q,x}\autoparent{-\frac{1}{\lambda}} = 0\right\rbrace} & =  & 1.
\end{IEEEeqnarray}
Finally, from the fact that the measures $P^{(h,Q,\lambda)}_{Y\mid X=x}$ and $Q$ are mutually absolutely continuous \cite[Lemma~3]{perlaza2024empirical}, it holds that 
\begin{IEEEeqnarray}{rcl}
Q\autoparent{\left\lbrace y \in \set{Y}:  h(x,y) - \mathsf{K}^{(1)}_{h,Q,x}\autoparent{-\frac{1}{\lambda}} = 0\right\rbrace} & =  & 1.
\end{IEEEeqnarray}
Thus, $Q\autoparent{\set{A}_{0}} = 1$ and shows that the function $h$ is nonseparable with respect to $Q$, given $x$. This completes the first part of the proof.

The second part of the proof is as follows. Consider that, given $x$, the function $h$ is nonseparable with respect to $Q$.  
Then, 
\begin{IEEEeqnarray}{rcl}
Q\autoparent{\left\lbrace y \in \set{Y}:  h(x,y) =   \int h(x,y) \mathrm{d} P^{(h,Q,\lambda)}_{Y\mid X=x} (y) \right\rbrace } & =  & 1,
\end{IEEEeqnarray}
which, from the fact that the measures $P^{(h,Q,\lambda)}_{Y\mid X=x}$ and $Q$ are mutually absolutely continuous \cite[Lemma~3]{perlaza2024empirical}, implies 
\begin{IEEEeqnarray}{rcl}
1 & =  & Q\autoparent{\left\lbrace y \in \set{Y}:  h(x,y) =   \int h(x,y) \mathrm{d} P^{(h,Q,\lambda)}_{Y\mid X=x} (y) \right\rbrace } \\
& = & P^{(h,Q,\lambda)}_{Y\mid X=x} \autoparent{\left\lbrace y \in \set{Y}:  h(x,y) =   \int h(x,y) \mathrm{d} P^{(h,Q,\lambda)}_{Y\mid X=x} (y) \right\rbrace }.
\end{IEEEeqnarray}
Hence, considering that the sets $\set{A}_{-}$, $\set{A}_{0}$, and $\set{A}_{+}$ form a partition of $\set{Y}$, it follows that
\begin{subequations}\label{EqSeptember7at14h44in2026Sophia}
\begin{IEEEeqnarray}{rcl}
\label{EqSeptember7at14h44in2026SophiaA}
P^{(h,Q,\lambda)}_{Y\mid X=x} \autoparent{ \set{A} _{-}} & = & 0;\\
\label{EqSeptember7at14h44in2026SophiaB}
P^{(h,Q,\lambda)}_{Y\mid X=x} \autoparent{ \set{A} _{+}} & = & 0; \mbox{ and } \\
\label{EqSeptember7at14h44in2026SophiaC}
P^{(h,Q,\lambda)}_{Y\mid X=x} \autoparent{ \set{A} _{0}} & = & 1.
\end{IEEEeqnarray}
\end{subequations}
Plugging \eqref{EqSeptember7at14h44in2026Sophia} in \eqref{EqSeptember7at11h25in2026SophiaAntipolisC}, it follows that $\mathsf{K}^{(2)}_{h,Q,x}\autoparent{-\frac{1}{\lambda}} = 0$, which completes the second part of the proof.
\end{IEEEproof}
Finally, from Lemma~\ref{LemmaSeptember7at14h56in2026Sophia} and the equality in~\eqref{EqApril30at17h07in2026Antibes}, under the assumption that the function $h$ is separable, it holds that
\begin{IEEEeqnarray}{rCl}
\lambda>0
&\Longrightarrow&
\frac{\mathrm{d}}{\mathrm{d}\lambda}
\KL{P^{(h,Q,\lambda)}_{Y\mid X=x}}{Q}
< 0,
\\
\lambda<0
&\Longrightarrow&
\frac{\mathrm{d}}{\mathrm{d}\lambda}
\KL{P^{(h,Q,\lambda)}_{Y\mid X=x}}{Q}
> 0,
\end{IEEEeqnarray}
which implies strict monotonicity. This completes the proof.

\section{Proof of Theorem~\ref{TheoremApril16at13h13in2026Antibes}}
\label{AppProofOfTheoremApril16at13h13in2026Antibes}

From Definition~\ref{DefRenormalization},  the $(\alpha, Q, Q_1)$-renormalization of the $(h,Q,\lambda)$-Gibbs probability measure $P^{(h,Q,\lambda)}_{Y\mid X=x}$ in~\eqref{EqGeneralGibbsRN},  denoted by $R_1$ in~\eqref{EqApril15at17h43in2026Sophia}, satisfies, for all $y \in \supp Q_1$,
\begin{IEEEeqnarray}{rCl}
\label{EqApril26at19h39in2026AntibesA}
\RND{R_1}{Q_1}\autoparent{y} &=& \frac{\autoparent{\RND{P^{(h,Q,\lambda)}_{Y\mid X=x}}{Q} \autoparent{y}}^{\alpha}}{\displaystyle\int \autoparent{\RND{P^{(h,Q,\lambda)}_{Y\mid X=x}}{Q}\autoparent{\hat{y}}}^{\alpha} \mathrm{d}Q_1\autoparent{\hat{y}}}\\
\label{EqApril26at19h39in2026AntibesB}
&=& \frac{\autoparent{\exp\autoparent{- \frac{1}{\lambda} h\autoparent{x,y} - \mathsf{K}_{h,Q,x}\autoparent{-\frac{1}{\lambda}}}}^{\alpha}}{\displaystyle\int \autoparent{\exp\autoparent{- \frac{1}{\lambda} h\autoparent{x,\hat{y}} - \mathsf{K}_{h,Q,x}\autoparent{-\frac{1}{\lambda}}}}^{\alpha} \mathrm{d}Q_1\autoparent{\hat{y}}}\\
\label{EqApril26at19h39in2026AntibesC}
&=& \frac{\exp\autoparent{- \frac{\alpha}{\lambda} h\autoparent{x,y}}}{\displaystyle\int \exp\autoparent{- \frac{\alpha}{\lambda} h\autoparent{x,\hat{y}}} \mathrm{d}Q_1\autoparent{\hat{y}}} \frac{\exp\autoparent{\alpha\mathsf{K}_{h,Q,x}\autoparent{-\frac{1}{\lambda}}}}{\exp\autoparent{\alpha\mathsf{K}_{h,Q,x}\autoparent{-\frac{1}{\lambda}}}}\\
\label{EqApril26at19h39in2026AntibesD}
&=& \frac{\exp\autoparent{- \frac{\alpha}{\lambda} h\autoparent{x,y}}}{\displaystyle\int \exp\autoparent{- \frac{\alpha}{\lambda} h\autoparent{x,\hat{y}}} \mathrm{d}Q_1\autoparent{\hat{y}}}\\
\label{EqApril26at19h39in2026AntibesE}
&=& \RND{P^{\autoparent{h,Q_1,\frac{\lambda}{\alpha}}}_{Y\mid X=x}}{Q_1}\autoparent{y},
\end{IEEEeqnarray}
where the function $\mathsf{K}_{h,Q,x}$ is defined in~\eqref{EqGeneralCumulant}; and  
the equalities in~\eqref{EqApril26at19h39in2026AntibesB} and~\eqref{EqApril26at19h39in2026AntibesE}  follow from Definition~\ref{DefGeneralGibbs}. This completes the proof.

\section{Proof of Theorem~\ref{TheoremMay20at13h57in2026Antibes}}
\label{AppProofOfTheoremMay20at13h57in2026Antibes}

From~\eqref{EqGeneralGibbsRN}, for all $y \in \supp Q$, it follows that
\begin{IEEEeqnarray}{rCl}
\label{EqJuly7in2026at13h20A}
\prod_{i=1}^{K}
\autoparent{\RND{P^{(h_i,Q_i,\lambda_i)}_{Y\mid X = x_i}}{Q_i}\autoparent{y}}^{\alpha_i}
&=&
\prod_{i=1}^{K}
\autoparent{\frac{
\exp\autoparent{
-\frac{1}{\lambda_i}h_i\autoparent{x_i,y}}}{\displaystyle \int \exp\autoparent{-\frac{1}{\lambda_i}h_i\autoparent{x_i,\tilde{y}}} \mathrm{d}Q_i\autoparent{\tilde{y}}}
}^{\alpha_i}\\
&=&
\prod_{i=1}^{K}
\autoparent{
\exp\autoparent{
-\frac{1}{\lambda_i}h_i\autoparent{x_i,y}
- \log \autoparent{\displaystyle \int \exp\autoparent{-\frac{1}{\lambda_i}h_i\autoparent{x_i,\tilde{y}}} \mathrm{d}Q_i\autoparent{\tilde{y}}}
}
}^{\alpha_i}
\\
\label{EqJuly7in2026at13h20B}
&=&
\prod_{i=1}^{K}
\exp\autoparent{
-\frac{\alpha_i}{\lambda_i}h_i\autoparent{x_i,y}
-\alpha_i \log \autoparent{\displaystyle \int \exp\autoparent{-\frac{1}{\lambda_i}h_i\autoparent{x_i,\tilde{y}}} \mathrm{d}Q_i\autoparent{\tilde{y}}}
}
\\
\label{EqJuly7in2026at13h20C}
&=&
\exp\autoparent{
-\displaystyle \sum_{i=1}^{K}\frac{\alpha_i}{\lambda_i}h_i\autoparent{x_i,y}
}
\exp\autoparent{
-\displaystyle \sum_{i=1}^{K}\alpha_i\log \autoparent{\displaystyle \int \exp\autoparent{-\frac{1}{\lambda_i}h_i\autoparent{x_i,\tilde{y}}} \mathrm{d}Q_i\autoparent{\tilde{y}}}
},
\end{IEEEeqnarray}
where the equality in~\eqref{EqJuly7in2026at13h20B} follows from the power rule of the exponential function; and the equality in~\eqref{EqJuly7in2026at13h20C} follows from the product rule of the exponential function.

Using~\eqref{EqJuly7in2026at13h20C} in~\eqref{EqMay19at15h40in2026Sophia} yields
\begin{IEEEeqnarray}{rCl}
\label{EqJuly7in2026at13h20D}
\RND{S_K}{Q}\autoparent{y}
&=&
\frac{\displaystyle \prod_{i=1}^{K}\autoparent{\RND{P_{Y\mid X = x_i}^{(h_i, Q_i, \lambda_i)}}{Q_i}\autoparent{y}}^{\alpha_i} }
{\displaystyle\int \displaystyle \prod_{i=1}^{K}\autoparent{\RND{P_{Y\mid X = x_i}^{(h_i, Q_i, \lambda_i)}}{Q_i}\autoparent{\hat{y}}}^{\alpha_i} \mathrm{d}Q\autoparent{\hat{y}}}
\\
\label{EqJuly7in2026at13h20E}
&=&
\frac{\exp\autoparent{
-\displaystyle \sum_{i=1}^{K}\frac{\alpha_i}{\lambda_i}h_i\autoparent{x_i,y}
}
\exp\autoparent{
-\displaystyle \sum_{i=1}^{K}\alpha_i \log \autoparent{\displaystyle \int \exp\autoparent{-\frac{1}{\lambda_i}h_i\autoparent{x_i,\tilde{y}}} \mathrm{d}Q_i\autoparent{\tilde{y}}}
} }
{\displaystyle\int \exp\autoparent{
-\displaystyle \sum_{i=1}^{K}\frac{\alpha_i}{\lambda_i}h_i\autoparent{x_i,\hat{y}}
}
\exp\autoparent{
-\displaystyle \sum_{i=1}^{K}\alpha_i \log \autoparent{\displaystyle \int \exp\autoparent{-\frac{1}{\lambda_i}h_i\autoparent{x_i,\tilde{y}}} \mathrm{d}Q_i\autoparent{\tilde{y}}}
} \mathrm{d}Q\autoparent{\hat{y}}}
\\
\label{EqJuly7in2026at13h20F}
&=&
\frac{
\exp\autoparent{
-\displaystyle \sum_{i=1}^{K}\frac{\alpha_i}{\lambda_i}h_i\autoparent{x_i,y}
}
}{
\displaystyle
\int
\exp\autoparent{
-\displaystyle \sum_{i=1}^{K}\frac{\alpha_i}{\lambda_i}h_i\autoparent{x_i,\hat{y}}
}
\mathrm{d}Q\autoparent{\hat{y}}
}\quad
\frac{
\exp\autoparent{
-\displaystyle \sum_{i=1}^{K}\alpha_i\log \autoparent{\displaystyle \int \exp\autoparent{-\frac{1}{\lambda_i}h_i\autoparent{x_i,\tilde{y}}} \mathrm{d}Q_i\autoparent{\tilde{y}}}
}
}{
\exp\autoparent{
-\displaystyle \sum_{i=1}^{K}\alpha_i\log \autoparent{\displaystyle \int \exp\autoparent{-\frac{1}{\lambda_i}h_i\autoparent{x_i,\tilde{y}}} \mathrm{d}Q_i\autoparent{\tilde{y}}}
}
}
\\
\label{EqJuly7in2026at13h20G}
&=&
\frac{
\exp\autoparent{
-\displaystyle \sum_{i=1}^{K}\frac{\alpha_i}{\lambda_i}h_i\autoparent{x_i,y}
}
}{
\displaystyle
\int
\exp\autoparent{
-\displaystyle \sum_{i=1}^{K}\frac{\alpha_i}{\lambda_i}h_i\autoparent{x_i,\hat{y}}
}
\mathrm{d}Q\autoparent{\hat{y}}
},
\end{IEEEeqnarray}
where the equality in~\eqref{EqJuly7in2026at13h20F} follows from the fact that
\begin{IEEEeqnarray}{rCl}
\label{EqJuly7in2026at13h20H}
\exp\autoparent{
-\displaystyle \sum_{i=1}^{K}\alpha_i\log \autoparent{\displaystyle \int \exp\autoparent{-\frac{1}{\lambda_i}h_i\autoparent{x_i,\tilde{y}}} \mathrm{d}Q_i\autoparent{\tilde{y}}}
}
\end{IEEEeqnarray}
does not depend on $\hat{y}$; and the equality in~\eqref{EqJuly7in2026at13h20G} follows by simplifying the common multiplicative factor in~\eqref{EqJuly7in2026at13h20H}.

From the definition of $\vect{\beta}$ in the statement of the theorem, it holds that
\begin{IEEEeqnarray}{rCl}
\label{EqJuly7in2026at13h20I}
\vect{\beta}
&=&
\autoparent{\frac{\alpha_1}{\lambda_1}, \frac{\alpha_2}{\lambda_2}, \ldots, \frac{\alpha_K}{\lambda_K}}.
\end{IEEEeqnarray}
From the definition of $\hat{h}_{\vect{\beta}}$ in~\eqref{EqApril29at13h53in2026Sophia}, for all $y\in\supp Q$, it follows that
\begin{IEEEeqnarray}{rCl}
\label{EqJuly7in2026at13h20J}
\hat{h}_{\vect{\beta}}\autoparent{\vect{x},y}
&=&
\displaystyle \sum_{i=1}^{K}\frac{\alpha_i}{\lambda_i}h_i\autoparent{x_i,y}, 
\end{IEEEeqnarray}
where $\vect{x} = \autoparent{x_1,x_2,\ldots,x_K}$.
Therefore, from Definition~\ref{DefGeneralGibbs}, for all $y\in\supp Q$,
\begin{IEEEeqnarray}{rCl}
\label{EqJuly7in2026at13h20K}
\RND{P_{Y\mid \vect{X} = \vect{x}}^{\autoparent{\hat{h}_{\vect{\beta}},Q,1}}}{Q}\autoparent{y}
&=&
\frac{
\exp\autoparent{
-\hat{h}_{\vect{\beta}}\autoparent{\vect{x},y}
}
}{
\displaystyle
\int
\exp\autoparent{
-\hat{h}_{\vect{\beta}}\autoparent{\vect{x},\hat{y}}
}
\mathrm{d}Q\autoparent{\hat{y}}
}
\\
\label{EqJuly7in2026at13h20L}
&=&
\frac{
\exp\autoparent{
-\displaystyle \sum_{i=1}^{K}\frac{\alpha_i}{\lambda_i}h_i\autoparent{x_i,y}
}
}{
\displaystyle
\int
\exp\autoparent{
-\displaystyle \sum_{i=1}^{K}\frac{\alpha_i}{\lambda_i}h_i\autoparent{x_i,\hat{y}}
}
\mathrm{d}Q\autoparent{\hat{y}}
},
\end{IEEEeqnarray}
where the equality in~\eqref{EqJuly7in2026at13h20L} follows from~\eqref{EqJuly7in2026at13h20J}. Combining~\eqref{EqJuly7in2026at13h20G} and~\eqref{EqJuly7in2026at13h20L} yields
\begin{IEEEeqnarray}{rCl}
\label{EqJuly7in2026at13h20M}
\RND{S_K}{Q}\autoparent{y}
&=&
\RND{P_{Y\mid \vect{X} = \vect{x}}^{\autoparent{\hat{h}_{\vect{\beta}},Q,1}}}{Q}\autoparent{y},
\end{IEEEeqnarray}
which completes the proof of~\eqref{EqMay20at13h58in2026Antibes} and~\eqref{EqApril15at22h01in2026AntibesB}.

\section{Proof of Theorem~\ref{TheoremApril16at15h23in2026Antibes}}
\label{AppProofOfTheoremApril16at15h23in2026Antibes}

From Theorem~\ref{TheoremMay20at13h57in2026Antibes}, the normalized log-linear combination $S_K$ in~\eqref{EqMay19at15h40in2026Sophia} is identical to the $\autoparent{\hat{h}_{\vect{\beta}},Q,1}$-Gibbs probability measure
$P_{Y\mid \vect{X} =\vect{x}}^{\autoparent{\hat{h}_{\vect{\beta}},Q,1}}$ in~\eqref{EqApril15at22h01in2026AntibesB}.
Moreover, taking into consideration that 
$P_{Y\mid \vect{X} =\vect{x}}^{\autoparent{\hat{h}_{\vect{\beta}},Q,1}}$
is a Gibbs probability measure with reference measure $Q$, from \cite[Lemma~3]{perlaza2024empirical}, it follows that
\begin{IEEEeqnarray}{rCl}
\label{EqJuly7in2026at13h35A}
Q
&\ll&
P_{Y\mid \vect{X} =\vect{x}}^{\autoparent{\hat{h}_{\vect{\beta}},Q,1}}
\ll
Q.
\end{IEEEeqnarray}
Thus, from \cite[Theorem~2]{InriaRR9591}, the Radon--Nikodym derivative
$\RND{Q}{P_{Y\mid \vect{X} =\vect{x}}^{\autoparent{\hat{h}_{\vect{\beta}},Q,1}}}$
is well defined and, from the multiplicative inverse of the Radon--Nikodym derivative \cite[Theorem~5]{InriaRR9591}, for all $y \in\supp Q$, it follows that
\begin{IEEEeqnarray}{rCl}
\label{EqJuly7in2026at13h35B}
\RND{Q}{P_{Y\mid \vect{X} =\vect{x}}^{\autoparent{\hat{h}_{\vect{\beta}},Q,1}}}\autoparent{y}
&=&
\autoparent{
\RND{P_{Y\mid \vect{X} =\vect{x}}^{\autoparent{\hat{h}_{\vect{\beta}},Q,1}}}{Q}\autoparent{y}
}^{-1}.
\end{IEEEeqnarray}
Therefore, from Theorem~\ref{TheoremMay20at13h57in2026Antibes} and the equality in~\eqref{EqJuly7in2026at13h35B}, for all $y\in\supp Q$,
\begin{IEEEeqnarray}{rCl}
\label{EqJuly7in2026at13h35C}
\RND{S_K}{P_{Y\mid \vect{X} =\vect{x}}^{\autoparent{\hat{h}_{\vect{\beta}},Q,1}}}\autoparent{y}
&=&
\RND{S_K}{Q}\autoparent{y}
\RND{Q}{P_{Y\mid \vect{X} =\vect{x}}^{\autoparent{\hat{h}_{\vect{\beta}},Q,1}}}\autoparent{y}
\\
\label{EqJuly7in2026at13h35D}
&=&
\RND{S_K}{Q}\autoparent{y}
\autoparent{
\RND{P_{Y\mid \vect{X} =\vect{x}}^{\autoparent{\hat{h}_{\vect{\beta}},Q,1}}}{Q}\autoparent{y}
}^{-1}
\\
\label{EqJuly7in2026at13h35E}
&=&
1,
\end{IEEEeqnarray}
where the equality in~\eqref{EqJuly7in2026at13h35C} follows from the chain rule of the Radon--Nikodym derivative \cite[Theorem~4]{InriaRR9591}; the equality in~\eqref{EqJuly7in2026at13h35D} follows from~\eqref{EqJuly7in2026at13h35B}; and the equality in~\eqref{EqJuly7in2026at13h35E} follows from Theorem~\ref{TheoremMay20at13h57in2026Antibes}. This proves~\emph{(i)}.

Assume that~\emph{(i)} holds. Then,
\begin{IEEEeqnarray}{rCl}
\label{EqJuly7in2026at13h35F}
\KL{S_K}{P_{Y\mid \vect{X} =\vect{x}}^{\autoparent{\hat{h}_{\vect{\beta}},Q,1}}}
&=&
\int
\log\autoparent{
\RND{S_K}{P_{Y\mid \vect{X} =\vect{x}}^{\autoparent{\hat{h}_{\vect{\beta}},Q,1}}}\autoparent{y}
}
 \mathrm{d}S_K(y)
\\
\label{EqJuly7in2026at13h35G}
&=&
0,
\end{IEEEeqnarray}
which proves~\emph{(ii)}.
Conversely, if~\emph{(ii)} holds, then from \cite[Theorem~1]{perlaza2024empirical} it follows that
\begin{IEEEeqnarray}{rCl}
\label{EqJuly7in2026at13h35H}
\RND{S_K}{P_{Y\mid \vect{X} =\vect{x}}^{\autoparent{\hat{h}_{\vect{\beta}},Q,1}}}\autoparent{y}
&=&
1.
\end{IEEEeqnarray}
Therefore,~\emph{(i)} and~\emph{(ii)} are equivalent.

Assume next that~\emph{(i)} holds. Then $S_K$ is identical to the $\autoparent{\hat{h}_{\vect{\beta}},Q,1}$-Gibbs probability measure
$P_{Y\mid \vect{X} =\vect{x}}^{\autoparent{\hat{h}_{\vect{\beta}},Q,1}}$.
Hence, from Lemma~\ref{LemmaJuly28at10h54in2026Sophia}, it is the unique solution, if it exists, to the optimization problem in~\eqref{EqApril22at22h51in2026Nice}. This proves that~\emph{(i)} implies~\emph{(iii)}.

Assume that~\emph{(iii)} holds. From Lemma~\ref{LemmaJuly28at10h54in2026Sophia}, the unique solution, if it exists, to the optimization problem in~\eqref{EqApril22at22h51in2026Nice} is precisely the $\autoparent{\hat{h}_{\vect{\beta}},Q,1}$-Gibbs probability measure
$P_{Y\mid \vect{X} =\vect{x}}^{\autoparent{\hat{h}_{\vect{\beta}},Q,1}}$.
Hence,
\begin{IEEEeqnarray}{rCl}
\label{EqJuly7in2026at13h35I}
S_K
&=&
P_{Y\mid \vect{X} =\vect{x}}^{\autoparent{\hat{h}_{\vect{\beta}},Q,1}}.
\end{IEEEeqnarray}
Thus,
\begin{IEEEeqnarray}{rCl}
\label{EqJuly7in2026at13h35J}
\KL{S_K}{Q}
&=&
\KL{P_{Y\mid \vect{X} =\vect{x}}^{\autoparent{\hat{h}_{\vect{\beta}},Q,1}}}{Q}.
\end{IEEEeqnarray}
Taking into account that 
$P_{Y\mid \vect{X} =\vect{x}}^{\autoparent{\hat{h}_{\vect{\beta}},Q,1}}$
is a $\autoparent{\hat{h}_{\vect{\beta}},Q,1}$-Gibbs probability measure, Lemma~\ref{LemmaMarch4at9h15in2026inSophia} implies that it is the unique solution, if it exists, to the optimization problem in~\eqref{EqApril22at22h52in2026Nice}. Therefore, $S_K$ is the unique solution, if it exists, to the same optimization problem. This proves that~\emph{(iii)} implies~\emph{(iv)}.

Assume finally that~\emph{(iv)} holds. From Lemma~\ref{LemmaMarch4at9h15in2026inSophia}, the unique solution, if it exists, to the optimization problem in~\eqref{EqApril22at22h52in2026Nice} is precisely the $\autoparent{\hat{h}_{\vect{\beta}},Q,1}$-Gibbs probability measure
$P_{Y\mid \vect{X} =\vect{x}}^{\autoparent{\hat{h}_{\vect{\beta}},Q,1}}$.
Hence,
\begin{IEEEeqnarray}{rCl}
\label{EqJuly7in2026at13h35K}
S_K
&=&
P_{Y\mid \vect{X} =\vect{x}}^{\autoparent{\hat{h}_{\vect{\beta}},Q,1}}.
\end{IEEEeqnarray}
Together with the absolute-continuity relation established above and the explicit Radon--Nikodym derivatives in Theorem~\ref{TheoremMay20at13h57in2026Antibes}, it follows that, for all $y\in\supp Q$,
\begin{IEEEeqnarray}{rCl}
\label{EqJuly7in2026at13h35L}
\RND{S_K}{P_{Y\mid \vect{X} =\vect{x}}^{\autoparent{\hat{h}_{\vect{\beta}},Q,1}}}\autoparent{y}
&=&
1.
\end{IEEEeqnarray}
This proves that~\emph{(iv)} implies~\emph{(i)}.

Therefore,~\emph{(i)}, \emph{(ii)}, \emph{(iii)}, and~\emph{(iv)} are equivalent.
This concludes the proof.

\section{Proof of Theorem~\ref{TheoremApril16at15h22in2026Antibes}}
\label{AppProofOfTheoremMarch9at14h42in2026Sophia}
From Definition~\ref{DefGeneralGibbs}, for all $y \in \supp Q_1$, it follows that
\begin{IEEEeqnarray}{rCl}
\label{EqMarch9at14h55in2026SophiaA}
\RND{P^{\autoparent{h_2, P^{(h_1,Q_1,\lambda_1)}_{Y\mid X  = x_1}, \lambda_2}}_{Y\mid X =x_2}}{P^{(h_1,Q_1,\lambda_1)}_{Y\mid X  = x_1}} (y) &=& \frac{\exp\autoparent{-\frac{1}{\lambda_2} h_2(x_2, y)}}{\displaystyle \int \exp\autoparent{-\frac{1}{\lambda_2} h_2(x_2, \tilde{y})} \mathrm{d}P^{(h_1,Q_1,\lambda_1)}_{Y\mid X  = x_1} (\tilde{y})}\\
\label{EqMarch9at14h55in2026SophiaB}
&=& \frac{\exp\autoparent{-\frac{1}{\lambda_2} h_2(x_2, y)}}{\displaystyle \int \exp\autoparent{-\frac{1}{\lambda_2} h_2(x_2, \tilde{y})} \RND{P^{(h_1,Q_1,\lambda_1)}_{Y\mid X  = x_1}}{Q_1} (\tilde{y}) \mathrm{d}Q_1 (\tilde{y})}\\
\label{EqMarch9at14h55in2026SophiaC}
&=& \frac{\exp\autoparent{-\frac{1}{\lambda_2} h_2(x_2, y)} \int \exp\autoparent{-\frac{1}{\lambda_1} h_1(x_1, \hat{y})} \mathrm{d}Q_1 (\hat{y})}{\displaystyle \int \exp\autoparent{-\frac{1}{\lambda_2} h_2(x_2, \tilde{y})} \exp\autoparent{-\frac{1}{\lambda_1} h_1(x_1, \tilde{y})} \mathrm{d}Q_1 (\tilde{y})}.
\end{IEEEeqnarray}
From \cite[Lemma~3]{perlaza2024empirical} and \cite[Theorem~2]{InriaRR9591} the Radon--Nikodym derivative $\RND{Q_1}{P^{(h_1,Q_1,\lambda_1)}_{Y\mid X  = x_1}}$
is well defined and from \cite[Theorem~4]{InriaRR9591}, for all $y \in\supp Q_1$, it follows that 
\begin{IEEEeqnarray}{rCl}
\label{EqMarch9at14h57in2026Sophia}
\RND{P^{\autoparent{h_2, P^{(h_1,Q_1,\lambda_1)}_{Y\mid X  = x_1}, \lambda_2}}_{Y\mid X =x_2}}{P^{(h_1,Q_1,\lambda_1)}_{Y\mid X  = x_1}} (y) &=& \RND{P^{\autoparent{h_2, P^{(h_1,Q_1,\lambda_1)}_{Y\mid X  = x_1}, \lambda_2}}_{Y\mid X =x_2}}{Q_1} (y) \RND{Q_1}{P^{(h_1,Q_1,\lambda_1)}_{Y\mid X  = x_1}}(y).
\end{IEEEeqnarray}

Combining~\eqref{EqMarch9at14h55in2026SophiaC} and~\eqref{EqMarch9at14h57in2026Sophia} yields for all $y \in \supp Q_1$,
\begin{IEEEeqnarray}{rCl}
\nonumber
\label{EqMarch9at14h58in2026SophiaA}
&& \RND{P^{\autoparent{h_2, P^{(h_1,Q_1,\lambda_1)}_{Y\mid X  = x_1}, \lambda_2}}_{Y\mid X =x_2}}{Q_1} (y)\\
  &=& \autoparent{\RND{Q_1}{P^{(h_1,Q_1,\lambda_1)}_{Y\mid X  = x_1}}(y)}^{-1} \frac{\exp\autoparent{-\frac{1}{\lambda_2} h_2(x_2, y)} \int \exp\autoparent{-\frac{1}{\lambda_1} h_1(x_1, \hat{y})} \mathrm{d}Q_1 (\hat{y})}{\displaystyle \int \exp\autoparent{-\frac{1}{\lambda_2} h_2(x_2, \tilde{y})-\frac{1}{\lambda_1} h_1(x_1, \tilde{y})} \mathrm{d}Q_1 (\tilde{y})}\\
 \label{EqMarch9at14h58in2026SophiaB}
 &=& \RND{P^{(h_1,Q_1,\lambda_1)}_{Y\mid X  = x_1}}{Q_1}(y) \frac{\exp\autoparent{-\frac{1}{\lambda_2} h_2(x_2, y)} \int \exp\autoparent{-\frac{1}{\lambda_1} h_1(x_1, \hat{y})} \mathrm{d}Q_1 (\hat{y})}{\displaystyle \int \exp\autoparent{-\frac{1}{\lambda_2} h_2(x_2, \tilde{y}) -\frac{1}{\lambda_1} h_1(x_1, \tilde{y})} \mathrm{d}Q_1 (\tilde{y})}\\
 \nonumber
&=& \frac{\exp\autoparent{-\frac{1}{\lambda_1} h_1(x_1, y)}}{\displaystyle \int \exp\autoparent{-\frac{1}{\lambda_1} h_1(x_1, \hat{y})} \mathrm{d}Q_1 (\hat{y})} \\
 \label{EqMarch9at14h58in2026SophiaC}
& & \frac{\exp\autoparent{-\frac{1}{\lambda_2} h_2(x_2, y)} \int \exp\autoparent{-\frac{1}{\lambda_1} h_1(x_1, \hat{y})} \mathrm{d}Q_1 (\hat{y})}{\displaystyle \int \exp\autoparent{-\frac{1}{\lambda_2} h_2(x_2, \tilde{y}) -\frac{1}{\lambda_1} h_1(x_1, \tilde{y})} \mathrm{d}Q_1 (\tilde{y})} %\middlesqueezeequ
\\
\label{EqMarch9at14h58in2026SophiaD}
&=& \frac{\exp\autoparent{-\frac{1}{\lambda_1} h_1(x_1, y) -\frac{1}{\lambda_2} h_2(x_2, y)}}{\displaystyle \int \exp\autoparent{-\frac{1}{\lambda_1} h_1(x_1, \tilde{y}) -\frac{1}{\lambda_2} h_2(x_2, \tilde{y})} \mathrm{d}Q_1 (\tilde{y})}\\
\label{EqMarch9at14h58in2026SophiaE}
&=& \frac{\exp\autoparent{- \hat{h}_{\vect{\beta}}(\vect{x}, y)}}{\displaystyle \int \exp\autoparent{- \hat{h}_{\vect{\beta}}(\vect{x}, \tilde{y})} \mathrm{d}Q_1 (\tilde{y})}\\
\label{EqMarch9at14h58in2026SophiaF}
&=&\RND{P_{Y\mid \vect{X} = \vect{x}}^{\autoparent{\hat{h}_{\vect{\beta}},Q_1,1}}}{Q_1} (y),
\end{IEEEeqnarray}
where the equality in~\eqref{EqMarch9at14h58in2026SophiaB} follows from the multiplicative inverse of the Radon-Nikodym derivative \cite[Theorem~5]{InriaRR9591}; the equality in~\eqref{EqMarch9at14h58in2026SophiaC} follows from Definition~\ref{DefGeneralGibbs}; the equality in~\eqref{EqMarch9at14h58in2026SophiaE} follows 
from the definition of the function $\hat{h}_{\vect{\beta}}$ in~\eqref{EqApril29at13h53in2026Sophia} with $\vect{x} = (x_1,x_2)$ and $\vect{\beta} = \left(\frac{1}{\lambda_1}, \frac{1}{\lambda_2}\right)$; and the equality in~\eqref{EqMarch9at14h58in2026SophiaF} follows from
Definition~\ref{DefGeneralGibbs}.  This completes the proof.

\section{Proof of Theorem~\ref{TheoremMay19at113h22in2026Antibes}}
\label{AppProofOfTheoremMay19at113h22in2026Antibes}

The proof of~\eqref{EqMay19at13h30in2026SophiaA}  is done by induction on $K \in \mathbb{N}$.

\medskip
\noindent\emph{Base cases ($K=1$ and $K=2$):}

\noindent When $K=1$, from~\eqref{EqNestedReferenceMeasures}, the reference measure is $Q_1$.
Hence, from Definition~\ref{DefGeneralGibbs}, for all $y\in\supp Q_1$, it follows that
\begin{IEEEeqnarray}{rCl}
\label{EqJuly7in2026at11h27A}
\RND{P^{\autoparent{h_1,Q_1,\lambda_1}}_{Y \mid X  = x_1}}{Q_1}\autoparent{y}
&=&
\frac{\exp\autoparent{-\frac{1}{\lambda_1}h_1\autoparent{x_1,y}}}
{\displaystyle \int \exp\autoparent{-\frac{1}{\lambda_1}h_1\autoparent{x_1,\tilde{y}}} \mathrm{d}Q_1(\tilde{y})}.
\end{IEEEeqnarray}
Thus, the equality holds for $K=1$.

\medskip
\noindent When $K=2$, from~\eqref{EqNestedReferenceMeasures}, the reference measure $Q_2$ satisfies $Q_2 = P^{\autoparent{h_1,Q_1,\lambda_1}}_{Y \mid X  = x_1}$.
Therefore, from Theorem~\ref{TheoremApril16at15h22in2026Antibes}, for all $y\in\supp Q_1$, it follows that
\begin{IEEEeqnarray}{rCl}
\label{EqJuly7in2026at11h27C}
\RND{P^{\autoparent{h_2,Q_2,\lambda_2}}_{Y\mid X =x_2}}{Q_1}\autoparent{y}
&=&
\frac{\exp\autoparent{-\frac{1}{\lambda_1}h_1\autoparent{x_1,y}
-\frac{1}{\lambda_2}h_2\autoparent{x_2,y}}}
{\displaystyle \int
\exp\autoparent{-\frac{1}{\lambda_1}h_1\autoparent{x_1,\tilde{y}}
-\frac{1}{\lambda_2}h_2\autoparent{x_2,\tilde{y}}}
 \mathrm{d}Q_1(\tilde{y})}.
\end{IEEEeqnarray}
Thus, the equality holds for $K=2$.
%Hence, it remains to prove the result for $k\geqslant 2$.

\medskip
\noindent\emph{Induction hypothesis:}
Assume that~\eqref{EqMay19at13h30in2026SophiaA} holds for $K\geqslant 2$, that is, for all $y\in\supp Q_1$,
\begin{IEEEeqnarray}{rCl}
\label{EqJuly7in2026at11h27E}
\RND{P^{\autoparent{h_K,Q_K,\lambda_K}}_{Y \mid X = x_K}}{Q_1}\autoparent{y}
&=&
\frac{\exp\autoparent{- \displaystyle \sum_{i=1}^{K} \frac{1}{\lambda_i} h_i\autoparent{x_i,y}}}
{\displaystyle \int \exp\autoparent{- \displaystyle \sum_{i=1}^{K} \frac{1}{\lambda_i} h_i\autoparent{x_i,\tilde{y}}} \mathrm{d}Q_1(\tilde{y})}.
\end{IEEEeqnarray}
%.

\medskip
\noindent\emph{Induction step ($K\rightarrow K+1$):}
It remains to prove~\eqref{EqMay19at13h30in2026SophiaA} for $K+1$. Consider the Gibbs probability measure
$P^{\autoparent{h_{K+1},Q_{K+1},\lambda_{K+1}}}_{Y \mid X =x_{K+1}}$
and its reference measure $Q_{K+1}$ satisfying~\eqref{EqNestedReferenceMeasures}.
From Definition~\ref{DefGeneralGibbs}, for all $y\in\supp Q_{K+1}$, it follows that
\begin{IEEEeqnarray}{rCl}
\label{EqJuly7in2026at11h28A}
\RND{P^{\autoparent{h_{K+1},Q_{K+1},\lambda_{K+1}}}_{Y \mid X =x_{K+1}}}{Q_{K+1}}\autoparent{y}
&=&
\frac{\exp\autoparent{-\frac{1}{\lambda_{K+1}}h_{K+1}\autoparent{x_{K+1},y}}}
{\displaystyle \int \exp\autoparent{-\frac{1}{\lambda_{K+1}}h_{K+1}\autoparent{x_{K+1},\tilde{y}}} \mathrm{d}Q_{K+1}(\tilde{y})},
%\\
%\label{EqJuly7in2026at11h28B}
%&=&
%\frac{\exp\autoparent{-\frac{1}{\lambda_{k+1}}h_{k+1}\autoparent{x_{k+1},y}}}
%{\displaystyle \int \exp\autoparent{-\frac{1}{\lambda_{k+1}}h_{k+1}\autoparent{x_{k+1},\tilde{y}}}\,
%\RND{Q_{k+1}}{Q_k}\autoparent{\tilde{y}}
%\RND{Q_k}{Q_{k-1}}\autoparent{\tilde{y}}
%\ldots
%\RND{Q_3}{Q_2}\autoparent{\tilde{y}}
%\RND{Q_2}{Q_1}\autoparent{\tilde{y}}
% \mathrm{d}Q_1(\tilde{y})},
\end{IEEEeqnarray}
%where the equality in~\eqref{EqJuly7in2026at11h28B} follows from recursively using the chain rule of the Radon--Nikodym derivative \cite[Theorem~4]{InriaRR9591} and 
The proof follows from the fact that Gibbs probability measures and their reference measures are mutually absolutely continuous~\cite[Lemma~3]{perlaza2024empirical}, which yields
\begin{IEEEeqnarray}{rCl}
\label{EqJuly7in2026at11h28G}
Q_{1}
&\ll&
Q_2
\ll
\ldots Q_{K-1}
\ll
Q_{K}
\ll Q_{K+1}
\ll
Q_1.
\end{IEEEeqnarray}
Thus, the measures $Q_{K+1}$ and $Q_1$ are mutually absolutely continuous and from \cite[Lemma~3]{perlaza2024empirical} and \cite[Theorem~2]{InriaRR9591}, the Radon--Nikodym derivatives
$\RND{Q_{k+1}}{Q_1}$ and $\RND{Q_{1}}{Q_{K+1}}$ are well defined.
From the chain rule of the Radon-Nikodym derivative \cite[Theorem~4]{InriaRR9591}, for all $y\in\supp Q_1$, it follows that 
\begin{IEEEeqnarray}{rCl}
\label{EqJuly7in2026at12h10}
\RND{P^{\autoparent{h_{K+1},Q_{K+1},\lambda_{K+1}}}_{Y \mid X =x_{K+1}}}{Q_{K+1}}\autoparent{y}
&=&
\RND{P^{\autoparent{h_{K+1},Q_{K+1},\lambda_{K+1}}}_{Y \mid X =x_{K+1}}}{Q_1}\autoparent{y}
\RND{Q_1}{Q_{K+1}}\autoparent{y}.
\end{IEEEeqnarray}

Moreover, from the multiplicative inverse of the Radon--Nikodym derivative \cite[Theorem~5]{InriaRR9591}, by combining~\eqref{EqJuly7in2026at11h28A} and~\eqref{EqJuly7in2026at12h10}, for all $y\in\supp Q_1$, it follows that

\begin{IEEEeqnarray}{rCl}
\label{EqJuly7in2026at11h28B}
\nonumber
&&\RND{P^{\autoparent{h_{K+1},Q_{K+1},\lambda_{K+1}}}_{Y \mid X =x_{K+1}}}{Q_1}\autoparent{y}\\
&=&
\RND{Q_{K+1}}{Q_1}\autoparent{y}
\frac{\exp\autoparent{-\frac{1}{\lambda_{K+1}}h_{K+1}\autoparent{x_{K+1},y}}}
{\displaystyle \int \exp\autoparent{-\frac{1}{\lambda_{K+1}}h_{K+1}\autoparent{x_{K+1},\tilde{y}}} \mathrm{d}Q_{K+1}(\tilde{y})}\\
\label{EqJuly7in2026at11h28C}
&=&
\RND{Q_{K+1}}{Q_1}\autoparent{y}
\frac{\exp\autoparent{-\frac{1}{\lambda_{K+1}}h_{K+1}\autoparent{x_{K+1},y}}}
{\displaystyle \int \exp\autoparent{-\frac{1}{\lambda_{K+1}}h_{K+1}\autoparent{x_{K+1},\tilde{y}}}\,
\RND{Q_{K+1}}{Q_1}\autoparent{\tilde{y}}
 \mathrm{d}Q_1(\tilde{y})}
\\
\label{EqJuly7in2026at11h28D}
&=&
\RND{P^{\autoparent{h_K,Q_K,\lambda_K}}_{Y \mid X = x_K}}{Q_1}\autoparent{y}
\frac{\exp\autoparent{-\frac{1}{\lambda_{K+1}}h_{K+1}\autoparent{x_{K+1},y}}}
{\displaystyle \int \exp\autoparent{-\frac{1}{\lambda_{K+1}}h_{K+1}\autoparent{x_{K+1},\tilde{y}}}\,
\RND{P^{\autoparent{h_K,Q_K,\lambda_K}}_{Y \mid X = x_K}}{Q_1}\autoparent{\tilde{y}}
 \mathrm{d}Q_1(\tilde{y})}
\\
\label{EqJuly7in2026at11h28E}
&=&
\frac{\exp\autoparent{- \displaystyle \sum_{i=1}^{K} \frac{1}{\lambda_i} h_i\autoparent{x_i,y}}}
{\displaystyle \int \exp\autoparent{- \displaystyle \sum_{i=1}^{K} \frac{1}{\lambda_i} h_i\autoparent{x_i,\tilde{y}}} \mathrm{d}Q_1(\tilde{y})}
\frac{\exp\autoparent{-\frac{1}{\lambda_{K+1}}h_{K+1}\autoparent{x_{K+1},y}}}
{\displaystyle \int \exp\autoparent{-\frac{1}{\lambda_{K+1}}h_{K+1}\autoparent{x_{K+1},\tilde{y}}}\,
\frac{\exp\autoparent{- \displaystyle \sum_{i=1}^{K} \frac{1}{\lambda_i} h_i\autoparent{x_i,\tilde{y}}}}
{\displaystyle \int \exp\autoparent{- \displaystyle \sum_{i=1}^{K} \frac{1}{\lambda_i} h_i\autoparent{x_i,\hat{y}}} \mathrm{d}Q_1(\hat{y})}
 \mathrm{d}Q_1(\tilde{y})}
\\
\label{EqJuly7in2026at11h28F}
&=&
\frac{
\exp\autoparent{- \displaystyle \sum_{i=1}^{K+1} \frac{1}{\lambda_i}h_i\autoparent{x_i,y}}
}
{\displaystyle \int \exp\autoparent{- \displaystyle \sum_{i=1}^{K+1} \frac{1}{\lambda_i}h_i\autoparent{x_i,\tilde{y}}}\,
\mathrm{d}Q_1(\tilde{y})},
\end{IEEEeqnarray}
where the equality in~\eqref{EqJuly7in2026at11h28C} follows from a change of measure of the Radon--Nikodym derivative \cite[Theorem~2]{InriaRR9591}; the equality in~\eqref{EqJuly7in2026at11h28D} follows from the definition of $Q_{K+1}$ in~\eqref{EqNestedReferenceMeasures}; the equality in~\eqref{EqJuly7in2026at11h28E} follows from the induction hypothesis in~\eqref{EqJuly7in2026at11h27E} and the equality in~\eqref{EqJuly7in2026at11h28F} follows by simplifying the common normalizing factor.
Thus,~\eqref{EqMay19at13h30in2026SophiaA} holds for $K+1$. This concludes the proof of~\eqref{EqMay19at13h30in2026SophiaA}.

It remains to prove~\eqref{EqMay19at13h30in2026SophiaB}. 
%From the definition of $\vect{\beta}$ in the statement of the theorem, it holds that
%\begin{IEEEeqnarray}{rCl}
%\label{EqJuly7in2026at11h27T}
%\vect{\beta}
%&=&
%\autoparent{\frac{1}{\lambda_1}, \frac{1}{\lambda_2}, \ldots, \frac{1}{\lambda_k}}.
%\end{IEEEeqnarray}
From the definition of $\vect{\beta}$ in the statement of Theorem~\ref{TheoremMay19at113h22in2026Antibes} and the definition of the function $\hat{h}_{\vect{\beta}}$ in~\eqref{EqApril29at13h53in2026Sophia}, for all $y\in\supp Q_1$, it follows that
\begin{IEEEeqnarray}{rCl}
\label{EqJuly7in2026at11h27U}
\hat{h}_{\vect{\beta}}\autoparent{\vect{x},y}
&=&
\displaystyle \sum_{i=1}^{K}\frac{1}{\lambda_i}h_i\autoparent{x_i,y}.
\end{IEEEeqnarray}
Therefore, from Definition~\ref{DefGeneralGibbs},
\begin{IEEEeqnarray}{rCl}
\nonumber
\label{EqJuly7in2026at11h27V}
\RND{P_{Y\mid \vect{X} = \vect{x}}^{\autoparent{\hat{h}_{\vect{\beta}},Q_1,1}}}{Q_1} (y)
&=&
\frac{
\exp\autoparent{
-\hat{h}_{\vect{\beta}}\autoparent{\vect{x},y}
}
}{
\int
\exp\autoparent{
-\hat{h}_{\vect{\beta}}\autoparent{\vect{x},\tilde{y}}
}
 \mathrm{d}Q_1(\tilde{y})
}
\\
\label{EqJuly7in2026at11h27W}
&=&
\frac{\exp\autoparent{- \displaystyle \sum_{i=1}^{K} \frac{1}{\lambda_i} h_i\autoparent{x_i,y}}}
{\displaystyle \int \exp\autoparent{- \displaystyle \sum_{i=1}^{K} \frac{1}{\lambda_i} h_i\autoparent{x_i,\tilde{y}}} \mathrm{d}Q_1(\tilde{y})}.
\end{IEEEeqnarray}
Combining~\eqref{EqMay19at13h30in2026SophiaA} and~\eqref{EqJuly7in2026at11h27W} yields
\begin{IEEEeqnarray}{rCl}
\label{EqJuly7in2026at11h27X}
\RND{P^{\autoparent{h_K,Q_K,\lambda_K}}_{Y \mid X = x_K}}{Q_1}\autoparent{y}
&=&
\RND{P_{Y\mid \vect{X} = \vect{x}}^{\autoparent{\hat{h}_{\vect{\beta}},Q_1,1}}}{Q_1} (y),
\end{IEEEeqnarray}
which proves~\eqref{EqMay19at13h30in2026SophiaB}. This completes the proof.

\section{Proof of Theorem~\ref{TheoremMay19at13h57in2026Sophia}}
\label{AppProofOfTheoremMay19at13h57in2026Sophia}

From \cite[Lemma~3]{perlaza2024empirical}, the nested Gibbs probability measure
$P^{\autoparent{h_K,Q_K,\lambda_K}}_{Y \mid X = x_K}$ in \eqref{EqAugust28at11h54in2026Sophia} is mutually absolutely continuous
with its reference measure $Q_K$. Moreover, from~\eqref{EqNestedReferenceMeasures} and
\cite[Lemma~3]{perlaza2024empirical}, it follows that
\begin{IEEEeqnarray}{rCl}
\label{EqJuly7in2026at12h42A}
P^{\autoparent{h_K,Q_K,\lambda_K}}_{Y \mid X = x_K}
&\ll&
Q_1
\ll
Q_{2}
\ll
\ldots
\ll
Q_{K-1}
\ll
Q_{K}
\ll
P^{\autoparent{h_K,Q_K,\lambda_K}}_{Y \mid X = x_K}.
\end{IEEEeqnarray}
Moreover, taking into account that 
$P_{Y\mid \vect{X} = \vect{x}}^{\autoparent{\hat{h}_{\vect{\beta}},Q_1,1}}$ in~\eqref{EqMay19at13h30in2026SophiaB}
is a Gibbs probability measure with reference measure $Q_1$, from
\cite[Lemma~3]{perlaza2024empirical}, it also follows that
\begin{IEEEeqnarray}{rCl}
\label{EqJuly7in2026at12h42B}
Q_1
&\ll&
P_{Y\mid \vect{X} = \vect{x}}^{\autoparent{\hat{h}_{\vect{\beta}},Q_1,1}} \ll Q_1.
\end{IEEEeqnarray}
Thus, from \cite[Theorem~2]{InriaRR9591}, the Radon--Nikodym derivative
\[
\RND{Q_1}{P_{Y\mid \vect{X} = \vect{x}}^{\autoparent{\hat{h}_{\vect{\beta}},Q_1,1}}}
\]
is well defined and from the multiplicative inverse of the Radon--Nikodym derivative
\cite[Theorem~5]{InriaRR9591}, for all $y\in\supp Q_1$, it follows that
\begin{IEEEeqnarray}{rCl}
\label{EqJuly7in2026at12h42C}
\RND{Q_1}{P_{Y\mid \vect{X} = \vect{x}}^{\autoparent{\hat{h}_{\vect{\beta}},Q_1,1}}}\autoparent{y}
&=&
\autoparent{
\RND{P_{Y\mid \vect{X} = \vect{x}}^{\autoparent{\hat{h}_{\vect{\beta}},Q_1,1}}}{Q_1}\autoparent{y}
}^{-1}.
\end{IEEEeqnarray}

Therefore, from Theorem~\ref{TheoremMay19at113h22in2026Antibes} and the equality in
\eqref{EqJuly7in2026at12h42C}, for all $y\in\supp Q_1$,
\begin{IEEEeqnarray}{rCl}
\label{EqJuly7in2026at12h42D}
&&
\RND{P^{\autoparent{h_K,Q_K,\lambda_K}}_{Y \mid X = x_K}}
{P_{Y\mid \vect{X} = \vect{x}}^{\autoparent{\hat{h}_{\vect{\beta}},Q_1,1}}}\autoparent{y}
\nonumber\\
&=&
\RND{P^{\autoparent{h_K,Q_K,\lambda_K}}_{Y \mid X = x_K}}{Q_1}\autoparent{y}
\RND{Q_1}{P_{Y\mid \vect{X} = \vect{x}}^{\autoparent{\hat{h}_{\vect{\beta}},Q_1,1}}}\autoparent{y}
\\
\label{EqJuly7in2026at12h42E}
&=&
1,
\end{IEEEeqnarray}
where the equality in~\eqref{EqJuly7in2026at12h42D} follows from the chain rule of the
Radon--Nikodym derivative \cite[Theorem~4]{InriaRR9591}; and the equality in
\eqref{EqJuly7in2026at12h42E} follows from Theorem~\ref{TheoremMay19at113h22in2026Antibes}.
This proves~\emph{(i)}.

Assume that~\emph{(i)} holds. Then,
\begin{IEEEeqnarray}{rCl}
\nonumber
&& \quad 
\KL{P^{\autoparent{h_K,Q_K,\lambda_K}}_{Y \mid X = x_K}}
{P_{Y\mid \vect{X} = \vect{x}}^{\autoparent{\hat{h}_{\vect{\beta}},Q_1,1}}}
\\
\nonumber
&=&
\int
\log\autoparent{
\RND{P^{\autoparent{h_K,Q_K,\lambda_K}}_{Y \mid X = x_K}}
{P_{Y\mid \vect{X} = \vect{x}}^{\autoparent{\hat{h}_{\vect{\beta}},Q_1,1}}}\autoparent{y}
}
 \mathrm{d}P^{\autoparent{h_K,Q_K,\lambda_K}}_{Y \mid X = x_K}(y)
\\
\label{EqJuly7in2026at12h42F}
&=&
0,
\end{IEEEeqnarray}
which proves~\emph{(ii)}. Conversely, if~\emph{(ii)} holds, then from
\cite[Theorem~1]{perlaza2024empirical}, it follows that
\begin{IEEEeqnarray}{rCl}
\label{EqJuly7in2026at12h42G}
\RND{P^{\autoparent{h_K,Q_K,\lambda_K}}_{Y \mid X = x_K}}
{P_{Y\mid \vect{X} = \vect{x}}^{\autoparent{\hat{h}_{\vect{\beta}},Q_1,1}}}\autoparent{y}
&=&
1.
\end{IEEEeqnarray}
Therefore,~\emph{(i)} and~\emph{(ii)} are equivalent.

Assume next that~\emph{(i)} holds. Then
$P^{\autoparent{h_K,Q_K,\lambda_K}}_{Y \mid X = x_K}$
is identical to the
$\autoparent{\hat{h}_{\vect{\beta}},Q_1,1}$-Gibbs probability measure
\[
P_{Y\mid \vect{X} = \vect{x}}^{\autoparent{\hat{h}_{\vect{\beta}},Q_1,1}}.
\]
Hence, from Lemma~\ref{LemmaJuly28at10h54in2026Sophia}, it is the unique solution, if it exists,
to the optimization problem in~\eqref{EqMarch17at6h43in2026BusToSophiaA}. This proves
that~\emph{(i)} implies~\emph{(iii)}.

Assume that~\emph{(iii)} holds. From Lemma~\ref{LemmaJuly28at10h54in2026Sophia}, the unique
solution, if it exists, to the optimization problem in~\eqref{EqMarch17at6h43in2026BusToSophiaA}
is precisely the $\autoparent{\hat{h}_{\vect{\beta}},Q_1,1}$-Gibbs probability measure
\[
P_{Y\mid \vect{X} = \vect{x}}^{\autoparent{\hat{h}_{\vect{\beta}},Q_1,1}}.
\]
Hence,
\begin{IEEEeqnarray}{rCl}
\label{EqJuly7in2026at12h42H}
P^{\autoparent{h_K,Q_K,\lambda_K}}_{Y \mid X = x_K}
&=&
P_{Y\mid \vect{X} = \vect{x}}^{\autoparent{\hat{h}_{\vect{\beta}},Q_1,1}}.
\end{IEEEeqnarray}
Thus,
\begin{IEEEeqnarray}{rCl}
\label{EqJuly7in2026at12h42I}
\KL{P^{\autoparent{h_K,Q_K,\lambda_K}}_{Y \mid X = x_K}}{Q_1}
&=&
\KL{P_{Y\mid \vect{X} = \vect{x}}^{\autoparent{\hat{h}_{\vect{\beta}},Q_1,1}}}{Q_1}.
\end{IEEEeqnarray}
Taking into account that 
$P_{Y\mid \vect{X} = \vect{x}}^{\autoparent{\hat{h}_{\vect{\beta}},Q_1,1}}$
is a $\autoparent{\hat{h}_{\vect{\beta}},Q_1,1}$-Gibbs probability measure,
Lemma~\ref{LemmaMarch4at9h15in2026inSophia} implies that it is the unique solution, if it exists,
to the optimization problem in~\eqref{EqMarch18at9h58in2026SophiaA}. Therefore,
$P^{\autoparent{h_K,Q_K,\lambda_K}}_{Y \mid X = x_K}$
is the unique solution, if it exists, to the same optimization problem. This proves
that~\emph{(iii)} implies~\emph{(iv)}.

Finally, assume that~\emph{(iv)} holds. From Lemma~\ref{LemmaMarch4at9h15in2026inSophia}, the unique
solution, if it exists, to the optimization problem in~\eqref{EqMarch18at9h58in2026SophiaA}
is precisely the $\autoparent{\hat{h}_{\vect{\beta}},Q_1,1}$-Gibbs probability measure
\[
P_{Y\mid \vect{X} = \vect{x}}^{\autoparent{\hat{h}_{\vect{\beta}},Q_1,1}}.
\]
Hence,
\begin{IEEEeqnarray}{rCl}
\label{EqJuly7in2026at12h42J}
P^{\autoparent{h_K,Q_K,\lambda_K}}_{Y \mid X = x_K}
&=&
P_{Y\mid \vect{X} = \vect{x}}^{\autoparent{\hat{h}_{\vect{\beta}},Q_1,1}}.
\end{IEEEeqnarray}
Therefore, for all $y\in\supp Q_1$,
\begin{IEEEeqnarray}{rCl}
\label{EqJuly7in2026at12h42K}
\RND{P^{\autoparent{h_K,Q_K,\lambda_K}}_{Y \mid X = x_K}}
{P_{Y\mid \vect{X} = \vect{x}}^{\autoparent{\hat{h}_{\vect{\beta}},Q_1,1}}}\autoparent{y}
&=&
1.
\end{IEEEeqnarray}
This proves that~\emph{(iv)} implies~\emph{(i)}.

Therefore,~\emph{(i)}, \emph{(ii)}, \emph{(iii)}, and~\emph{(iv)} are equivalent.
This concludes the proof.

\section{Proof Of Lemma~\ref{LemmaSept11at15h55in2026Sophia}}
\label{AppProofOfLemmaSept11at15h55in2026Sophia}

For all $i\in\{1,2,\ldots,K-1\}$, using~\cite[Lemma~3]{PerlazaEntropy2025}
with the Gibbs probability measures
$P^{\autoparent{h_i,Q_i,\lambda_i}}_{Y\mid X=x_i}$
 and 
$P^{\autoparent{h_K,Q_K,\lambda_K}}_{Y\mid X=x_K}$,
it follows that 
\begin{IEEEeqnarray}{rCl}
\label{EqSept22ProofGapsA}
&&
\frac{1}{\lambda_i}
\left(
\int h_i(x_i,y)
 \mathrm{d}
P^{\autoparent{h_K,Q_K,\lambda_K}}_{Y\mid X=x_K}(y)
-
\int h_i(x_i,y)
 \mathrm{d}Q_{i+1}(y)
\right)
+
\KL{
P^{\autoparent{h_K,Q_K,\lambda_K}}_{Y\mid X=x_K}
}{Q_i}
\nonumber\\
&=&
\KL{Q_{i+1}}{Q_i}
+
\KL{
P^{\autoparent{h_K,Q_K,\lambda_K}}_{Y\mid X=x_K}
}{Q_{i+1}},
\end{IEEEeqnarray}
where
$Q_{i+1}=P^{\autoparent{h_i,Q_i,\lambda_i}}_{Y\mid X=x_i}$
from~\eqref{EqNestedReferenceMeasures}.
Using the definition of $\delta_i$
in~\eqref{EqSept11at15h41in2026Sophia},
the equality in~\eqref{EqSept22ProofGapsA}
can be written as
\begin{IEEEeqnarray}{rCl}
\label{EqSept22ProofGapsB}
\KL{
P^{\autoparent{h_K,Q_K,\lambda_K}}_{Y\mid X=x_K}
}{Q_i}
&=&
\KL{
P^{\autoparent{h_i,Q_i,\lambda_i}}_{Y\mid X=x_i}
}{Q_i}
-\delta_i
+
\KL{
P^{\autoparent{h_K,Q_K,\lambda_K}}_{Y\mid X=x_K}
}{Q_{i+1}}.
\end{IEEEeqnarray}

Hence, from \eqref{EqSept22ProofGapsB},  it follows that 
\begin{IEEEeqnarray}{rCl}
\label{EqSept22ProofGapsC}
\KL{
P^{\autoparent{h_K,Q_K,\lambda_K}}_{Y\mid X=x_K}
}{Q_1}
&=&
\label{EqSept22ProofGapsD}
\KL{
P^{\autoparent{h_1,Q_1,\lambda_1}}_{Y\mid X=x_1}
}{Q_1}
-\delta_1+
\KL{
P^{\autoparent{h_K,Q_K,\lambda_K}}_{Y\mid X=x_K}
}{Q_2}
\\
\label{EqSept22ProofGapsE}
&=&
\KL{
P^{\autoparent{h_1,Q_1,\lambda_1}}_{Y\mid X=x_1}
}{Q_1}
-\delta_1+
\KL{
P^{\autoparent{h_2,Q_2,\lambda_2}}_{Y\mid X=x_2}
}{Q_2}
-\delta_2
+
\KL{
P^{\autoparent{h_K,Q_K,\lambda_K}}_{Y\mid X=x_K}
}{Q_{3}}
\\
\label{EqSept22ProofGapsF}
&=&
\sum_{i=1}^{2}
\left(
\KL{
P^{\autoparent{h_i,Q_i,\lambda_i}}_{Y\mid X=x_i}
}{Q_i}
-\delta_i
\right)
+
\KL{
P^{\autoparent{h_3,Q_3,\lambda_3}}_{Y\mid X=x_3}
}{Q_3}
-\delta_3
+
\KL{
P^{\autoparent{h_K,Q_K,\lambda_K}}_{Y\mid X=x_K}
}{Q_{4}}
\\
\nonumber
& \vdots &\\
\nonumber
\label{EqSept22ProofGapsG}
&=&
\sum_{i=1}^{K-2}
\left(
\KL{
P^{\autoparent{h_i,Q_i,\lambda_i}}_{Y\mid X=x_i}
}{Q_i}
-\delta_i
\right)
+\KL{
P^{\autoparent{h_{K-1},Q_{K-1},\lambda_{K-1}}}_{Y\mid X=x_{K-1}}
}{Q_{K-1}}
-\delta_{K-1}\\
&&
+
\KL{
P^{\autoparent{h_K,Q_K,\lambda_K}}_{Y\mid X=x_K}
}{Q_K}
\\
\label{EqSept22ProofGapsH}
&=&
\sum_{k=1}^{K}
\KL{
P^{\autoparent{h_k,Q_k,\lambda_k}}_{Y\mid X=x_k}
}{Q_k}
-
\sum_{i=1}^{K-1}\delta_i, 
\end{IEEEeqnarray}
where
the equality in~\eqref{EqSept22ProofGapsE}
follows by applying~\eqref{EqSept22ProofGapsB}
to the term
$\KL{
P^{\autoparent{h_K,Q_K,\lambda_K}}_{Y\mid X=x_K}
}{Q_2}$
in~\eqref{EqSept22ProofGapsD};
the equality in~\eqref{EqSept22ProofGapsF}
follows by applying~\eqref{EqSept22ProofGapsB}
to the term
$\KL{
P^{\autoparent{h_K,Q_K,\lambda_K}}_{Y\mid X=x_K}
}{Q_3}$
in~\eqref{EqSept22ProofGapsE}
and rearranging the terms;
the equality in~\eqref{EqSept22ProofGapsG}
follows by successively applying~\eqref{EqSept22ProofGapsB}
to the last relative entropy until the application
with $i=K-1$, which yields
$\KL{
P^{\autoparent{h_K,Q_K,\lambda_K}}_{Y\mid X=x_K}
}{Q_K}$
as the last relative entropy;
and the equality in~\eqref{EqSept22ProofGapsH}
follows by rearranging the terms.
This completes the proof.

\section{Proof Of Lemma~\ref{LemmaSept24at20h24in2026Antibes}}\label{AppProofOfLemmaSept24at20h24in2026Antibes}
Taking into consideration that $\vect{z}_0$ is the aggregation of the $K$ local datasets,
for all $\vect{\theta}\in\set{M}$, it follows from~\eqref{EqEmpiricalRiskLn} that
\begin{IEEEeqnarray}{rCl}
\label{EqSept24at18h44in2026Antibes}
\sum_{i=1}^{K-1}
n_i\mathsf{L}_{n_i}
\autoparent{\vect{z}_i,\vect{\theta}}
&=&
n_0\mathsf{L}_{n_0}
\autoparent{\vect{z}_0,\vect{\theta}}
-
n_K\mathsf{L}_{n_K}
\autoparent{\vect{z}_K,\vect{\theta}}.
\end{IEEEeqnarray}
Hence,
\begin{IEEEeqnarray}{rCl}
\sum_{i=1}^{K-1}n_i
\mathsf{R}_{\vect{z}_i}\autoparent{
P^{(\mathsf{L}_{n_0},Q_1,\lambda_0)}_{
\vect{\Theta}\mid\vect{Z}=\vect{z}_0}}
\label{EqSept24at18h50in2026AntibesA}
&=&
\int
\left(
\sum_{i=1}^{K-1}
n_i\mathsf{L}_{n_i}
\autoparent{\vect{z}_i,\vect{\theta}}
\right)
\mathrm{d}
P^{(\mathsf{L}_{n_0},Q_1,\lambda_0)}_{
\vect{\Theta}\mid\vect{Z}=\vect{z}_0}
\autoparent{\vect{\theta}}
\\
\label{EqSept24at18h50in2026AntibesB}
&=&
n_0\mathsf{R}_{\vect{z}_0}\autoparent{
P^{(\mathsf{L}_{n_0},Q_1,\lambda_0)}_{
\vect{\Theta}\mid\vect{Z}=\vect{z}_0}}
-
n_K\mathsf{R}_{\vect{z}_K}\autoparent{
P^{(\mathsf{L}_{n_0},Q_1,\lambda_0)}_{
\vect{\Theta}\mid\vect{Z}=\vect{z}_0}}
\\
\label{EqSept24at18h50in2026AntibesC}
&=&
n_0\mathsf{R}_{\vect{z}_0}\autoparent{
P^{(\mathsf{L}_{n_0},Q_1,\lambda_0)}_{
\vect{\Theta}\mid\vect{Z}=\vect{z}_0}}
-
n_K\mathsf{R}_{\vect{z}_K}\autoparent{
P^{\autoparent{\mathsf{L}_{n_K},Q_K,\lambda_K}}_{
\vect{\Theta}\mid\vect{Z}=\vect{z}_K}},
\end{IEEEeqnarray}
where the equality in~\eqref{EqSept24at18h50in2026AntibesA}
follows from~\eqref{EqExpEmpRiskModels}
and the linearity of the integral;
the equality in~\eqref{EqSept24at18h50in2026AntibesB}
follows from~\eqref{EqSept24at18h44in2026Antibes};
and the equality in~\eqref{EqSept24at18h50in2026AntibesC}
follows from~\eqref{EqJanuary1at16h30in2026atHomeNice}.
Substituting~\eqref{EqSept24at18h50in2026AntibesC}
into~\eqref{EqDecentralizedTriangleEquality} yields
\begin{IEEEeqnarray}{rCl}
\sum_{k=1}^{K}\gamma_k-\eta
&=&
\frac{1}{\lambda_0}
\Bigg(
\mathsf{R}_{\vect{z}_0}\autoparent{
P^{(\mathsf{L}_{n_0},Q_1,\lambda_0)}_{
\vect{\Theta}\mid\vect{Z}=\vect{z}_0}}
-
\sum_{i=1}^{K}\frac{n_i}{n_0}
\mathsf{R}_{\vect{z}_i}\autoparent{
P^{\autoparent{\mathsf{L}_{n_i},Q_i,\lambda_i}}_{
\vect{\Theta}\mid\vect{Z}=\vect{z}_i}}
\Bigg),
\end{IEEEeqnarray}
which completes the proof.

\end{document}

%% file: Figures/Journal-TIT2026/renormalization_geometry.tikz
\begin{tikzpicture}[
  x=1cm, y=1cm,
  line cap=round, line join=round,
  level/.style={circle, inner sep=0pt, outer sep=0pt},
  levelone/.style={
    level, draw=red!75!black, fill=red!6,
    solid, line width=0.85pt
  },
  leveltwo/.style={
    level, draw=blue!70!black, fill=blue!6,
    dotted, line width=1.1pt
  },
renormalization/.style={
  draw=violet!80!black,
  line width=1.0pt,
  -{Stealth[length=6pt,width=5pt]},
  shorten <=4pt,
  shorten >=5pt
},
  entropyone/.style={draw=red!75!black, thick},
  entropytwo/.style={draw=blue!70!black, thick},
  point/.style={font=\Large, inner sep=0pt},
  measure/.style={font=\large, inner sep=1pt},
  setlabel/.style={font=\large, inner sep=1pt}
]

% Layout parameters: lengths are in centimetres.
\def\RadiusQ{2.15}
\def\RadiusOne{2.35}
\def\CircleGap{4.50}
\def\AngleP{0}
\def\AngleR{180}
\def\CurveRise{1.20}
\def\SetLabelGap{-0.8}
\def\CurveLabelGap{0.35}
\def\FramePadding{1.00}

% Circle diameters.
\pgfmathsetlengthmacro{\DiameterQ}{2*\RadiusQ*1cm}
\pgfmathsetlengthmacro{\DiameterOne}{2*\RadiusOne*1cm}

% Relative-entropy level sets.
\coordinate (CQ) at (0,0);
\coordinate (C1)
  at ($(CQ)+({\RadiusQ+\CircleGap+\RadiusOne},0)$);

\node[leveltwo, minimum size=\DiameterQ]
  (GQ) at (CQ) {};
\node[levelone, minimum size=\DiameterOne]
  (GOne) at (C1) {};

% Points anchored to the circle boundaries.
\coordinate (P) at (GQ.\AngleP);
\coordinate (R1) at (GOne.\AngleR);

% Relative-entropy radii.
\draw[entropytwo] (GQ.center) --
  node[midway, fill=white, inner sep=1pt, text=blue!70!black]
    {$\gamma$}
  (P);

\draw[entropyone] (R1) --
  node[midway, fill=white, inner sep=1pt, text=red!75!black]
    {$\eta_1$}
  (GOne.center);

% Renormalization curve.
\coordinate (ControlP)
  at ($(P)!0.333333!(R1)+(0,\CurveRise)$);
\coordinate (ControlR)
  at ($(P)!0.666667!(R1)+(0,\CurveRise)$);

\draw[renormalization]
  (P) .. controls (ControlP) and (ControlR) .. (R1)
  node[midway, above=\CurveLabelGap cm,
    measure, text=violet!80!black]
    {$(\alpha,Q,Q_1)$-renormalization};

% Reference probability measures.
\node[point, text=blue!70!black]
  at (GQ.center) {$\times$};
\node[measure, anchor=north east,
  xshift=-5pt, yshift=-6pt, text=blue!70!black]
  at (GQ.center) {$Q$};

\node[point, text=red!75!black]
  at (GOne.center) {$\times$};
\node[measure, anchor=north west,
  xshift=5pt, yshift=-6pt, text=red!75!black]
  at (GOne.center) {$Q_1$};

% Boundary measures and adjacent labels.
\node[point, text=blue!70!black] at (P) {$\times$};
\node[measure, anchor=north west,
  xshift=-3pt, yshift=-8pt, text=blue!70!black]
  at (P)
  {$P^{\autoparent{h,Q,\lambda}}_{Y\mid X=x}$};

\node[point, text=red!75!black] at (R1) {$\times$};
\node[measure, anchor=north east,
  xshift=-3pt, yshift=-8pt, text=red!75!black]
  at (R1)
  {$P^{\autoparent{h,Q_1,\frac{\lambda}{\alpha}}}_{Y\mid X=x}$};

% Set labels anchored above the circles.
\node[setlabel, above=\SetLabelGap cm of GQ.north,
  text=blue!70!black]
  {$\mathcal{G}_{Q,\gamma}$};

\node[setlabel, above=\SetLabelGap cm of GOne.north,
  text=red!75!black]
  {$\mathcal{G}_{Q_1,\eta_1}$};

% Automatic frame: no fill and no background layer.
\node[draw=black, fill=none, line width=0.75pt,
  fit=(current bounding box), inner sep=\FramePadding cm]
  (Ambient) {};

\node[font=\Large, anchor=north east, inner sep=8pt]
  at (Ambient.north east)
  {$\triangle\autoparent{\mathcal{Y}}$};

\end{tikzpicture}

%% file: Figures/Journal-TIT2026/renormalization_geodesic_commonQ_two_cases.tikz
% Input-only file. Main preamble:
% \usepackage{tikz,graphicx}
% \usetikzlibrary{arrows.meta,calc,positioning,fit}
% \autoparent is supplied by Bermudez_IEEE.
% Arguments: radius for eta_1, radius for gamma, panel caption.
\providecommand{\RenormCommonPanel}[3]{}
\renewcommand{\RenormCommonPanel}[3]{%
\begin{tikzpicture}[
  x=1cm, y=1cm, baseline=(CQ),
  line cap=round, line join=round,
  level/.style={circle, inner sep=0pt, outer sep=0pt},
  leveleta/.style={level, draw=red!75!black, fill=red!6,
    solid, line width=0.85pt},
  levelgamma/.style={level, draw=blue!70!black, fill=blue!6,
    dotted, line width=1.1pt},
  curve/.style={draw=violet!80!black, line width=1.0pt},
  point/.style={font=\Large, inner sep=0pt},
  measure/.style={font=\large, inner sep=1pt},
  setlabel/.style={font=\large, inner sep=1pt}
]

% Layout parameters (centimetres and degrees).
\def\RadiusEta{#1}
\def\RadiusGamma{#2}
\def\AngleR{130}
\def\AngleP{60}
\def\FramePadding{1.00}
\def\FrameLabelRoom{1.50}
\pgfmathsetmacro{\MaxRadius}{max(\RadiusEta,\RadiusGamma)}
\pgfmathsetlengthmacro{\DiameterEta}{2*\RadiusEta*1cm}
\pgfmathsetlengthmacro{\DiameterGamma}{2*\RadiusGamma*1cm}

% Common reference measure; draw the larger filled set first.
\coordinate (CQ) at (0,0);
\ifdim\RadiusEta pt>\RadiusGamma pt
  \node[leveleta, minimum size=\DiameterEta] (GEta) at (CQ) {};
  \node[levelgamma, minimum size=\DiameterGamma] (GGamma) at (CQ) {};
\else
  \node[levelgamma, minimum size=\DiameterGamma] (GGamma) at (CQ) {};
  \node[leveleta, minimum size=\DiameterEta] (GEta) at (CQ) {};
\fi

% Each measure retains its angle in both panels.
\coordinate (R1) at (GEta.\AngleR);
\coordinate (P) at (GGamma.\AngleP);

% Relative-entropy radii.
\draw[thick, red!75!black] (CQ) --
  node[midway, fill=white, inner sep=1pt] {$\eta_1$} (R1);
\draw[thick, blue!70!black] (CQ) --
  node[midway, fill=white, inner sep=1pt] {$\gamma$} (P);

% Keep the original single smooth quadratic Bezier construction.
% Visit the inner measure first and end at the outer measure.
\ifdim\RadiusEta pt<\RadiusGamma pt
  \coordinate (CurveInner) at (R1);
  \coordinate (CurveEnd) at (P);
\else
  \coordinate (CurveInner) at (P);
  \coordinate (CurveEnd) at (R1);
\fi
\coordinate (CurveMid) at ($(CQ)!0.5!(CurveEnd)$);
\coordinate (CurveControl) at ($(CurveMid)!2!(CurveInner)$);
\draw[curve]
  (CQ) .. controls ($(CQ)!0.666666667!(CurveControl)$)
                  and ($(CurveEnd)!0.666666667!(CurveControl)$)
       .. (CurveEnd);

% Crosses and measure labels; label anchors do not change by case.
\node[point] at (CQ) {$\times$};
\node[measure, anchor=north, yshift=-8pt]
  at (CQ) {$Q=Q_1$};
\node[point, text=red!75!black] at (R1) {$\times$};
\node[measure, anchor=south east, xshift=-4pt, yshift=5pt,
  text=red!75!black,
  fill=white, fill opacity=0.75, text opacity=1,
  inner sep=2pt] at (R1)
  {$P^{\autoparent{h,Q,\frac{\lambda}{\alpha}}}_{Y\mid X=x}$};
\node[point, text=blue!70!black] at (P) {$\times$};
\node[measure, anchor=south west, xshift=6pt, yshift=5pt,
  text=blue!70!black, fill=white, fill opacity=0.75, text opacity=1,
  inner sep=2pt] at (P)
  {$P^{\autoparent{h,Q,\lambda}}_{Y\mid X=x}$};

% Labels inside the sets: the larger-set label lies in the annulus.
\ifdim\RadiusEta pt>\RadiusGamma pt
  \coordinate (EtaLabel) at ($(GGamma.south)!0.50!(GEta.south)$);
  \coordinate (GammaLabel) at ($(CQ)!0.60!(GGamma.south)$);
\else
  \coordinate (EtaLabel) at ($(CQ)!0.60!(GEta.south)$);
  \coordinate (GammaLabel) at ($(GEta.south)!0.50!(GGamma.south)$);
\fi
\node[setlabel, text=red!75!black] at (EtaLabel)
  {$\mathcal{G}_{Q,\eta_1}$};
\node[setlabel, text=blue!70!black] at (GammaLabel)
  {$\mathcal{G}_{Q,\gamma}$};

% Equal rectangles when the two radii are exchanged.
% No background layer or dependence on moving label bounds.
\pgfmathsetmacro{\FrameHalfWidth}{\MaxRadius+\FramePadding+\FrameLabelRoom}
\pgfmathsetmacro{\FrameHalfHeight}{\MaxRadius+\FramePadding}
\coordinate (FrameSW) at ($(CQ)+(-\FrameHalfWidth,-\FrameHalfHeight)$);
\coordinate (FrameNE) at ($(CQ)+(\FrameHalfWidth,\FrameHalfHeight)$);
\node[rectangle, draw=black, fill=none, line width=0.75pt,
  fit=(FrameSW)(FrameNE), inner sep=0pt, outer sep=0pt] (Ambient) {};
\node[font=\Large, anchor=north east, inner sep=8pt]
  at (Ambient.north east) {$\triangle\autoparent{\mathcal{Y}}$};
\node[font=\large, below=5mm of Ambient.south] {#3};
\end{tikzpicture}%
}%

%% file: Figures/Journal-TIT2026/renormalization_geodesic_two_cases.tikz
% Input this file from the main document; it defines the drawing command only.
% Main preamble: \usepackage{tikz,graphicx}
%                \usetikzlibrary{arrows.meta,calc,positioning,fit}
% The main document supplies \autoparent (Bermudez_IEEE).
\providecommand{\RenormPanel}[3]{%
\begin{tikzpicture}[
  x=1cm, y=1cm, baseline=(CQ),
  line cap=round, line join=round,
  level/.style={circle, inner sep=0pt, outer sep=0pt},
  levelone/.style={level, draw=red!75!black, fill=red!6,
    solid, line width=0.85pt},
  leveltwo/.style={level, draw=blue!70!black, fill=blue!6,
    dotted, line width=1.1pt},
  levelouter/.style={level, draw=black!75, fill=black!3,
    dashed, line width=0.85pt},
  geodesic/.style={draw=violet!80!black, line width=1.0pt},
  entropyone/.style={draw=red!75!black, thick},
  entropytwo/.style={draw=blue!70!black, thick},
  entropyouter/.style={draw=black!75, thick},
  point/.style={font=\Large, inner sep=0pt},
  measure/.style={font=\large, inner sep=1pt},
  setlabel/.style={font=\large, inner sep=1pt}
]

% Layout parameters: lengths are in centimetres.
% Arguments: eta_1 radius, eta_2 radius, panel caption.
\def\RadiusQ{2.15}
\def\RadiusOne{#1}
\def\RadiusOuter{#2}
\def\CircleGap{3.00}
\def\AngleP{0}
% Fixed angles and label positions for each measure in both cases.
\def\AngleR{130}
\def\AngleOuter{60}
\def\RAnchor{south east}\def\RShift{-6pt}
\def\PAnchor{south west}\def\PShift{6pt}
\def\FramePadding{1.00}
% Extra room on the right for the measure label (centimetres).
\def\FrameRightExtra{1.00}

\pgfmathsetmacro{\MaxRadius}{max(\RadiusOne,\RadiusOuter)}
\pgfmathsetlengthmacro{\DiameterQ}{2*\RadiusQ*1cm}
\pgfmathsetlengthmacro{\DiameterOne}{2*\RadiusOne*1cm}
\pgfmathsetlengthmacro{\DiameterOuter}{2*\RadiusOuter*1cm}

% Reference centers: the gap is between the left and outer circles.
\coordinate (CQ) at (0,0);
\coordinate (C1)
  at ($(CQ)+({\RadiusQ+\CircleGap+\MaxRadius},0)$);

% Neighborhoods: draw the outer filled circle before the inner one.
\node[leveltwo, minimum size=\DiameterQ] (GQ) at (CQ) {};
\ifdim\RadiusOne pt<\RadiusOuter pt
  \node[levelouter, minimum size=\DiameterOuter] (GOuter) at (C1) {};
  \node[levelone, minimum size=\DiameterOne] (GOne) at (C1) {};
\else
  \node[levelone, minimum size=\DiameterOne] (GOne) at (C1) {};
  \node[levelouter, minimum size=\DiameterOuter] (GOuter) at (C1) {};
\fi

% Measures anchored to their circle boundaries.
\coordinate (P) at (GQ.\AngleP);
\coordinate (R1) at (GOne.\AngleR);
\coordinate (P1) at (GOuter.\AngleOuter);

% Relative-entropy radii.
\draw[entropytwo] (CQ) --
  node[midway, fill=white, inner sep=1pt, text=blue!70!black]
    {$\gamma$} (P);
\draw[entropyone] (C1) --
  node[midway, fill=white, inner sep=1pt, text=red!75!black]
    {$\eta_1$} (R1);
\draw[entropyouter] (C1) --
  node[midway, fill=white, inner sep=1pt, text=black!75]
    {$\eta_2$} (P1);

% Keep the original single quadratic Bezier construction.
% Only the order of the two boundary measures changes with the radii.
\ifdim\RadiusOne pt<\RadiusOuter pt
  \coordinate (CurveInner) at (R1);
  \coordinate (CurveEnd) at (P1);
\else
  \coordinate (CurveInner) at (P1);
  \coordinate (CurveEnd) at (R1);
\fi
\coordinate (CurveMid) at ($(C1)!0.5!(CurveEnd)$);
\coordinate (CurveControl) at ($(CurveMid)!2!(CurveInner)$);
\draw[geodesic]
  (C1) .. controls ($(C1)!0.666666667!(CurveControl)$)
                  and ($(CurveEnd)!0.666666667!(CurveControl)$)
       .. (CurveEnd);

% Reference probability measures.
\node[point, text=blue!70!black] at (CQ) {$\times$};
\node[measure, anchor=north east,
  xshift=-5pt, yshift=-6pt, text=blue!70!black]
  at (CQ) {$Q$};
\node[point, text=red!75!black] at (C1) {$\times$};
\node[measure, anchor=north west,
  xshift=5pt, yshift=-6pt, text=red!75!black]
  at (C1) {$Q_1$};

% Boundary measure labels placed next to the crosses.
\node[point, text=blue!70!black] at (P) {$\times$};
\node[measure, anchor=north east,
  xshift=-3pt, yshift=-8pt, text=blue!70!black]
  at (P) {$P^{\autoparent{h,Q,\lambda}}_{Y\mid X=x}$};
\node[point, text=red!75!black] at (R1) {$\times$};
\node[measure, anchor=\RAnchor,
  xshift=\RShift, yshift=5pt, text=red!75!black,  fill=white, fill opacity=0.75, text opacity=1,
  inner sep=2pt]
  at (R1)
  {$P^{\autoparent{h,Q_1,\frac{\lambda}{\alpha}}}_{Y\mid X=x}$};
\node[point, text=black!75] at (P1) {$\times$};
\node[measure, anchor=\PAnchor,
  xshift=\PShift, yshift=5pt, text=black!75,  fill=white, fill opacity=0.75, text opacity=1,
  inner sep=2pt]
  at (P1) {$P^{\autoparent{h,Q_1,\lambda}}_{Y\mid X=x}$};

% Set labels inside their neighborhoods.
\node[setlabel, text=blue!70!black]
  at ($(CQ)!0.60!(GQ.south)$) {$\mathcal{G}_{Q,\gamma}$};
\ifdim\RadiusOne pt<\RadiusOuter pt
  \coordinate (EtaLabel) at ($(C1)!0.60!(GOne.south)$);
  \coordinate (GammaLabel) at ($(GOne.south)!0.50!(GOuter.south)$);
\else
  \coordinate (GammaLabel) at ($(C1)!0.60!(GOuter.south)$);
  \coordinate (EtaLabel) at ($(GOuter.south)!0.50!(GOne.south)$);
\fi
\node[setlabel, text=red!75!black] at (EtaLabel)
  {$\mathcal{G}_{Q_1,\eta_1}$};
\node[setlabel, text=black!75] at (GammaLabel)
  {$\mathcal{G}_{Q_1,\eta_2}$};

% Equal frames when the two radii are exchanged.
% Frame dimensions depend on the circle sizes, not the moving labels.
\pgfmathsetmacro{\FrameHalfHeight}{max(\RadiusQ,\MaxRadius)+\FramePadding}
\coordinate (FrameSW)
  at ($(CQ)+({-\RadiusQ-\FramePadding},-\FrameHalfHeight)$);
\coordinate (FrameNE)
  at ($(C1)+({\MaxRadius+\FramePadding+\FrameRightExtra},\FrameHalfHeight)$);
\node[rectangle, draw=black, fill=none, line width=0.75pt,
  fit=(FrameSW)(FrameNE), inner sep=0pt, outer sep=0pt]
  (Ambient) {};
\node[font=\Large, anchor=north east, inner sep=8pt]
  at (Ambient.north east) {$\triangle\autoparent{\mathcal{Y}}$};

\node[font=\large, below=5mm of Ambient.south] {#3};
\end{tikzpicture}
}%

%% file: Figures/Journal-TIT2026/log-linear-geodesic_commonQ.tikz
% Input-only file: defines a panel command; draws nothing on input.
% Preamble requirements: tikz, graphicx, and \usetikzlibrary{calc}.
% Uses your existing \autoparent macro.
% Arguments: radius gamma_1, radius eta, radius gamma_2, caption.
% The same angles, label offsets, and frame are used in all panels.
\providecommand{\LogLinearCommonPanel}[4]{}
\renewcommand{\LogLinearCommonPanel}[4]{%
\begin{tikzpicture}[
  x=1cm,y=1cm,line cap=round,line join=round,
  first/.style={draw=red!75!black,densely dotted,line width=1.05pt},
  middle/.style={draw=black!75,dash pattern=on 5pt off 2pt on 1pt off 2pt,line width=0.85pt},
  last/.style={draw=blue!70!black,dashed,line width=0.90pt},
  curve/.style={draw=violet!80!black,solid,line width=1.1pt},
  point/.style={font=\Large,inner sep=0pt},
  measure/.style={font=\large,inner sep=1pt},
  setlabel/.style={font=\large,inner sep=1pt},
  radiuslabel/.style={fill=white,inner sep=1.5pt}
]
\coordinate (CQ) at (0,0);
\coordinate (P1) at (165:#1);
\coordinate (S2) at (90:#2);
\coordinate (P2) at (15:#3);

% Transparent fills first, so every boundary remains visible.
\fill[red!75!black,opacity=0.035] (CQ) circle[radius=#1];
\fill[black,opacity=0.025] (CQ) circle[radius=#2];
\fill[blue!70!black,opacity=0.035] (CQ) circle[radius=#3];
\draw[first] (CQ) circle[radius=#1];
\draw[middle] (CQ) circle[radius=#2];
\draw[last] (CQ) circle[radius=#3];

% All radii are non-solid, with names at their midpoints.
\draw[first] (CQ) -- node[midway,radiuslabel,text=red!75!black]
  {$\gamma_1$} (P1);
\draw[middle] (CQ) -- node[midway,radiuslabel] {$\gamma_0$} (S2);
\draw[last] (CQ) -- node[midway,radiuslabel,text=blue!70!black]
  {$\gamma_2$} (P2);

% One smooth quadratic arc through the three nonaligned measures.
% The same point angles and curve construction apply in every case.
% At t=1/2 the arc passes through S2, with no piecewise join.
\coordinate (CurveMid) at ($(P1)!0.5!(P2)$);
\coordinate (CurveControl) at ($(CurveMid)!2!(S2)$);
\draw[curve] (P1)
  .. controls ($(P1)!0.666666667!(CurveControl)$)
          and ($(P2)!0.666666667!(CurveControl)$) .. (P2);

% Crosses are drawn over the radii and the curve.
\node[point] at (CQ) {$\times$};
\node[measure,anchor=north,yshift=-8pt] at (CQ) {$Q=Q_1=Q_2$};
\node[point,text=red!75!black] at (P1) {$\times$};
\node[point,text=violet!80!black] at (S2) {$\times$};
\node[point,text=blue!70!black] at (P2) {$\times$};
\node[measure,anchor=south,yshift=9pt,text=red!75!black, fill=white, fill opacity=0.75, text opacity=1,
  inner sep=2pt] at (P1)
  {$P^{\autoparent{h_1,Q,\lambda_1}}_{Y\mid X=x_1}$};
\node[measure,anchor=south,yshift=10pt,text=violet!80!black,  fill=white, fill opacity=0.75, text opacity=1,
  inner sep=2pt] at (S2)
  {$P^{\autoparent{\hat{h}_{\vect{\beta}},Q,1}}_{Y\mid X=\vect{x}}$};
\node[measure,anchor=south,yshift=9pt,text=blue!70!black,  fill=white, fill opacity=0.75, text opacity=1,
  inner sep=2pt] at (P2)
  {$P^{\autoparent{h_2,Q,\lambda_2}}_{Y\mid X=x_2}$};

% Each label sits inside its own circle, in a separate lower band.
\node[setlabel,text=red!75!black] at (0,{-#1+0.55})
  {$\mathcal{G}_{Q,\gamma_1}$};
\node[setlabel,text=black!75] at (0,{-#2+0.55})
  {$\mathcal{G}_{Q,\gamma_0}$};
\node[setlabel,text=blue!70!black] at (0,{-#3+0.55})
  {$\mathcal{G}_{Q,\gamma_2}$};

% Identical frames, with case captions below the rectangles.
% The enclosing rectangle always has a solid black boundary.
\draw[black,solid,line width=0.75pt] (-5.8,-5.45) rectangle (5.8,6.0);
\node[font=\Large,anchor=north east,inner sep=8pt] at (5.8,6.0)
  {$\triangle\autoparent{\mathcal{Y}}$};
\node[font=\large,anchor=north,yshift=-5mm] at (0,-5.45) {#4};
\end{tikzpicture}%
}

%% file: Figures/Journal-TIT2026/log-linear-geodesic_diffQ.tikz
\begin{tikzpicture}[
  x=1cm,
  y=1cm,
  line cap=round,
  line join=round,
  levelone/.style={draw=red!75!black, solid, line width=0.95pt},
  levelcommon/.style={draw=black!75,
    dash pattern=on 7pt off 2.5pt on 1.2pt off 2.5pt,
    line width=0.85pt},
  leveltwo/.style={draw=blue!70!black,
    dash pattern=on 10pt off 3.5pt,
    line width=0.95pt},
  etaone/.style={draw=red!75!black, densely dashed, line width=0.90pt},
  etatwo/.style={draw=blue!70!black, densely dotted, line width=1.00pt},
  geodesic/.style={draw=violet!80!black, solid, line width=1.0pt},
  entropyone/.style={draw=red!75!black, thick},
  entropytwo/.style={draw=blue!70!black, thick},
  etaoneradius/.style={draw=red!75!black, densely dashed, thick},
  etatworadius/.style={draw=blue!70!black, densely dotted, line width=1.0pt},
  point/.style={font=\Large, inner sep=0pt},
  measure/.style={font=\large, inner sep=1pt},
  setlabel/.style={font=\large, inner sep=1pt}
]

% Layout parameters (cm). Points and labels follow their centers.
\def\RadiusOne{2.15}
\def\RadiusTwo{2.15}
\def\RadiusEtaOne{2.15}
\def\RadiusEta{3.30}
\def\RadiusEtaTwo{4.40}
\def\SideGap{1.50}
\coordinate (CQ) at (0,0);
\coordinate (C1) at ($(CQ)+({-\RadiusEtaTwo-\SideGap-\RadiusOne},0)$);
\coordinate (C2) at ($(CQ)+({\RadiusEtaTwo+\SideGap+\RadiusTwo},0)$);

% Light fills first, then all boundaries.
\fill[red!6] (C1) circle[radius=\RadiusOne];
\fill[blue!6] (C2) circle[radius=\RadiusTwo];
\fill[blue!4] (CQ) circle[radius=\RadiusEtaTwo];
\fill[black!4] (CQ) circle[radius=\RadiusEta];
\fill[red!6] (CQ) circle[radius=\RadiusEtaOne];
\draw[levelone] (C1) circle[radius=\RadiusOne];
\draw[leveltwo] (C2) circle[radius=\RadiusTwo];
\draw[levelcommon] (CQ) circle[radius=\RadiusEta];
\draw[etaone] (CQ) circle[radius=\RadiusEtaOne];
\draw[etatwo] (CQ) circle[radius=\RadiusEtaTwo];

% Reference probability measures.
\node[point, text=red!75!black] at (C1) {$\times$};
\node[measure, text=red!75!black] at ($(C1)+(-0.35,-0.45)$) {$Q_1$};
\node[point] at (CQ) {$\times$};
\node[measure] at ($(CQ)+(0,-0.45)$) {$Q$};
\node[point, text=blue!70!black] at (C2) {$\times$};
\node[measure, text=blue!70!black] at ($(C2)+(0.25,-0.45)$) {$Q_2$};

% Gibbs probability measures on the original Q_1- and Q_2-centered sets.
\coordinate (P1) at ($(C1)+(90:\RadiusOne)$);
\coordinate (P2) at ($(C2)+(50:\RadiusTwo)$);
\node[point, text=red!75!black] at (P1) {$\times$};
\node[point, text=blue!70!black] at (P2) {$\times$};

\node[measure, align=center, anchor=south, text=red!75!black] at ($(P1)+(0,0.30)$)
  {$P^{\autoparent{h_1,Q_1,\lambda_1}}_{Y\mid X=x_1}$};
\node[measure, align=center, anchor=south, text=blue!70!black] at ($(P2)+(0,0.30)$)
  {$P^{\autoparent{h_2,Q_2,\lambda_2}}_{Y\mid X=x_2}$};

% Gibbs probability measures on the Q-centered sets.
\coordinate (P1Q) at ($(CQ)+(140:\RadiusEtaOne)$);
\coordinate (S2) at ($(CQ)+(84:\RadiusEta)$);
\coordinate (P2Q) at ($(CQ)+(35:\RadiusEtaTwo)$);

\node[point, text=red!75!black] at (P1Q) {$\times$};
\node[point, text=violet!80!black] at (S2) {$\times$};
\node[point, text=blue!70!black] at (P2Q) {$\times$};

\node[measure, align=center, anchor=south east, text=red!75!black,  fill=white, fill opacity=0.75, text opacity=1,
  inner sep=2pt]
  at ($(P1Q)+(-0.08,0.22)$)
  {$P^{\autoparent{h_1,Q,\lambda_1}}_{Y\mid X=x_1}$};
\node[measure, align=center, anchor=south, text=violet!80!black,  fill=white, fill opacity=0.75, text opacity=1,
  inner sep=2pt]
  at ($(S2)+(0,0.30)$)
  {$P^{\autoparent{\hat{h}_{\vect{\beta}},Q,1}}_{Y\mid X=\vect{x}}$};
\node[measure, align=center, anchor=south west, text=blue!70!black,  fill=white, fill opacity=0.75, text opacity=1,
  inner sep=2pt]
  at ($(P2Q)+(0.14,0.16)$)
  {$P^{\autoparent{h_2,Q,\lambda_2}}_{Y\mid X=x_2}$};

% Curved line from P^(h_1,Q,lambda_1) to P^(h_2,Q,lambda_2), through S_2.
% One quadratic Bezier arc, represented exactly as a cubic.
% At parameter t=1/2 the curve passes exactly through S2.
% A single polynomial arc avoids any join or change of curvature at S2.
\coordinate (CurveMid) at ($(P1Q)!0.5!(P2Q)$);
\coordinate (CurveControl) at ($(CurveMid)!2!(S2)$);
\draw[geodesic]
  (P1Q) .. controls ($(P1Q)!0.666666667!(CurveControl)$)
                   and ($(P2Q)!0.666666667!(CurveControl)$) .. (P2Q);

% Relative-entropy radii.
\draw[entropyone] (P1) --
  node[midway, fill=white, inner sep=1pt, text=red!75!black] {$\gamma_1$}
  (C1);
\draw[entropytwo] (P2) --
  node[midway, fill=white, inner sep=1pt, text=blue!70!black] {$\gamma_2$}
  (C2);
\draw[etaoneradius] (CQ) --
  node[midway, fill=white, inner sep=1pt, text=red!75!black] {$\eta_1$}
  (P1Q);
\draw[etatworadius] (CQ) --
  node[midway, fill=white, inner sep=1pt, text=blue!70!black] {$\eta_2$}
  (P2Q);

% Radius from Q to S2; eta is centered on the segment.
\draw[thick, black!75] (CQ) --
  node[midway, fill=white, inner sep=1pt] {$\gamma_0$} (S2);

% Set labels lie inside their own sets. The central labels occupy
% separate lower bands, away from the measures and radii.
\node[setlabel, text=red!75!black]
  at ($(C1)+(0,{-0.60*\RadiusOne})$)
  {$\mathcal{G}_{Q_1,\gamma_1}$};
\node[setlabel, text=blue!70!black]
  at ($(C2)+(0,{-0.60*\RadiusTwo})$)
  {$\mathcal{G}_{Q_2,\gamma_2}$};
\node[setlabel, text=red!75!black]
  at ($(CQ)+(0,{-0.62*\RadiusEtaOne})$)
  {$\mathcal{G}_{Q,\eta_1}$};
\node[setlabel, text=black!75]
  at ($(CQ)+(0,{-0.5*(\RadiusEtaOne+\RadiusEta)})$)
  {$\mathcal{G}_{Q,\gamma_0}$};
\node[setlabel, text=blue!70!black]
  at ($(CQ)+(0,{-0.5*(\RadiusEta+\RadiusEtaTwo)})$)
  {$\mathcal{G}_{Q,\eta_2}$};

% Fit the entire drawing, including every measure label.
% No background-layer library is required.
\node[rectangle, draw=black, fill=none, line width=0.75pt,
  fit=(current bounding box), inner sep=0.80cm] (Ambient) {};
\node[font=\Large, anchor=north east, inner sep=8pt]
  at (Ambient.north east) {$\triangle\autoparent{\mathcal{Y}}$};

\end{tikzpicture}

%% file: Figures/Journal-TIT2026/nesting_geometric_inclusions.tikz
% Input-only file. Load tikz, graphicx and the calc, positioning, fit libraries.
% The main document supplies autoparent and vect through Bermudez_IEEE.
% Radii are schematic; the blue radius is derived to meet the final measure.
\providecommand{\NestingPanel}[3]{}
\renewcommand{\NestingPanel}[3]{%
\begin{tikzpicture}[
  x=1cm, y=1cm, baseline=(Q0), line cap=round, line join=round,
  level/.style={circle, inner sep=0pt, outer sep=0pt},
  levelone/.style={level, draw=black!75, dashed, line width=0.85pt},
  leveltwo/.style={level, draw=blue!70!black, dotted, line width=1.1pt},
  levelfinal/.style={level, draw=red!75!black, solid, line width=0.85pt},
  point/.style={font=\Large, inner sep=0pt},
  measure/.style={font=\large, inner sep=1pt},
  setlabel/.style={font=\large, inner sep=1pt}
]

% Arguments: graphical radius for gamma_1, radius for eta, caption.
% Only the radii change between the two panels.
\def\FirstRadius{#1}
\def\FinalRadius{#2}
\def\FirstAngle{16}
\def\FinalAngle{64}
\def\FramePadding{1.00}
\pgfmathsetmacro{\FirstX}{\FirstRadius*cos(\FirstAngle)}
\pgfmathsetmacro{\FirstY}{\FirstRadius*sin(\FirstAngle)}
\pgfmathsetmacro{\FinalX}{\FinalRadius*cos(\FinalAngle)}
\pgfmathsetmacro{\FinalY}{\FinalRadius*sin(\FinalAngle)}
\pgfmathsetmacro{\MaxRadius}{max(\FirstRadius,\FinalRadius)}
\pgfmathsetmacro{\SecondRadius}{veclen(\FinalX-\FirstX,\FinalY-\FirstY)}
\coordinate (Q0) at (0,0);
\coordinate (P0) at ($(Q0)+(\FirstAngle:\FirstRadius)$);
\coordinate (P2) at ($(Q0)+(\FinalAngle:\FinalRadius)$);
\pgfmathsetlengthmacro{\RadiusOne}{\FirstRadius*1cm}
\pgfmathsetlengthmacro{\RadiusTwo}{\SecondRadius*1cm}
\pgfmathsetlengthmacro{\RadiusFinal}{\FinalRadius*1cm}
\pgfmathsetlengthmacro{\DiameterOne}{2*\RadiusOne}
\pgfmathsetlengthmacro{\DiameterTwo}{2*\RadiusTwo}
\pgfmathsetlengthmacro{\DiameterFinal}{2*\RadiusFinal}

% Light fills first, so all circle boundaries remain visible.
\fill[red!75!black, opacity=0.06] (Q0) circle[radius=\RadiusFinal];
\fill[black, opacity=0.03] (Q0) circle[radius=\RadiusOne];
\fill[blue!70!black, opacity=0.06] (P0) circle[radius=\RadiusTwo];

% Distinct solid, dashed and dotted boundaries for black-and-white printing.
\node[levelfinal, minimum size=\DiameterFinal] (GEta) at (Q0) {};
\node[levelone, minimum size=\DiameterOne] (GOne) at (Q0) {};
\node[leveltwo, minimum size=\DiameterTwo] (GTwo) at (P0) {};

% Relative-entropy radii.
\draw[thick, black!75] (Q0) --
  node[midway, fill=white, inner sep=1pt] {$\gamma_1$} (P0);
\draw[thick, blue!70!black] (P0) --
  node[midway, fill=white, inner sep=1pt] {$\gamma_2$} (P2);
\draw[thick, red!75!black] (Q0) --
  node[midway, fill=white, inner sep=1pt] {$\eta$} (P2);

% Cross markers and adjacent measure labels.
\node[point, text=black!75] at (Q0) {$\times$};
\node[measure, anchor=north west, xshift=5pt, yshift=-6pt]
  at (Q0) {$Q_1$};
\node[point, text=blue!70!black] at (P0) {$\times$};
\node[measure, anchor=north west, xshift=8pt, yshift=-8pt,
  text=blue!70!black,fill=white, fill opacity=0.75, text opacity=1,
  inner sep=2pt] at (P0)
  {$Q_2=P_{Y_1\mid X_1=x_1}^{\autoparent{h_1,Q_1,\lambda_1}}$};
\node[point, text=red!75!black] at (P2) {$\times$};
% One math node keeps the full Gibbs equality on a single line.
\node[measure, anchor=south west, xshift=8pt, yshift=10pt,  fill=white, fill opacity=0.75, text opacity=1,
  inner sep=2pt]
  at (P2)
  {$\textcolor{green!50!black}{P_{Y_2\mid X_2=x_2}^{\autoparent{h_2,Q_2,\lambda_2}}}
   =\textcolor{red!75!black}{P_{Y\mid\vect{X}=\vect{x}}^{\autoparent{\hat{h}_{\beta},Q_1,1}}}$};

% Labels inside the corresponding sets.
\ifdim\FirstRadius pt<\FinalRadius pt
  \coordinate (GammaLabel) at ($(Q0)!0.65!(GOne.220)$);
  \coordinate (EtaLabel) at ($(GOne.145)!0.55!(GEta.145)$);
\else
  \coordinate (EtaLabel) at ($(Q0)!0.65!(GEta.220)$);
  \coordinate (GammaLabel) at ($(GEta.145)!0.55!(GOne.145)$);
\fi
\node[setlabel, text=black!75] at (GammaLabel)
  {$\mathcal{G}_{Q_1,\gamma_1}$};
\node[setlabel, text=red!75!black] at (EtaLabel)
  {$\mathcal{G}_{Q_1,\eta}$};
\node[setlabel, text=blue!70!black]
  at ($(P0)!0.70!(GTwo.30)$) {$\mathcal{G}_{Q_2,\gamma_2}$};

% Common frame dimensions for exchanged radii.
\pgfmathsetmacro{\FrameLeft}{-\MaxRadius-\FramePadding}
\pgfmathsetmacro{\FrameRight}{max(\MaxRadius*cos(\FirstAngle)+\SecondRadius+\FramePadding,\MaxRadius*cos(\FinalAngle)+7)}
\pgfmathsetmacro{\FrameTop}{max(\MaxRadius,\MaxRadius*sin(\FirstAngle)+\SecondRadius)+1.5}
\pgfmathsetmacro{\FrameBottom}{-max(\MaxRadius,\SecondRadius)-\FramePadding}
\coordinate (FrameSW) at (\FrameLeft,\FrameBottom);
\coordinate (FrameNE) at (\FrameRight,\FrameTop);
\node[rectangle, draw=black, fill=none, line width=0.75pt,
  fit=(FrameSW)(FrameNE), inner sep=0pt, outer sep=0pt] (Ambient) {};
\node[font=\Large, anchor=north east, inner sep=8pt]
  at (Ambient.north east) {$\triangle\autoparent{\mathcal{Y}}$};
\node[font=\large, below=5mm of Ambient.south] {#3};
\end{tikzpicture}%
}%

%% file: Figures/Journal-TIT2026/Triangle_Geo_rep.tikz
% Triangle_Unlearning.tikz
% Include in a LaTeX document with:
%   \usepackage{tikz}
%   \usetikzlibrary{arrows.meta,positioning,calc,fit,shapes.multipart}
%   \input{Triangle_Unlearning.tikz}
% This figure also uses custom macros from Bermudez_IEEE.sty:
% \KL, \vect, \matx, \gen, and \autoparent.

% --- Federated Learning diagram (TikZ) ---
\begingroup
% Tunable layout
\newcommand*{\LaneH}{4.2}   % lane height (in cm since x=y=1cm)
\newcommand*{\LeftW}{1.5}   % left label strip width
\newcommand*{\BodyW}{14.8}  % main drawing width (to the right of label strip)
\newcommand*{\OuterW}{15.0} % total width (LeftW + BodyW)
\newcommand*{\Corner}{2pt}  % corner radius for boxes

% Styles
\tikzset{
  lane/.style={draw,thick},
  laneTitle/.style={font=\sffamily\bfseries,rotate=90,align=center},
  bigbox/.style={draw,thick,rounded corners=\Corner,minimum height=2.8cm,minimum width=10cm},
  block/.style={draw,thick,minimum height=1.5cm,minimum width=3.0cm,align=center},
  smblock/.style={draw,thick,minimum height=1.5cm,minimum width=2.5cm,align=center},
  sq/.style={draw,thick,minimum size=8mm,inner sep=0pt},
  outcirc/.style={circle,draw,thick,minimum size=3mm,inner sep=0pt},
  lab/.style={font=\sffamily},
  arr/.style={-{Latex[length=2.4mm]},thick}
}

\begin{tikzpicture}[x=1cm,y=1cm]

%--- Geometry ----------------------------------------------------
\coordinate (A) at (0,0);      % bottom-left
\coordinate (B) at (20,0);     % bottom-right
\coordinate (C) at (0,10);      % top-left

% Outer triangle (thick frame)
\draw[line width=2.6pt] (A)--(B)--(C)--cycle;

%---------------------------
\draw [line width=1pt]  (0.5,0) -- (0.5,0.5) -- (0,0.5);

%--- Labels ------------------------------------------------------

% Bottom expression
\path (A) -- (B)
  node[midway, below=10pt, align=center, scale=2] {$\displaystyle
  \sqrt{
\KL{
P^{\autoparent{h_1,Q_1,\lambda_1}}_{Y_1\mid X_1=x_1}
}{Q_1}
}
  $};

% Left vertical expression (rotated)
  \path (A) -- (C)
  node[pos=0.45, left=20pt, rotate=90, above=5pt, align=center, scale =1.8] {$\displaystyle
  \sqrt{
\KL{
P^{\autoparent{
h_2,
P^{\autoparent{h_1,Q_1,\lambda_1}}_{Y_1\mid X_1=x_1},
\lambda_2
}}_{Y_2\mid X_2=x_2}
}{
P^{\autoparent{h_1,Q_1,\lambda_1}}_{Y_1\mid X_1=x_1}
}
}
  $};

% Hypotenuse expression inside a thin, rotated white box
\path (B) -- (C)
  node[midway, sloped, above=10pt,align=center, scale=1.6] {$\displaystyle
 \sqrt{
\KL{
P^{\autoparent{
h_2,
P^{\autoparent{h_1,Q_1,\lambda_1}}_{Y_1\mid X_1=x_1},
\lambda_2
}}_{Y_2\mid X_2=x_2}
}{Q_1}
+
\frac{1}{\lambda_1}
\left[
\int h_1\autoparent{x_1,y}\,
\mathrm{d}
P^{\autoparent{
h_2,
P^{\autoparent{h_1,Q_1,\lambda_1}}_{Y_1\mid X_1=x_1},
\lambda_2
}}_{Y_2\mid X_2=x_2}
\autoparent{y}
-
\int h_1\autoparent{x_1,y}\,
\mathrm{d}
P^{\autoparent{h_1,Q_1,\lambda_1}}_{Y_1\mid X_1=x_1}
\autoparent{y}
\right]
}
  $};

\end{tikzpicture}
\endgroup

%% file: Figures/Journal-TIT2026/Nested_Decentralized/recursive_nesting_inclusions.tikz
% Input-only file. Main preamble:
% \usepackage{tikz,graphicx}
% \usetikzlibrary{arrows.meta,calc,positioning,fit}
% The main document supplies \autoparent and \vect.
\ifdefined\RecursiveNestingBoxA\else
  \newsavebox{\RecursiveNestingBoxA}
  \newsavebox{\RecursiveNestingBoxB}
\fi
\providecommand{\RecursiveNestingArtwork}[2]{}
\renewcommand{\RecursiveNestingArtwork}[2]{%
\begin{tikzpicture}[
  x=1cm, y=1cm, baseline=(Q1),
  line cap=round,
  line join=round,
  >=Stealth
]

% Arguments: graphical radii for gamma_1 and eta.
% Extra geometric space improves readability without scaling the text.
\pgfmathsetmacro{\FirstRadius}{1.75*(#1)}
\pgfmathsetmacro{\FinalRadius}{1.75*(#2)}
\def\FirstAngle{150}
\def\FinalAngle{55}
% Vertical offsets: Q3 is high and Q4 is low.
\def\QThreeRise{1.45}
\def\QFourDrop{0.65}
\pgfmathsetmacro{\MaxRadius}{max(\FirstRadius,\FinalRadius)}
\coordinate (Q1) at (0,0);
\coordinate (Q2) at ($(Q1)+(\FirstAngle:\FirstRadius)$);
\coordinate (Q5) at ($(Q1)+(\FinalAngle:\FinalRadius)$);
\coordinate (Q3) at ($(Q2)!0.33!(Q5)+(0,\QThreeRise)$);
\coordinate (Q4) at ($(Q2)!0.67!(Q5)+(0,-\QFourDrop)$);

%--- Styles ---
\tikzset{
  GiBase/.style={line width=0.85pt},
  G0/.style={GiBase, draw=red!75!black, solid},
  G1/.style={GiBase, draw=blue!70!black, dotted, line width=1.1pt},
  G2/.style={GiBase, draw=orange!75!black, dashed},
  G3/.style={GiBase, draw=violet!80!black, dash dot},
  G4/.style={GiBase, draw=green!60!black, dash dot dot},
  GiLbl/.style={font=\fontsize{14}{17}\selectfont, fill=white, inner sep=1.5pt},
  measure/.style={font=\fontsize{18}{21}\selectfont, inner sep=1pt},
  ArrSeg/.style={
    line width=0.9pt, black!75,
    -{Stealth[length=5pt,width=4pt]},
    shorten <=3pt, shorten >=3pt
  }
}

% Light fills first; boundaries are drawn afterwards.
\foreach \Center/\Boundary/\Color in {
  Q1/Q5/red!75!black,
  Q1/Q2/blue!70!black,
  Q2/Q3/orange!75!black,
  Q3/Q4/violet!80!black,
  Q4/Q5/green!60!black}{
  \fill[\Color, opacity=0.025]
    let \p1=(\Center), \p2=(\Boundary),
        \n1={veclen(\x2-\x1,\y2-\y1)}
    in (\Center) circle[radius=\n1];
}

%--- Circles ---
\draw[G0]
  let
    \p1=(Q1),
    \p2=(Q5),
    \n1={veclen(\x2-\x1,\y2-\y1)}
  in
    (Q1) circle (\n1);

\draw[G1]
  let
    \p1=(Q1),
    \p2=(Q2),
    \n1={veclen(\x2-\x1,\y2-\y1)}
  in
    (Q1) circle (\n1);

\draw[G2]
  let
    \p1=(Q2),
    \p2=(Q3),
    \n1={veclen(\x2-\x1,\y2-\y1)}
  in
    (Q2) circle (\n1);

\draw[G3]
  let
    \p1=(Q3),
    \p2=(Q4),
    \n1={veclen(\x2-\x1,\y2-\y1)}
  in
    (Q3) circle (\n1);

\draw[G4]
  let
    \p1=(Q4),
    \p2=(Q5),
    \n1={veclen(\x2-\x1,\y2-\y1)}
  in
    (Q4) circle (\n1);

%--- Path / iterates ---
\draw[ArrSeg]
  (Q1) --
  node[midway, font=\fontsize{14}{17}\selectfont, fill=white, inner sep=1pt] {$\gamma_1$}
  (Q2);

\draw[ArrSeg]
  (Q2) --
  node[midway, font=\fontsize{14}{17}\selectfont, fill=white, inner sep=1pt] {$\gamma_2$}
  (Q3);

\draw[ArrSeg]
  (Q3) --
  node[midway, font=\fontsize{14}{17}\selectfont, fill=white, inner sep=1.5pt] {$\gamma_3$}
  (Q4);

\draw[ArrSeg]
  (Q4) --
  node[midway, font=\fontsize{14}{17}\selectfont, fill=white, inner sep=1.5pt] {$\gamma_4$}
  (Q5);

% Auxiliary connections, subdued to keep the nesting path prominent.
\draw[draw=black!30, line width=0.45pt] (Q1) -- (Q3);
\draw[draw=black!30, line width=0.45pt] (Q1) -- (Q4);

\draw[thick, red!75!black]
  (Q1) --
  node[
    midway,
    font=\fontsize{14}{17}\selectfont,
    fill=white,
    inner sep=1pt,
    text=red!75!black
  ]
  {$\eta$}
  (Q5);

%--- Points ---
\draw[draw=red!75!black, line width=0.8pt]
  ($(Q1)+(-1.7pt,-1.7pt)$) -- ($(Q1)+(1.7pt,1.7pt)$)
  ($(Q1)+(-1.7pt,1.7pt)$) -- ($(Q1)+(1.7pt,-1.7pt)$);
\draw[draw=black!75, line width=0.8pt]
  ($(Q2)+(-1.7pt,-1.7pt)$) -- ($(Q2)+(1.7pt,1.7pt)$)
  ($(Q2)+(-1.7pt,1.7pt)$) -- ($(Q2)+(1.7pt,-1.7pt)$);
\draw[draw=black!75, line width=0.8pt]
  ($(Q3)+(-1.7pt,-1.7pt)$) -- ($(Q3)+(1.7pt,1.7pt)$)
  ($(Q3)+(-1.7pt,1.7pt)$) -- ($(Q3)+(1.7pt,-1.7pt)$);
\draw[draw=black!75, line width=0.8pt]
  ($(Q4)+(-1.7pt,-1.7pt)$) -- ($(Q4)+(1.7pt,1.7pt)$)
  ($(Q4)+(-1.7pt,1.7pt)$) -- ($(Q4)+(1.7pt,-1.7pt)$);
\draw[draw=black!75, line width=0.8pt]
  ($(Q5)+(-1.7pt,-1.7pt)$) -- ($(Q5)+(1.7pt,1.7pt)$)
  ($(Q5)+(-1.7pt,1.7pt)$) -- ($(Q5)+(1.7pt,-1.7pt)$);

% Labels are drawn last, so no later boundary runs over their text.
\node[measure,text=red!75!black] at ($(Q1)+(0,-0.65)$) {$Q_1$};
\node[measure] at ($(Q2)+(-0.75,0)$) {$Q_2$};
\node[measure] at ($(Q3)+(0,0.65)$) {$Q_3$};
\node[measure] at ($(Q4)+(0,-0.70)$) {$Q_4$};
\node[measure,anchor=west] at ($(Q5)+(0.40,0.65)$)
  {$Q_5=P^{\autoparent{\hat{h}_{\vect{\beta}},Q_1,1}}_
    {Y\mid\vect{X}=\vect{x}}$};

% Interior positions, relative to each circle's center and radius.
% The two Q1 labels occupy distinct lower bands.
\ifdim\FirstRadius pt<\FinalRadius pt
  \path let \p1=(Q1), \p2=(Q5),
    \n1={veclen(\x2-\x1,\y2-\y1)} in
    node[GiLbl,text=red!75!black]
      at ($(Q1)+(270:{0.780000*\n1})$)
      {$\mathcal{G}_{Q_1,\eta}$};
  \path let \p1=(Q1), \p2=(Q2),
    \n1={veclen(\x2-\x1,\y2-\y1)} in
    node[GiLbl,text=blue!70!black]
      at ($(Q1)+(249:{0.387391*\n1})$)
      {$\mathcal{G}_{Q_1,\gamma_1}$};
  \path let \p1=(Q2), \p2=(Q3),
    \n1={veclen(\x2-\x1,\y2-\y1)} in
    node[GiLbl,text=orange!75!black]
      at ($(Q2)+(291:{0.611739*\n1})$)
      {$\mathcal{G}_{Q_2,\gamma_2}$};
  \path let \p1=(Q3), \p2=(Q4),
    \n1={veclen(\x2-\x1,\y2-\y1)} in
    node[GiLbl,text=violet!80!black]
      at ($(Q3)+(15:{0.574348*\n1})$)
      {$\mathcal{G}_{Q_3,\gamma_3}$};
  \path let \p1=(Q4), \p2=(Q5),
    \n1={veclen(\x2-\x1,\y2-\y1)} in
    node[GiLbl,text=green!60!black]
      at ($(Q4)+(306:{0.705217*\n1})$)
      {$\mathcal{G}_{Q_4,\gamma_4}$};
\else
  \path let \p1=(Q1), \p2=(Q5),
    \n1={veclen(\x2-\x1,\y2-\y1)} in
    node[GiLbl,text=red!75!black]
      at ($(Q1)+(270:{0.350000*\n1})$)
      {$\mathcal{G}_{Q_1,\eta}$};
  \path let \p1=(Q1), \p2=(Q2),
    \n1={veclen(\x2-\x1,\y2-\y1)} in
    node[GiLbl,text=blue!70!black]
      at ($(Q1)+(270:{0.780000*\n1})$)
      {$\mathcal{G}_{Q_1,\gamma_1}$};
  \path let \p1=(Q2), \p2=(Q3),
    \n1={veclen(\x2-\x1,\y2-\y1)} in
    node[GiLbl,text=orange!75!black]
      at ($(Q2)+(189:{0.574348*\n1})$)
      {$\mathcal{G}_{Q_2,\gamma_2}$};
  \path let \p1=(Q3), \p2=(Q4),
    \n1={veclen(\x2-\x1,\y2-\y1)} in
    node[GiLbl,text=violet!80!black]
      at ($(Q3)+(102:{0.649130*\n1})$)
      {$\mathcal{G}_{Q_3,\gamma_3}$};
  \path let \p1=(Q4), \p2=(Q5),
    \n1={veclen(\x2-\x1,\y2-\y1)} in
    node[GiLbl,text=green!60!black]
      at ($(Q4)+(96:{0.761304*\n1})$)
      {$\mathcal{G}_{Q_4,\gamma_4}$};
\fi

\end{tikzpicture}%
}%

\providecommand{\RecursiveNestingPanel}[3]{}
\renewcommand{\RecursiveNestingPanel}[3]{%
\begingroup
% Measure both complete drawings, including every circle and label.
% Both cases consequently use the same frame size.
\sbox{\RecursiveNestingBoxA}{\RecursiveNestingArtwork{#1}{#2}}%
\sbox{\RecursiveNestingBoxB}{\RecursiveNestingArtwork{#2}{#1}}%
\pgfmathsetlengthmacro{\RecursiveFrameWidth}{max(\the\wd\RecursiveNestingBoxA,\the\wd\RecursiveNestingBoxB)+1.0cm}%
\pgfmathsetlengthmacro{\RecursiveFrameHeight}{max(\the\ht\RecursiveNestingBoxA+\the\dp\RecursiveNestingBoxA,\the\ht\RecursiveNestingBoxB+\the\dp\RecursiveNestingBoxB)+1.8cm}%
\begin{tikzpicture}[baseline=(Ambient.center)]
\node[rectangle, draw=black, fill=none, line width=0.75pt,
  inner sep=0pt, outer sep=0pt,
  minimum width=\RecursiveFrameWidth,
  minimum height=\RecursiveFrameHeight] (Ambient) at (0,0) {};
\node[inner sep=0pt, outer sep=0pt] at (Ambient.center)
  {\usebox{\RecursiveNestingBoxA}};
\node[font=\Large, anchor=north east, inner sep=8pt]
  at (Ambient.north east) {$\triangle\autoparent{\mathcal{Y}}$};
\node[font=\fontsize{14}{17}\selectfont, anchor=north, yshift=-4mm, inner sep=1pt]
  at (Ambient.south) {#3};
\end{tikzpicture}%
\endgroup
}%
 

%% file: Figures/Journal-TIT2026/Triangle_Geo_repGen.tikz
% Triangle_Unlearning.tikz
% Include in a LaTeX document with:
%   \usepackage{tikz}
%   \usetikzlibrary{arrows.meta,positioning,calc,fit,shapes.multipart}
%   \input{Triangle_Unlearning.tikz}
% This figure also uses custom macros from Bermudez_IEEE.sty:
% \KL, \vect, \matx, \gen, and \autoparent.

% --- Federated Learning diagram (TikZ) ---
\begingroup
% Tunable layout
\newcommand*{\LaneH}{4.2}   % lane height (in cm since x=y=1cm)
\newcommand*{\LeftW}{1.5}   % left label strip width
\newcommand*{\BodyW}{14.8}  % main drawing width (to the right of label strip)
\newcommand*{\OuterW}{15.0} % total width (LeftW + BodyW)
\newcommand*{\Corner}{2pt}  % corner radius for boxes

% Styles
\tikzset{
  lane/.style={draw,thick},
  laneTitle/.style={font=\sffamily\bfseries,rotate=90,align=center},
  bigbox/.style={draw,thick,rounded corners=\Corner,minimum height=2.8cm,minimum width=10cm},
  block/.style={draw,thick,minimum height=1.5cm,minimum width=3.0cm,align=center},
  smblock/.style={draw,thick,minimum height=1.5cm,minimum width=2.5cm,align=center},
  sq/.style={draw,thick,minimum size=8mm,inner sep=0pt},
  outcirc/.style={circle,draw,thick,minimum size=3mm,inner sep=0pt},
  lab/.style={font=\sffamily},
  arr/.style={-{Latex[length=2.4mm]},thick}
}

\begin{tikzpicture}[x=1cm,y=1cm]

%--- Geometry ----------------------------------------------------
\coordinate (A) at (0,0);      % bottom-left
\coordinate (B) at (12,0);     % bottom-right
\coordinate (C) at (0,6);      % top-left

% Outer triangle (thick frame)
\draw[line width=2.6pt] (A)--(B)--(C)--cycle;

%---------------------------
\draw [line width=1pt]  (0.5,0) -- (0.5,0.5) -- (0,0.5);

%--- Labels ------------------------------------------------------

% Bottom expression
\path (A) -- (B)
  node[midway, below=10pt, align=center, scale=1.4] {$\displaystyle
  \sqrt{
\displaystyle\sum_{k=2}^{K}\KL{P^{\autoparent{h_k,Q_k,\lambda_k}}_{Y \mid X = x_k}}{Q_k}
}
  $};

% Left vertical expression (rotated)
  \path (A) -- (C)
  node[pos=0.45, left=20pt, rotate=90, above=5pt, align=center, scale =1.4] {$\displaystyle
  \sqrt{
\KL{P^{\autoparent{h_1,Q_1,\lambda_1}}_{Y \mid X = x_1}}{Q_1}}
  $};

% Hypotenuse expression inside a thin, rotated white box
\path (B) -- (C)
  node[midway, sloped, above=10pt,align=center, scale=1.4] {$\displaystyle
 \sqrt{\displaystyle\sum_{i=1}^{K-1}\delta_i + \KL{P^{\autoparent{h_K,Q_K,\lambda_K}}_{Y \mid X = x_K}}{Q_1}}
  $};

\end{tikzpicture}
\endgroup

%% file: Figures/Journal-TIT2026/geometric_interpretation.tikz
% Single-panel version of the construction in
% nesting_geometric_inclusions.tikz, with the labels specific to this figure.
\begin{tikzpicture}[
  x=1cm,
  y=1cm,
  line cap=round,
  line join=round,
  level/.style={circle, inner sep=0pt, outer sep=0pt},
  levelone/.style={level, draw=black!75, dashed, line width=0.85pt},
  leveltwo/.style={level, draw=blue!70!black, dotted, line width=1.1pt},
  levelfinal/.style={level, draw=red!75!black, solid, line width=0.85pt},
  point/.style={font=\Large, inner sep=0pt},
  measure/.style={font=\large, inner sep=1pt},
  setlabel/.style={font=\large, inner sep=1pt}
]

% One case: eta > gamma_1.
\def\FirstRadius{2.35}
\def\FinalRadius{3.50}
\def\FirstAngle{16}
\def\FinalAngle{64}

\pgfmathsetmacro{\FirstX}{\FirstRadius*cos(\FirstAngle)}
\pgfmathsetmacro{\FirstY}{\FirstRadius*sin(\FirstAngle)}
\pgfmathsetmacro{\FinalX}{\FinalRadius*cos(\FinalAngle)}
\pgfmathsetmacro{\FinalY}{\FinalRadius*sin(\FinalAngle)}
\pgfmathsetmacro{\SecondRadius}{
  veclen(\FinalX-\FirstX,\FinalY-\FirstY)
}

\coordinate (Q0) at (0,0);
\coordinate (P0) at ($(Q0)+(\FirstAngle:\FirstRadius)$);
\coordinate (P2) at ($(Q0)+(\FinalAngle:\FinalRadius)$);

\pgfmathsetlengthmacro{\RadiusOne}{\FirstRadius*1cm}
\pgfmathsetlengthmacro{\RadiusTwo}{\SecondRadius*1cm}
\pgfmathsetlengthmacro{\RadiusFinal}{\FinalRadius*1cm}
\pgfmathsetlengthmacro{\DiameterOne}{2*\RadiusOne}
\pgfmathsetlengthmacro{\DiameterTwo}{2*\RadiusTwo}
\pgfmathsetlengthmacro{\DiameterFinal}{2*\RadiusFinal}

% Light fills first, so all neighborhood boundaries remain visible.
\fill[red!75!black, opacity=0.06]
  (Q0) circle[radius=\RadiusFinal];
\fill[black, opacity=0.03]
  (Q0) circle[radius=\RadiusOne];
\fill[blue!70!black, opacity=0.06]
  (P0) circle[radius=\RadiusTwo];

% Distinct solid, dashed, and dotted boundaries.
\node[levelfinal, minimum size=\DiameterFinal]
  (GEta) at (Q0) {};
\node[levelone, minimum size=\DiameterOne]
  (GOne) at (Q0) {};
\node[leveltwo, minimum size=\DiameterTwo]
  (GTwo) at (P0) {};

% Relative-entropy radii.
\draw[thick, black!75] (Q0) --
  node[midway, fill=white, inner sep=1pt] {$\gamma_1$}
  (P0);

\draw[thick, blue!70!black] (P0) --
  node[midway, fill=white, inner sep=1pt] {$\gamma_2$}
  (P2);

\draw[thick, red!75!black] (Q0) --
  node[midway, fill=white, inner sep=1pt, text=red!75!black]
  {$\eta$}
  (P2);

% Cross markers and adjacent measure labels.
\node[point, text=black!75] at (Q0) {$\times$};
\node[measure, anchor=north west, xshift=5pt, yshift=-6pt]
  at (Q0) {$Q$};

\node[point, text=blue!70!black] at (P0) {$\times$};
\node[
  measure,
  anchor=north west,
  xshift=8pt,
  yshift=-8pt,
  text=blue!70!black,
  fill=white,
  fill opacity=0.75,
  text opacity=1,
  inner sep=2pt
] at (P0)
  {$P^{(\mathsf{L}_{n_0},Q,\lambda_0)}_{
    \vect{\Theta}\mid\vect{Z}=\vect{z}_0}$};

\node[point, text=red!75!black] at (P2) {$\times$};
\node[
  measure,
  anchor=south west,
  xshift=8pt,
  yshift=10pt,
  fill=white,
  fill opacity=0.75,
  text opacity=1,
  inner sep=2pt
] at (P2)
  {$
    \textcolor{red!75!black}{
      P^{\autoparent{
        \mathsf{L}_{n_2},Q,
        \frac{n_0}{n_2}\lambda_0
      }}_{\vect{\Theta}\mid\vect{Z}=\vect{z}_2}
    }
    =
    \textcolor{green!50!black}{
      P^{\autoparent{
        \mathsf{L}_{n_1},
        P^{(\mathsf{L}_{n_0},Q,\lambda_0)}_{
          \vect{\Theta}\mid\vect{Z}=\vect{z}_0},
        -\frac{n_0}{n_1}\lambda_0
      }}_{\vect{\Theta}\mid\vect{Z}=\vect{z}_1}
    }
  $};

% Labels inside their corresponding neighborhoods.
\coordinate (GammaLabel) at ($(Q0)!0.65!(GOne.220)$);
\coordinate (EtaLabel) at ($(GOne.145)!0.55!(GEta.145)$);

\node[setlabel, text=black!75]
  at (GammaLabel)
  {$\mathcal{G}_{Q,\gamma_1}$};

\node[setlabel, text=red!75!black]
  at (EtaLabel)
  {$\mathcal{G}_{Q,\eta}$};

\node[setlabel, text=blue!70!black,   fill=white,
  fill opacity=0.75,
  text opacity=1,
  inner sep=2pt]
  at ($(P0)!0.70!(GTwo.30)$)
  {$\mathcal{G}_{
    P^{(\mathsf{L}_{n_0},Q,\lambda_0)}_{
      \vect{\Theta}\mid\vect{Z}=\vect{z}_0},
    \gamma_2
  }$};

% Ambient statistical manifold.
\node[
  rectangle,
  draw=black,
  fill=none,
  line width=0.75pt,
  fit=(current bounding box),
  inner sep=1cm
] (Ambient) {};

\node[font=\Large, anchor=north east, inner sep=8pt]
  at (Ambient.north east)
  {$\triangle\autoparent{\mathcal{M}}$};

\end{tikzpicture}

%% file: Figures/Journal-TIT2026/Nested_Decentralized/Nested_structure.tikz
\begin{tikzpicture}[
  line cap=round, line join=round,
  arr/.style={-{Stealth[length=2.0mm,width=2.0mm]}, line width=0.7pt},
  msgarr/.style={-{Stealth[length=2.0mm,width=2.0mm]}, line width=0.7pt, blue},
  backarr/.style={-{Stealth[length=2.0mm,width=2.0mm]}, line width=0.7pt, red}, % <- change color here
  node distance=18mm and 26mm,
  prior/.style={align=center, inner sep=6pt},
  client/.style={draw, rounded corners=2pt, align=left, inner sep=7pt, minimum width=2.5cm},
  data/.style={text depth=0.50ex},
  msg/.style={midway, above, align=center, font=\scriptsize, text depth=0.30ex, blue},
  msgback/.style={midway, below, align=center, font=\scriptsize, text depth=0.30ex, red},
]

% --- Initial reference --------------------------------------------------------
\node[prior] (Q1) {$Q_1$};

% --- Clients ------------------------------------------------------------------
\node[client, right=10mm of Q1] (C1) {%
  \textbf{Client 1}\\[1mm]
  $
  P^{\autoparent{\mathsf{L}_{n_1}, Q_1,\lambda_1}}_{\vect{\Theta}\mid \vect{Z}=\vect{z}_1}$};

\node[client, right=26mm of C1] (C2) {%
  \textbf{Client 2}\\[1mm]
  $
  P^{\autoparent{\mathsf{L}_{n_2},Q_2,\lambda_2}}_{\vect{\Theta}\mid \vect{Z}=\vect{z}_2}$};

\node[right=26mm of C2, font=\Large] (dots) {$\cdots$};

\node[client, right=30mm of dots] (CK) {%
  \textbf{Client $K$}\\[1mm]
  $P^{\autoparent{\mathsf{L}_{n_K},Q_K,\lambda_K}}_{\vect{\Theta}\mid \vect{Z}=\vect{z}_K}$};

% --- Local datasets -----------------------------------------------------------
\node[data, below=8mm of C1] (Z1) {$\vect{z}_1$};
\node[data, below=8mm of C2] (Z2) {$\vect{z}_2$};
\node[data, below=8mm of CK] (ZK) {$\vect{z}_K$};

\draw[arr] (Z1) -- (C1.south);
\draw[arr] (Z2) -- (C2.south);
\draw[arr] (ZK) -- (CK.south);

% --- Forward measure messages -------------------------------------------------
\draw[arr] (Q1) -- (C1);

\draw[msgarr]  ([yshift=2.5mm]C1.east) --([yshift=2.5mm]C2.west)
node[msg] {$
P^{\autoparent{\mathsf{L}_{n_1}, Q_1,\lambda_1}}_{\vect{\Theta}\mid \vect{Z}=\vect{z}_1}=Q_2
$\\[-2mm]};

\draw[msgarr]   ([yshift=2.5mm]C2.east)
 --  ([yshift=2.5mm]dots.west)
node[msg] {$
P^{\autoparent{\mathsf{L}_{n_2}, Q_2,\lambda_2}}_{\vect{\Theta}\mid \vect{Z}=\vect{z}_2}=Q_3
$\\[-2mm]};

\draw[msgarr] ([yshift=2.5mm]dots.east) -- ([yshift=2.5mm]CK.west) 
node[msg] {$
P^{\autoparent{\mathsf{L}_{n_{K-1}}, Q_{K-1},\lambda_{K-1}}}_{\vect{\Theta}\mid \vect{Z}=\vect{z}_{K-1}}=Q_K
$\\[-2mm]};

% --- Back arrows (under the forward ones, different color) --------------------
% (drawn slightly below each forward arrow using yshift)
\draw[backarr] ([yshift=-2.5mm]C2.west) -- ([yshift=-2.5mm]C1.east)
node[msgback] {$P^{\autoparent{\mathsf{L}_{n_K},Q_K,\lambda_K}}_{\vect{\Theta}\mid \vect{Z}=\vect{z}_K}$};

\draw[backarr] ([yshift=-2.5mm]dots.west) -- ([yshift=-2.5mm]C2.east)
node[msgback] {$P^{\autoparent{\mathsf{L}_{n_K},Q_K,\lambda_K}}_{\vect{\Theta}\mid \vect{Z}=\vect{z}_K}$};

\draw[backarr] ([yshift=-2.5mm]CK.west) -- ([yshift=-2.5mm]dots.east)
node[msgback] {$P^{\autoparent{\mathsf{L}_{n_K},Q_K,\lambda_K}}_{\vect{\Theta}\mid \vect{Z}=\vect{z}_K}$};

\end{tikzpicture}

%% file: Figures/Journal-TIT2026/Nested_Decentralized/geometric_interpretation.tikz
\begin{tikzpicture}[
  x=1cm, y=1cm,
  line cap=round,
  line join=round,
  >=Stealth
]

%--- Coordinates: first case of recursive_nesting_inclusions ---
\pgfmathsetmacro{\FirstRadius}{1.75*2.35}
\pgfmathsetmacro{\FinalRadius}{1.75*3.50}
\def\FirstAngle{150}
\def\FinalAngle{55}
\def\QThreeRise{1.45}
\def\QFourDrop{0.65}

\coordinate (Q1) at (0,0);
\coordinate (Q2) at ($(Q1)+(\FirstAngle:\FirstRadius)$);
\coordinate (Q5) at ($(Q1)+(\FinalAngle:\FinalRadius)$);
\coordinate (Q3) at ($(Q2)!0.33!(Q5)+(0,\QThreeRise)$);
\coordinate (Q4) at ($(Q2)!0.67!(Q5)+(0,-\QFourDrop)$);

%--- Styles ---
\tikzset{
  GiBase/.style={line width=0.85pt},
  G0/.style={GiBase, draw=red!75!black, solid},
  G1/.style={GiBase, draw=blue!70!black, dotted, line width=1.1pt},
  G2/.style={GiBase, draw=orange!75!black, dashed},
  G3/.style={GiBase, draw=violet!80!black, dash dot},
  G4/.style={GiBase, draw=green!60!black, dash dot dot},
  GiLbl/.style={font=\normalsize, fill=white, inner sep=1.5pt},
  measure/.style={font=\large, inner sep=1pt},
  ArrSeg/.style={
    line width=0.9pt, black!75,
    -{Stealth[length=5pt,width=4pt]},
    shorten <=3pt, shorten >=3pt
  }
}

% Light fills first; boundaries are drawn afterwards.
\foreach \Center/\Boundary/\Color in {
  Q1/Q5/red!75!black,
  Q1/Q2/blue!70!black,
  Q2/Q3/orange!75!black,
  Q3/Q4/violet!80!black,
  Q4/Q5/green!60!black}{
  \fill[\Color, opacity=0.025]
    let \p1=(\Center), \p2=(\Boundary),
        \n1={veclen(\x2-\x1,\y2-\y1)}
    in (\Center) circle[radius=\n1];
}

%--- Circles ---
\draw[G0]
  let
    \p1=(Q1),
    \p2=(Q5),
    \n1={veclen(\x2-\x1,\y2-\y1)}
  in
    (Q1) circle (\n1);

\draw[G1]
  let
    \p1=(Q1),
    \p2=(Q2),
    \n1={veclen(\x2-\x1,\y2-\y1)}
  in
    (Q1) circle (\n1);

\draw[G2]
  let
    \p1=(Q2),
    \p2=(Q3),
    \n1={veclen(\x2-\x1,\y2-\y1)}
  in
    (Q2) circle (\n1);

\draw[G3]
  let
    \p1=(Q3),
    \p2=(Q4),
    \n1={veclen(\x2-\x1,\y2-\y1)}
  in
    (Q3) circle (\n1);

\draw[G4]
  let
    \p1=(Q4),
    \p2=(Q5),
    \n1={veclen(\x2-\x1,\y2-\y1)}
  in
    (Q4) circle (\n1);

%--- Path / iterates ---
\draw[ArrSeg]
  (Q1) --
  node[midway, fill=white, inner sep=1pt] {$\gamma_1$}
  (Q2);

\draw[ArrSeg]
  (Q2) --
  node[midway, fill=white, inner sep=1pt] {$\gamma_2$}
  (Q3);

\draw[ArrSeg]
  (Q3) --
  node[midway, fill=white, inner sep=1.5pt] {$\gamma_3$}
  (Q4);

\draw[ArrSeg]
  (Q4) --
  node[midway, fill=white, inner sep=1.5pt] {$\gamma_4$}
  (Q5);

% Auxiliary connections, subdued to keep the nesting path prominent.
\draw[draw=black!30, line width=0.45pt] (Q1) -- (Q3);
\draw[draw=black!30, line width=0.45pt] (Q1) -- (Q4);

\draw[thick, red!75!black]
  (Q1) --
  node[
    midway,
    fill=white,
    inner sep=1pt,
    text=red!75!black
  ]
  {$\eta$}
  (Q5);

%--- Points ---
\draw[draw=red!75!black, line width=0.8pt]
  ($(Q1)+(-1.7pt,-1.7pt)$) -- ($(Q1)+(1.7pt,1.7pt)$)
  ($(Q1)+(-1.7pt,1.7pt)$) -- ($(Q1)+(1.7pt,-1.7pt)$);
\draw[draw=black!75, line width=0.8pt]
  ($(Q2)+(-1.7pt,-1.7pt)$) -- ($(Q2)+(1.7pt,1.7pt)$)
  ($(Q2)+(-1.7pt,1.7pt)$) -- ($(Q2)+(1.7pt,-1.7pt)$);
\draw[draw=black!75, line width=0.8pt]
  ($(Q3)+(-1.7pt,-1.7pt)$) -- ($(Q3)+(1.7pt,1.7pt)$)
  ($(Q3)+(-1.7pt,1.7pt)$) -- ($(Q3)+(1.7pt,-1.7pt)$);
\draw[draw=black!75, line width=0.8pt]
  ($(Q4)+(-1.7pt,-1.7pt)$) -- ($(Q4)+(1.7pt,1.7pt)$)
  ($(Q4)+(-1.7pt,1.7pt)$) -- ($(Q4)+(1.7pt,-1.7pt)$);
\draw[draw=black!75, line width=0.8pt]
  ($(Q5)+(-1.7pt,-1.7pt)$) -- ($(Q5)+(1.7pt,1.7pt)$)
  ($(Q5)+(-1.7pt,1.7pt)$) -- ($(Q5)+(1.7pt,-1.7pt)$);

% Labels are drawn last, so no later boundary runs over their text.
\node[measure,text=red!75!black] at ($(Q1)+(0,-0.50)$) {$Q_1$};
\node[measure] at ($(Q2)+(-0.60,0)$) {$Q_2$};
\node[measure] at ($(Q3)+(0,0.50)$) {$Q_3$};
\node[measure] at ($(Q4)+(0,-0.55)$) {$Q_4$};
\node[measure,anchor=west] at ($(Q5)+(0.40,0.50)$)
  {$Q_5=P^{(\mathsf{L}_{n_0},Q_1,\lambda_0)}_{
\vect{\Theta}\mid\vect{Z}=\vect{z}_0}$};

% Interior label positions from the first recursive-nesting case.
\path let \p1=(Q1), \p2=(Q5),
  \n1={veclen(\x2-\x1,\y2-\y1)} in
  node[GiLbl,text=red!75!black]
    at ($(Q1)+(270:{0.780000*\n1})$)
    {$\mathcal{G}_{Q_1,\eta}$};

\path let \p1=(Q1), \p2=(Q2),
  \n1={veclen(\x2-\x1,\y2-\y1)} in
  node[GiLbl,text=blue!70!black]
    at ($(Q1)+(249:{0.387391*\n1})$)
    {$\mathcal{G}_{Q_1,\gamma_1}$};

\path let \p1=(Q2), \p2=(Q3),
  \n1={veclen(\x2-\x1,\y2-\y1)} in
  node[GiLbl,text=orange!75!black]
    at ($(Q2)+(291:{0.611739*\n1})$)
    {$\mathcal{G}_{Q_2,\gamma_2}$};

\path let \p1=(Q3), \p2=(Q4),
  \n1={veclen(\x2-\x1,\y2-\y1)} in
  node[GiLbl,text=violet!80!black]
    at ($(Q3)+(15:{0.574348*\n1})$)
    {$\mathcal{G}_{Q_3,\gamma_3}$};

\path let \p1=(Q4), \p2=(Q5),
  \n1={veclen(\x2-\x1,\y2-\y1)} in
  node[GiLbl,text=green!60!black]
    at ($(Q4)+(306:{0.705217*\n1})$)
    {$\mathcal{G}_{Q_4,\gamma_4}$};

% Ambient statistical manifold, without background layers.
\node[rectangle, draw=black, fill=none, line width=0.75pt,
  fit=(current bounding box), inner sep=1cm] (Ambient) {};
\node[font=\Large, anchor=north east, inner sep=8pt]
  at (Ambient.north east) {$\triangle\autoparent{\mathcal{M}}$};

\end{tikzpicture}

%% file: Figures/Journal-TIT2026/FL.tikz
\begin{tikzpicture}[
  line cap=round,
  line join=round,
  >=Stealth,
  msgarr/.style={
    ->,
    line width=0.8pt,
    blue,
    shorten <=0.5mm,
    shorten >=0.5mm
  },
  backarr/.style={
    ->,
    line width=0.8pt,
    red,
    densely dotted,
    shorten <=0.5mm,
    shorten >=0.5mm
  },
  prior/.style={
    draw,
    rounded corners=3pt,
    align=center,
    inner sep=8pt,
    minimum width=5.8cm,
    minimum height=1.8cm
  },
  client/.style={
    draw,
    rounded corners=3pt,
    align=center,
    inner sep=8pt,
    minimum width=3.0cm,
    minimum height=1.4cm
  },
  upmsg/.style={
    font=\scriptsize,
    text=blue,
    fill=white,
    inner sep=1pt
  },
  downmsg/.style={
    font=\scriptsize,
    text=red,
    fill=white,
    inner sep=1pt
  },
  clabel/.style={font=\Large},
  dots/.style={font=\Large}
]

% ===================== Nodes =====================

\node[prior] (Top) at (0,4) {%
Server};

\node[client] (C1) at (-6,0) {%
Client 1};

\node[client] (Ck) at (0,0) {%
Client k};

\node[client] (CK) at (6,0) {%
Client K};

\node[dots] at (-3,0) {$\cdots$};
\node[dots] at ( 3,0) {$\cdots$};

%\node[clabel] at (-6,-1.35) {Client 1};
%\node[clabel] at ( 0,-1.35) {Client k};
%\node[clabel] at ( 6,-1.35) {Client K};

% ===================== Arrow coordinates =====================

% --- Left client pair (unchanged / balanced) ---
\coordinate (LupA)   at ($(C1.north)+(0.10,0.40)$);
\coordinate (LupB)   at ($(Top.south west)+(0.22,-0.42)$);

\coordinate (LdownA) at ($(LupB)+(0.55,0)$);
\coordinate (LdownB) at ($(LupA)+(0.55,0)$);

% --- Middle client pair (fixed) ---
% vertical pair, clearly separated and away from the boxes
\coordinate (MupA)   at ($(Ck.north)+(-0.38,0.42)$);
\coordinate (MupB)   at ($(Top.south)+(-0.38,-0.42)$);

\coordinate (MdownA) at ($(Top.south)+(0.38,-0.42)$);
\coordinate (MdownB) at ($(Ck.north)+(0.38,0.42)$);

% --- Right client pair (Client K) ---
\coordinate (RupA)   at ($(CK.north)+(-0.20,0.42)$);
\coordinate (RupB)   at ($(Top.south east)+(-1.00,-0.42)$);

% red arrow = parallel shifted copy of blue arrow
\coordinate (RdownA) at ($(RupB)+(0.58,0)$);
\coordinate (RdownB) at ($(RupA)+(0.58,0)$);

% Client K
\draw[msgarr] (RupA) -- (RupB)
  node[pos=.48, below, sloped, upmsg] {$\displaystyle
  P^{\autoparent{\mathsf{L}_{n_K},Q_K,\lambda_K}}_{\vect{\Theta}\mid Z=z_K}$};

\draw[backarr] (RdownA) -- (RdownB)
  node[pos=.52, above, sloped, downmsg] {$S_K$};

% ===================== Arrows + messages =====================

% Client 1
\draw[msgarr] (LupA) -- (LupB)
  node[pos=.48, above, sloped, upmsg] {$\displaystyle
  P^{\autoparent{\mathsf{L}_{n_1},Q_1,\lambda_1}}_{\vect{\Theta}\mid Z=z_1}$};

\draw[backarr] (LdownA) -- (LdownB)
  node[pos=.60, below, sloped, downmsg] {$S_K$};

% Client k  (fixed)
\draw[msgarr] (MupA) -- (MupB)
  node[pos=.52, left=2pt, rotate=90, anchor=south, upmsg] {$\displaystyle
  P^{\autoparent{\mathsf{L}_{n_k},Q_k,\lambda_k}}_{\vect{\Theta}\mid Z=z_k}$};

\draw[backarr] (MdownA) -- (MdownB)
  node[pos=.50, right=2pt, rotate=-90, anchor=south, downmsg] {$S_K$};
%
%% Client K  (fixed)
%\draw[msgarr] (RupA) -- (RupB)
%  node[pos=.44, above, sloped, upmsg] {$\displaystyle
%  P^{\autoparent{\mathsf{L}_{n_K},Q_K,\lambda_K}}_{\vect{\Theta}\mid Z=z_K}$};
%
%\draw[backarr] (RdownA) -- (RdownB)
%  node[pos=.58, below, sloped, downmsg] {$S_K$};

\end{tikzpicture}